\documentclass[draftcls, onecolumn]{IEEEtran}
\IEEEoverridecommandlockouts
\usepackage{algorithm}
\usepackage{algpseudocode}
\usepackage{graphicx}
\usepackage{textcomp}
\usepackage{xcolor}
\def\BibTeX{{\rm B\kern-.05em{\sc i\kern-.025em b}\kern-.08em
    T\kern-.1667em\lower.7ex\hbox{E}\kern-.125emX}}

\usepackage{amssymb}
\usepackage{amsfonts}
\usepackage{amssymb}
\usepackage{amsbsy}
\usepackage{amsfonts}
\usepackage{amsthm, mathrsfs, amsmath}
\usepackage{longtable}
\usepackage{stackengine}
\usepackage{mathtools}
\usepackage{stmaryrd}
\usepackage{relsize}
\usepackage{multirow}
\usepackage{bm}
\usepackage[T1]{fontenc}
\usepackage{subcaption}
\usepackage{caption}
\usepackage{dsfont}
\usepackage{cite}
\usepackage{url}
\usepackage{color}   
\usepackage{epsfig}
\usepackage{xfrac}
\usepackage{nicefrac}
\usepackage{setspace}
\usepackage{array}
\usepackage{booktabs,tabularx}

\usepackage{cuted}
\usepackage{multicol}
\usepackage{lipsum}

\usepackage{float}% http://ctan.org/pkg/float
\usepackage{dblfloatfix}

\usepackage{hyperref}
\hypersetup{
    colorlinks=true,% make the links colored
}

\usepackage[T1]{fontenc}         % Not always necessary, but recommended.
\usepackage{amstext}
\usepackage{amssymb,amsmath,amsthm,enumitem}
\usepackage{sansmath}

\usepackage{mathrsfs}
\DeclareMathAlphabet{\mathpzc}{OT1}{pzc}{m}{it}

\def \H {\mathrm{H}} % Entropy (discrete)
\def \I {\mathrm{I}} % Mutual Information
\def \D {\mathrm{D}} % Divergence

\newcommand{\orcid}[1]{\href{https://orcid.org/#1}{\includegraphics[width=10pt]{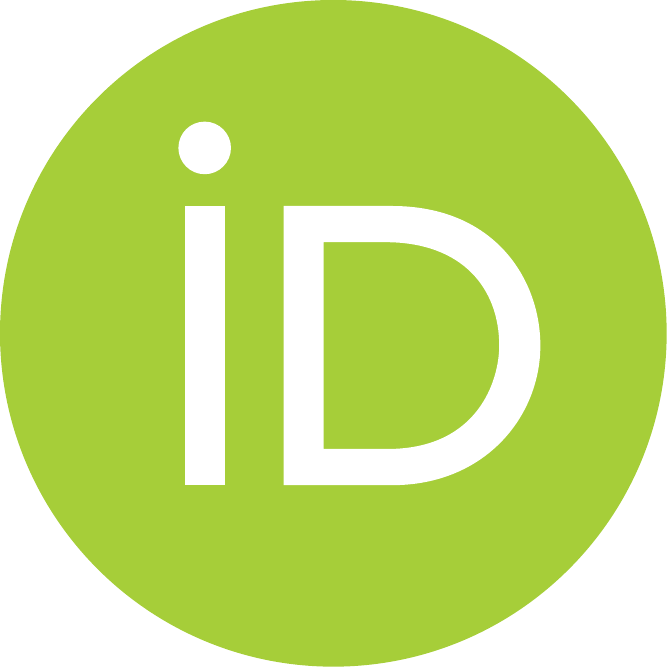}}}

\theoremstyle{definition}

\newtheorem{definition}{Definition}%[section]

\newtheorem{remark}{Remark}

\def\markov{\hbox{$\--$}\kern-1.5pt\hbox{$\circ$}\kern-1.5pt\hbox{$\--$}}

\makeatletter
\newcommand*{\centernot}{%
  \mathpalette\@centernot
}
\def\@centernot#1#2{%
  \mathrel{%
    \rlap{%
      \settowidth\dimen@{$\m@th#1{#2}$}%
      \kern.5\dimen@
      \settowidth\dimen@{$\m@th#1=$}%
      \kern-.5\dimen@
      $\m@th#1\not$%
    }%
    {#2}%
  }%
}
\makeatother

\renewcommand\thesection{\arabic{section}}
\renewcommand \thesubsection{\arabic{section}.\arabic{subsection}}
\def\thesectiondis{\thesection.} 
\def\thesubsectiondis{\thesectiondis\arabic{subsection}.} 
\makeatletter
\def\@seccntformatinl#1{\csname the#1dis\endcsname\hskip 1em\relax}
\makeatother
\usepackage{tocloft}
\makeatletter
\newcommand*{\rom}[1]{\expandafter\@slowromancap\romannumeral #1@}
\makeatother

\usepackage{makecell}

\makeatletter
\def\ps@headings{%
\def\@oddhead{\mbox{}\scriptsize\rightmark \hfil \thepage}%
\def\@evenhead{\scriptsize\thepage \hfil \leftmark\mbox{}}%
\def\@oddfoot{\scriptsize \@date\hfil DRAFT, Version 1.0}%
\def\@evenfoot{\scriptsize DRAFT, Version 1.0\hfil \@date}}
\makeatother

\let\counterwithin\relax
\usepackage{chngcntr}

\usepackage{footmisc}

\begin{document}

\title{\huge %Taxonomy of Bottleneck Problems:\\ 
Bottlenecks CLUB:\\
Unifying Information-Theoretic Trade-offs Among\\Complexity, Leakage, and Utility 
%\thanks{XXX}
}
%
%%% author block for TIFS submission
%
\author{Behrooz~Razeghi$^\ast$,~%\IEEEmembership{Student~Member,~IEEE,}
			Flavio~P.~Calmon,~%\IEEEmembership{Senior~Member,~IEEE,}
            Deniz~G\"{u}nd\"{u}z,~%,~\IEEEmembership{Senior~Member,~IEEE}% <-this % stops a space
            and~Slava~Voloshynovskiy$^\ast$%\IEEEmembership{Senior~Member,~IEEE,}
\thanks{$^\ast$ Corresponding Authors.}
\thanks{B.~Razeghi and S.~Voloshynovskiy are with the University of Geneva, Switzerland (e-mail:~\{behrooz.razeghi, svolos\}@unige.ch).}%
\thanks{F.~P.~Calmon is with the Harvard University, US (e-mail:~flavio@seas.harvard.edu).}% <-this % stops a space
%\thanks{}%
\thanks{D.~G\"{u}nd\"{u}z is with the Imperial College London, UK (e-mail:~d.gunduz@imperial.ac.uk).}% <-this % stops a space
%\thanks{A part of the material in this manuscript has been accepted in 2020 IEEE Workshop on Information Forensics and Security (WIFS'20) \cite{razeghi2020perfectobfuscation}.}
\thanks{\urlstyle{sf}Implementation codes available at: \scriptsize{\url{https://github.com/BehroozRazeghi}}}
}

\maketitle

%---------------------------------------------------
%		          	Abstract
%---------------------------------------------------
\vspace{-30pt}
\begin{abstract}
Bottleneck problems are an important class of optimization problems that have recently gained increasing attention in the domain of machine learning and information theory.
They are widely used in generative models, fair machine learning algorithms, design of privacy-assuring mechanisms, and appear as information-theoretic performance bounds in various multi-user communication problems. In this work, we propose a general family of optimization problems, termed as \textit{complexity-leakage-utility bottleneck (CLUB)} model, which (i) provides a unified theoretical framework that generalizes most of the state-of-the-art literature for the information-theoretic privacy models, (ii) establishes a new interpretation of the popular generative and discriminative models, (iii) constructs new insights to the generative compression models, and (iv) can be used in the fair generative models. We first formulate the CLUB model as a complexity-constrained privacy-utility optimization problem. We then connect it with the closely related bottleneck problems, namely information bottleneck (IB), privacy funnel (PF), deterministic IB (DIB), conditional entropy bottleneck (CEB), and conditional PF (CPF). We show that the CLUB model generalizes all these problems as well as most other information-theoretic privacy models. Then, we construct the deep variational CLUB (DVCLUB) models by employing neural networks to parameterize variational approximations of the associated information quantities. Building upon these information quantities, we present unified objectives of the \textit{supervised} and \textit{unsupervised} DVCLUB models. Leveraging the DVCLUB model in an unsupervised setup, we then connect it with state-of-the-art generative models, such as variational auto-encoders (VAEs), generative adversarial networks (GANs), as well as the Wasserstein GAN (WGAN), Wasserstein auto-encoder (WAE), and adversarial auto-encoder (AAE) models through the optimal transport (OT) problem. We then show that the DVCLUB model can also be used in fair representation learning problems, where the goal is to mitigate the undesired bias during the training phase of a machine learning model. We conduct extensive quantitative experiments on colored-MNIST and CelebA datasets, with a public implementation available, to evaluate and analyze the CLUB model. 
%
%
%This shed some light on the connections between unsupervised Complexity-Utility-Leakage Bottleneck and generative models. 
%
%\keywords{authentication, privacy, ambiguization, ...}
%
%
%For an up-to-date and complete version of this paper, please see ... (arXiv link)
%
%
\end{abstract}

% Constructing an $\I$-measure, consistent with Shannon's information measures, we geometrically represent the relationship among Shannon's information measures in these problems, interpreting the differences in their objectives. 
%%% supervised and unsupervised CLUB
% We then generalize the CLUB model to comprise both \textit{supervised} and \textit{unsupervised} setups. 
%%% Deep Variational CLUB and approximations
%Using neural networks to parameterize the variational bounds of information quantities,
%
\vspace{-10pt}

\begin{IEEEkeywords}%
\vspace{-5pt}
\hspace{-16pt}Information-theoretic privacy, statistical inference, information bottleneck, information obfuscation, generative models.
\end{IEEEkeywords}

\pagebreak
{
%\singlespacing
  \hypersetup{linkcolor=blue!60!black}
  \tableofcontents
  \thispagestyle{headings}
}

%---------------------------------------------------
%				 Introduction
%---------------------------------------------------
\pagebreak
\section{Introduction}
\label{Sec:Introduction}

%$\CircleftarrowCirc$

\subsection{Motivation}
\label{ssec:Motivation}

%%-----------------------
%        First paragraph
%%-----------------------
%
% What’s the problem/opportunity?
% Why is it important?
% What is wrong with current solutions?

Releasing an \textit{`optimal' representation} of data for a given task while simultaneously assuring \textit{privacy} of the individuals' identity and their associated data is an important challenge in today's highly connected and data-driven world, and has been widely studied in the information theory, signal processing, data mining and machine learning communities. 
An optimal representation is the most useful (sufficient), compressed (compact), and least privacy-breaching (minimal) representation of data. 
Indeed, an optimal representation of data can be obtained subject to constraints on the target task and its computational and storage complexities. 

%Note that, a maximally informative representation of data is not necessarily a maximally useful one. 

%%-----------------------
%        Second paragraph
%%-----------------------
%
% What is our approach? Describe what the system/analysis/paper does

\begin{figure}[!t]
\centering
\includegraphics[scale=0.82]{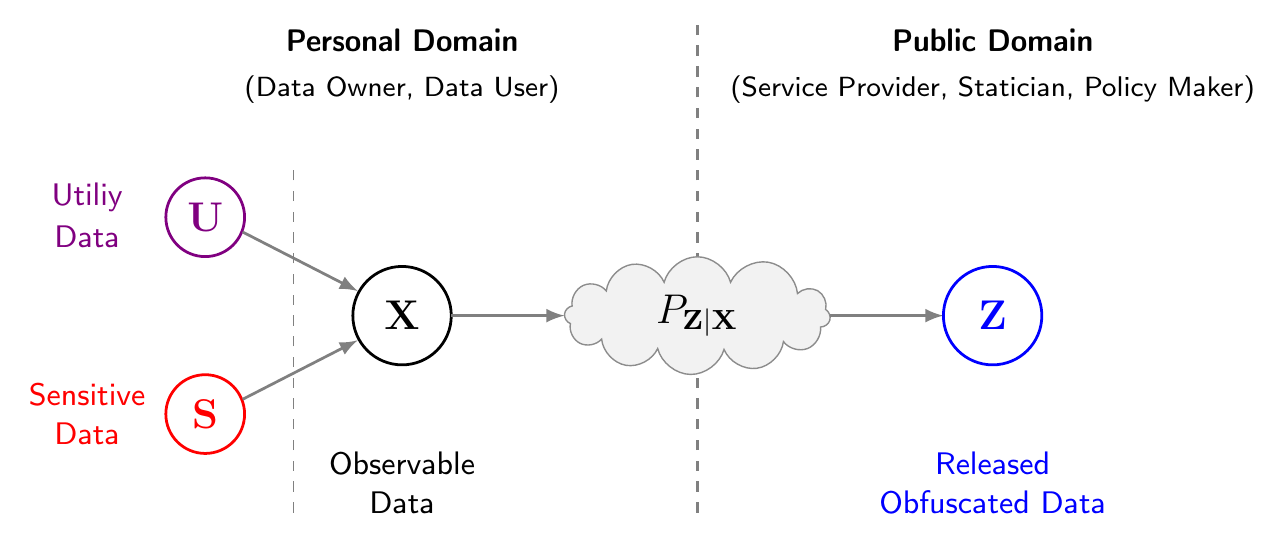}
%\vspace{-4pt}
\caption{The general complexity-leakage-utility bottleneck framework.}
%\vspace{-8pt}
\label{Fig:Fist_diagram}
\end{figure}

\begin{figure}[!t]
\centering
\includegraphics[scale=1]{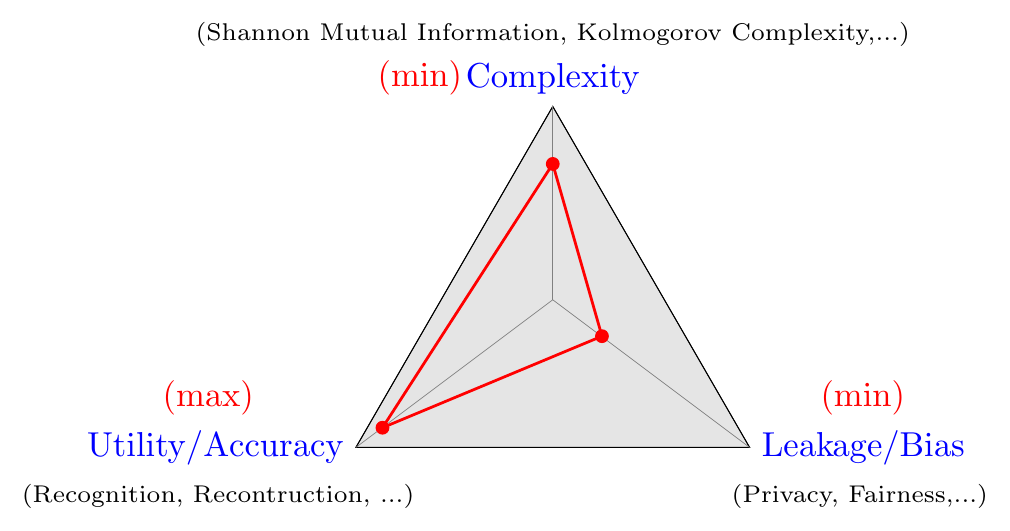}
\vspace{-5pt}
\caption{The schematic diagram of the trade-off among complexity, leakage and utility.}
\vspace{-5pt}
\label{Fig:triangle-opt-tradeoffs}
\end{figure}

We investigate the problem of privacy-preserving data representation for a specific \textit{utility task}, e.g., classification, identification, or reconstruction. 
Treating utility and privacy in a statistical framework \cite{du2012privacy} with mutual information as both the utility and obfuscation\footnote{Since the unified notion of privacy in  the computer science community is the ``differential privacy'', we try to use ``obfuscation'' instead of `privacy' when this does not cause confusion with common terminology in the literature, e.g., privacy funnel.} measures, we generalize the privacy funnel (PF) \cite{makhdoumi2014information} and information bottleneck (IB) \cite{tishby2000information} models, and introduce a new and more general model called \textit{complexity-leakage-utility bottleneck (CLUB)}. 
Consider two parties, a data owner and a utility service provider. The data owner observes a random variable $\mathbf{X}$ and acquires some utility from the service provider based on the information he discloses. 
%For instance, 
Simultaneously, the data owner wishes to limit the amount of information revealed about a sensitive random variable $\mathbf{S}$ that depends on $\mathbf{X}$. Therefore, instead of revealing $\mathbf{X}$ directly to the service provider, the data owner releases a new representation, denoted by $\mathbf{Z}$. 
The amount of information leaked to the service provider (public domain) about the sensitive variable $\mathbf{S}$ is measured by the mutual information $\I \left( \mathbf{S}; \mathbf{Z} \right)$. 
Moreover, the data owner is subjected to a constraint on information complexity of representation that is revealed to the service provider. This imposed information complexity is measured by $\I \left( \mathbf{X}; \mathbf{Z} \right)$. 
Moreover, in general, the acquired utility depends on a utility random variable $\mathbf{U}$ that is dependent on $\mathbf{X}$ and may also be correlated with $\mathbf{S}$. 
The amount of useful information revealed to the service provider is measured by $\I \left( \mathbf{U}; \mathbf{Z} \right)$. 
Therefore, considering a Markov chain $\left( \mathbf{U}, \mathbf{S}\right) \markov \mathbf{X} \markov \mathbf{Z}$, 
our aim is to share a \textit{privacy-preserving (sanitized) representation} $\mathbf{Z}$ of \textit{observed data} $\mathbf{X}$, through a stochastic mapping $P_{\mathbf{Z} \mid \mathbf{X}}$, while preserving information about \textit{utility attribute} $\mathbf{U}$ and obfuscating information about \textit{sensitive attribute} $\mathbf{S}$. 
The stochastic mapping $P_{\mathbf{Z} \mid \mathbf{X}}$ is called \textit{complexity-constrained obfuscation-utility-assuring mapping}. 
The general diagram of our setup is depicted in Fig.~\ref{Fig:Fist_diagram}. The schematic diagram of the trade-off among complexity, leakage and utility is depicted in Fig.~\ref{Fig:triangle-opt-tradeoffs}.

\subsection{Contributions}
\label{ssec:Contributions}

%\textcolor{gray}{(Behrooz: to complete and revise when paper finalized)}
%
%%We adopt an information-theoretic framework of deep networks and propose
%
\begin{itemize}
\item
Inspired by \cite{li2017extended}, we propose the CLUB model and provide a characterization of the complexity-leakage-utility trade-off for this model. 
The CLUB model provides a statistical inference framework that generalizes most of the state-of-the-art literature in the information-theoretic privacy models. 
To the best of our knowledge, this is the first unified formulation, which can bridge machine-learning and information-theoretic privacy communities. 
%This unified model can serve as a main reference for researchers and practitioners interested in designing privatizer mechanisms. 
%
\item
We provide new insights into the representation learning problems by bridging information-theoretic privacy with the generative models through the bottleneck principle. We start from a purely information-theoretic framework that has roots in the classical Shannon rate-distortion theory. Next we demonstrate the connection of our model to several recent research trends in generative models and representation learning. 
In particular, we show that the CLUB model has connections to the variational fair auto-encoder (VFAE) model \cite{louizos2015variational} to learn representations for a prediction problem while removing potential biases against some variable (e.g., gender, ethnicity)
%, sexual orientation
from the results of a learned model \cite{dwork2012fairness, kamiran2012data, zemel2013learning, hardt2016equality, louizos2015variational, calmon2017optimized, zafar2017fairness, louppe2017learning, wadsworth2018achieving, zhang2018mitigating}. 
%Moreover, we show that the CLUB model can be view as an adversarial model, in which their goal is to obtain an invariant representation using an iterative mini-max game. 
\item
We generalize the CLUB perspective to comprise both \textit{supervised} and \textit{unsupervised} setups. 
In the supervised setup, $\mathbf{U}$ is a generic attribute of data $\mathbf{X}$, while in the unsupervised setup, the data owner wishes to release the original domain data $\mathbf{X}$ (e.g., facial images) as accurately as possible (i.e., $\mathbf{U} \equiv \mathbf{X}$), without revealing a specific sensitive attribute $\mathbf{S}$ (e.g., gender, emotion, etc.). In the unsupervised setup, the deep variational CLUB (DVCLUB) model is shown to have interesting connections with the several generative models such as variational auto-encoder (VAE) \cite{kingma2014auto}, $\beta$-VAE \cite{higgins2016beta}, InfoVAE \cite{zhao2017infovae}, generative adversarial network (GAN) \cite{goodfellow2014generative}, Wasserstein GAN (WGAN) \cite{arjovsky2017wasserstein}, Wasserstein auto-encoders \cite{tolstikhin2018wasserstein}, adversarial auto-encoder (AAE) \cite{makhzani2015adversarial}, VAE-GAN \cite{larsen2016autoencoding}, and BIB-AE \cite{Voloshynovskiy2019NeurIPS}. 
%From one hand, our analysis show that how the 
%
% \item
% We study perfect information obfuscation under a local information geometry analysis. 
% This analysis provides us useful insight into information coupling and trajectory construction of perturbation of probability mass functions. Next, we characterize the necessary and sufficient conditions under which a non-trivial amount of useful information is feasible under perfect information obfuscation constraint. 
% Furthermore, we develop the notion of perfect obfuscation based on $\chi^2$-divergence and Kullback–Leibler divergence in the Euclidean information space.
%
\item
We conduct extensive qualitative and quantitative experiments on several real-world datasets to evaluate and validate the effectiveness of the proposed CLUB model. %Experimental results show that the CLUB model can .....
%
% Data driven .... to learn representations for high-dimensional distributions. 
%
%
% This analysis allows us to construct the modal decomposition of the joint distributions, divergence transfer matrices and mutual information. 
%By decomposing the mutual information into orthogonal modes, we obtain the locally sufficient statistics for inferences about utility attributes, while satisfying perfect obfuscation constraints. 
%
%
%
%\item
%Next, we adopt the general statistical inference framework and address a new privacy concept, named ``nested leakage''.   
%The nested privacy model (i) takes into account the inherent privacy leakage in data release mechanisms by removing the pre-defined sensitive attribute for a particular utility service at hand, (ii) proposes fundamental required injected obfuscation in order to avoid adversarial inference based on the nature of attribute, (iii) ...
%%\item
%%%Finally, we propose the \textit{Wyner Compressive Adversarial Privacy} model to address the ...
%\item
%We propose a data-driven ...
%\item 
%We....
\end{itemize}

%\textcolor{red}{Behrooz: to write}
%The main objective ....

%---------------------------------------------------
%  Figure: Complexity-Leakage-Utility Bottleneck General setup
%---------------------------------------------------
%
%---------------------------------------------------
%---------------------------------------------------

%\vspace{-9pt}

%---------------------------------------------------
%
%				 State of the Art
%
%---------------------------------------------------
%\subsection{Privacy Protection Mechanisms: State-of-the-Art}
\subsection{State-of-the-Art}
\label{ssec:RelatedWork}

%\vspace{-3pt}
%---------------------------------------------------
%%       Classical popular Models / Statistical Models
%---------------------------------------------------
%
%\paragraph*{Statistical Privacy Models}. 
Addressing data anonymization in \cite{sweeney2000simple}, various well-known statistical formulations and schemes were proposed for preserving-privacy of data, such as $k$-anonymity \cite{sweeney2002k}, $\ell$-diversity \cite{machanavajjhala2006diversity}, $t$-closeness \cite{li2007t},  differential privacy (DP) \cite{dwork2006calibrating}, and pufferfish \cite{kifer2012rigorous}, which are basically based on some form of data perturbation. These mechanisms mainly focus on the querying data, inference algorithms and transporting probability measures. DP is the most popular \textit{context-free} notion of privacy, which is characterized in terms of the distinguishability of `neighboring' databases. 
%bounds the variation of the distribution of the released data 
However, DP does not provide any guarantee on the (average or maximum) information leakage \cite{du2012privacy}. 
Pufferfish framework is a generalized version of DP that can capture data correlation; however, it does not focus on preserving data utility.  
% Fundamentally, these mechanisms 
%This weakness leads the researchers to propose  the context-aware notions of privacy which are categorized in the domain of information theoretic (IT) privacy. 
%The classical research was mainly focus on ...

%\textcolor{gray}{Behrooz: to complete}

\pagebreak

%---------------------------------------------------
%			  IT based Models 
%---------------------------------------------------
%
%\paragraph*{Information Theoretic Privacy Models}. 
Information theoretic (IT) privacy approaches 
%\cite{reed1973information,yamamoto1983source, evfimievski2003limiting, rebollo2009t, du2012privacy, sankar2013utility, calmon2013bounds, makhdoumi2013privacy, asoodeh2014notes, calmon2015fundamental, salamatian2015managing, basciftci2016privacy, kalantari2017information, asoodeh2018estimation, Hsu2019watchdogs, rassouli2018perfect, rassouli2019data}, 
\cite{reed1973information,yamamoto1983source, evfimievski2003limiting, rebollo2009t, du2012privacy, sankar2013utility, calmon2013bounds, makhdoumi2013privacy, asoodeh2014notes, calmon2015fundamental, salamatian2015managing, basciftci2016privacy, asoodeh2016information, kalantari2017information, rassouli2018latent, asoodeh2018estimation, rassouli2018perfect, liao2018privacy, osia2018deep, tripathy2019privacy, Hsu2019watchdogs, liao2019tunable,  sreekumar2019optimal, xiao2019maximal, diaz2019robustness, rassouli2019data, rassouli2019optimal, rassouli2021perfect} 
model and analyze privacy-utility trade-offs using IT metrics to provide 
asymptomatic 
%\cite{reed1973information, yamamoto1983source, evfimievski2003limiting, sankar2013utility} 
or non-asymptotic 
%\cite{rebollo2009t, du2012privacy, basciftci2016privacy, Hsu2019watchdogs} 
privacy-utility-guarantees. 
Following Shannon's information-theoretic notion of secrecy \cite{shannon1949communication}, where security is measured through the \textit{equivocation rate} at the eavesdropper\footnote{A secret listener (wiretapper) to private conversations.}, 
Reed \cite{reed1973information} and Yamamoto \cite{yamamoto1983source} treated security and privacy from a lossy source coding standpoint, i.e., from the rate-distortion theory. 
Inspired by \cite{yamamoto1983source}, in the most general form, the IT privacy framework is based on the presence of a specific `private' variable (or attribute, information) and a correlated non-private variable, and the knowledge of exact joint distribution or partial statistical knowledge of private and/or non-private variables. 
In this setup, the goal is to design a privacy assuring mapping that transforms the pair of these variables into a new representation that achieves a specific application-based target utility, while simultaneously minimizing the information inferred about the private variable. Although IT privacy approaches provide \textit{context-aware} notion of privacy, which can explicitly model the capability of data users and adversaries, it requires statistical knowledge of data, i.e., priors.

%
%
%
% privacy-assuring mappings with provable information-theoretic security guarantees.
% that transforms X into a new data Y that achieves a certain target utility, while minimizing the information revealed about S.
%
%
%
%
%Another related problem to us is secure lossy source coding with side information at the decoder. 
%The problem of source coding with side information at the decoder introduced by Slepian-Wolf \cite{slepian1973noiseless} and Wyner-Ziv \cite{wyner1976rate}. 
%Shannon \cite{shannon1949communication} introduced the information-theoretic notion of secrecy, where security is measured through the \textit{equivocation rate} at the eavesdropper. ...

%---------------------------------------------------
%			 Privacy-Preserving Adversarial Model
%---------------------------------------------------
%
%\subsection{Generative Adversarial Privacy Model}
%\subsection{Privacy-Preserving Adversarial Model}
%\paragraph{Generative Adversarial Privacy Models}

Inspired by GANs \cite{goodfellow2014generative}, the recent data-driven privacy mechanisms address the privacy-utility trade-off as a constrained mini-max game between two players: 
a \textit{defender (privatizer)} that encodes the dataset to minimize the inference leakage on the individual's private/sensitive variables, and an \textit{adversary} that tries to infer the `private' variables from the released dataset \cite{edwards2016censoring, hamm2017enhancing, huang2017context, tripathy2019privacy, huang2018generative}. The adversarial training algorithms proposed in \cite{edwards2016censoring, hamm2017enhancing}, considered deterministic approaches for optimizing privacy-preserving mechanisms. 
The authors in \cite{huang2017context} and \cite{tripathy2019privacy}, independently and simultaneously, addressed randomness in their training algorithms. Our model subsumes these models. 
We establish a more precise connection with the state-of-the-art after introducing the CLUB model. 

%%  recent trends : Information Bottleneck and GAN inspired models

%Recent research trends adopted the Information Bottleneck (IB) problem \cite{tishby2000information} and generative adversarial networks (GANs) \cite{goodfellow2014generative} to address information-theoretic trade-offs and address potentially new data-driven frameworks for privacy-assuring data release mechanisms. 

% Behrooz: to revise this senetence
%
Our model is inspired by \cite{li2017extended}, where the authors established the rate region of the extended Gray-Wyner system for two discrete memoryless sources, which include Wyner's common information, G\'{a}cs-K\"{o}rner common information, IB, K\"{o}rner graph entropy, necessary conditional entropy, and the PF, as extreme points. 
We extend and unify most of the previously proposed objectives in the state-of-the-art literature based on IT privacy models. We hope that our unified model can serve as a reference for researchers and practitioners interested in designing privacy-sensitive data release mechanisms. 
Our research is also closely related to \cite{bertran2019adversarially, sreekumar2019optimal, razeghi2020perfectobfuscation, atashin2021variational}. 
Considering the Markov chain $ \left( \mathbf{U}, \mathbf{S} \right) \markov \mathbf{X} \markov \mathbf{Z}$, the authors in \cite{bertran2019adversarially}  addressed the problem of privacy-preserving representation learning in a scenario where the goal is to share a sanitized representation $\mathbf{Z}$ of high-dimensional data $\mathbf{X}$ while preserving information about utility attribute $\mathbf{U}$ and obfuscate information about private (sensitive) attribute $\mathbf{S}$. 
Inspired by GANs, the framework is formulated as a distribution matching problem. Compared with \cite{bertran2019adversarially}, our formulation is more general as it addresses a key missing component in their formulation, i.e., the rate (description length, information complexity) constraint. 
Another fundamental related work to ours is \cite{sreekumar2019optimal}, which studied the rate-constrained privacy-utility trade-off problem and considered a similar model as \cite{bertran2019adversarially}, independently. The proposed framework is restricted to discrete alphabets and studied the necessary and sufficient conditions for the existence of positive utility, i.e., $\I (U; Z) > 0$, under a  perfect obfuscation regime, i.e., $\I (S; Z) = 0$. 
Analogous to \cite{sreekumar2019optimal}, considering the Markov chain $ \left( U, S \right) \markov X \markov Z$, the authors in \cite{razeghi2020perfectobfuscation} adopted a local information geometry analysis to construct the modal decomposition of the joint distributions, divergence transfer matrices, and mutual information. Next, they obtained the locally sufficient statistics for inferences about the utility attribute, while satisfying perfect obfuscation constraint. 
Furthermore, they developed the notion of perfect obfuscation based on $\chi^2$-divergence and Kullback–Leibler divergence in the Euclidean information space. 
% they characterized the necessary and sufficient conditions under which a non-trivial solution is feasible. 
Considering the Markov chain $ \left( \mathbf{U}, \mathbf{S} \right) \markov \mathbf{X} \markov \mathbf{Z}$, the role of information complexity in privacy leakage about an attribute of an adversary's interest is studied in \cite{atashin2021variational}. In contrast to the PF and generative adversarial privacy models, they considered the setup in which the adversary's interest is \textit{not known a priori} to the data owner. 
More detailed connections between the CLUB model and the state-of-the-art bottleneck models, generative models, modern data compression models, and fair machine learning models is presented in Sec.~\ref{Sec:ConnectionsSOTA}.

%---------------------------------------------------
%
%			 Outline
%
%---------------------------------------------------
\subsection{Outline}
\label{ssec:Outline}

%Before proposing our problem formulation, 

In Sec.~\ref{Sec:Preliminaries}, we present the general problem statement of the CLUB model, and next we briefly review and introduce a few preliminary concepts that are necessary to understand the problem formulation. 
%
% we present the main concepts necessary to understand the fundamentals of our model. In particular, we concretely express our inference threat model and explain why we consider mutual information as the obfuscation and utility measures. 
% Next, we briefly address the concept of relevant information and minimal sufficient statistics. 
%
We then present the CLUB model in Sec.~\ref{Sec:CLUB_model}. 
%
%and discuss its connection to a couple of state-of-the-art bottleneck models in Sec.~\ref{Sec:CLUB_model}. 
% Due to lack of space we addressed a more detailed descriptions in the Appendices. 
The variational bounds of information measures are derived in Sec.~\ref{Sec:VariationalCLUB}. 
% The DVCLUB model as well as the connections between the CLUB model and the state-of-the-art generative models is presented in Sec.~\ref{Sec:DVCLUB}. 
The DVCLUB model is presented in Sec.~\ref{Sec:DVCLUB}. 
%is addressed in Sec.~\ref{Sec:VariationalCLUB}. 
More detailed connections between the CLUB model and the literature is presented in Sec.~\ref{Sec:ConnectionsSOTA}. Experimental results are provided in Sec.~\ref{Sec:Experiments}. Finally, conclusions are drawn in Sec.~\ref{Sec:Conclusion}.

%---------------------------------------------------
%
%			 Notations
%
%---------------------------------------------------
\subsection{Notations}

%Throughout this paper, we use $\mathbb{E}\left[.\right]$ to denote the expectation operator; superscript $(.)^t$ stands for the transpose.
Throughout this paper, 
random variables are denoted by capital letters (e.g., $X$, $Y$), deterministic values are denoted by small letters (e.g., $x$, $y$), random vectors are denoted by capital bold letter (e.g., $\mathbf{X}$, $\mathbf{Y}$), deterministic vectors are denoted by small bold letters (e.g., $\mathbf{x}$, $\mathbf{y}$), alphabets (sets) are denoted by calligraphic fonts (e.g., $\mathcal{X, Y}$), and for specific quantities/values we use sans serif font (e.g., $\mathsf{x}$, $\mathsf{y}$, $\mathsf{C}$, $\mathsf{D}$, $\mathsf{P}$, $\Omega)$. 
Superscript $(.)^T$ stands for the transpose. % and $(.)^\dagger$ stands for pseudo-inverse.  $x(i)$ denotes the $i$-th entry of deterministic vector $\mathbf{x}$. 
%$\mathbf{x}{\left(  j \right)}$ denotes the $j$-th column of $\mathbf{X}$. 
% and $\mathbf{x}{\left(  i,: \right)}$ denotes the $i$-th row of $\mathbf{X}$. 
Also, we use the notation $\left[ N \right]$ for the set $\{ 1, 2, 3, ..., N\}$. % and $\mathrm{card} \left(\mathcal{S}\right)$ for the cardinality of a set $\mathcal{S}$. 
%Moreover, we write $\mathcal{\bar{S}}$ for the complement $\left[ N \right] \backslash \mathcal{S}$ of a set $\mathcal{S}$ in $\left[ N \right]$. 
%
$\H \left( P_{\mathbf{X}} \right) \coloneqq \mathbb{E}_{P_{\mathbf{X}}} \left[ - \log P_{\mathbf{X}} \right]$ denotes the Shannon entropy; $\H \left( P_{\mathbf{X}} \Vert Q_{\mathbf{X}} \right) \coloneqq \mathbb{E}_{P_{\mathbf{X}}} \left[ - \log Q_{\mathbf{X}} \right]$ denotes the cross-entropy of the distribution $P_{\mathbf{X}}$ relative to a distribution $Q_{\mathbf{X}}$; and $\H \left( P_{\mathbf{Z} \mid \mathbf{X}} \Vert Q_{\mathbf{Z} \mid \mathbf{X}} \mid P_{\mathbf{X}}  \right) \coloneqq \mathbb{E}_{P_{\mathbf{X}}}  \mathbb{E}_{P_{\mathbf{Z} \mid \mathbf{X} }} \left[ - \log Q_{\mathbf{Z} \mid \mathbf{X}} \right]$ denotes the cross-entropy loss for $Q_{\mathbf{Z} \mid \mathbf{X}}$. The relative entropy is defined as $\D_{\mathrm{KL}} \left( P_{\mathbf{X}} \Vert Q_{\mathbf{X}} \right)  \coloneqq \mathbb{E}_{P_{\mathbf{X}}} \big[ \log \frac{P_{\mathbf{X}}}{Q_{\mathbf{X}}} \big] $. 
%\begin{eqnarray}
%\D_{\mathrm{KL}} \left( P_{\mathbf{X}} \Vert Q_{\mathbf{X}} \right)  \coloneqq \left\{ 
%\begin{array}{ll}
%\mathbb{E}_{P_{\mathbf{X}}} \left[ \log \frac{P_{\mathbf{X}}}{Q_{\mathbf{X}}} \right]  & \mathrm{if} \;  P_{\mathbf{X}} \ll Q_{\mathbf{X}}
%\\
%+\infty &  \mathrm{otherwise},
%\end{array}
%\right. 
%\end{eqnarray}
%where the notation $P_{\mathbf{X}} \ll Q_{\mathbf{X}}$ is used to denote that the probability measure $P_{\mathbf{X}} $ is absolutely continuous with respect to $Q_{\mathbf{X}}$. 
The conditional relative entropy is defined by $\D_{\mathrm{KL}} \left( P_{\mathbf{Z} \mid \mathbf{X}} \Vert  Q_{\mathbf{Z} \mid \mathbf{X}}  \mid P_{\mathbf{X}} \right) \coloneqq  \mathbb{E}_{P_{\mathbf{X}}} \left[ \D_{\mathrm{KL}} \left( P_{\mathbf{Z} \mid \mathbf{X}= \mathbf{x}} \Vert  Q_{\mathbf{Z} \mid \mathbf{X} = \mathbf{x}} \right)  \right]$ and the mutual information is defined by $\I \left( P_{\mathbf{X}}; P_{\mathbf{Z} \mid \mathbf{X}} \right)  \coloneqq \D_{\mathrm{KL}} \left( P_{\mathbf{Z} \mid \mathbf{X}} \Vert  P_{\mathbf{Z} }  \mid P_{\mathbf{X}} \right)$. We abuse notation to write $\H \left( \mathbf{X} \right) = \H \left( P_{\mathbf{X}} \right)$ and $\I \left( \mathbf{X}; \mathbf{Z} \right) = \I \left( P_{\mathbf{X}}; P_{\mathbf{Z} \mid \mathbf{X}} \right)$ for random objects\footnote{The name random object includes random variables, random vectors and random processes.} $\mathbf{X} \sim P_{\mathbf{X}}$ and $\mathbf{Z} \sim P_{\mathbf{Z}}$. We use the same notation for the probability distributions and the associated densities. 
%
%For discrete random variable $X$, let consider a finite support set $\mathcal{X} \triangleq \{1, ..., \vert \mathcal{X} \vert \}$ with $2 \leq \vert \mathcal{X} \vert < + \infty$. 
%We denote by $\mathcal{P}\left( \mathcal{X} \right)$ the set of all possible probability distributions of a random variable $X$ with range $\mathcal{X}$. 
%We denote by $\mathbf{p}_X$ the probability mass function (pmf) vector with $i$-th entry equal to $p_X(i)$. 

%We review the fundamentals of these research in %the follow.
%Sec.~\ref{Sec:Preliminaries}.  

%---------------------------------------------------
%			 Preliminaries
%---------------------------------------------------
\section{Problem Statement and Preliminaries}
\label{Sec:Preliminaries}
 
% Before we proceed with problem formulation, we briefly review a few concepts. 
% and present the two main research trends which are necessary to understand our new concept. 

In this section, first we present the general problem statement of the CLUB model, and next we introduce the main concepts necessary to understand the fundamentals of our model. In particular, we concretely express our inference threat model and explain why we consider mutual information as the obfuscation and utility measures. 
Next, we briefly address the concepts of relevant information and minimal sufficient statistics.

%first we present the general problem statement of the CLUB model, and next we briefly review and introduce a few preliminary concepts in Sec.~\ref{Ssec:ObfuscationUtilityUnderLogLoss} and Sec.~\ref{Ssec:Relevant Information} which are necessary to understand the origins of problem formulation. 

% In Sec.~\ref{Sec:Preliminaries}, we present the main concepts necessary to understand the fundamentals of our model. 
% In particular, we concretely express our inference threat model and explain why we consider mutual information as the obfuscation and utility measures. 
% Next, we briefly address the concept of relevant information and minimal sufficient statistics. 

%---------------------------------------------------
%
%		     General Problem Statement
%
%---------------------------------------------------
\subsection{General Problem Statement}
\label{Ssec:GeneralProblemStatement}

Consider the problem of releasing a sanitized representation $\mathbf{Z}$ of observed data $\mathbf{X}$, through a stochastic mapping $P_{\mathbf{Z} \mid \mathbf{X}}$, while preserving information about a utility attribute $\mathbf{U}$ and obfuscating a sensitive attribute $\mathbf{S}$. In this scenario, $\left( \mathbf{U}, \mathbf{S}\right) \markov \mathbf{X} \markov \mathbf{Z}$ forms a Markov chain. 
In general, one can consider a well-defined generic obfuscation measure as a functional of the joint distribution $P_{\mathbf{S}, \mathbf{Z}}$ that captures the amount of leakage about $\mathbf{S}$ by releasing $\mathbf{Z}$. Let $\mathcal{C}_{\mathsf{L}}  : \mathcal{P}\left( \mathcal{S} \times  \mathcal{Z}\right) \rightarrow \mathbb{R}^+ \cup \{ 0\}$ denote this generic leakage (obfuscation) measure. 
On the other hand, one can consider a well-defined and application-specific generic utility measure as a functional of the joint distribution $P_{\mathbf{U}, \mathbf{Z}}$ that captures the amount of utility about $\mathbf{U}$ by releasing $\mathbf{Z}$, instead of original data $\mathbf{X}$. Let $\mathcal{C}_{\mathsf{U}}: \mathcal{P}\left( \mathcal{U} \times \mathcal{Z}\right) \rightarrow \mathbb{R}^+ \cup \{ 0\}$ denote a generic utility \textit{loss} measure. 
Finally, let $\mathcal{C}_{\mathsf{C}} : \mathcal{P}\left( \mathcal{X} \times  \mathcal{Z}\right) \rightarrow \mathbb{R}^+ \cup \{ 0\}$ denote a generic measure of \textit{complexity} of the distribution $P_{\mathbf{Z} \mid \mathbf{X}}$ for the data distribution $P_{\mathbf{X}}$. 
Then, one can consider the generic CLUB functional as:
\begin{equation}\label{Eq:CUI_functional}
     \mathsf{CLUB} \left(  P_{\mathbf{U}, \mathbf{S}, \mathbf{X}}   \right)   \coloneqq  \mathop{\inf}_{\substack{ P_{\mathbf{Z} \mid \mathbf{X}}: \\\left( \mathbf{U}, \mathbf{S}\right) \markov \mathbf{X} \markov \mathbf{Z}}}
    \mathcal{C}_{\mathsf{C}}    \left( P_{\mathbf{X}},  P_{  \mathbf{Z} \mid \mathbf{X} }\right)  + 
     \mathcal{C}_{\mathsf{L}}   \left( P_{\mathbf{S}},  P_{ \mathbf{Z}\mid \mathbf{X} } \right) +
    \mathcal{C}_{\mathsf{U}}   \left( P_{\mathbf{U}},  P_{ \mathbf{Z} \mid \mathbf{X} }\right) .
\end{equation}

%---------------------------------------------------
%  Figure: Complexity-Leakage-Utility Bottleneck General setup
%---------------------------------------------------
%
%---------------------------------------------------
%---------------------------------------------------

\vspace{-3pt}

%---------------------------------------------------
%
%		 Obfuscation and Utility Measures under Logarithm Loss
%
%---------------------------------------------------
\subsection{Obfuscation and Utility Measures under Logarithmic Loss}
\label{Ssec:ObfuscationUtilityUnderLogLoss}
%
%%  The aim of this section is get insight to reader why MI measure? and where this measure is coming from

We consider obfuscation-utility trade-off model where both utility and obfuscation are measured under \textit{logarithmic loss} (also often referred to as the self-information loss). 
The logarithmic loss function has been widely used in learning theory \cite{cesa2006prediction}, image processing \cite{andre2006entropy}, IB \cite{harremoes2007information}, multi-terminal source coding \cite{courtade2011multiterminal}, as well as PF \cite{makhdoumi2014information}. 
In this case, both the obfuscation and utility measures can be modelled by the mutual information. 
By minimizing the obfuscation measure under the logarithmic loss, one actually minimizes an upper bound on any bounded loss function \cite{makhdoumi2014information}. 
%Consider the problem of releasing a sanitized representation $\mathbf{Z}$ of observed data $\mathbf{X}$, through a stochastic mapping $P_{\mathbf{Z} \mid \mathbf{X}}$, while preserving information about utility attribute $\mathbf{U}$ and obfuscate information about sensitive attribute $\mathbf{S}$. In this scenario, $\left( \mathbf{U}, \mathbf{S}\right) \markov \mathbf{X} \markov \mathbf{Z}$ forms a Markov chain. 
%In general, one can consider a well-defined generic obfuscation measure as a functional of the joint distribution $P_{\mathbf{S}, \mathbf{Z}}$ that captures the amount of leakage about $\mathbf{S}$ by releasing $\mathbf{Z}$. On the other hand, one can consider a well-defined and application-specific generic utility measure as a functional of the joint distribution $P_{\mathbf{U}, \mathbf{Z}}$ that captures the amount of utility about $\mathbf{S}$ by releasing $\mathbf{Z}$, instead of original data $\mathbf{X}$. 

%property of logarithmic loss distortion measure is that the expected distortion is lower-bounded by a conditional entropy. 
% 

Consider the \textit{inference threat model} introduced in \cite{du2012privacy}, which models a broad class of adversaries that perform statistical inference attacks on the sensitive data. Consider an inference cost function $\mathsf{C}: \mathcal{S} \times \mathcal{P}\left( \mathcal{S}\right) \rightarrow \mathbb{R}^+ \cup \{ 0 \}$. Prior to observing $\mathbf{Z}$, the adversary chooses a belief distribution $Q$ from the set $\mathcal{P} \! \left( \mathcal{S} \right)$ of all possible distributions over $\mathcal{S}$, that minimizes the expected inference cost function $\mathsf{C}\left( \mathbf{S} , Q \right)$. Therefore:\vspace{-2pt}
%$Q^{\ast} = \arg \mathop{\min}_{Q \in \mathcal{P}\left( \mathcal{S} \right)} \mathbb{E}_{P_{\mathbf{S}}} \left[ C \! \left( \mathbf{S} , Q \right) \right]$.
\begin{equation}
Q^{\ast} = \arg \mathop{\min}_{Q \in \mathcal{P}\left( \mathcal{S} \right)} \mathbb{E}_{P_{\mathbf{S}}} \left[ \mathsf{C}   \left( \mathbf{S} , Q \right) \right]. 
\end{equation}
Let $c_0^{\ast} = \mathbb{E}_{P_\mathbf{S}} \left[ \mathsf{C}  \left( \mathbf{S} , Q^{\ast}\right) \right]$ denote the corresponding minimum average cost.
% Let $c_0^{\ast} =  \mathop{\min}_{Q \in \mathcal{P}\left( \mathcal{S} \right)} \mathbb{E}_{P_\mathbf{S}} \left[ \mathsf{C}  \left( \mathbf{S} , Q\right) \right]$ denote the corresponding minimum average cost.
%of inferring $\mathbf{S}$ prior to observing $\mathbf{Z}$. 
After observing $\mathbf{Z}=\mathbf{z}$, the adversary revises his belief distribution as:\vspace{-2pt}
%$Q_{\mathbf{z}}^{\ast} = \arg \mathop{\min}_{Q  \in \mathcal{P}\left( \mathcal{S} \right)} \mathbb{E}_{ P_{\mathbf{S} \mid \mathbf{Z}}} \left[ C \! \left( \mathbf{S} , Q \right)  \mid \mathbf{Z}= \mathbf{z}\right]$. 
\begin{equation}
Q_{\mathbf{z}}^{\ast} = \arg \mathop{\min}_{Q  \in \mathcal{P}\left( \mathcal{S} \right)} \mathbb{E}_{ P_{\mathbf{S} \mid \mathbf{Z}}} \left[ \mathsf{C}  \left( \mathbf{S} , Q \right)  \mid \mathbf{Z}= \mathbf{z}\right]. 
\end{equation}
% Let $c_{\mathbf{z}}^{\ast} =  \mathop{\min}_{Q \in \mathcal{P}\left( \mathcal{S} \right)} \mathbb{E}_{P_{\mathbf{S\mid Z}} } \left[ \mathsf{C} \left( \mathbf{S} , Q\right) \mid \mathbf{Z}= \mathbf{z} \right]$ 
% denote the corresponding minimum average cost of inferring $\mathbf{S}$ after observing $\mathbf{Z=z}$. 
Let $c_{\mathbf{z}}^{\ast} = \mathbb{E}_{P_{\mathbf{S\mid Z}} } \left[ \mathsf{C} \left( \mathbf{S} , Q_{\mathbf{z}}^{\ast}\right) \right]$ denote the corresponding minimum average cost of inferring $\mathbf{S}$ after observing $\mathbf{Z=z}$. 
Therefore, the adversary obtains an average gain of $\Delta \mathsf{C} = c_0^{\ast} - \mathbb{E}_{P_{\mathbf{Z}}} \left[ c_{\mathbf{Z}}^{\ast} \right]$ in inference cost. 
This cost gain measures the improvement in the quality of the inference of sensitive data $\mathbf{S}$ due to observation of $\mathbf{Z}$. Under the self-information loss cost function $\mathsf{C} \left( \mathbf{s} , Q\right) = - \log Q (\mathbf{s}), \forall \mathbf{s} \in \mathcal{S}$, the information leakage $\Delta \mathsf{C}$ can be measured by the Shannon mutual information $\I \left( \mathbf{S}; \mathbf{Z} \right)$. 

On the other hand, the stochastic mapping $P_{\mathbf{Z} \mid \mathbf{X}}$ should maintain the utility of desired data $\mathbf{U}$. 
%Consider a utility (accuracy) measure $d : \mathcal{U} \times \mathcal{Z}  \rightarrow \mathbb{R}^+$. 
Under the self-information loss, the utility measure is defined as 
$d\left( \mathbf{u} , \mathbf{z} \right) = - \log P_{\mathbf{U} \mid \mathbf{Z}} \left( \mathbf{u} \mid \mathbf{z}\right)$, 
%$\mathcal{C}_{\mathsf{U}} \! \left( P_{\mathbf{U}},  P_{ \! \mathbf{Z} \mid \mathbf{X}} \right) = - \log P_{\mathbf{U} \mid \mathbf{Z}} \left( \mathbf{u} \mid \mathbf{z}\right)$, 
which is a function of $\mathbf{u}$ and $\mathbf{z}$ as well as stochastic mapping $P_{\mathbf{Z} \mid \mathbf{X}}$. Hence, the average utility loss is $\mathbb{E}_{P_{\mathbf{U}, \mathbf{Z}}} \left[ - \log P_{\mathbf{U} \mid \mathbf{Z}} \right] = \H \left( \mathbf{U} \mid \mathbf{Z} \right)$ that can be minimized by designing the stochastic mapping $P_{\mathbf{Z} \mid \mathbf{X}}$. Consider some utility level $E^{\mathrm{u}}\geq 0$, such that we constrain to have $\H \left( \mathbf{U} \mid \mathbf{Z} \right) \leq E^{\mathrm{u}}$. Given $P_{\mathbf{U}}$, and therefore $\H \left( \mathbf{U} \right)$, and assuming that $R^{\mathrm{u}} = \H ( \mathbf{U} ) - E^{\mathrm{u}} \geq 0$, the utility constraint can be recast as $\I \left( \mathbf{U} ; \mathbf{Z} \right) \geq R^{\mathrm{u}}$. 

In Sec.~\ref{Sec:CLUB_model}, built upon this general threat model and considering the self-information loss as both utility and obfuscation measures, we present a unified formulation for the complexity-leakage-utility trade-off that generalizes most of the state-of-the-art literature.

\vspace{-4pt}

%---------------------------------------------------
%
%		 Relevant Information
%
%---------------------------------------------------
\subsection{Relevant Information}
\label{Ssec:Relevant Information}
%
%%  The aim of this section is get insight to reader about relevant information? 
%%  Here my main target is 'ML community'. Since I saw many articles from highly cited people, that they didn't have have a clear vision about Sufficient Statistic

The relevant information is a common concept in information theory and statistics which captures the information that a random object $\mathbf{X}$ contains about another random object $\mathbf{U}$. 
Below, we present statistical and information-theoretical formulations proposed for measuring relevant information. 

\begin{definition}[Sufficient Statistic]
Let $\mathbf{U} \in \mathcal{U}$ be an unknown parameter and $\mathbf{X} \in \mathcal{X}$ be a random variable with conditional probability distribution function $P_{\mathbf{X} \mid \mathbf{U}}$. Given a function $f : \mathcal{X} \rightarrow \mathcal{Z}$, the random variable $\mathbf{Z} = f(\mathbf{X})$ is called a sufficient statistics for $\mathbf{U}$ if and  only if (iff) \cite{cover1999elements, polyanskiy2014lecture}:\vspace{-4pt}
\begin{equation}
P_{\mathbf{X} \mid {\mathbf{U},\mathbf{Z}}} \, (\mathbf{x} \mid \mathbf{u}, \mathbf{z}) = P_{\mathbf{X} \mid {\mathbf{Z}}} \, (\mathbf{x} \mid \mathbf{z}) , \;\; \forall (\mathbf{x}, \mathbf{u}) \in \mathcal{X} \times \mathcal{U}. 
\end{equation}
Equivalently, quantifying with mutual information, we have:\vspace{-4pt}
\begin{equation}
 \I \left( \mathbf{U}; \mathbf{Z} \right) = \I \left( \mathbf{U}; \mathbf{X} \right).\vspace{-3pt}
\end{equation}
\end{definition}

Note that $\mathbf{U} \markov \mathbf{X} \markov f \!\left( \mathbf{X}\right)$ forms a Markov chain, and by the data processing inequity (DPI), for a general statistic $f \! \left( \mathbf{X} \right)$, we have $\I \left( \mathbf{U};  f \! \left( \mathbf{X}\right) \right) \leq \I \left( f \! \left( \mathbf{X}\right); \mathbf{X} \right)$. If equality holds, a sufficient statistic $\mathbf{Z}$ captures all the information in $\mathbf{X}$ about $\mathbf{U}$. 
%once we know the statistic $Z=z$, we cannot obtain any additional information about $Y=y$ from knowing the observed $X$. 

\vspace{-3pt}

\begin{remark}[Measure-Theoretic Interpretation]
Consider a probability space $\left( \Omega, \mathcal{B}, \mathsf{P} \right)$, where $\mathcal{B}$ denotes the $\sigma$-algebra of all subsets of $\Omega$, and $\mathsf{P}$ is a probability measure defined on the measurable space $\left( \Omega, \mathcal{B} \right)$. 
Given two measurable spaces $\left( \Omega, \mathcal{B} \right)$ and $\left( \mathcal{X}, \mathcal{F}\right)$, a random object (or measurable space) is a mapping (or function) $X: \Omega \rightarrow \mathcal{X}$, i.e., a measurable function defined on $\left( \Omega, \mathcal{B} \right)$ and taking values in $\left( \mathcal{X}, \mathcal{F} \right)$, with the property that if 
$\mathpzc{E} \in \mathcal{F}$, then $X^{-1} \left( \mathpzc{E} \right) = \{ \omega: X(\omega) \in \mathpzc{E} \} \in \mathcal{B}$. In this case, this random object is also called $\mathcal{F}$-measurable. 
Let $\mathcal{Z}$ be a proper subset of $\mathcal{X}$. A sufficient statistic is defined to be a measurable mapping $f$ from the measurable space $\left( \mathcal{X}, \mathcal{F}\right)$ onto a measurable space $\left( \mathcal{Z}, \mathcal{A}_f\right)$, i.e., $f: \left( \mathcal{X}, \mathcal{F}\right) \rightarrow \left( \mathcal{Z}, \mathcal{A}_f\right)$. 
The classical problem of data reduction is to find a partition $\mathcal{A}_f$ of $\mathcal{X}$ determined by some measurable mapping $f$. In other words, the problem is to find a new random object $Z\left( \omega \right) = f \left( X \left( \omega \right)\right)$. 
\end{remark}

\vspace{-3pt}

\begin{remark}[Simple Interpretation]
A statistic $\mathbf{Z} = f \!\left( \mathbf{X}\right)$ induces a partition on sample space $\mathcal{X}$. For each $\mathbf{Z} = \mathbf{z}$, the sample set $\{ \mathbf{x}: f(\mathbf{x}) = \mathbf{z} \}$ is one element of the partition, i.e., $\mathbf{z} \in \mathcal{Z}= \{  \mathbf{z} : \mathbf{z} = f\! \left( \mathbf{x} \right) \; \mathrm{for~some}\; \mathbf{x} \in \mathcal{X}  \}$. 
The statistic $f \!\left( \mathbf{X}\right)$ is sufficient iff the assigned samples in each partition do not depend on $\mathbf{U}$.  
Maximum data reduction is achieved when $\vert \mathcal{Z}\vert$ is minimal.
%Given an induced partition by $g \left( \mathbf{X} \right)$, if 
\end{remark}

\vspace{-3pt}

\begin{definition}[Minimal Sufficient Statistic]\label{Def:MinimalSufficientStatistic}
A sufficient statistics $\mathbf{Z}$ is said to be minimal if it is a function of all other sufficient statistics, that is, for all sufficient statistics $\mathbf{Z}^{\prime}$, there exists $f$ such that $\mathbf{Z} = f \! \left( \mathbf{Z}^{\prime} \right)$. 
\end{definition}

\vspace{-3pt}

It means that $\mathbf{Z} $ induces the coarsest sufficient partition. In other words, $\mathbf{Z}$ achieves the maximum data reduction, while assuring $\I \left( \mathbf{U};  \mathbf{Z}  \right) = \I \left( \mathbf{U}; \mathbf{X} \right)$. 

Suppose that nature chooses a parameter $\mathbf{U}$ at random, after which the sample $\mathbf{X}$ is drawn from the distribution $P_{\mathbf{X} \mid \mathbf{U}}$. One can show that the statistic $\mathbf{Z}$ is a minimal sufficient statistic for $\mathbf{U}$, iff it is a solution of one of the two equivalent optimization problems:\vspace{-5pt}
\begin{equation}
\mathbf{Z} =  \arg \min_{\substack{\mathbf{Z}^{\prime}:~\mathrm{sufficient~statistic}\\\mathrm{for~}\mathbf{U}}} \I \left( \mathbf{X}; \mathbf{Z}^{\prime} \right) \; \equiv  \;
 \mathop{\arg \min}_{\mathbf{Z}^{\prime}:~\I \left(\mathbf{U}; \mathbf{Z}^{\prime} \right) = \I \left( \mathbf{U}; \mathbf{X}  \right)} \I \left( \mathbf{X}; \mathbf{Z}^{\prime} \right). 
\end{equation}
It means that minimal sufficient statistic $\mathbf{Z}$ is the best compression of $\mathbf{X}$, with the zero information loss about parameter $\mathbf{U}$.

\vspace{-3pt}

\vspace{-3pt}

%\section{Complexity-Leakage-Utility Bottleneck Model}
\section{CLUB Model}
\label{Sec:CLUB_model}

\vspace{-3pt}

%---------------------------------------------------
%
%			  General Setup
%
%---------------------------------------------------
\subsection{General Setup}
\label{Ssec:GeneralSetup}

\begin{figure}[!t]
\centering
\includegraphics[scale=0.71]{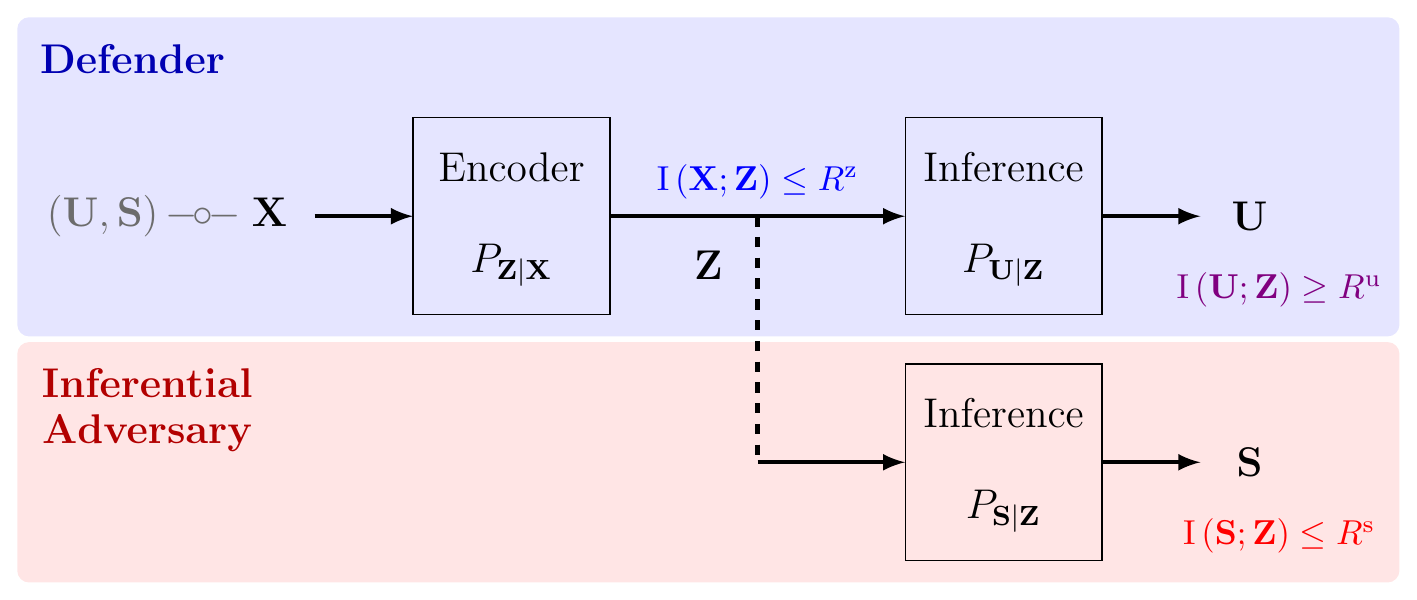}
\caption{The CLUB model.}
\label{Fig:CLUB_diagram}
\end{figure}

The CLUB model (Fig~\ref{Fig:CLUB_diagram}) is a generalization of the sufficient statistic methods that formulate the problem of extracting the relevant information from a random object (data) $\mathbf{X}$ about the random object $\mathbf{U}$ that is of interest, while limiting statistical inference about a sensitive random object $\mathbf{S}$ that depends on $\mathbf{X}$ and is possibly depended on $\mathbf{U}$. 
Consider a scenario in which $P_{\mathbf{U}, \mathbf{S}, \mathbf{X}}$ is fixed and known by both the defender and the adversary, where $\mathbf{X}  \in  \mathcal{X}$ represents the data observed by the defender (e.g., high dimensional facial image), $\mathbf{U} \in \mathcal{U}$ denotes the utility attribute of our interest for a utility service provider (e.g., person identity), and $\mathbf{S} \in \mathcal{S}$ denotes the sensitive attribute (e.g., gender) that we wish to restrict its statistical inference. 
Intuitively, built upon our introduction of the concept of relevant information above, we intend to find a stochastic mapping $P_{\mathbf{Z}\mid \mathbf{X}}$ such that the posterior distribution of the utility attribute $\mathbf{U}$ is similar given the released representation $\mathbf{Z}$ and the original data $\mathbf{X}$, i.e., $P_{\mathbf{U} \mid \mathbf{Z}} \approx P_{\mathbf{U} \mid \mathbf{X}}$, while the posterior of private attribute $\mathbf{S}$ given released representation $\mathbf{Z}$ is as close as possible to its prior, i.e., $P_{\mathbf{S} \mid \mathbf{Z}} \approx P_{\mathbf{S}}$. 
In the rest of the paper, we assume our attributes are supported on a finite space.

%Given the observed data $\mathbf{X}$ the defender (data owner) wishes to release a representation $\mathbf{Z}$ for a utility task $\mathbf{U}$ (e.g. identity verification) while keeps another attribute $\mathbf{S}$ (e.g. emotion) as sensitive. 
%
%In this section, 
%, and Markov chain $\left( \mathbf{U}, \mathbf{S}\right) \markov \mathbf{X} \markov \mathbf{Z}$ holds. 
%Moreover, we consider the non-interactive, one-shot regime, where data owner discloses the representation $\mathbf{Z}$ once, and no additional information is released.  
%Based on that, we concretely formulate our CLUB model and express its connection with the SOTA information-theoretic privacy literature. 

%\pagebreak
%---------------------------------------------------
%
%			  Threat Model
%
%---------------------------------------------------
\subsection{Threat Model}
\label{Ssec:ThreatModel}

We consider the inference threat model described in Sec.~\ref{Ssec:ObfuscationUtilityUnderLogLoss}. 
In particular, we have the following assumptions:
\begin{itemize}[leftmargin=1em]
\item
%We consider two game scenarios associated with the sensitive attribute $\mathbf{S}$. 
%In the \textit{pre-defined} game scenario, 
We assume the adversary is interested in an attribute $\mathbf{S}$ of data $\mathbf{X}$. The attribute $\mathbf{S}$ can be any (possibly randomized) function of $\mathbf{X}$. 
%In the \textit{undefined} game scenario, 
%we assume the adversary is interested in an attribute $\mathbf{S}$ of data $\mathbf{X}$ which is \textit{not known a priori} to the defender\footnote{In other words, the distribution $P_{\mathbf{S} \mid \mathbf{X}}$ is unknown to the defender.}.
We restrict attribute $\mathbf{S}$ to be discrete, which captures the most scenarios of interest, e.g., a facial attribute, an identity, etc.  
\item
The adversary observes released representation $\mathbf{Z}$ and the Markov chain $\mathbf{S} \markov \mathbf{X} \markov \mathbf{Z}$ holds. 
\item
We assume that the adversary knows the \textit{complexity-constrained obfuscation-utility-assuring mapping} $P_{\mathbf{Z} \mid \mathbf{X}}$ designed by the data owner (defender), i.e., the defence mechanism is public. 
\end{itemize}

%---------------------------------------------------
%	    Figure: Complexity-Leakage-Utility Bottleneck model
%---------------------------------------------------
%
%---------------------------------------------------
%---------------------------------------------------

%---------------------------------------------------
%
%				Problem Formulation
%
%---------------------------------------------------
\subsection{Problem Formulation}
\label{Ssec:ProblemFormulation}

%Let us first consider the \textit{pre-defined} game scenario, where the utility and sensitive (private) attributes are specified \textit{a priori}. 
%In this scenario, 
%Given the observed data $\mathbf{X}$ the defender (designer) wish to release a representation $\mathbf{Z}$ for a utility task (e.g. identity verification) $\mathbf{U}$ while keep another attribute $\mathbf{S}$ (e.g. emotion) as sensitive.   
%
%More concretely, g

Given three dependent (correlated) random variables $\mathbf{U}$, $\mathbf{S}$ and $\mathbf{X}$ with joint distribution $P_{\mathbf{U ,S , X}}\,$, the goal of the CLUB model is to find a representation $\mathbf{Z}$ of $\mathbf{X}$ using a stochastic mapping $P_{\mathbf{Z}\mid \mathbf{X}}$ such that: (i) $\left( \mathbf{U}, \mathbf{S} \right) \markov \mathbf{X} \markov \mathbf{Z}$, and (ii) representation $\mathbf{Z}$ is maximally informative about $\mathbf{U}$ (maximizing $\I \left( \mathbf{U}; \mathbf{Z} \right)$) while being minimally informative about $\mathbf{X}$ (minimizing $\I \left( \mathbf{X}; \mathbf{Z}\right)$) and minimally informative about $\mathbf{S}$ (minimizing $\I \left( \mathbf{S}; \mathbf{Z}\right)$). 
%$\I \left( \mathbf{S}; \mathbf{Z}\right)$ measures average amount of private information leakage. 
%
%It allows a model to smoothly trade-off the maximality of the informativeness of the bottleneck variable ($\mathbf{Z}$) for the utility task at hand ($\mathbf{U}$), against the compressiveness of the bottleneck variable ($\mathbf{Z}$) from data ($\mathbf{X}$). 
%
%Let $f: \left( 0, \infty \right) \rightarrow \mathbb{R}$ be a convex function which $f(1) =0$. The  $f$-information between two random variables $X$ and $Z$ is defined as $I_f \left( X ; Z \right) = D_f \left(  p(x,z) \| p(x) p(z) \right)$, where $D_f \left( \cdot \| \cdot \right)$ is $f$-divergence \cite{polyanskiy2014lecture}. 
%
%
We can formulate this three-dimensional trade-off by imposing constraints on the two of them. That is, for a given information complexity and information leakage
constraints, $R^{\mathrm{z}} \geq 0$ and $R^{\mathrm{s}} \geq 0$, respectively, this trade-off can be formulated by a CLUB functional\footnote{Note that one can generalize this formulation and consider Arimoto's mutual information \cite{arimoto1977information} and/or $f$-information \cite{csiszar1967information} in the CLUB model \eqref{CLUB_functional}. 
%instead of $\I (\mathbf{X}; \mathbf{Z})$, $\I (\mathbf{U}; \mathbf{Z})$ and $\I (\mathbf{S}; \mathbf{Z})$, where, 
For instance, $\I_f \! \left( \mathbf{X}; \mathbf{Z} \right) = \D_f \! \left( P_{\mathbf{X, Z}} \| P_{\mathbf{X}} P_{\mathbf{Z}}\right)$ and $D_f \! \left( \cdot \| \cdot \right)$ is $f$-divergence \cite{polyanskiy2014lecture, duchi2016lecture}. 
For the sake of brevity, we consider Kullback–Leibler divergence, a special case of the family of $f$-divergences between probability distributions associated with $f(t) = t \log t$. 
Moreover, note that although all $\D_f \! \left( \cdot \| \cdot \right)$ quantify the dissimilarity between a pair of distributions, their operational meanings are different. For example, $f(t) = \frac{1}{2} \vert t - 1 \vert$ results in total variation (TV), which is utilized in hypothesis testing and considered as a privacy measure in \cite{rassouli2019optimal}, while $f(t) = (t-1)^2$ (or $t^2 - 1$) results in $\chi^2$-information, which is useful in estimation problems.\label{footnote_generalize_CLUB}}: 
\begin{eqnarray}\label{CLUB_functional}
\mathsf{CLUB} \left( R^{\mathrm{z}}, R^{\mathrm{s}}, P_{\mathbf{U}, \mathbf{S}, \mathbf{X}}\right)  \coloneqq \mathop{\sup}_{\substack{ P_{\mathbf{Z} \mid \mathbf{X}}: \\\left( \mathbf{U}, \mathbf{S}\right) \markov \mathbf{X} \markov \mathbf{Z}}} \I  \left( \mathbf{U}; \mathbf{Z} \right) \; \quad  \; \mathrm{s.t.} \quad   \I \left( \mathbf{X}; \mathbf{Z} \right) \leq R^{\mathrm{z}}, \;\; 
\I \left( \mathbf{S}; \mathbf{Z} \right) \leq R^{\mathrm{s}}. 
\end{eqnarray}

%We refer $\I  \left( \mathbf{U}; \mathbf{Z} \right)$ to as the relevance of $\mathbf{Z}$, $\I \left( \mathbf{X}; \mathbf{Z} \right)$ to as the information complexity of $\mathbf{Z}$, and $\I \left( \mathbf{S}; \mathbf{Z} \right)$ to as the information leakage about $\mathbf{S}$. 
%% say about role of  I (  S ;  Z )
The constraint $\I \left( \mathbf{S}; \mathbf{Z} \right) \leq R^{\mathrm{s}}$ ensures that $\H \left( \mathbf{S} \mid \mathbf{Z} \right) \geq \H \left( \mathbf{S} \right) - R^{\mathrm{s}} = E^{\mathrm{a}}$, where $E^{\mathrm{a}}$ quantifies the amount of uncertainty for adversary. 
On the other hand, note that $\I \left( \mathbf{S}; \mathbf{Z}\right) = \mathbb{E}_{P_{\mathbf{Z}}} \left[  \D_{\mathrm{KL}} \left( P_{\mathbf{S}\mid \mathbf{Z}} \Vert P_{\mathbf{S}} \right) \right]$. 
This means that for small values of $R^{\mathrm{s}}$, the posterior of private attribute $\mathbf{S}$ given released representation $\mathbf{Z}$, i.e., $P_{\mathbf{S} \mid \mathbf{Z}}$, is as close as possible to the prior $P_{\mathbf{S}}$. 
%
% analogous to equivocation rate introduced by Shannon\cite{shannon1949communication}. 
%
%
%
%% say about role of  I (  X ;  Z )
The constraint $\I \left( \mathbf{X}; \mathbf{Z} \right) \leq R^{\mathrm{z}}$ controls \textit{information complexity}\footnote{The notion of `information complexity' inspired by the concept of \textit{stochastic complexity} of the data relative to a model \cite{rissanen1986stochastic,rissanen1987stochastic, rissanen1996fisher, grunwald2005advances, grunwald2004tutorial}, which can be interpreted as the shortest code-length of the data given a model. We will elaborate on this notion in Sec.~\ref{Sec:VariationalCLUB}} (aka \textit{compactness}) of representations and goes beyond a simple regularization term. Indeed, it is related to the notion of \textit{encoder capacity} \cite{vera2018role,vera2018roleISIT}, which is a measure of distinguishability among input data samples from their released representations. Moreover, the information complexity $\I \left( \mathbf{X}; \mathbf{Z} \right) $ establishes a fundamental relation to the generalization capability of the stochastic encoder model $P_{\mathbf{Z} \mid \mathbf{X}}$. The trade-off in \eqref{CLUB_functional}  was studied in \cite{sreekumar2019optimal}, where the constraint on $\I \left(\mathbf{U}; \mathbf{Z}\right)$ is motivated as a rate-constraint, and the trade-off is studied under perfect privacy regime (i.e., $\I \left(\mathbf{S}; \mathbf{Z}\right) =0$). Note that by controlling the rate $R^{\mathrm{z}}$, the defender (data owner) exploits the imposed distortion at the utility service provider (authorized decoder) to control the uncertainty for the adversary (non-authorized decoder). 
%
%Our first objective is to provide a characterization of the \textit{rate-utility-leakage} for the case of known joint distribution $P_{\mathbf{U, S, X}}$. 
%
The values $\mathsf{CLUB} \left( R^{\mathrm{z}}, R^{\mathrm{s}}, P_{\mathbf{U}, \mathbf{S}, \mathbf{X}}\right)$ for different $R^{\mathrm{z}}$ and $R^{\mathrm{s}}$ specify the CLUB curve.

Alternatively, one may impose constraints on useful information $\I \left( \mathbf{U}; \mathbf{Z} \right)$ and disclosed private information $\I \left( \mathbf{S}; \mathbf{Z}\right)$, which leads to\footnote{For the sake of brevity, we do not consider a new notation for this functional. The CLUB arguments clearly distinguish them.}:\vspace{-5pt}
\begin{equation}\label{CLUB_functional_infimum}
\mathsf{CLUB} \left( R^{\mathrm{u}}, R^{\mathrm{s}}, P_{\mathbf{U}, \mathbf{S}, \mathbf{X}}\right)  \coloneqq \mathop{\inf}_{\substack{ P_{\mathbf{Z} \mid \mathbf{X}}: \\\left( \mathbf{U}, \mathbf{S}\right) \markov \mathbf{X} \markov \mathbf{Z}}} \I  \left( \mathbf{X}; \mathbf{Z} \right)  \; \quad  \; \mathrm{s.t.}  \quad   \I \left( \mathbf{U}; \mathbf{Z} \right) \geq R^{\mathrm{u}}, \;\; 
\I \left( \mathbf{S}; \mathbf{Z} \right) \leq R^{\mathrm{s}}. 
\end{equation}

In order to explore the CLUB curve, one must find the optimal bottleneck representation $\mathbf{Z}$ for different values of $R^{\mathrm{z}}$ and $R^{\mathrm{s}}$ ($R^{\mathrm{u}}$ and $R^{\mathrm{s}}$). 
In practice, the CLUB curve is explored by maximizing (minimizing) its associated Lagrangian functional. 
Therefore, equivalently, with the introduction of a Lagrange multipliers $\beta , \alpha \in \left[ 0, 1\right]$ and $\gamma , \lambda \in \left( 0, + \infty \right)$\footnote{Note that by the DPI \cite{polyanskiy2014lecture, duchi2016lecture}, $\I  \left( \mathbf{U}; \mathbf{Z} \right) \leq \I  \left( \mathbf{X}; \mathbf{Z} \right)$ and $\I  \left( \mathbf{S}; \mathbf{Z} \right) \leq \I  \left( \mathbf{X}; \mathbf{Z} \right)$. Hence, for $\beta , \alpha > 1$, based on the considered Lagrangian functional the model may learn a trivial representation, independent of $\mathbf{X}$.}, we can formulate the CLUB problem by the associated Lagrangian functional\footnote{Note that depending on the focus of the optimization problem, one can consider different optimization objectives along with corresponding information regularization terms, which lead to different Lagrangian functionals. The range of Lagrange multipliers and the behaviour of solutions differ.}. 
%
%
%
%Finally, with the introduction of two Lagrange multipliers, we can quantify Complexity-Leakage-Utility Bottleneck problem by the associated Lagrangian functional\footnote{Note that depending on the focus of optimization problem, one can consider different optimization objective along with corresponding information regularization terms, which leads to different Lagrangian functional. In this case, the range of Lagrange multipliers and behaviour of solution are differ.}. 
In particular, the two associated CLUB Lagrangian functionals of our interest are given as follows:
\begin{subequations}\label{CLUB_Lagrangian}
\begin{align}
\mathcal{L}_{\mathrm{CLUB}} \left( P_{\mathbf{Z}\mid \mathbf{X}}, \beta , \alpha  \right) 
&  \coloneqq      \I  \left( \mathbf{U}; \mathbf{Z} \right) - \beta \,   \I \left( \mathbf{X}; \mathbf{Z} \right) - \alpha  \, \I \left( \mathbf{S}; \mathbf{Z} \right), \label{CLUB_Lagrangian_1} \\
\mathcal{L}_{\mathrm{CLUB}} \left( P_{\mathbf{Z}\mid \mathbf{X}}, \gamma , \lambda \right)   
&  \coloneqq     \I  \left( \mathbf{X}; \mathbf{Z} \right) - \gamma \,   \I \left( \mathbf{U}; \mathbf{Z} \right) +\lambda  \, \I \left( \mathbf{S}; \mathbf{Z} \right).\label{CLUB_Lagrangian_2}
\end{align}
\end{subequations}
%where $\beta , \alpha \in \left[ 0, 1\right]$ and $\gamma , \lambda \in \left( 0, + \infty \right)$. 

%Our first objective is to provide a characterization of the \textit{complexity-leakage-utility} trade-off for the case of known joint distribution $P_{\mathbf{U, S, X}}$. 
Consider the set of 3-dimensional mutual information region for $\left( \mathbf{U}, \mathbf{S}, \mathbf{X} \right) \sim P_{\mathbf{U}, \mathbf{S}, \mathbf{X}}$, defined as follows: 
\begin{equation}\label{CLUB_InformationRegions}
%\mathcal{I}_{\mathbf{U}, \mathbf{S}, \mathbf{X}} 
\mathcal{I} \left( P_{\mathbf{U}, \mathbf{S}, \mathbf{X}} \right) 
\coloneqq \bigcup_{P_{\mathbf{Z} \mid \mathbf{X}}} \Big\{ \I \left( \mathbf{U}; \mathbf{Z} \right), \I \left( \mathbf{S}; \mathbf{Z} \right), \I \left( \mathbf{X}; \mathbf{Z} \right) : \left( \mathbf{U}, \mathbf{S} \right) \markov \mathbf{X} \markov \mathbf{Z} \Big\} \subseteq \mathbb{R}^3. 
\end{equation}
Also, consider the constrained information regions $\mathcal{R} \left( P_{\mathbf{U}, \mathbf{S}, \mathbf{X}} \right)$, defined as follows:
\begin{eqnarray}
\mathcal{R} \left( P_{\mathbf{U}, \mathbf{S}, \mathbf{X}} \right) \coloneqq \big\{ \left( R^{\mathrm{u}}, R^{\mathrm{s}}, R^{\mathrm{z}} \right) \in  \mathcal{I}_{\mathbf{U}, \mathbf{S}, \mathbf{X}} :   \;  \I \left( \mathbf{U}; \mathbf{Z} \right) \geq R^{\mathrm{u}}, \I \left( \mathbf{S}; \mathbf{Z} \right) \leq R^{\mathrm{s}},  \I \left( \mathbf{X}; \mathbf{Z} \right) \leq R^{\mathrm{z}}
\big\}. 
\end{eqnarray}
%\begin{eqnarray}
%\mathcal{R} \left( \mathbf{P}_{\mathbf{U}, \mathbf{S}, \mathbf{X}} \right) \coloneqq \big\{ \left( R^{\mathrm{u}}, R^{\mathrm{s}}, R^{\mathrm{x}} \right) \in  \mathbb{R}^3 : \exists
%\; \mathbf{Z} \; \mathrm{s.t.} \;  \I \left( \mathbf{U}; \mathbf{Z} \right) \geq R^{\mathrm{u}}, \I \left( \mathbf{S}; \mathbf{Z} \right) \leq R^{\mathrm{s}},  \I \left( \mathbf{X}; \mathbf{Z} \right) \leq R^{\mathrm{z}}, \left( \mathbf{U}, \mathbf{S}\right) \markov \mathbf{X} \markov \mathbf{Z}
%\big\}. 
%\end{eqnarray}

%It is clear that Information Bottleneck \cite{tishby2000information} and Privacy Funnel \cite{makhdoumi2014information} are special cases of Complexity-Leakage-Utility Bottleneck $\mathcal{I}_{\mathbf{U}, \mathbf{S}, \mathbf{X}} $. 
In Sec.~\ref{Ssec:ConnectionToOtherProblems}, we show that the CLUB model subsumes most previously reported information-theoretic privacy models. % \cite{du2012privacy, sankar2013utility, calmon2013bounds, makhdoumi2013privacy, asoodeh2014notes, calmon2015fundamental, salamatian2015managing, basciftci2016privacy, asoodeh2016information, kalantari2017information, rassouli2018latent, asoodeh2018estimation, rassouli2018perfect, liao2018privacy, osia2018deep, tripathy2019privacy, Hsu2019watchdogs, liao2019tunable,  sreekumar2019optimal, xiao2019maximal, diaz2019robustness, rassouli2019data, rassouli2019optimal, rassouli2021perfect}. 
More specifically, we show that the mutual information region $\mathcal{I} \left( P_{\mathbf{U}, \mathbf{S}, \mathbf{X}} \right) $ subsumes the recent models proposed in  \cite{bertran2019adversarially, fischer2020conditional, rodriguez2020variational}. 
\section{Variational CLUB}
\label{Sec:VariationalCLUB}

Direct optimization of \eqref{CLUB_Lagrangian_1} to obtain the optimal stochastic mapping $P_{\mathbf{Z} \mid \mathbf{X}}$ is generally challenging. Instead, a tight variational bound can be optimized. 
Let $Q_{\mathbf{U}\mid \mathbf{Z}} : \mathcal{Z} \rightarrow \mathcal{P}\left( \mathcal{U}\right)$, %$Q_{\mathbf{S}\mid \mathbf{Z}} : \mathcal{Z} \rightarrow \mathcal{P}\left( \mathcal{S}\right)$, 
$Q_{\mathbf{X}\mid \mathbf{S}, \mathbf{Z}}: \mathcal{S} \times \mathcal{Z}  \rightarrow \mathcal{P}\left( \mathcal{X}\right)$ and $Q_{\mathbf{Z}}: \mathcal{Z} \rightarrow \mathcal{P}\left( \mathcal{Z}\right)$ be variational approximations of the optimal \textit{utility} decoder distribution $P_{\mathbf{U} \mid \mathbf{Z}}$, 
%adversary decoder distribution $P_{\mathbf{S} \mid \mathbf{Z}}$, 
\textit{uncertainty} decoder distribution $P_{\mathbf{X}\mid \mathbf{S}, \mathbf{Z}}$, and latent space distribution $P_{\mathbf{Z}}$, respectively. 
In the sequel, we will obtain a variational bound $\mathcal{L}_{\mathrm{VCLUB}} \left( P_{\mathbf{Z}\mid \mathbf{X}}, Q_{\mathbf{U}\mid \mathbf{Z}}, Q_{\mathbf{Z}}, Q_{\mathbf{X}\mid \mathbf{S}, \mathbf{Z}}, \beta , \alpha  \right)$, such that for any valid mapping $P_{\mathbf{Z} \mid \mathbf{X}}$ satisfying the Markov chain condition $ \left( \mathbf{U}, \mathbf{S} \right) \markov \mathbf{X} \markov \mathbf{Z}$, we have:\vspace{-7pt}
\begin{equation}\label{CLUB_vs_VCLUB_1}
\mathcal{L}_{\mathrm{CLUB}} \left( P_{\mathbf{Z}\mid \mathbf{X}}, \beta , \alpha  \right)  \geq \mathcal{L}_{\mathrm{VCLUB}} \left( P_{\mathbf{Z}\mid \mathbf{X}}, Q_{\mathbf{U}\mid \mathbf{Z}},  Q_{\mathbf{Z}}, Q_{\mathbf{X}\mid \mathbf{S}, \mathbf{Z}}, \beta , \alpha  \right).
\end{equation}
The inequality holds with equality, if the variational approximations match the true distributions,
%(Bayes optimal decoders)
i.e., when $Q_{\mathbf{U} \mid \mathbf{Z}} = P_{\mathbf{U}\mid \mathbf{Z}}$, $Q_{\mathbf{X}\mid \mathbf{S}, \mathbf{Z}} = P_{\mathbf{X}\mid \mathbf{S}, \mathbf{Z}}$, and $Q_{\mathbf{Z}}= P_{\mathbf{Z}}$. Since \eqref{CLUB_vs_VCLUB_1} holds for any $P_{\mathbf{Z} \mid \mathbf{X}}$, instead of maximizing $\mathcal{L}_{\mathrm{CLUB}} \left( P_{\mathbf{Z}\mid \mathbf{X}}, \beta , \alpha  \right)$, we will maximize $\mathcal{L}_{\mathrm{VCLUB}} \left( P_{\mathbf{Z}\mid \mathbf{X}}, Q_{\mathbf{U}\mid \mathbf{Z}},  Q_{\mathbf{Z}}, Q_{\mathbf{X}\mid \mathbf{S}, \mathbf{Z}}, \beta , \alpha  \right)$ over the variational distributions.

Alternatively, consider the CLUB Lagrangian functional \eqref{CLUB_Lagrangian_2} and let $\mathcal{L}_{\mathrm{VCLUB}} \left( P_{\mathbf{Z}\mid \mathbf{X}}, Q_{\mathbf{U}\mid \mathbf{Z}}, Q_{\mathbf{Z}}, Q_{\mathbf{X}\mid \mathbf{S}, \mathbf{Z}}, \gamma , \lambda  \right)$ be the obtained variational bound. 
For any valid mapping $P_{\mathbf{Z} \mid \mathbf{X}}$ satisfying the Markov chain condition $ \left( \mathbf{U}, \mathbf{S} \right) \markov \mathbf{X} \markov \mathbf{Z}$, we have:\vspace{-7pt}
\begin{equation}\label{CLUB_vs_VCLUB_2}
\mathcal{L}_{\mathrm{CLUB}} \left( P_{\mathbf{Z}\mid \mathbf{X}}, \gamma , \lambda  \right)  \leq \mathcal{L}_{\mathrm{VCLUB}} \left( P_{\mathbf{Z}\mid \mathbf{X}}, Q_{\mathbf{U}\mid \mathbf{Z}},  Q_{\mathbf{Z}}, Q_{\mathbf{X}\mid \mathbf{S}, \mathbf{Z}}, \gamma , \lambda  \right).
\end{equation}
%In the following, we focus on the var

\vspace{-6pt}

%%------------------------------
%       Variational bound of  I ( X; Z )
%%------------------------------
%
\subsection{Variational Bound on Information Complexity} 
The information complexity $\I \left( \mathbf{X}; \mathbf{Z} \right)$ can be decomposed as:
\begin{eqnarray}\label{I_xz_decomposition}
\I \left( \mathbf{X}; \mathbf{Z} \right) = \mathbb{E}_{P_{\mathbf{X}, \mathbf{Z}}} \left[ \log \frac{P_{\mathbf{X}, \mathbf{Z} }}{P_{\mathbf{X}} P_{\mathbf{Z}}} \right] = 
\mathbb{E}_{P_{\mathbf{X}, \mathbf{Z}}} \left[ \log \frac{P_{\mathbf{Z} \mid \mathbf{X} }}{ P_{\mathbf{Z}}} \right] &=& 
\mathbb{E}_{P_{\mathbf{X}, \mathbf{Z}}} \left[ \log \frac{P_{\mathbf{Z} \mid \mathbf{X} }}{ Q_{\mathbf{Z}}} \right] - \D_{\mathrm{KL}} \left( P_{\mathbf{Z}} \Vert Q_{\mathbf{Z}}\right)\\
&=&  \D_{\mathrm{KL}} \left( P_{\mathbf{Z} \mid \mathbf{X}} \Vert Q_{\mathbf{Z}} \mid P_{\mathbf{X}} \right) - \D_{\mathrm{KL}} \left( P_{\mathbf{Z}} \Vert Q_{\mathbf{Z}}\right),
\end{eqnarray}
where $\D_{\mathrm{KL}} \left( P_{\mathbf{Z} \mid \mathbf{X}} \Vert Q_{\mathbf{Z}} \mid P_{\mathbf{X}} \right)$ is the conditional divergence. Since $\D_{\mathrm{KL}} \left( P_{\mathbf{Z}} \Vert Q_{\mathbf{Z}}\right) \geq 0$, we can upper bound \eqref{I_xz_decomposition} as:\vspace{-6pt}
\begin{eqnarray}\label{I_XZ_upperBound}
\I \left( \mathbf{X}; \mathbf{Z} \right)  \leq 
%\mathbb{E}_{P_{\mathbf{X}, \mathbf{Z}}} \left[ \log \frac{P_{\mathbf{Z} \mid \mathbf{X} }}{ Q_{\mathbf{Z}}} \right] = 
% \mathbb{E}_{P_{\mathbf{X}}} \left[ \D_{\mathrm{KL}}  \left( P_{\mathbf{Z} \mid \mathbf{X}} \Vert Q_{\mathbf{Z}} \right) \right] = 
\D_{\mathrm{KL}} \left( P_{\mathbf{Z} \mid \mathbf{X}} \Vert Q_{\mathbf{Z}} \mid P_{\mathbf{X}} \right). 
\end{eqnarray}

Hence, one can minimize the information complexity upper bound \eqref{I_XZ_upperBound}, instead of minimizing \eqref{I_xz_decomposition}. 
Note that the information complexity is completely characterized using two terms $ \D_{\mathrm{KL}} \left( P_{\mathbf{Z} \mid \mathbf{X}} \Vert Q_{\mathbf{Z}} \mid P_{\mathbf{X}} \right) $ and $\D_{\mathrm{KL}} \left( P_{\mathbf{Z}} \Vert Q_{\mathbf{Z}}\right)$.

\vspace{-6pt}

%%------------------------------
%       Variational bound of  I ( U; Z )
%%------------------------------
%
\subsection{Variational Bound on Information Utility} 
\vspace{-3pt}
We can decompose the mutual information between the released representation $\mathbf{Z}$ and the \textit{utility attribute} $\mathbf{U}$ in two analytically equivalent ways:\vspace{-7pt}
\begin{equation}
\I \left( \mathbf{U}; \mathbf{Z} \right) = \underbrace{\H \left( \mathbf{U} \right) - \H \left( \mathbf{U} \mid \mathbf{Z} \right)}_{\mathrm{Discriminative~View}} = \underbrace{\H \left( \mathbf{Z} \right) - \H \left( \mathbf{Z} \mid \mathbf{U} \right)}_{\mathrm{Generative~View}} .
\end{equation}
Therefore, maximizing $\I \left( \mathbf{U}; \mathbf{Z} \right)$ can have two different interpretations. The discriminative view expresses that (i) the utility attribute needs to be distributed as uniformly as possible in the data space\footnote{We assumed our attributes are supported on a finite space.
%Also, note that this distribution is out of our control.
}, and (ii) the utility attribute should be confidently inferred from the released representation $\mathbf{Z}$. 
On the other hand, the generative view expresses that (i) the released representations should be spread as much as possible in the latent space (i.e., high entropy $\H(\mathbf{Z})$), and (ii) the released representation corresponding to the same utility attribute should be close together\footnote{Note that this interpretation is aligned with the partitioning interpretation of sufficient statistics as discussed in Sec.~\ref{Ssec:Relevant Information}.} (i.e., minimizing conditional entropy $\H (\mathbf{Z} \! \mid \! \mathbf{U})$). 
Since our goal is to identify the utility attributes based on the revealed representation, the discriminative view of this decomposition is more aligned with our model. We rewrite the conditional entropy term as follows:\vspace{-8pt}
\begin{subequations}
\begin{align}\label{Decom_H_UmidZ}
\H \left( \mathbf{U} \mid \mathbf{Z} \right) &= 
\mathbb{E}_{P_{\mathbf{U}, \mathbf{Z}}} \left[ \log P_{\mathbf{U}\mid \mathbf{Z}} \right] \\
&= \mathbb{E}_{P_{\mathbf{U}, \mathbf{Z}}} \left[ \log Q_{\mathbf{U}\mid \mathbf{Z}} \right] + 
\mathbb{E}_{P_{\mathbf{U}, \mathbf{Z}}} \left[ \log \frac{P_{\mathbf{U}\mid \mathbf{Z}}}{Q_{\mathbf{U}\mid \mathbf{Z}}} \right] 
\\
&= \mathbb{E}_{P_{\mathbf{U}, \mathbf{Z}}} \left[ \log Q_{\mathbf{U}\mid \mathbf{Z}} \right] +   \D_{\mathrm{KL}} \left( P_{\mathbf{U}\mid \mathbf{Z} }  \Vert Q_{\mathbf{U} \mid \mathbf{Z}} \mid P_{\mathbf{Z}} \right).
\end{align}
\end{subequations}
Note that $\H \left( P_{\mathbf{U} \mid \mathbf{Z}} \Vert Q_{\mathbf{U} \mid \mathbf{Z}}  \mid P_{\mathbf{Z}} \right) \coloneqq \mathbb{E}_{P_{\mathbf{Z}}} \left[  \mathbb{E}_{P_{\mathbf{U} \mid \mathbf{Z}}} \left[  \log Q_{\mathbf{U} \mid \mathbf{Z}}  \right] \right] =
\mathbb{E}_{P_{\mathbf{U}, \mathbf{Z}}} \left[ \log Q_{\mathbf{U}\mid\mathbf{Z}} \right]$ is the cross-entropy loss function. Since $\D_{\mathrm{KL}} \left( P_{\mathbf{U}\mid \mathbf{Z} }  \Vert Q_{\mathbf{U} \mid \mathbf{Z}}  \mid P_{\mathbf{Z}} \right) \geq 0$, we can lower bound information utility as:\vspace{-6pt}
\begin{eqnarray}\label{I_UZ_lowerBound}
\I \left( \mathbf{U}; \mathbf{Z} \right) \geq  \H \left( \mathbf{U} \right) - \H \left( P_{\mathbf{U} \mid \mathbf{Z}} \Vert Q_{\mathbf{U} \mid \mathbf{Z}}  \mid P_{\mathbf{Z}} \right) . 
\end{eqnarray}
According to \eqref{I_UZ_lowerBound}, the minimization of the average cross-entropy loss leads to the maximization of (a lower bound of) the information gain $\I \left( \mathbf{U}; \mathbf{Z} \right)$. 
Therefore, if the conditional distributions are very close to each other, i.e., if the gap $\D_{\mathrm{KL}} \left( P_{\mathbf{U}\mid \mathbf{Z} }  \Vert Q_{\mathbf{U} \mid \mathbf{Z}} \mid P_{\mathbf{Z}} \right)$ is small, by minimizing the average cross-entropy loss $\H \left( P_{\mathbf{U} \mid \mathbf{Z}} \Vert Q_{\mathbf{U} \mid \mathbf{Z}} \mid P_{\mathbf{Z}}  \right)$ over representations, we can achieve a tight upper bound on the decoder's uncertainty $\H \left( \mathbf{U} \!  \mid \! \mathbf{Z} \right)$. Furthermore, note that minimizing $\H \left( \mathbf{U} \! \mid \! \mathbf{Z} \right)$ leads to a small probability of error in the prediction of the (discrete) utility attribute from released representation. 

\vspace{-5pt}

\begin{remark}[From supervised CLUB to unsupervised CLUB]
Let us consider a scenario in which the data owner wishes to release the original domain data $\mathbf{X}$ (e.g., facial images) as accurately as possible (i.e., $\mathbf{U} \equiv \mathbf{X}$), without revealing a specific sensitive attribute $\mathbf{S}$ (e.g., gender, emotion, etc.). 
%Hence, the data owner shares the sanitized representation $\mathbf{Z}$ for utility task, which is defined as to \textit{reconstruct} (generate) $\mathbf{X}$ from $\mathbf{Z}$, i.e., $\mathbf{U} \equiv \mathbf{X}$. 
We call this setup as the \textit{unsupervised} CLUB model and the formerly described general case, where $\mathbf{U}$ is a generic attribute of data $\mathbf{X}$ as the \textit{supervised} CLUB model. 
In the unsupervised scenario, we can proceed with a similar decomposition as derived above, but with a few modifications. 
%In particular, we need to distinguish the mutual information utility measure in this setup from the information complexity \eqref{I_xz_decomposition}. 
For the sake of brevity, we address this decomposition in Sec.~\ref{Sec:DVCLUB}, after introducing our parameterized variational approximation of the distributions. 
\end{remark}

\vspace{-6pt}

%%------------------------------
%       Variational bound of  I ( S; Z )
%%------------------------------
%
\subsection{Variational Bound on Information Leakage} 
\vspace{-3pt}
One can obtain a variational lower bound on $\I \left( \mathbf{S}; \mathbf{Z}\right)$ similarly to the one obtained in \eqref{I_UZ_lowerBound}, that is:\vspace{-6pt}
\begin{eqnarray}\label{I_SZ_LowerBound}
\I \left( \mathbf{S}; \mathbf{Z} \right) \geq  \H \left( \mathbf{S} \right) + \mathbb{E}_{P_{\mathbf{S}, \mathbf{X}}} \left[ \mathbb{E}_{P_{\mathbf{Z} \mid \mathbf{X}}} \left[ \log Q_{\mathbf{S} \mid \mathbf{Z}} \right] \right]. 
\end{eqnarray}
However, we want to minimize $\I \left( \mathbf{S}; \mathbf{Z} \right)$; therefore, we instead need a variational upper bound.
% for it, in which we able to minimize this upper bound. 
Alternatively, we can express $\I \left( \mathbf{S}; \mathbf{Z} \right)$ as:\vspace{-6pt}
\begin{eqnarray}\label{I_sz_decomposition}
\I \left( \mathbf{S}; \mathbf{Z} \right) &=& \I \left( \mathbf{X}; \mathbf{Z} \right) - \I \left( \mathbf{X}; \mathbf{Z} \mid \mathbf{S} \right) \nonumber \\
&=& 
\I \left( \mathbf{X}; \mathbf{Z} \right) - \H \left( \mathbf{X} \mid \mathbf{S} \right) + \H \left(  \mathbf{X} \mid \mathbf{S}, \mathbf{Z}\right). 
%&=& 
%\I \left( \mathbf{X}; \mathbf{Z} \right) - \H \left( \mathbf{Z} \mid \mathbf{S} \right) + \H \left(  \mathbf{Z} \mid  \mathbf{X}\right)
\end{eqnarray}
By utilizing the variational approximation $Q_{\mathbf{X}\mid \mathbf{S}, \mathbf{Z}}$ for the conditional entropy $\H \left( \mathbf{X} \! \mid \! \mathbf{S}, \mathbf{Z} \right)$, we have:\vspace{-6pt}
\begin{equation}\label{ConditionalEntropy_X_given_SZ}
\H \left( \mathbf{X} \! \mid \! \mathbf{S}, \mathbf{Z} \right) = - \mathbb{E}_{P_{\mathbf{S}, \mathbf{X}, \mathbf{Z}}} \left[ \log P_{\mathbf{X} \mid \mathbf{S}, \mathbf{Z}} \right] 
= -\mathbb{E}_{P_{\mathbf{S}, \mathbf{X}}} \left[  \mathbb{E}_{P_{\mathbf{Z} \mid \mathbf{X}}} \left[ \log Q_{\mathbf{X} \mid \mathbf{S}, \mathbf{Z}} \right] \right] - 
\D_{\mathrm{KL}} \left( P_{\mathbf{X}\mid \mathbf{S}, \mathbf{Z} }  \Vert Q_{\mathbf{X} \mid \mathbf{S}, \mathbf{Z}} \right).\vspace{-4pt}
\end{equation}
Considering $\D_{\mathrm{KL}} \left( P_{\mathbf{X}\mid \mathbf{S}, \mathbf{Z} }  \Vert Q_{\mathbf{X} \mid \mathbf{S}, \mathbf{Z}} \right) \geq 0$ and using \eqref{I_XZ_upperBound}, we can obtain an upper bound on $\I \left( \mathbf{S}; \mathbf{Z} \right) $ as:\vspace{-6pt}
\begin{equation}\label{I_SZ_upperBound}
\I \left( \mathbf{S}; \mathbf{Z} \right)  \leq  \D_{\mathrm{KL}} \left( P_{\mathbf{Z} \mid \mathbf{X}} \Vert Q_{\mathbf{Z}} \mid P_{\mathbf{X}} \right) + \H \left( \mathbf{X} \mid \mathbf{S} \right) +
\H \left( P_{\mathbf{X} \mid \mathbf{S,Z}} \Vert  Q_{\mathbf{X} \mid \mathbf{S,Z}}  \mid P_{\mathbf{S,Z}} \right),\vspace{-4pt}
\end{equation}
where $\H \left( P_{\mathbf{X} \mid \mathbf{S,Z}} \Vert  Q_{\mathbf{X} \mid \mathbf{S,Z}}  \mid P_{\mathbf{S,Z}} \right) \coloneqq \mathbb{E}_{P_{\mathbf{S}, \mathbf{X}, \mathbf{Z}}} \left[ \log Q_{\mathbf{X} \mid \mathbf{S}, \mathbf{Z}} \right]$. 
%
%Note that the first term has also appeared in the upper bound of \eqref{I_XZ_upperBound}. 
%
Eqn.~\eqref{I_sz_decomposition} depicts the critical role of information complexity $\I \left( \mathbf{X}; \mathbf{Z}\right)$ in controlling the information leakage $\I \left( \mathbf{S}; \mathbf{Z}\right)$. 
The less the information complexity, the less distinguishable the revealed representations, and hence the less the information leakage. 
The term $\H \! \left( P_{\mathbf{X} \mid \mathbf{S,Z}} \Vert  Q_{\mathbf{X} \mid \mathbf{S,Z}} \! \mid \! P_{\mathbf{S,Z}} \right)$ can be interpreted as follows. Consider the inference threat model presented in Sec.~\ref{Ssec:ObfuscationUtilityUnderLogLoss}. The inferential adversary has access to the revealed representation $\mathbf{Z}$ and is interested in inferring the sensitive attribute $\mathbf{S}$. Note that the adversary's inference is through reconstructing the original data $\mathbf{X}$. 
Therefore, after observing $\mathbf{Z}$, the adversary chooses a belief distribution over $\mathcal{S}$ and then tries to reconstruct the original data $\mathbf{X}$ to revise his belief. If the released representation is statistically independent of the sensitive attribute $\mathbf{S}$, then the adversary cannot revise his belief by injecting various $\mathbf{S}$ to his inferential model. 
Hence, by maximizing the average log-likelihood $ \log Q_{\mathbf{X} \mid \mathbf{S}, \mathbf{Z}} $ over $P_{\mathbf{Z} \mid \mathbf{X}}$, the defender minimizes the average adversarial inference about $\mathbf{S}$. 

Considering \eqref{I_XZ_upperBound}, \eqref{I_UZ_lowerBound}, and \eqref{I_SZ_upperBound}, in the \textit{supervised} scenario, the VCLUB bound in \eqref{CLUB_vs_VCLUB_1} can be written as\footnote{Note that we use the superscripts $(\cdot)^{\mathsf{S}}$ and $(\cdot)^{\mathsf{U}}$ for the (Deep) VCLUB Lagrangian functionals associated with supervised and unsupervised setups, respectively, while we use the superscripts $(\cdot)^{\mathrm{s}}$ and $(\cdot)^{\mathrm{u}}$ for the quantities associated to sensitive and utility attributes, respectively.}:\vspace{-8pt}
\begin{eqnarray}\label{supervised_VCLUB}
 && \hspace{-40pt} \mathcal{L}_{\mathrm{VCLUB}}^{\mathsf{S}} \left( P_{\mathbf{Z}\mid \mathbf{X}}, Q_{\mathbf{U}\mid \mathbf{Z}}, Q_{\mathbf{Z}}, Q_{\mathbf{X}\mid \mathbf{S}, \mathbf{Z}}, \beta , \alpha  \right) \nonumber \\
& & = \mathbb{E}_{P_{\mathbf{U}, \mathbf{X}}} \left[ \mathbb{E}_{P_{\mathbf{Z} \mid \mathbf{X}}} \left[ \log Q_{\mathbf{U} \mid \mathbf{Z}} \right] \right] - \left( \beta + \alpha \right)  \mathbb{E}_{P_{\mathbf{X}, \mathbf{Z}}} \left[ \log \frac{P_{\mathbf{Z} \mid \mathbf{X} }}{ Q_{\mathbf{Z}}} \right]  + \alpha \, \mathbb{E}_{P_{\mathbf{S}, \mathbf{X}}} \left[  \mathbb{E}_{P_{\mathbf{Z} \mid \mathbf{X}}} \left[ \log Q_{\mathbf{X} \mid \mathbf{S}, \mathbf{Z}} \right] \right] + \mathrm{c}, \vspace{-4pt}
\end{eqnarray}
where $\mathrm{c}$ is a constant term. 
For the sake of brevity, we directly address the \textit{unsupervised} DVCLUB in Sec.~\ref{Sec:DVCLUB}.

%\vspace{-2pt}

%---------------------------------------------------
%		 Deep Variational Complexity-Leakage-Utility Bottleneck
%---------------------------------------------------
%%------------------------------------
%      		Deep Variational CLUB  
%%------------------------------------
%
\section{Deep Variational CLUB (DVCLUB)}
\label{Sec:DVCLUB}

The common approach is to use neural networks to parameterize variational inference bounds. To this goal, we parameterize the encoding distribution $P_{\mathbf{Z}\mid \mathbf{X}}$, utility decoding distribution $Q_{\mathbf{U}\mid \mathbf{Z}}$, uncertainty decoding distribution $Q_{\mathbf{X}\mid \mathbf{S,Z}}$, and prior $Q_{\mathbf{Z}}$. 
Let $P_{\boldsymbol{\phi}} (\mathbf{Z} \! \mid \! \mathbf{X})$ denote the family of encoding probability distributions $P_{\mathbf{Z} \mid \mathbf{X}}$ over $\mathcal{Z}$ for each element of space $\mathcal{X}$, parameterized by the output of a deep neural network $f_{\boldsymbol{\phi}}$ with parameters $\boldsymbol{\phi}$. In the context of inference problems, $P_{\boldsymbol{\phi}} (\mathbf{Z} \! \mid \! \mathbf{X})$ is called the amortized\footnote{The term `amortized' comes from the fact that 
instead of optimizing a set of free (explicit) parameters per datum, the model learns a parameterized mapping \textit{from} samples \textit{to} the target distribution. This allows the variational parameters to remain constant with the data size.} variational inference (AVI) distribution or variational posterior distribution. 
Analogously, let $P_{\boldsymbol{\theta}} (\mathbf{U} \! \mid \! \mathbf{Z})$ and $P_{\boldsymbol{\varphi}} \! \left( \mathbf{X} \! \mid \! \mathbf{S}, \mathbf{Z} \right)$ denote the corresponding family of decoding probability distributions $Q_{\mathbf{U} \mid \mathbf{Z}}$ and $Q_{\mathbf{X} \mid \mathbf{S}, \mathbf{Z}}$, respectively, parameterized by the output of the deep neural networks $g_{\boldsymbol{\theta}}$ and $g_{\boldsymbol{\varphi}}$. 
Moreover, for the prior distribution $Q_{\mathbf{Z}}$ we consider the family of distributions $Q_{\boldsymbol{\psi}} \! \left( \mathbf{Z} \right)$, which can be interpreted as the target (proposal) distribution in the latent space. The choice for these distributions is considered by trading off computational complexity with model expressiveness. 
Let $P_{\mathsf{D}} \! \left( \mathbf{X} \right) \! = \! \! \frac{1}{N} \! \sum_{n=1}^{N} \! \delta ( \mathbf{x}   -   \mathbf{x}_n )$, $\mathbf{x}_n \!\! \in \! \mathcal{X}$, denote the empirical data distribution. In this case, $P_{\! \boldsymbol{\phi}} \! \left( \mathbf{X}, \mathbf{Z} \right) \! = \! P_{\mathsf{D}} (\mathbf{X}) P_{\! \boldsymbol{\phi}} \! \left( \mathbf{Z} \! \mid \!  \mathbf{X} \right)$ denotes our joint inference data distribution, and $P_{\! \boldsymbol{\phi}} (\mathbf{Z}) \! = \! \mathbb{E}_{\! P_{\mathsf{D}} (\mathbf{X})} \! \left[ P_{\! \boldsymbol{\phi}} (\mathbf{Z} \! \mid \! \mathbf{X}) \right]$ denotes the learned \textit{aggregated} posterior distribution over latent space~$\mathcal{Z}$.

\noindent
\textbf{Parameterized Information Complexity}: 
%Under this parameterization, 
The parameterized variational approximation of \textit{information complexity} \eqref{I_xz_decomposition} can be defined~as:\vspace{-7pt}
%$\I_{\boldsymbol{\phi}} \! \left( \mathbf{X}; \mathbf{Z} \right) =  \D_{\mathrm{KL}} \! \left( P_{\boldsymbol{\phi}} (\mathbf{Z} \! \mid \! \mathbf{X}) \, \Vert \, Q_{\boldsymbol{\psi}}   (\mathbf{Z}) \mid P_{\mathsf{D}}(\mathbf{X}) \right) + \D_{\mathrm{KL}} \! \left( P_{\boldsymbol{\phi}}(\mathbf{Z}) \, \Vert \, Q_{\boldsymbol{\psi}} (\mathbf{Z})\right)$. 
\begin{eqnarray}\label{Eq:I_XZ_phi}
\I_{\boldsymbol{\phi}, \boldsymbol{\psi}} \! \left( \mathbf{X}; \mathbf{Z} \right) \coloneqq  \D_{\mathrm{KL}} \! \left( P_{\boldsymbol{\phi}} (\mathbf{Z} \! \mid \! \mathbf{X}) \, \Vert \, Q_{\boldsymbol{\psi}}   (\mathbf{Z}) \mid P_{\mathsf{D}}(\mathbf{X}) \right) -  \D_{\mathrm{KL}} \! \left( P_{\boldsymbol{\phi}}(\mathbf{Z}) \, \Vert \, Q_{\boldsymbol{\psi}} (\mathbf{Z})\right).  
\end{eqnarray}
Indeed, the information complexity $\I_{\boldsymbol{\phi}, \boldsymbol{\psi}} \! \left( \mathbf{X}; \mathbf{Z} \right)$ measures the amount of Shannon's mutual information between the parameters of the model and the dataset $\mathsf{D}$, given a prior $Q_{\boldsymbol{\psi}} (\mathbf{Z})$ and stochastic map $P_{\boldsymbol{\phi}}(\mathbf{Z} \! \mid \! \mathbf{X}): \mathcal{X} \rightarrow \mathcal{P} (\mathcal{Z})$. Note that the posterior distribution $P_{\boldsymbol{\phi}}(\mathbf{Z} \! \mid \! \mathbf{X})$ depends on the choice of the optimization algorithm, therefore, the information complexity implicitly depends on this choice. 
The relative entropy $\D_{\mathrm{KL}} \! \left( P_{\boldsymbol{\phi}}(\mathbf{Z}) \, \Vert \, Q_{\boldsymbol{\psi}} (\mathbf{Z})\right)$ is usually ignored in the literature. % specifically, in the supervised scenarios. 
A critical challenge is to guarantee that the learned aggregated posterior distribution $P_{\boldsymbol{\phi}} (\mathbf{Z})$ conforms well to the proposed prior $Q_{\boldsymbol{\psi}} \! \left( \mathbf{Z} \right)$ \cite{kingma2016improved, rezende2015variational, rosca2018distribution, tomczak2018vae, bauer2019resampled}. We can tackle this issue by employing a more \textit{expressive} form for $Q_{\boldsymbol{\psi}} \! \left( \mathbf{Z} \right)$, which would allow us to provide a good fit for an arbitrary space $\mathcal{Z}$, at the expense of additional \textit{computational complexity}.

From \eqref{Eq:I_XZ_phi}, we have:\vspace{-7pt}
\begin{subequations}\label{Eq:Lemma_InfoComplexity_KLD_CE}
\begin{align}
\!\!\!\!\!\! \I_{\boldsymbol{\phi}, \boldsymbol{\psi}} \! \left( \mathbf{X}; \mathbf{Z} \right) & \leq \D_{\mathrm{KL}} \! \left( P_{\boldsymbol{\phi}} (\mathbf{Z} \! \mid \! \mathbf{X}) \, \Vert \, Q_{\boldsymbol{\psi}}   (\mathbf{Z}) \mid P_{\mathsf{D}}(\mathbf{X}) \right) \\ 
&=  \mathbb{E}_{P_{\mathsf{D}} (\mathbf{X})} \Big[ \mathbb{E}_{P_{\boldsymbol{\phi}}(\mathbf{Z} \mid \mathbf{X})} \Big[ \log \frac{P_{\boldsymbol{\phi}}(\mathbf{Z} \! \mid \! \mathbf{X})}{Q_{\boldsymbol{\psi}}   (\mathbf{Z})}  \Big] \Big]   \\
&=  \mathbb{E}_{P_{\mathsf{D}} (\mathbf{X})} \left[ \mathbb{E}_{P_{\boldsymbol{\phi}}(\mathbf{Z} \mid \mathbf{X})} \big[ \log P_{\boldsymbol{\phi}}(\mathbf{Z} \! \mid \! \mathbf{X}) \big] \right] 
-  \mathbb{E}_{P_{\boldsymbol{\phi}} (\mathbf{Z})} \left[ \log  Q_{\boldsymbol{\psi}}(\mathbf{Z} )  \right] 
  \\
&= - \mathbb{E}_{P_{\mathsf{D}} (\mathbf{X})} \left[ \, \H \left( P_{\boldsymbol{\phi}}(\mathbf{Z} \! \mid \! \mathbf{X} = \mathbf{x}) \right)\,  \right] + \H  \left( P_{\boldsymbol{\phi}} (\mathbf{Z}) \Vert Q_{\boldsymbol{\psi}} (\mathbf{Z}) \right).
\end{align}
\end{subequations}

% Therefore, if $\D_{\mathrm{KL}} \! \left( P_{\boldsymbol{\phi}}(\mathbf{Z}) \, \Vert \, Q_{\boldsymbol{\psi}} (\mathbf{Z})\right)$ is small, by  minimizing the cross-entropy of the averaged (over training data) variational posterior $P_{\boldsymbol{\phi}}(\mathbf{Z}) $ relative to the proposed prior $Q_{\boldsymbol{\psi}} (\mathbf{Z})$, we can achieve a tight upper bound on the information complexity. Note that by imposing a constraint on the information complexity $\I_{\boldsymbol{\phi}, \boldsymbol{\psi}} \! \left( \mathbf{X}; \mathbf{Z} \right)$, we are imposing a constraint on the entropy of the AVI distribution $P_{\boldsymbol{\phi}}(\mathbf{Z}) $. 

Therefore, if $\D_{\mathrm{KL}} \! \left( P_{\boldsymbol{\phi}}(\mathbf{Z}) \, \Vert \, Q_{\boldsymbol{\psi}} (\mathbf{Z})\right)$ is small, minimizing the cross-entropy term $\H  \left( P_{\boldsymbol{\phi}} (\mathbf{Z}) \Vert Q_{\boldsymbol{\psi}} (\mathbf{Z}) \right)$ provides an upper bound on the information complexity. Note that by imposing a constraint on the information complexity $\I_{\boldsymbol{\phi}, \boldsymbol{\psi}} \! \left( \mathbf{X}; \mathbf{Z} \right)$, we are imposing a constraint on the entropy of the AVI distribution $P_{\boldsymbol{\phi}}(\mathbf{Z}) $.

\noindent
\textbf{Parameterized Information Utility}: 
%In the case in which the utility task is to infer some \textit{attribute} from the original data $\mathbf{X}$, 
The parameterized variational approximation associated to the information utility bound in \eqref{I_UZ_lowerBound} can be defined as:\vspace{-6pt}
\begin{subequations}\label{Eq:I_UZ_phi_theta}
\begin{align}
\I_{\boldsymbol{\phi}, \boldsymbol{\theta}} \! \left( \mathbf{U}; \mathbf{Z} \right) 
& \coloneqq  
%\H (\mathbf{U}) + \mathbb{E}_{P_{\mathbf{U}, \mathbf{X}}} \left[ \mathbb{E}_{P_{\boldsymbol{\phi}} \left( \mathbf{Z} \mid \mathbf{X} \right) } \left[ \log P_{\boldsymbol{\theta}} \! \left( \mathbf{U} \! \mid \! \mathbf{Z} \right) \right] \right] + \D_{\mathrm{KL}} \left( P_{ \mathbf{U}  \mid   \mathbf{Z}} \, \Vert \, P_{\boldsymbol{\theta}} \! \left( \mathbf{U} \! \mid \! \mathbf{Z} \right)  \mid P_{\boldsymbol{\phi}} (\mathbf{Z}) \right)   \\
\H (\mathbf{U}) - \H \left( P_{ \mathbf{U}  \mid   \mathbf{Z}} \, \Vert \, P_{\boldsymbol{\theta}} \! \left( \mathbf{U} \! \mid \! \mathbf{Z} \right)  \mid P_{\boldsymbol{\phi}} (\mathbf{Z}) \right) + \D_{\mathrm{KL}} \left( P_{ \mathbf{U}  \mid   \mathbf{Z}} \, \Vert \, P_{\boldsymbol{\theta}} \! \left( \mathbf{U} \! \mid \! \mathbf{Z} \right)  \mid P_{\boldsymbol{\phi}} (\mathbf{Z}) \right)  \\
& \geq  \H (\mathbf{U}) - \H \left( P_{ \mathbf{U}  \mid   \mathbf{Z}} \, \Vert \, P_{\boldsymbol{\theta}} \! \left( \mathbf{U} \! \mid \! \mathbf{Z} \right)  \mid P_{\boldsymbol{\phi}} (\mathbf{Z}) \right) = 
\H (\mathbf{U})
- \H_{\boldsymbol{\phi}, \boldsymbol{\theta}} \left( \mathbf{U} \! \mid  \! \mathbf{Z} \right),
\end{align}
\end{subequations}
where $\H_{\boldsymbol{\phi}, \boldsymbol{\theta}} \left( \mathbf{U} \! \mid  \! \mathbf{Z} \right) \coloneqq \H \left( P_{ \mathbf{U}  \mid   \mathbf{Z}} \, \Vert \, P_{\boldsymbol{\theta}} \! \left( \mathbf{U} \! \mid \! \mathbf{Z} \right)  \mid P_{\boldsymbol{\phi}} (\mathbf{Z}) \right)$. 

%Note that based on \eqref{Eq:I_UZ_phi_theta} and \eqref{I_uz_decomposition}, we have $\H \left( \mathbf{U} \! \mid \! \mathbf{Z} \right) = \H_{\boldsymbol{\phi}, \boldsymbol{\theta}} \left( \mathbf{U} \! \mid  \! \mathbf{Z} \right) - \D_{\mathrm{KL}} \left( P_{ \mathbf{U}  \mid   \mathbf{Z}} \, \Vert \, P_{\boldsymbol{\theta}} \! \left( \mathbf{U} \! \mid \! \mathbf{Z} \right)  \mid P_{\boldsymbol{\phi}} (\mathbf{Z}) \right)$. 

Alternatively, we can decompose $\I_{\boldsymbol{\phi}, \boldsymbol{\theta}} \! \left( \mathbf{U}; \mathbf{Z} \right) $ as follows:\vspace{-5pt}
\begin{subequations}\label{Eq:I_UZ_phi_theta_SecondDecomposition}
\begin{align}
\I_{\boldsymbol{\phi}, \boldsymbol{\theta}} \! \left( \mathbf{U}; \mathbf{Z} \right) 
& \coloneqq  
 \mathbb{E}_{P_{\mathbf{U}, \mathbf{X}}} \left[ \mathbb{E}_{P_{\boldsymbol{\phi}} \left( \mathbf{Z} \mid \mathbf{X} \right) } \Big[ \log \frac{P_{\boldsymbol{\theta}} \! \left( \mathbf{U} \! \mid \! \mathbf{Z} \right) }{P_{\mathbf{U}}}  \cdot \frac{P_{\boldsymbol{\theta}} (\mathbf{U})}{P_{\boldsymbol{\theta}} (\mathbf{U})} \Big] \right]   \\
&= 
\mathbb{E}_{P_{\mathbf{U}, \mathbf{X}}} \left[  \mathbb{E}_{P_{\boldsymbol{\phi}} \left( \mathbf{Z} \mid \mathbf{X} \right) } \left[ \log P_{\boldsymbol{\theta}} \! \left( \mathbf{U} \! \mid \! \mathbf{Z} \right) \right] \right]
- \mathbb{E}_{P_{\mathbf{U}}} \Big[ \log \frac{P_{\mathbf{U}}}{P_{\boldsymbol{\theta}} (\mathbf{U})}\Big]
- \mathbb{E}_{P_{\mathbf{U}}} \left[ \log P_{\boldsymbol{\theta}} (\mathbf{U}) \right]
\\
& =    
- \H_{\boldsymbol{\phi}, \boldsymbol{\theta}} \left( \mathbf{U} \! \mid  \! \mathbf{Z} \right) 
- \D_{\mathrm{KL}} \left( P_{\mathbf{U}} \, \Vert \, P_{\boldsymbol{\theta}} (\mathbf{U}) \right)
+ \H \left( P_{\mathbf{U}} \, \Vert \, P_{\boldsymbol{\theta}} (\mathbf{U}) \right) 
  \\
& \geq 
 \underbrace{- \H_{\boldsymbol{\phi}, \boldsymbol{\theta}} \left( \mathbf{U} \! \mid  \! \mathbf{Z} \right)}_{\mathrm{Prediction~Fidelity}}
\,\,\,  - \!\!\! \underbrace{\D_{\mathrm{KL}} \left( P_{\mathbf{U}} \, \Vert \, P_{\boldsymbol{\theta}} (\mathbf{U}) \right)}_{\mathrm{Distribution~Discrepancy~Loss}} \!\! \!\!  \eqqcolon \,\, 
\I_{\boldsymbol{\phi}, \boldsymbol{\theta}}^{\mathrm{L}} \! \left( \mathbf{U}; \mathbf{Z} \right) .
\end{align}
\end{subequations}
This decomposition will lead us to a \textit{unified} formulation for both the \textit{supervised} and \textit{unsupervised} DVCLUB objectives.

When the utility task is to \textit{reconstruct} (generate) the original data $\mathbf{X}$, i.e., $P_{\boldsymbol{\theta}} (\mathbf{X} \! \mid \! \mathbf{Z}) = P_{\boldsymbol{\theta}} (\mathbf{U} \! \mid \! \mathbf{Z})$, let us denote the generated data distribution as $P_{\boldsymbol{\theta}} (\mathbf{X}) \! \coloneqq   \! \mathbb{E}_{Q_{\boldsymbol{\psi}}(\mathbf{Z})} \left[ P_{\boldsymbol{\theta}} (\mathbf{X} \! \mid  \! \mathbf{Z} ) \right]$. 
The parameterized variational approximation associated to the information utility $\I (\mathbf{U}; \mathbf{Z}) = \I (\mathbf{X}; \mathbf{Z})$ can be defined as:
\begin{subequations}\label{Eq:I_XZ_phi_theta}
\begin{align}
\I_{\boldsymbol{\phi}, \boldsymbol{\theta}} \! \left( \mathbf{X}; \mathbf{Z} \right) 
& \coloneqq  
 \mathbb{E}_{P_{\mathsf{D}} (\mathbf{X})} \left[ \mathbb{E}_{P_{\boldsymbol{\phi}} \left( \mathbf{Z} \mid \mathbf{X} \right) } \Big[ \log \frac{P_{\boldsymbol{\theta}} \! \left( \mathbf{X} \! \mid \! \mathbf{Z} \right) }{P_{\mathsf{D}} (\mathbf{X})}  \Big] \right]   \\
&= 
\mathbb{E}_{P_{\mathsf{D}} (\mathbf{X})} \left[  \mathbb{E}_{P_{\boldsymbol{\phi}} \left( \mathbf{Z} \mid \mathbf{X} \right) } \left[ \log P_{\boldsymbol{\theta}} \! \left( \mathbf{X} \! \mid \! \mathbf{Z} \right) \right] \right]
- \mathbb{E}_{P_{\mathsf{D}} (\mathbf{X})} \Big[ \log \frac{P_{\mathsf{D}} (\mathbf{X})}{P_{\boldsymbol{\theta}} (\mathbf{X})}\Big]
+ \mathbb{E}_{P_{\mathsf{D}} (\mathbf{X})} \left[ \log P_{\boldsymbol{\theta}} (\mathbf{X}) \right]
\\
& =    
\mathbb{E}_{P_{\mathsf{D}} (\mathbf{X})} \left[  \mathbb{E}_{P_{\boldsymbol{\phi}} \left( \mathbf{Z} \mid \mathbf{X} \right) } \left[ \log P_{\boldsymbol{\theta}} \! \left( \mathbf{X} \! \mid \! \mathbf{Z} \right) \right] \right] 
- \D_{\mathrm{KL}} \left( P_{\mathsf{D}} (\mathbf{X}) \, \Vert \, P_{\boldsymbol{\theta}} (\mathbf{X}) \right)
+ \H \left( P_{\mathsf{D}} (\mathbf{X}) \, \Vert \, P_{\boldsymbol{\theta}} (\mathbf{X}) \right) 
  \\
& \geq 
  \underbrace{- \H_{\boldsymbol{\phi}, \boldsymbol{\theta}} \left( \mathbf{X} \! \mid  \! \mathbf{Z} \right)}_{\mathrm{Reconstruction~Fidelity}}
\,\, - \!\! \underbrace{\D_{\mathrm{KL}} \left( P_{\mathsf{D}} (\mathbf{X}) \, \Vert \, P_{\boldsymbol{\theta}} (\mathbf{X}) \right)}_{\mathrm{Distribution~Discrepancy~Loss}}
\!\!\! \eqqcolon \,\,\, \I_{\boldsymbol{\phi}, \boldsymbol{\theta}}^{\mathrm{L}} \! \left( \mathbf{X}; \mathbf{Z} \right),
\end{align}
\end{subequations}
where $\H_{\boldsymbol{\phi}, \boldsymbol{\theta}} \left( \mathbf{X} \! \mid  \! \mathbf{Z} \right) \coloneqq \mathbb{E}_{P_{\mathsf{D}} (\mathbf{X})} \left[  \mathbb{E}_{P_{\boldsymbol{\phi}} \left( \mathbf{Z} \mid \mathbf{X} \right) } \left[ \log P_{\boldsymbol{\theta}} \! \left( \mathbf{X} \! \mid \! \mathbf{Z} \right) \right] \right] = \H \left( P_{ \mathbf{X}  \mid   \mathbf{Z}} \, \Vert \, P_{\boldsymbol{\theta}} \! \left( \mathbf{X} \! \mid \! \mathbf{Z} \right)  \mid P_{\boldsymbol{\phi}} (\mathbf{Z}) \right)$. The information utility $\I_{\boldsymbol{\phi}, \boldsymbol{\theta}} \! \left( \mathbf{X}; \mathbf{Z} \right) $ measures how perceptually similar the original data $\mathbf{X}$ and reconstructed data $\mathbf{\widehat{X}}$ through the bottleneck representation $\mathbf{Z}$.  
%Note that information utility \eqref{Eq:I_XZ_phi_theta} depends on the choice of generating distribution $P_{\boldsymbol{\theta}} \! \left( \mathbf{X} \! \mid \! \mathbf{Z} \right)$, while the information complexity \eqref{Eq:I_XZ_phi} does not. 

\noindent
\textbf{Parameterized Information Leakage}: 
Let $P_{\boldsymbol{\xi}} (\mathbf{S} \! \mid \! \mathbf{Z})$ denote the corresponding family of decoding probability distribution $Q_{\mathbf{S} \mid \mathbf{Z}}$, where $Q_{\mathbf{S} \mid \mathbf{Z}} : \mathcal{Z} \rightarrow \mathcal{P} (\mathcal{S})$ is a variational approximation of optimal decoder distribution $P_{\mathbf{S} \mid \mathbf{Z}}$. 
Similarly to \eqref{Eq:I_UZ_phi_theta_SecondDecomposition}, one can recast the parameterized variational approximation associated with the information leakage:\vspace{-6pt}
\begin{eqnarray}\label{Eq:I_XZ_phi_xi}
\I_{\boldsymbol{\phi}, \boldsymbol{\xi}} (\mathbf{S}; \mathbf{Z})
\geq 
 - \H_{\boldsymbol{\phi}, \boldsymbol{\xi}} \left( \mathbf{S} \! \mid  \! \mathbf{Z} \right) 
- \D_{\mathrm{KL}} \left( P_{\mathbf{S}} \, \Vert \, P_{\boldsymbol{\xi}} (\mathbf{S}) \right) \eqqcolon
\I_{\boldsymbol{\phi}, \boldsymbol{\xi}}^{\mathrm{L}} \! \left( \mathbf{S}; \mathbf{Z} \right) .
\end{eqnarray}

On the other hand, the parameterized variational approximation of conditional entropy \eqref{ConditionalEntropy_X_given_SZ} can be defined as:\vspace{-6pt}
\begin{eqnarray}\label{Eq:H_XgivenSZ_phi_varphi}
\H_{\boldsymbol{\phi}, \boldsymbol{\varphi}} \! \left( \mathbf{X} \! \mid \! \mathbf{S}, \mathbf{Z} \right) 
& \coloneqq & \!\!  - \;  \mathbb{E}_{P_{\mathbf{S}, \mathbf{X}}} \left[  \mathbb{E}_{P_{\boldsymbol{\phi}} (\mathbf{Z} \mid \mathbf{X})} \left[ \log P_{\boldsymbol{\varphi}}  (\mathbf{X} \! \mid \! \mathbf{S}, \mathbf{Z}) \right] \right] 
- \D_{\mathrm{KL}} \left( P_{\mathbf{X}\mid \mathbf{S}, \mathbf{Z} } \, \Vert \, P_{\boldsymbol{\varphi}}\!  \left( \mathbf{X} \! \mid \! \mathbf{S}, \mathbf{Z} \right) \right) \nonumber \\
& \leq &  \!\!  - \; \mathbb{E}_{P_{\mathbf{S}, \mathbf{X}}} \left[  \mathbb{E}_{P_{\boldsymbol{\phi}} (\mathbf{Z} \mid \mathbf{X})} \left[ \log P_{\boldsymbol{\varphi}}  (\mathbf{X} \! \mid \! \mathbf{S}, \mathbf{Z}) \right] \right] \eqqcolon \H_{\boldsymbol{\phi}, \boldsymbol{\varphi}}^{\mathrm{U}} \! \left( \mathbf{X} \! \mid \! \mathbf{S}, \mathbf{Z} \right) . 
\end{eqnarray}

%----------------------------------------------------------
%      Figure:  Deep Variational CLUB  >> P1, P2, P3, P4
%----------------------------------------------------------
%
\begin{figure}[t!]
    \centering
   %  \hspace{-9pt} 
        \begin{subfigure}[h]{0.485\textwidth}
        \includegraphics[scale=0.32]{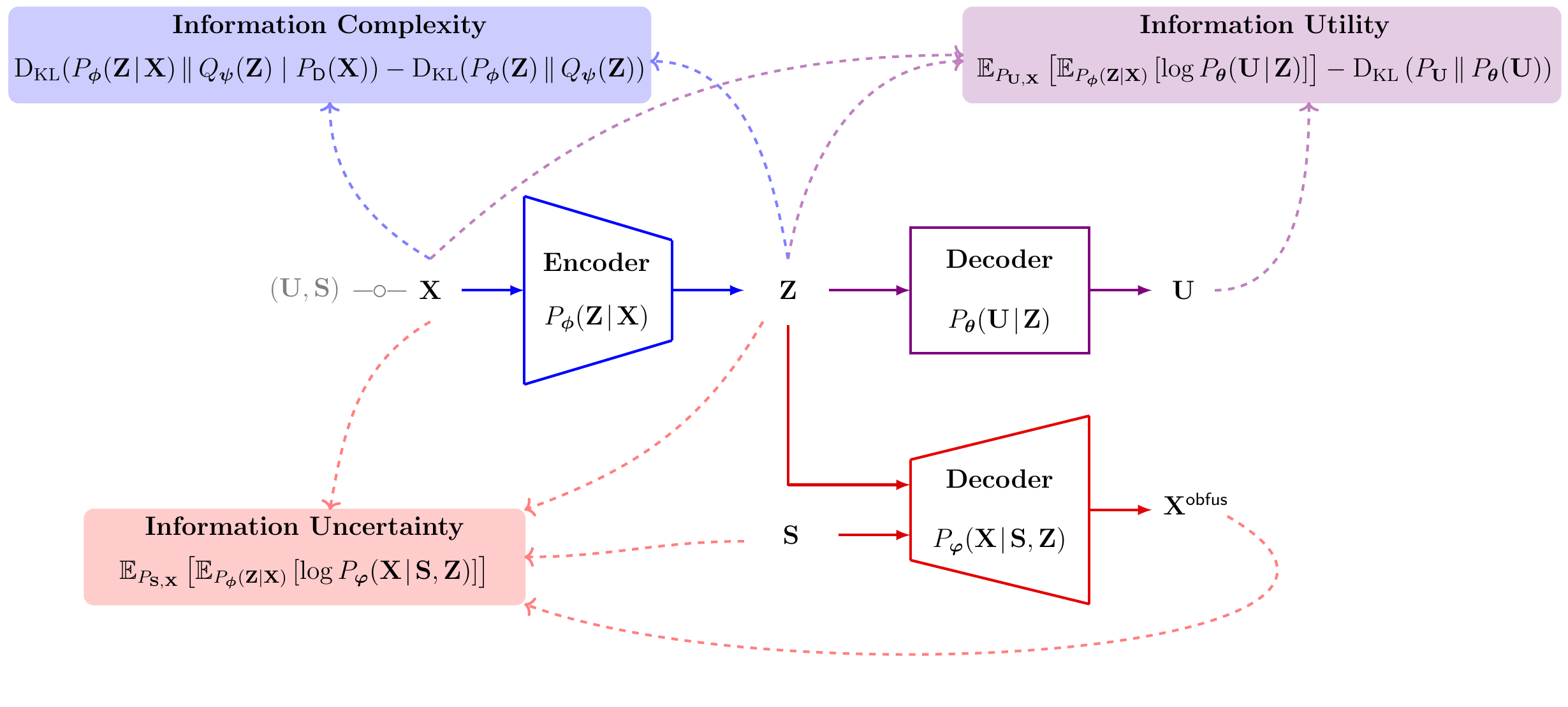}%
 %\includegraphics[width=cm, height=3.3cm]{TeXFig/GeneralDiag2.pdf}
%   \begin{subfigure}[h]{0.48\textwidth}
   %     \includegraphics[width=6cm, height=3.3cm]{TeXFig/GeneralDiag.pdf}%
           \vspace{-5pt}
        \caption{}
        %   \vspace{-10pt}
        \label{fig:DeepVariationalCLUB_Supervised}
    \end{subfigure}%
~
       \begin{subfigure}[h]{0.485\textwidth}
     %  \vspace{5pt}
        \includegraphics[scale=0.32]{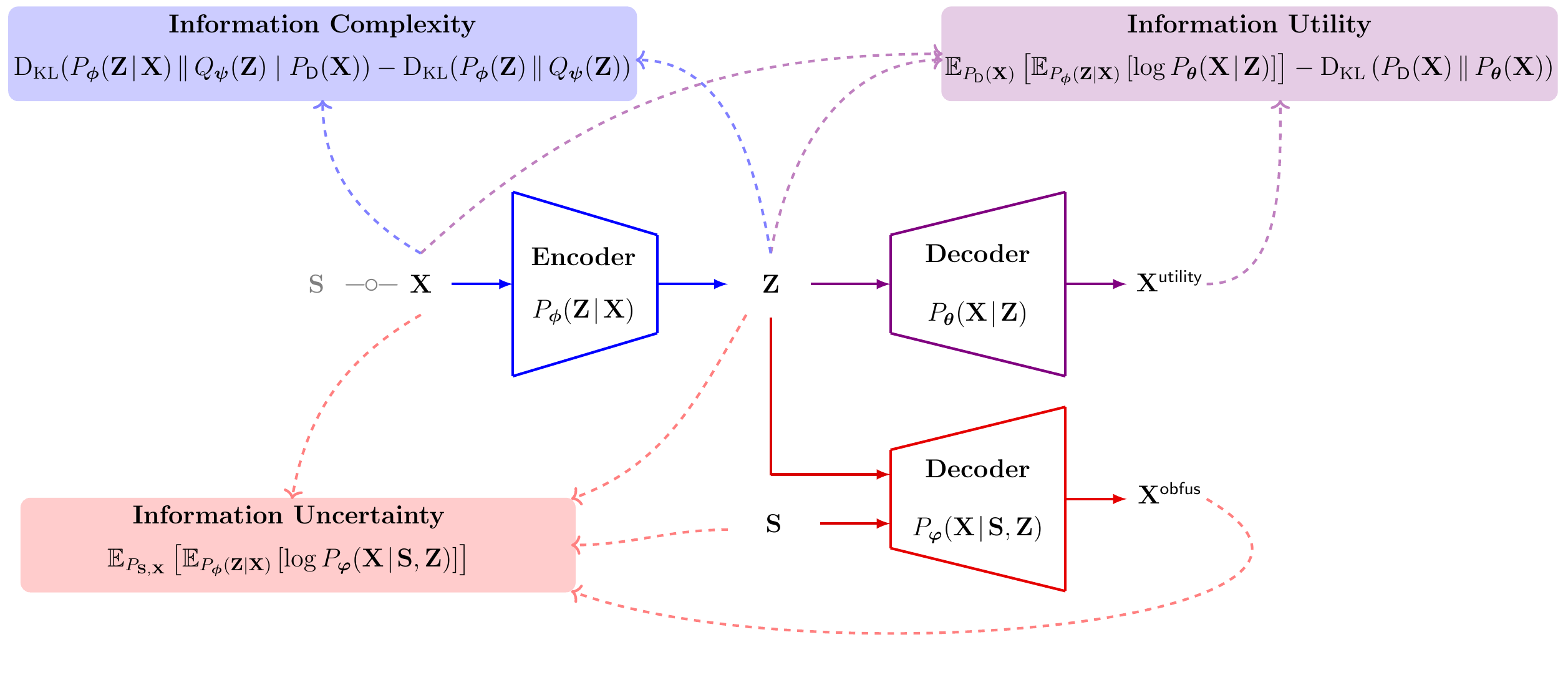}%
%     \begin{subfigure}[h]{0.48\textwidth}
  %   \includegraphics[width=6cm, height=3.3cm]{TeXFig/RegionOverlap.pdf}%
%\includegraphics[scale=0.2]{pdfFig/DisPreQuerySide.pdf}%
        \vspace{-5pt}
        \caption{}
       %    \vspace{-10pt}
        \label{fig:DeepVariationalCLUB_UnSupervised}
    \end{subfigure}
    \vspace{-6pt}
    \caption{Block diagram of the DVCLUB associated with $(\text{P1})$ and $(\text{P2})$ in the (a) supervised setup and (b) unsupervised setup.}
    \vspace{-8pt}
    \label{Fig:DeepVariationalCLUB_P1_P2_P3_P4}
%   \vspace{-13pt}
\end{figure}
%---------------------------------------------------
%---------------------------------------------------

%---------------------------------------------------
%	    Figure: Supervised Deep Variational CLUB
%---------------------------------------------------
%
%\begin{figure}[!t]
%\centering
%\includegraphics[scale=0.5]{TeXFig/DeepVariationalCLUB_Supervised.pdf}
%\vspace{-15pt}
%\caption{Block diagram of the Deep Variational Complexity-Utility-Leakage Bottleneck in `supervised' setup.}
%\label{Fig:DeepVariationalCLUB_Supervised}
%\vspace{-10pt}
%\end{figure} 
%---------------------------------------------------
%---------------------------------------------------

%\begin{remark}[Parametrized Variational Lower and Upper Bounds on Information Leakage]
%
Consider the lower bound \eqref{I_SZ_LowerBound} and upper bound \eqref{I_SZ_upperBound} on information leakage $\I \left( \mathbf{S}; \mathbf{Z} \right)$. Using \eqref{Eq:I_XZ_phi_xi} and \eqref{Eq:H_XgivenSZ_phi_varphi}, the lower and upper bounds on $\I_{\boldsymbol{\phi}, \boldsymbol{\xi}} (\mathbf{S}; \mathbf{Z})$ are given as:\vspace{-7pt}
\begin{multline}\label{Eq:I_SZ_phi_Xi_LowerUpperBounds}
\I_{\boldsymbol{\phi}, \boldsymbol{\xi}}^{\mathrm{L}} \! \left( \mathbf{S}; \mathbf{Z} \right) \!  \coloneqq \! - \H_{\boldsymbol{\phi}, \boldsymbol{\xi}} \left( \mathbf{S} \! \mid  \! \mathbf{Z} \right) 
 -  \D_{\mathrm{KL}} \left( P_{\mathbf{S}} \, \Vert \, P_{\boldsymbol{\xi}} (\mathbf{S}) \right)\\%
\hspace{-40pt} \leq \; \I_{\boldsymbol{\phi}, \boldsymbol{\xi}} (\mathbf{S}; \mathbf{Z})\; \leq \; \\
\I_{\boldsymbol{\phi}, \boldsymbol{\psi}} \! \left( \mathbf{X}; \mathbf{Z} \right) +  \H_{\boldsymbol{\phi}, \boldsymbol{\varphi}}^{\mathrm{U}} \! \left( \mathbf{X} \! \mid \! \mathbf{S}, \mathbf{Z} \right) + \mathrm{c} \eqqcolon \I_{\boldsymbol{\phi}, \boldsymbol{\psi}, \boldsymbol{\varphi}}^{\mathrm{U}}   \left( \mathbf{S}; \mathbf{Z} \right) + \mathrm{c} ,
\end{multline}
where $\mathrm{c}$ is a constant term, independent of the neural network parameters. 
%
%\end{remark}
The above lower and upper bounds lead us towards two alternative models. The upper bound in \eqref{Eq:I_SZ_phi_Xi_LowerUpperBounds} encourages the model to directly minimize the information complexity $\I_{\boldsymbol{\phi}, \boldsymbol{\psi}} \! \left( \mathbf{X}; \mathbf{Z} \right)$ as well as the information uncertainty $\H_{\boldsymbol{\phi}, \boldsymbol{\varphi}}^{\mathrm{U}} \! \left( \mathbf{X} \! \mid \! \mathbf{S}, \mathbf{Z} \right)$. By minimizing the information uncertainty $\H_{\boldsymbol{\phi}, \boldsymbol{\varphi}}^{\mathrm{U}} \! \left( \mathbf{X} \! \mid \! \mathbf{S}, \mathbf{Z} \right)$ the model forces to forget the sensitive attribute $\mathbf{S}$ at the expense of reducing the uncertainty about the original data $\mathbf{X}$, i.e., encourages the model to reconstruct the original data $\mathbf{X}$. 
In contrast, the lower bound in \eqref{Eq:I_SZ_phi_Xi_LowerUpperBounds} encourages the model to maximizes (i) uncertainty about the sensitive attribute $\mathbf{S}$ given the released representation $\mathbf{Z}$, i.e., $\H_{\boldsymbol{\phi}, \boldsymbol{\xi}} \left( \mathbf{S} \! \mid  \! \mathbf{Z} \right)$, as well as (ii) the distribution discrepancy measure $\D_{\mathrm{KL}} \left( P_{\mathbf{S}} \, \Vert \, P_{\boldsymbol{\xi}} (\mathbf{S}) \right)$. 
Note that minimizing the lower bound on information leakage may not necessarily minimize the average maximal possible leakage. Furthermore, although the lower bound in \eqref{Eq:I_SZ_phi_Xi_LowerUpperBounds} does not explicitly depend on the information complexity $\I_{\boldsymbol{\phi}, \boldsymbol{\psi}} \! \left( \mathbf{X}; \mathbf{Z} \right)$, it depends implicitly through the encoder $f_{\boldsymbol{\phi}}$.

% \noindent
% \textbf{DVCLUB Objectives}:
\subsection{DVCLUB Objectives}

Considering \eqref{CLUB_vs_VCLUB_1} and \eqref{CLUB_vs_VCLUB_2}, and using the addressed parameterized approximations, we have:
\begin{eqnarray}\label{Eq:VCLUB_vs_DVCLUB_max}
\mathop{\max}_{P_{\mathbf{Z} \mid \mathbf{X}}}  \; \mathop{\max}_{Q_{\mathbf{U}\mid \mathbf{Z}}, Q_{\mathbf{Z}}, Q_{\mathbf{X}\mid \mathbf{S}, \mathbf{Z}}}\;
\mathcal{L}_{\mathrm{VCLUB}}^{\mathsf{S/U}} \left( P_{\mathbf{Z}\mid \mathbf{X}}, Q_{\mathbf{U}\mid \mathbf{Z}}, Q_{\mathbf{Z}}, Q_{\mathbf{X}\mid \mathbf{S}, \mathbf{Z}}, \beta , \alpha  \right)  \geq   \mathop{\max}_{\boldsymbol{\phi}, \boldsymbol{\theta}, \boldsymbol{\psi}, \boldsymbol{\varphi}}  \mathcal{L}_{\mathrm{DVCLUB}}^{\mathsf{S/U}} \left( \boldsymbol{\phi}, \boldsymbol{\theta},  \boldsymbol{\psi} ,\boldsymbol{\varphi}, \beta, \alpha \right)\!, 
\end{eqnarray}
or alternatively, we have:
\begin{eqnarray}\label{Eq:VCLUB_vs_DVCLUB_min}
\mathop{\min}_{P_{\mathbf{Z} \mid \mathbf{X}}}  \; \mathop{\min}_{Q_{\mathbf{U}\mid \mathbf{Z}}, Q_{\mathbf{Z}}, Q_{\mathbf{X}\mid \mathbf{S}, \mathbf{Z}}}\;
\mathcal{L}_{\mathrm{VCLUB}}^{\mathsf{S/U}} \left( P_{\mathbf{Z}\mid \mathbf{X}}, Q_{\mathbf{U}\mid \mathbf{Z}}, Q_{\mathbf{Z}}, Q_{\mathbf{X}\mid \mathbf{S}, \mathbf{Z}}, \gamma , \lambda  \right)  \leq   \mathop{\min}_{\boldsymbol{\phi}, \boldsymbol{\theta}, \boldsymbol{\psi}, \boldsymbol{\varphi}}  \mathcal{L}_{\mathrm{DVCLUB}}^{\mathsf{S/U}} \left( \boldsymbol{\phi}, \boldsymbol{\theta},  \boldsymbol{\psi} ,\boldsymbol{\varphi}, \gamma, \lambda \right)\!, 
\end{eqnarray}
where $\mathcal{L}_{\mathrm{DVCLUB}}^{\mathsf{S/U}} \left( \boldsymbol{\phi}, \boldsymbol{\theta}, \boldsymbol{\psi} , \boldsymbol{\varphi}, \beta, \alpha \right) $ and $\mathcal{L}_{\mathrm{DVCLUB}}^{\mathsf{S/U}} \left( \boldsymbol{\phi}, \boldsymbol{\theta},  \boldsymbol{\psi} ,\boldsymbol{\varphi}, \gamma, \lambda \right)$ denote the associated \textit{Deep Variational CLUB (DVCLUB)} Lagrangian functionals in the `\textit{supervised}' or `\textit{unsupervised}' scenarios, respectively, which are given as:\vspace{-4pt}
\begin{multline}\label{Eq:DVCLUB_Supervised}
\!\!(\text{P1}\!\!:\!  \mathsf{S})\!:\;  \mathcal{L}_{\mathrm{DVCLUB}}^{\mathsf{S}} \left( \boldsymbol{\phi}, \boldsymbol{\theta}, \boldsymbol{\psi} , \boldsymbol{\varphi}, \beta, \alpha \right)  \coloneqq 
\overbrace{ \mathbb{E}_{P_{\mathbf{U}, \mathbf{X}}} \left[ \mathbb{E}_{P_{\boldsymbol{\phi}} \left( \mathbf{Z} \mid \mathbf{X} \right) } \left[ \log P_{\boldsymbol{\theta}} \! \left( \mathbf{U} \! \mid \! \mathbf{Z} \right) \right] \right] - \D_{\mathrm{KL}} \left( P_{\mathbf{U}} \, \Vert \, P_{\boldsymbol{\theta}} (\mathbf{U}) \right) }^{\textcolor{violet}{\mathrm{Information~Utility:}~\I_{\boldsymbol{\phi}, \boldsymbol{\theta}}^{\mathrm{L}} \left( \mathbf{U}; \mathbf{Z} \right) }} \\
 \qquad \quad\qquad \quad \qquad \quad \qquad \; \; -  \left( \beta + \alpha \right) \!
\underbrace{ \Big( \D_{\mathrm{KL}} \! \left( P_{\boldsymbol{\phi}} (\mathbf{Z} \! \mid \! \mathbf{X}) \, \Vert \, Q_{\boldsymbol{\psi}}   (\mathbf{Z}) \mid P_{\mathsf{D}}(\mathbf{X})  \right)- 
 \D_{\mathrm{KL}} \! \left( P_{\boldsymbol{\phi}}(\mathbf{Z}) \, \Vert \, Q_{\boldsymbol{\psi}} (\mathbf{Z})\right) \! \Big) }_{\textcolor{blue}{\mathrm{Information~Complexity:}~\I_{\boldsymbol{\phi}, \boldsymbol{\psi}}   \left( \mathbf{X}; \mathbf{Z} \right)}} \\
 \; \; - \,   \alpha \, \underbrace{ \big( - \mathbb{E}_{P_{\mathbf{S}, \mathbf{X}}} \left[  
\mathbb{E}_{P_{\boldsymbol{\phi}} \left( \mathbf{Z}  \mid \mathbf{X} \right)} \left[ \log P_{\boldsymbol{\varphi}} \! \left( \mathbf{X} \! \mid \! \mathbf{S}, \mathbf{Z} \right) \right] \right] \big) }_{\textcolor{red}{\mathrm{Information~Uncertainty:}~\H_{\boldsymbol{\phi}, \boldsymbol{\varphi}}^{\mathrm{U}}  \left( \mathbf{X} \mid  \mathbf{S}, \mathbf{Z} \right) }} . 
\end{multline}

\begin{multline}\label{Eq:DVCLUB_UnSupervised}
\!\! (\text{P1}\!\!:\! \mathsf{U})\!:\;   \mathcal{L}_{\mathrm{DVCLUB}}^{\mathsf{U}} \left( \boldsymbol{\phi}, \boldsymbol{\theta}, \boldsymbol{\psi} , \boldsymbol{\varphi}, \beta, \alpha \right)  \coloneqq \overbrace{ \mathbb{E}_{P_{\mathsf{D}} (\mathbf{X})} \left[  \mathbb{E}_{P_{\boldsymbol{\phi}} \left( \mathbf{Z} \mid \mathbf{X} \right) } \left[ \log P_{\boldsymbol{\theta}} \! \left( \mathbf{X} \! \mid \! \mathbf{Z} \right) \right] \right] - \D_{\mathrm{KL}} \left( P_{\mathsf{D}} (\mathbf{X}) \, \Vert \, P_{\boldsymbol{\theta}} (\mathbf{X}) \right)
}^{\textcolor{violet}{\mathrm{Information~Utility:}~\I_{\boldsymbol{\phi}, \boldsymbol{\theta}}^{\mathrm{L}}  \left( \mathbf{X}; \mathbf{Z} \right) }}  \\
\qquad \qquad \quad \qquad \qquad \quad\quad    \; \; - \left( \beta + \alpha \right) \!
 \underbrace{ \Big( \! \D_{\mathrm{KL}} \! \left( P_{\boldsymbol{\phi}} (\mathbf{Z} \! \mid \! \mathbf{X}) \, \Vert \, Q_{\boldsymbol{\psi}}   (\mathbf{Z}) \mid P_{\mathsf{D}}(\mathbf{X}) \right)
-  \D_{\mathrm{KL}} \! \left( P_{\boldsymbol{\phi}}(\mathbf{Z}) \, \Vert \, Q_{\boldsymbol{\psi}} (\mathbf{Z})\right) \! \Big)  }_{\textcolor{blue}{\mathrm{Information~Complexity:}~\I_{\boldsymbol{\phi}, \boldsymbol{\psi}} \left( \mathbf{X}; \mathbf{Z} \right)}}   \\
  \; \; -    \alpha \; \underbrace{ \big( - \mathbb{E}_{P_{\mathbf{S}, \mathbf{X}}} \left[  
\mathbb{E}_{P_{\boldsymbol{\phi}} \left( \mathbf{Z}  \mid \mathbf{X} \right)} \left[ \log P_{\boldsymbol{\varphi}} \! \left( \mathbf{X} \! \mid \! \mathbf{S}, \mathbf{Z} \right) \right] \right] \big) }_{\textcolor{red}{\mathrm{Information~Uncertainty:}~\H_{\boldsymbol{\phi}, \boldsymbol{\varphi}}^{\mathrm{U}}  \left( \mathbf{X}  \mid  \mathbf{S}, \mathbf{Z} \right) }}\!  . \!
\end{multline}
Alternatively, one can obtain the DVCLUB Lagrangian functionals associated to \eqref{CLUB_vs_VCLUB_2} in both the supervised and unsupervised scenarios. 
We will discuss the alternative objectives in Sec.~\ref{Sec:Experiments}. 
% Fig.~\ref{Fig:DeepVariationalCLUB_P1_P2_P3_P4} and Fig.~\ref{Fig:DeepVariationalCLUB_P5_P6} demonstrate the general block diagram for the supervised and unsupervised DVCLUB models. 
Fig.~\ref{Fig:DeepVariationalCLUB_P1_P2_P3_P4} demonstrates the general block diagram for the supervised and unsupervised DVCLUB models. 
In practice, we need to train the model using alternating block coordinate descent algorithms. 
In the following, we gradually build the fundamentals of the learning model.% and establish its connections to other generative models.

\subsection{Learning Algorithm}
\label{Ssec:Algorithm}

Given a collection of i.i.d. training samples $\{ \left( \mathbf{u}_n, \mathbf{s}_n, \mathbf{x}_n  \right) \}_{n=1}^{N} \subseteq \mathcal{U} \times \mathcal{S} \times \mathcal{X}$, and using the stochastic gradient descent (SGD)-type algorithms, the deep neural networks $f_{\boldsymbol{\phi}}$, $g_{\boldsymbol{\theta}}$, and $g_{\boldsymbol{\varphi}}$ (or $g_{\boldsymbol{\xi}}$) can be trained jointly to maximize a Monte-Carlo approximation of the DVCLUB functionals over parameters $\boldsymbol{\phi}$, $\boldsymbol{\theta}$, $\boldsymbol{\varphi}$, and $\boldsymbol{\xi}$. 
In order to have a stable gradient with respect to the encoder, the reparameterization trick \cite{kingma2014auto} is used to sample from the \textit{learned posterior} distribution $P_{\boldsymbol{\phi}} (\mathbf{Z} \! \mid \! \mathbf{X})$. To do this, we need to \textit{explicitly} consider $P_{\boldsymbol{\phi}} (\mathbf{Z} \! \mid \! \mathbf{X})$ to belong to a tractable parametric family of distributions (e.g., Gaussian distributions) such that we are able to sample from $P_{\boldsymbol{\phi}} (\mathbf{Z} \!\! \mid \!\! \mathbf{X})$ by: 
(i) sampling a random vector $\boldsymbol{\mathcal{E}}$ with distribution $P_{\boldsymbol{\mathcal{E}}} (\boldsymbol{\varepsilon)}$, $\boldsymbol{\varepsilon} \in \mathcal{E}$, which does not depend on $\boldsymbol{\phi}$\footnote{Hence, does not impact differentiation of the network.}; 
(ii) transforming the samples using some parametric function $f_{\boldsymbol{\phi}} : \mathcal{X} \times \mathcal{E} \rightarrow \mathcal{Z}$, such that $\mathbf{Z} = f_{\boldsymbol{\phi}} \left( \mathbf{x}, \boldsymbol{\mathcal{E}} \right) \sim P_{\boldsymbol{\phi}} \! \left( \mathbf{Z} \mid \mathbf{X} = \mathbf{x} \right)$. 
One can consider multivariate Gaussian parametric encoders of mean $\boldsymbol{\mu}_{\boldsymbol{\phi}} (\mathbf{x})$, and co-variance $\boldsymbol{\Sigma}_{\boldsymbol{\phi}} (\mathbf{x})$, i.e., $P_{\boldsymbol{\phi}} (\mathbf{Z} \! \mid \! \mathbf{x}) = \mathcal{N} \left( \boldsymbol{\mu}_{\boldsymbol{\phi}} (\mathbf{x}), \boldsymbol{\Sigma}_{\boldsymbol{\phi}} (\mathbf{x}) \right)$, where $\boldsymbol{\mu}_{\boldsymbol{\phi}} (\mathbf{x})$ and $\boldsymbol{\Sigma}_{\boldsymbol{\phi}} (\mathbf{x})$ are the two output vectors of the network $f_{\boldsymbol{\phi}}$ for the given input sample $\mathbf{x}$. 
The inferred posterior distribution is typically a multi-variate Gaussian with diagonal co-variance, i.e., $P_{\boldsymbol{\phi}} (\mathbf{Z} \! \mid \! \mathbf{x}) = \mathcal{N} \big( \boldsymbol{\mu}_{\boldsymbol{\phi}} (\mathbf{x}), \mathsf{diag} ( \boldsymbol{\sigma}_{\boldsymbol{\phi}} (\mathbf{x}) ) \big)$. 
Suppose $\mathcal{Z} = \mathbb{R}^{d_{\mathbf{z}}}$, therefore, we first sample a random variable $\boldsymbol{\mathcal{E}}$ i.i.d. from $\mathcal{N} \! \left( \boldsymbol{0}, \mathbf{I}_{d_{\mathbf{z}}}\right)$, then given a data sample $\mathbf{x} \in \mathcal{X}$, we generate the sample $\mathbf{z} = \boldsymbol{\mu}_{\boldsymbol{\phi}} (\mathbf{x}) + \boldsymbol{\sigma}_{\boldsymbol{\phi}} (\mathbf{x}) \odot \boldsymbol{\varepsilon}$, where $\odot$ is the element-wise (Hadamard) product. 
Noting that $f_{\boldsymbol{\phi}}$ is a deterministic mapping, the stochasticity of encoder $P_{\boldsymbol{\phi}} \! \left( \mathbf{Z} \! \mid \! \mathbf{X} \right)$ is relegated to the random variable $\boldsymbol{\mathcal{E}}$. 
%In this case, one can interpret the stochastic encoder $P_{\boldsymbol{\psi}} \left( \mathbf{Z} \mid \mathbf{X} \right)$,
%
The decoder $P_{\boldsymbol{\theta}} (\mathbf{U} \! \! \mid \! \! \mathbf{Z})$ (likewise $P_{\boldsymbol{\theta}} (\mathbf{X} \! \! \mid \! \! \mathbf{Z})$) utilizes a suitable distribution for the data and task under consideration.  
%

%To approximate the cost function \eqref{Eq:DVCLUB_Supervised}, for each training sample $\left( \mathbf{u}_n, \mathbf{s}_n, \mathbf{x}_n  \right), n \in \left[ N \right]$, we first draw $M$ independent samples $\{ \boldsymbol{\varepsilon}_m \}_{m=1}^M$ i.i.d. from $P_{\boldsymbol{\mathcal{E}}} (\boldsymbol{\varepsilon)}$, which give us $M$ independent samples $\mathbf{z}_{n, m} = f_{\boldsymbol{\phi}} \left( \mathbf{x}_n, \boldsymbol{\varepsilon}_m \right)$. Hence, for the $n$-th training sample the empirical supervised DVCLUB functional  $\mathcal{L}_{\mathrm{DVCLUB}_{\mathsf{emp}_{(n)}}}^{\mathsf{S}} \!\!\! \left( \boldsymbol{\phi}, \boldsymbol{\theta}, \boldsymbol{\psi} , \boldsymbol{\varphi}, \beta, \alpha \right) $ is computed. 
%Accordingly, we can maximize the emperical DVCLUB functional over the deep neural network parameters $\boldsymbol{\phi}, \boldsymbol{\theta}, \boldsymbol{\psi} , \boldsymbol{\varphi}$, as $\mathop{ \max}_{\boldsymbol{\phi}, \boldsymbol{\theta}, \boldsymbol{\psi} , \boldsymbol{\varphi}} \frac{1}{N} \sum_{n=1}^{N} \mathcal{L}_{\mathrm{DVCLUB}_{\mathsf{emp}_{(n)}}}^{\mathsf{S}} \!\! \left( \boldsymbol{\phi}, \boldsymbol{\theta}, \boldsymbol{\psi} , \boldsymbol{\varphi}, \beta, \alpha \right) $. Note that due to memory and time limitations, only 1-sample approximation is generally accomplished. 

%---------------------------------------------------
%	  Figure: Unsupervised DVCLUB >> Complete Diagram
%---------------------------------------------------
%
\begin{figure}[t!]
\centering
\includegraphics[height=5.5cm]{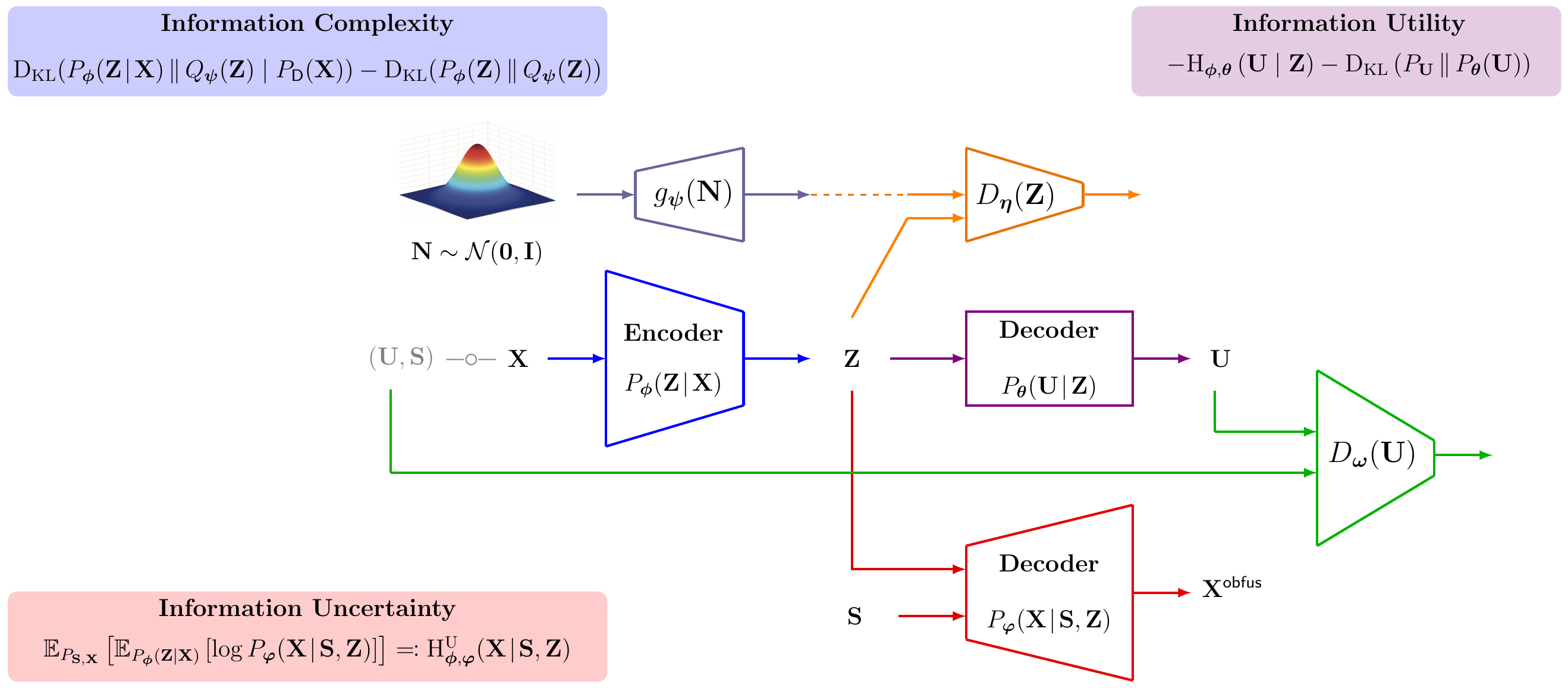}
%\vspace{-15pt}
\caption{Supervised DVCLUB training architecture associated with $(\text{P1})$.}%~and~$(\text{P2})$.

\label{Fig:DVCLUB_Supervised_complete}
\vspace{5pt}
%\vspace{-10pt} 
%\end{figure} 
%---------------------------------------------------
%---------------------------------------------------
%
%\vspace*{\floatsep}% https://tex.stackexchange.com/q/26521/5764
%
%---------------------------------------------------
%	  Figure: Unsupervised DVCLUB >> Complete Diagram
%---------------------------------------------------
%
%\begin{figure}[t]
%\centering
\includegraphics[height=5.5cm]{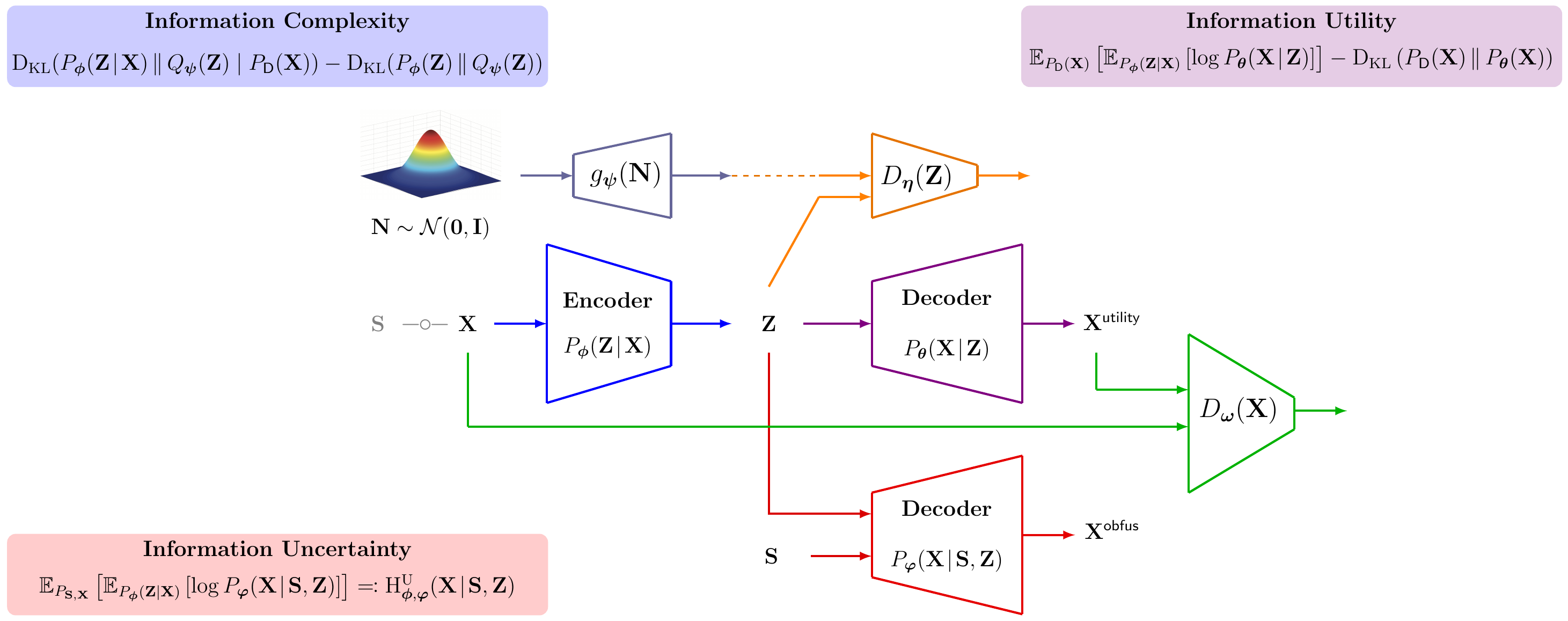}
%\vspace{-15pt}
\caption{Unsupervised DVCLUB training architecture associated with $(\text{P1})$.}%~and~$(\text{P2})$.}
\label{Fig:DVCLUB_UnSupervised_complete}
\vspace{-10pt} 
\end{figure} 
%---------------------------------------------------
%---------------------------------------------------

\sloppy The latent space prior distribution is typically considered as a \textit{fixed} $d$-dimensional standard isotropic multi-variate Gaussian, i.e., $Q_{\boldsymbol{\psi}}\! \left( \mathbf{Z} \right) = Q_{ \mathbf{Z}} = \mathcal{N} \! \left( \boldsymbol{0}, \mathbf{I}_d \right)$. 
For this simple \textit{explicit} choice, the information complexity upper bound $\mathbb{E}_{P_{\boldsymbol{\phi}} (\mathbf{X}, \mathbf{Z})} \big[ \log \frac{P_{\boldsymbol{\phi}} \left( \mathbf{Z}  \mid  \mathbf{X}\right) }{ Q_{ \mathbf{Z}}} \big] = 
\mathbb{E}_{P_{\mathsf{D}} (\mathbf{X})} \left[ \D_{\mathrm{KL}}  \left( P_{\boldsymbol{\phi}} \! \left( \mathbf{Z} \! \mid \! \mathbf{X}\right) \Vert Q_{ \mathbf{Z}}) \right) \right]$ has a closed-form expression for a given sample $\mathbf{x}$, which reads as $ 2 \, \D_{\mathrm{KL}}  \left( P_{\boldsymbol{\phi}} \! \left( \mathbf{Z} \! \mid \! \mathbf{X} = \mathbf{x}\right) \Vert Q_{ \mathbf{Z}} \right) = {\Vert \boldsymbol{\mu}_{\boldsymbol{\phi}} (\mathbf{x})  \Vert}_2^2 + d + \sum_{i=1}^{d} ( \boldsymbol{\sigma}_{\boldsymbol{\phi}} (\mathbf{x})_i - \log \boldsymbol{\sigma}_{\boldsymbol{\phi}} (\mathbf{x})_i)$. 
However, simple prior can lead to under-fitting, and, as a consequence, poor representations. 
On the other hand, having a complete match, i.e., $Q_{ \mathbf{Z}} \! = \! \frac{1}{N} \sum_{n=1}^{N} P_{\boldsymbol{\phi}} (\mathbf{Z} \! \mid \! \mathbf{x}_n)$, may potentially lead to over-fitting. Moreover, it is a computationally expensive task due to the summation over all training samples.  
One possible solution to overcome this issue is to explicitly consider a mixture of diagonal Gaussian distributions as the proposal prior \cite{dilokthanakul2016deep, nalisnick2016approximate}, i.e., $Q_{\boldsymbol{\psi}} (\mathbf{Z}) \! = \! \sum_{k=1}^{K}  \pi_k \, \mathcal{N} \big( \boldsymbol{\mu}_{{\boldsymbol{\psi}}_k} , \mathsf{diag} ( \boldsymbol{\sigma}_{{\boldsymbol{\psi}}_k} ) \big)$, $\sum_{k=1}^{K} \pi_k \! = \! 1$, where  $K \! \ll \! N$, and learn the mixture weights. In \cite{tomczak2018vae}, the authors considered the proposal prior as a mixture of equi-probable variational posteriors, such that $Q_{\boldsymbol{\psi}} (\mathbf{Z}) = \frac{1}{K} \sum_{k=1}^{K} P_{\boldsymbol{\phi}} (\mathbf{Z} \! \mid \! \widetilde{\mathbf{x}}_k)$, where $\{ \widetilde{\mathbf{x}}_k \}_{k=1}^{K}$, $K \ll N$, are referred to as the pseudo-inputs, which are learned through back-propagation. 
In the same line of research, in \cite{bauer2019resampled}, the authors constructed the proposal prior by multiplying a simple prior with a learned acceptance probability function, which re-weights the considered simple prior. 
Alternatively, one can adversarially learn the prior distribution $Q_{\boldsymbol{\psi}} (\mathbf{Z})$ through a generator model $g_{\boldsymbol{\psi}} (\mathbf{N})$, where $\mathbf{N} \sim \mathcal{N} \left( \boldsymbol{0}, \mathbf{I}\right)$. This choice gives us an \textit{implicit} prior distribution $Q_{\boldsymbol{\psi}} (\mathbf{Z})$. Implicit distributions are probability distributions that are learned via passing noise through a deterministic function which is parameterized by a neural network \cite{makhzani2015adversarial, mescheder2017adversarial, dumoulin2017adversarially, donahue2016adversarial, huszar2017variational}. This allows us easily sample from them and take derivatives of samples with respect to the model parameters.

\textbf{Divergence Estimation:} 
We can estimate the $\mathrm{KL}$-divergences in \eqref{Eq:DVCLUB_Supervised} and \eqref{Eq:DVCLUB_UnSupervised} using the \textit{density-ratio trick} \cite{nguyen2010estimating, sugiyama2012density}, utilized in the GAN framework to directly match the data distribution $P_{\mathbf{X}}$ and the marginal model distribution $P_{\boldsymbol{\theta}}(\mathbf{X})$. 
The trick is to express two distributions as conditional distributions, conditioned on a label $C \in \{ 0, 1 \}$, and reduce the task to binary classification. The key point is that we can estimate the KL-divergence, and indeed all the well-defined $f$-divergences, by estimating the \textit{ratio} of the two distributions without modeling each distribution explicitly. 
Consider $\D_{\mathrm{KL}} \! \left( P_{\boldsymbol{\phi}}(\mathbf{Z}) \, \Vert \, Q_{\boldsymbol{\psi}} (\mathbf{Z})\right)= \mathbb{E}_{ P_{\boldsymbol{\phi}}(\mathbf{Z})} \! \left[ \log \frac{ P_{\boldsymbol{\phi}}(\mathbf{Z})}{Q_{\boldsymbol{\psi}} (\mathbf{Z})} \right]$. Define $\rho_{\mathbf{Z}} (\mathbf{z} \mid c)$ as follows:
\begin{eqnarray}
    \rho_{\mathbf{Z}} (\mathbf{z} \! \mid \! c) = \left\{
    \begin{array}{ll}
    P_{\boldsymbol{\phi}}(\mathbf{Z})     &  ,~\mathrm{if}\quad c=1\\
    Q_{\boldsymbol{\psi}} (\mathbf{Z})    &  ,~\mathrm{if}\quad c=0
    \end{array}
\right. .
\end{eqnarray}
Suppose that a perfect binary classifier (discriminator) $D_{\boldsymbol{\eta}} (\mathbf{z})$, with parameters $\boldsymbol{\eta}$, is trained to associate label $c=1$ to samples from distribution $P_{\boldsymbol{\phi}}(\mathbf{Z}) $ and label $c=0$ to samples from $Q_{\boldsymbol{\psi}} (\mathbf{Z})$. Using the Bayes' rule and assuming that the marginal class probabilities are equal, i.e., $\rho (c=1) = \rho (c=0)$, the density ratio can be expressed as:
\begin{eqnarray}\label{Eq:DensityRatioTrick}
    \frac{P_{\boldsymbol{\phi}}(\mathbf{Z} = \mathbf{z})}{Q_{\boldsymbol{\psi}} (\mathbf{Z} = \mathbf{z})} = 
    \frac{\rho_{\mathbf{Z}} (\mathbf{z} \mid c=1)}{\rho_{\mathbf{Z}} (\mathbf{z} \mid c=0)} =
    \frac{\rho_{\mathbf{Z}} (c=1 \mid \mathbf{z} )}{\rho_{\mathbf{Z}} ( c=0 \mid \mathbf{z})} \approx \frac{D_{\boldsymbol{\eta}} (\mathbf{z})}{1 - D_{\boldsymbol{\eta}} (\mathbf{z})}. 
\end{eqnarray}
Therefore, given a trained discriminator $D_{\boldsymbol{\eta}} (\mathbf{z})$ and $M$ i.i.d. samples $\{ \mathbf{z}_m \}_{m=1}^{M}$ from $P_{\boldsymbol{\phi}}(\mathbf{Z})$, one can estimate the divergence $\D_{\mathrm{KL}} \! \left( P_{\boldsymbol{\phi}}(\mathbf{Z}) \, \Vert \, Q_{\boldsymbol{\psi}} (\mathbf{Z})\right)$ as:
\begin{equation}\label{Eq:D_KL_Z_estimation}
    \D_{\mathrm{KL}} \! \left( P_{\boldsymbol{\phi}}(\mathbf{Z}) \, \Vert \, Q_{\boldsymbol{\psi}} (\mathbf{Z})\right) \approx \frac{1}{M} \sum_{m=1}^{M} \log \frac{D_{\boldsymbol{\eta}} (\mathbf{z}_m)}{1 - D_{\boldsymbol{\eta}} (\mathbf{z}_m)}. 
\end{equation}
The density ratio trick opens the door to \textit{implicit} prior and posterior distributions. 
%The implicit distributions are probability distributions that are learned via passing noise through a deterministic function which is parametrized by a neural network \cite{makhzani2015adversarial, mescheder2017adversarial, dumoulin2017adversarially, donahue2016adversarial}. 
The implicit generative models provide likelihood-free inference models. 
Interestingly, this trick allows us to \textit{learn} the parameterized prior distribution $Q_{\boldsymbol{\psi}} (\mathbf{Z})$ through a generator model.  

Given the discriminator (parameterized scoring function) $D_{\boldsymbol{\eta}} (\mathbf{z}) \approx \rho_{\mathbf{Z}} (c = 1 \mid \mathbf{z}) = P_{\boldsymbol{\phi}} (\mathbf{Z} = \mathbf{z})$, we now need to specify a proper scoring rule for binary discrimination to allow us for parameter learning. Binary cross-entropy loss is typically considered to this end. In this case, the \textit{latent space discriminator} $D_{\boldsymbol{\eta}} (\mathbf{z})$ minimizes the following loss function:%\vspace{-4pt}
\begin{subequations}\label{Eq:Discriminator_Z_loss}
\begin{align}
\mathcal{L}_{\mathrm{disc}}^{\mathsf{z}} (\boldsymbol{\eta}, \boldsymbol{\phi})  & \coloneqq   \mathbb{E}_{\rho_{\mathbf{Z}} (\mathbf{z} \mid c) \rho(c)} \left[ \, - c \,  \log D_{\boldsymbol{\eta}} (\mathbf{Z}) - (1 - c) \, \log (1 - D_{\boldsymbol{\eta}} (\mathbf{Z}))\, \right]  \\
  & =   \mathbb{E}_{P_{\mathsf{D}}(\mathbf{X})} \left[ \, \mathbb{E}_{P_{\boldsymbol{\phi}}(\mathbf{Z} \mid \mathbf{X})} \left[ - \log D_{\boldsymbol{\eta}} (\mathbf{Z}) \right] \, \right] + \mathbb{E}_{Q_{\boldsymbol{\psi}} (\mathbf{Z})} \left[ \, - \log (1 - D_{\boldsymbol{\eta}} (\mathbf{Z})) \, \right].
\end{align}
\end{subequations}

% %---------------------------------------------------
% %	  Figure: Unsupervised DVCLUB >> Complete Diagram
% %---------------------------------------------------
% %
% \begin{figure}[!t]
% \centering
% \includegraphics[height=5.8cm]{TeXFig/DVCLUB_Supervised_Complete_SLowerBound.pdf}
% %\vspace{-15pt}
% \caption{Supervised Deep Variational CLUB training architecture: considering the lower bound of information leakage.}
% \label{Fig:DVCLUB_Supervised_complete_SLowerBound}
% \vspace{-10pt} 
% \end{figure} 
% %---------------------------------------------------
% %---------------------------------------------------
% %
% %---------------------------------------------------
% %	  Figure: Unsupervised DVCLUB >> Complete Diagram
% %---------------------------------------------------
% %
% \begin{figure}[!t]
% \centering
% \includegraphics[height=5.8cm]{TeXFig/DVCLUB_UnSupervised_Complete_SLowerBound.pdf}
% %\vspace{-15pt}
% \caption{Unsupervised Deep Variational CLUB training architecture: considering the lower bound of information leakage.}
% \label{Fig:DVCLUB_UnSupervised_complete_SLowerBound}
% \vspace{-10pt} 
% \end{figure} 
% %---------------------------------------------------
% %---------------------------------------------------

Analogously, we can estimate the KL-divergence $\D_{\mathrm{KL}} \left( P_{\mathsf{D}} (\mathbf{X}) \, \Vert \, P_{\boldsymbol{\theta}} (\mathbf{X}) \right)$, the discrepancy measure in the visible (perceptual) space in the unsupervised DVCLUB functional \eqref{Eq:DVCLUB_UnSupervised}. Let us introduce a random variable $y$, and assign a label $y=1$ to sample drawn from $P_{\mathsf{D}} (\mathbf{X})$ and $y=0$ to samples drawn from $P_{\boldsymbol{\theta}} (\mathbf{X})$. Also, consider the discriminator $D_{\boldsymbol{\omega}} (\mathbf{x})  \approx \rho_{\mathbf{X}} (y = 1 \mid \mathbf{x}) = P_{\mathsf{D}} (\mathbf{X} = \mathbf{x}) $, with parameters $\boldsymbol{\omega}$. Given a trained discriminator $D_{\boldsymbol{\omega}} (\mathbf{x})$ and $M$ i.i.d. samples $\{ \mathbf{x}_m \}_{m=1}^{M}$ from $P_{\mathsf{D}}(\mathbf{X})$, we have:\vspace{-4pt}
\begin{equation}\label{Eq:D_KL_X_estimation}
    \D_{\mathrm{KL}} \left( P_{\mathsf{D}} (\mathbf{X}) \, \Vert \, P_{\boldsymbol{\theta}} (\mathbf{X}) \right) \approx \frac{1}{M} \sum_{m=1}^{M} \log \frac{D_{\boldsymbol{\omega}} (\mathbf{x}_m)}{1 - D_{\boldsymbol{\omega}} (\mathbf{x}_m)}. 
\end{equation}
In this case, the \textit{visible space discriminator} $D_{\boldsymbol{\omega}} (\mathbf{x})$ minimizes the following loss function:
\begin{subequations}\label{Eq:Discriminator_X_loss}
\begin{align}
    \mathcal{L}_{\mathrm{disc}}^{\mathsf{x}} (\boldsymbol{\omega}, \boldsymbol{\theta})  & \coloneqq   \mathbb{E}_{\rho_{\mathbf{X}} (\mathbf{x} \mid y) \rho(y)} \left[ \, - y \,  \log D_{\boldsymbol{\omega}} (\mathbf{X}) - (1 - y) \, \log (1 - D_{\boldsymbol{\omega}} (\mathbf{X}))\, \right]  \\
     & =   \mathbb{E}_{P_{\mathsf{D}}(\mathbf{X})} \left[ \, - \log D_{\boldsymbol{\omega}} (\mathbf{X})  \, \right] + \mathbb{E}_{P_{\boldsymbol{\theta}} (\mathbf{X})} \left[ \, - \log (1 - D_{\boldsymbol{\omega}} (\mathbf{X})) \, \right]   \\
     & =    \mathbb{E}_{P_{\mathsf{D}}(\mathbf{X})} \left[ \, - \log D_{\boldsymbol{\omega}} (\mathbf{X})  \, \right] + \mathbb{E}_{Q_{\boldsymbol{\psi}} (\mathbf{Z})} \left[ \, - \log \left( 1 - D_{\boldsymbol{\omega}} ( \, g_{\boldsymbol{\theta}}(\mathbf{Z} ) \, ) \right) \, \right], \label{Eq:Discriminator_X_loss_c}
\end{align}
\end{subequations}
or equivalently, maximizes $\mathbb{E}_{P_{\mathsf{D}}(\mathbf{X})} \left[ \, \log D_{\boldsymbol{\omega}} (\mathbf{X})  \, \right] + \mathbb{E}_{Q_{\boldsymbol{\psi}} (\mathbf{Z})} \left[ \, \log \left( 1 - D_{\boldsymbol{\omega}} ( \, g_{\boldsymbol{\theta}}(\mathbf{Z} ) \, ) \right) \, \right] $. % This is the well-known GAN objective. The connection between DVCLUB and GANs family is addressed in Sec.~\ref{Ssec:ConnectionOtherGenerativeModels}.

\pagebreak

\textbf{Learning Procedure:}
The DVCLUB models $(\text{P1}\!\!:\! \mathsf{S})$ and $(\text{P1}\!\!:\! \mathsf{U})$ are trained using alternating block coordinate descent across five steps:
\vspace{10pt}

\noindent
(1) {\small\textbf{\textsf{Train the Encoder, Utility Decoder and Uncertainty Decoder}}}.
\begin{itemize}
    \item Supervised Setup:
        \begin{multline}
              \mathop{\max}_{\boldsymbol{\phi}, \boldsymbol{\theta}, \boldsymbol{\varphi}} \quad   \mathbb{E}_{P_{\mathbf{U}, \mathbf{X}}} \left[ \mathbb{E}_{P_{\boldsymbol{\phi}} (\mathbf{Z} \mid \mathbf{X})} \left[ \log P_{\boldsymbol{\theta}} (\mathbf{U} \! \mid \! \mathbf{Z} ) \right]  \right] - ( \beta + \alpha ) \; \; \D_{\mathrm{KL}} \left( P_{\boldsymbol{\phi}} (\mathbf{Z} \mid \mathbf{X}) \Vert Q_{\boldsymbol{\psi}} (\mathbf{Z}) \! \mid \! P_{\mathsf{D}} (\mathbf{X})\right)   \\ 
               -  \alpha \;\;  \mathbb{E}_{P_{\mathbf{S}, \mathbf{X}}} \left[ \mathbb{E}_{P_{\boldsymbol{\phi}}(\mathbf{Z} \mid \mathbf{X})} \left[ - \log P_{\boldsymbol{\varphi}} (\mathbf{X} \! \mid \! \mathbf{S}, \mathbf{Z})\right] \right].  
        \end{multline}
    \item Unsupervised Setup:
        \begin{multline}
              \mathop{\max}_{\boldsymbol{\phi}, \boldsymbol{\theta}, \boldsymbol{\varphi}} \quad   \mathbb{E}_{P_{\mathsf{D}}(\mathbf{X})} \left[ \mathbb{E}_{P_{\boldsymbol{\phi}} (\mathbf{Z} \mid \mathbf{X})} \left[ \log P_{\boldsymbol{\theta}} (\mathbf{X} \! \mid \! \mathbf{Z} ) \right]  \right]  -  ( \beta + \alpha ) \; \; \D_{\mathrm{KL}} \left( P_{\boldsymbol{\phi}} (\mathbf{Z} \mid \mathbf{X}) \Vert Q_{\boldsymbol{\psi}} (\mathbf{Z}) \! \mid \! P_{\mathsf{D}} (\mathbf{X})\right)  \\ 
               -  \alpha \;\;  \mathbb{E}_{P_{\mathbf{S}, \mathbf{X}}} \left[ \mathbb{E}_{P_{\boldsymbol{\phi}}(\mathbf{Z} \mid \mathbf{X})} \left[ - \log P_{\boldsymbol{\varphi}} (\mathbf{X} \! \mid \! \mathbf{S}, \mathbf{Z})\right] \right]. 
        \end{multline}
\end{itemize}
\noindent
(2) {\small\textbf{\textsf{Train the Latent Space Discriminator}}}.
    \begin{eqnarray}
        \mathop{\min}_{\boldsymbol{\eta}}  \quad   \mathbb{E}_{P_{\mathsf{D}}(\mathbf{X})} \left[ \, \mathbb{E}_{P_{\boldsymbol{\phi}}(\mathbf{Z} \mid \mathbf{X})} \left[ - \log D_{\boldsymbol{\eta}} (\mathbf{Z}) \right] \, \right] + \mathbb{E}_{Q_{\boldsymbol{\psi}} (\mathbf{Z})} \left[ \, - \log (1 - D_{\boldsymbol{\eta}} (\mathbf{Z})) \, \right]. 
    \end{eqnarray}
\noindent
(3) {\small\textbf{\textsf{Train the Encoder and Prior Distribution Generator Adversarially}}}.
        \begin{eqnarray}
        \mathop{\max}_{\boldsymbol{\phi}, \boldsymbol{\psi}}  \quad   \mathbb{E}_{P_{\mathsf{D}}(\mathbf{X})} \left[ \, \mathbb{E}_{P_{\boldsymbol{\phi}}(\mathbf{Z} \mid \mathbf{X})} \left[ - \log D_{\boldsymbol{\eta}} (\mathbf{Z}) \right] \, \right] + \mathbb{E}_{Q_{\boldsymbol{\psi}} (\mathbf{Z})} \left[ \, - \log (1 - D_{\boldsymbol{\eta}} (\mathbf{Z})) \, \right].
    \end{eqnarray}
\noindent
(4) {\small\textbf{\textsf{Train the Output Space Discriminator}}}.
\begin{itemize}
    \item Supervised Scenario: 
    the \textit{Attribute Class Discriminator} $D_{\omega} (\mathbf{U})$ is updated as:
        \begin{eqnarray}
            \mathop{\min}_{\boldsymbol{\omega}}  \quad  \mathbb{E}_{P_{\mathbf{U}}} \left[ \, - \log D_{\boldsymbol{\omega}} (\mathbf{U})  \, \right] + \mathbb{E}_{Q_{\boldsymbol{\psi}} (\mathbf{Z})} \left[ \, - \log \left( 1 - D_{\boldsymbol{\omega}} ( \, g_{\boldsymbol{\theta}}(\mathbf{Z} ) \, ) \right) \, \right].
        \end{eqnarray}
    \item Unsupervised Scenario: 
    the \textit{Visible Space Discriminator} $D_{\omega} (\mathbf{X})$ is updated as:
        \begin{eqnarray}\label{Eq:TrainVisibleSpaceDiscriminator}
            \mathop{\min}_{\boldsymbol{\omega}}  \quad  \mathbb{E}_{P_{\mathsf{D}}(\mathbf{X})} \left[ \, - \log D_{\boldsymbol{\omega}} (\mathbf{X})  \, \right] + \mathbb{E}_{Q_{\boldsymbol{\psi}} (\mathbf{Z})} \left[ \, - \log \left( 1 - D_{\boldsymbol{\omega}} ( \, g_{\boldsymbol{\theta}}(\mathbf{Z} ) \, ) \right) \, \right].
        \end{eqnarray}
 \end{itemize}
\noindent
(5) {\small\textbf{\textsf{Train the Prior Distribution Generator and Utility Decoder Adversarially}}}.
\begin{eqnarray}\label{TrainPriorDistributionGeneratorAndUtilityDecoderAdversarially}
    \mathop{\max}_{\boldsymbol{\psi}, \boldsymbol{\theta}}  \quad  \mathbb{E}_{Q_{\boldsymbol{\psi}} (\mathbf{Z})} \left[ \, - \log \left( 1 - D_{\boldsymbol{\omega}} ( \, g_{\boldsymbol{\theta}}(\mathbf{Z} ) \, ) \right) \, \right] .
\end{eqnarray}

The complete training algorithm of the supervised and unsupervised DVCLUB models are shown in the Algorithm~\ref{Algorithm:P1_Supervised_DVCLUB} and Algorithm~\ref{Algorithm:P1_Unsupervised_DVCLUB}, respectively. 
%
%The iterative alternating block coordinate descent algorithm associated with $(\text{P2}\!\!:\! \mathsf{S})$ and $(\text{P2}\!\!:\! \mathsf{U})$ (will be discussed in Sec.~\ref{Sec:Experiments}) are analogous to the above steps with a few modifications which are omitted for brevity. The training algorithm of the Supervised and Unsupervised DVCLUB models $(\text{P3}\!\!:\! \mathsf{S})$ and $(\text{P3}\!\!:\! \mathsf{U})$ are provided in Appendix~\ref{Appendix:NetworkArchitecture_TrainingDetails}. 
%
We will discuss the alternative objectives in Sec.~\ref{Sec:Experiments}, and address their associated training algorithms in Appendix.~\ref{Appendix:SupplementaryResults}. 
An overview of our implementation is provided in Appendix.~\ref{Appendix:ImplementationOverview}. 
%---------------------------------------------------
%
%       TABLE:  Supervised DVCLUB Training Algorithm
%
%---------------------------------------------------
%
\begin{center}
%\begin{minipage}{0.9\linewidth}
\centering
\begin{spacing}{1}
\begin{algorithm}
    \setstretch{1.2}
    \begin{algorithmic}[1]
        \State \textbf{Input:} Training Dataset: $\{ \left( \mathbf{u}_n, \mathbf{s}_n, \mathbf{x}_n  \right) \}_{n=1}^{N}$; Hyper-Parameters: $\alpha,\beta$
        \State $\boldsymbol{\phi}, \boldsymbol{\theta}, \boldsymbol{\psi} , \boldsymbol{\varphi}, \boldsymbol{\eta}, \boldsymbol{\omega}\; \gets$ Initialize Network Parameters
        
        \Repeat
        
        \hspace{-15pt}(1) {\small\textbf{\textsf{Train the Encoder, Utility Decoder, Uncertainty Decoder}} $\left( \boldsymbol{\phi}, \boldsymbol{\theta}, \boldsymbol{\varphi} \right)$}
         \State  ~~Sample a mini-batch $\{ \mathbf{u}_m, \mathbf{s}_m, \mathbf{x}_m \}_{m=1}^{M} \sim P_{\mathsf{D}} (\mathbf{X}) P_{\mathbf{U}, \mathbf{S} 
         \mid \mathbf{X}}$
         \State  ~~Compute $\mathbf{z}_m \sim f_{\boldsymbol{\phi}} (\mathbf{x}_m), \forall m \in [M]$
        %  \Comment{\textcolor{blue!60!gray}{$f_{\boldsymbol{\phi}}$: Stochastic or Deterministic Encoder}}
         \State  ~~Sample $\{ \mathbf{\widehat{z}}_m \}_{m=1}^{M} \sim Q_{\boldsymbol{\psi}} (\mathbf{Z})$
         \State  ~~Compute $\mathbf{\widehat{x}}_m =  g_{\boldsymbol{\varphi}} (\mathbf{\widehat{z}}_m, \mathbf{s}_m), \forall m \in [M]$
         \State  ~~Back-propagate loss:\vspace{3pt}
         
                $ \mathcal{L} \! \left( \boldsymbol{\phi}, \! \boldsymbol{\theta}, \! \boldsymbol{\varphi} \right) \! = \!\! - \frac{1}{M} \! \sum_{m=1}^{M} \!\! \Big( \! \log P_{\boldsymbol{\theta}} ( \mathbf{u}_m \! \mid  \! \mathbf{z}_m ) \! - \! ( \beta + \alpha ) \D_{\mathrm{KL}} \! \left( P_{\boldsymbol{\phi}} (\mathbf{z}_m \! \mid \! \mathbf{x}_m ) \Vert  Q_{\boldsymbol{\psi}} (\mathbf{z}_m) \right) \! + \! \alpha \log P_{\boldsymbol{\varphi}} (\mathbf{\widehat{x}}_m \! \mid \! \mathbf{s}_m, \mathbf{z}_m) \Big)$
                
        \vspace{3pt}  
        
        \hspace{-15pt}(2) {\small\textbf{\textsf{Train the Latent Space Discriminator}} $ \boldsymbol{\eta} $}
        \State  ~~Sample $\{ \mathbf{x}_m \}_{m=1}^{M} \sim P_{\mathsf{D}} (\mathbf{X})$
        \State  ~~Sample $\{ \mathbf{n}_m \}_{m=1}^{M} \sim \mathcal{N}(\boldsymbol{0}, \mathbf{I})$
        \State  ~~Compute $\mathbf{z}_m \sim f_{\boldsymbol{\phi}} (\mathbf{x}_m), \forall m \in [M]$
        \State  ~~Compute $\mathbf{\widetilde{z}}_m \sim g_{\boldsymbol{\psi}} (\mathbf{n}_m), \forall m \in [M]$
        \State  ~~Back-propagate loss:\vspace{3pt}
        
                $\;\;\;\; \mathcal{L} \! \left( \boldsymbol{\eta} \right) \! = \!  \left( \beta + \alpha \right) \left[ - \frac{1}{M} \sum_{m=1}^{M} \! \Big( \log D_{\boldsymbol{\eta}} (\mathbf{z}_m)   +  \log \left( 1- D_{\boldsymbol{\eta}} (\, \mathbf{\widetilde{z}}_m \,) \right) \Big) \right]$
       % \State  ~~Clip discriminator $\boldsymbol{\eta}$ to $\left[ - \epsilon , \epsilon \right]^d$      
          
        \vspace{3pt}
        
        \hspace{-15pt}(3) {\small\textbf{\textsf{Train the Encoder and Prior Distribution Generator $\left( \boldsymbol{\phi}, \boldsymbol{\psi} \right)$ Adversarially}}} 
        \State  ~~Sample $\{ \mathbf{x}_m \}_{m=1}^{M} \sim P_{\mathsf{D}} (\mathbf{X})$
        \State  ~~Sample $\{ \mathbf{n}_m \}_{m=1}^{M} \sim \mathcal{N}(\boldsymbol{0}, \mathbf{I})$
        \State  ~~Compute $\mathbf{z}_m \sim f_{\boldsymbol{\phi}} (\mathbf{x}_m), \forall m \in [M]$
        \State  ~~Compute $\mathbf{\widetilde{z}}_m \sim g_{\boldsymbol{\psi}} (\mathbf{n}_m), \forall m \in [M]$
        \State  ~~Back-propagate loss:\vspace{3pt}
        
                $\;\;\;\; \mathcal{L} \! \left( \boldsymbol{\phi}, \boldsymbol{\psi} \right) \! = \!  \left( \beta + \alpha \right) \left[ \frac{1}{M} \sum_{m=1}^{M} \! \Big( \log D_{\boldsymbol{\eta}} (\mathbf{z}_m)   +  \log \left( 1- D_{\boldsymbol{\eta}} (\, \mathbf{\widetilde{z}}_m \,) \right) \Big) \right]$
        \vspace{3pt}

        \hspace{-15pt}(4) {\small\textbf{\textsf{Train the Attribute Class Discriminator}} $ \boldsymbol{\omega} $}
        \State  ~~Sample $\{ \mathbf{u}_m \}_{m=1}^{M} \sim P_{\mathbf{U}} $
        \State  ~~Sample $\{ \mathbf{n}_m \}_{m=1}^{M} \sim \mathcal{N} \! \left( \boldsymbol{0}, \mathbf{I}\right)$
        %\State  ~~Compute $\mathbf{z}_m \sim g_{\boldsymbol{\psi}} (\mathbf{n}_m), \forall m \in [M]$
        \State  ~~Compute $\mathbf{\widetilde{u}}_m \sim g_{\boldsymbol{\theta}} \left( g_{\boldsymbol{\psi}} (\mathbf{n}_m) \right), \forall m \in [M]$
        \State  ~~Back-propagate loss:\vspace{3pt}
        
                $\;\;\;\; \mathcal{L} \! \left( \boldsymbol{\omega} \right) \! = \!   - \frac{1}{M} \sum_{m=1}^{M} \! \Big( \log D_{\boldsymbol{\omega}} (\mathbf{u}_m)  +  \log \left( 1- D_{\boldsymbol{\omega}} (\, \mathbf{\widetilde{u}}_m \,) \right) \Big) $
        
        \vspace{3pt} 
        
        \hspace{-15pt}(5) {\small\textbf{\textsf{Train the Prior Distribution Generator and Utility Decoder $\left( \boldsymbol{\psi}, \boldsymbol{\theta} \right)$ Adversarially}}}
        %\Comment{\textcolor{blue!60!gray}{Utility Decoder Task: Attribute Classification}}
        \State  ~~Sample $\{ \mathbf{n}_m \}_{m=1}^{M} \sim \mathcal{N} \! \left( \boldsymbol{0}, \mathbf{I}\right)$
        %\State  ~~Compute $\mathbf{z}_m \sim g_{\boldsymbol{\psi}} (\mathbf{n}_m), \forall m \in [M]$
        \State  ~~Compute $\mathbf{\widetilde{u}}_m \sim g_{\boldsymbol{\theta}} \left( g_{\boldsymbol{\psi}} (\mathbf{n}_m) \right), \forall m \in [M]$
        \State  ~~Back-propagate loss:\vspace{3pt}
        
                $\;\;\;\; \mathcal{L} \! \left( \boldsymbol{\psi}, \boldsymbol{\theta} \right) \! = \!   \frac{1}{M} \sum_{m=1}^{M} \! \log \left( 1- D_{\boldsymbol{\omega}} (\, \mathbf{\widetilde{u}}_m \,) \right)$
        
        \vspace{3pt}  
        
        \Until{Convergence}
        \State \textbf{return} $\boldsymbol{\phi}, \boldsymbol{\theta}, \boldsymbol{\psi}, \boldsymbol{\varphi}, \boldsymbol{\eta}, \boldsymbol{\omega}$
    \end{algorithmic}
    \caption{Supervised Deep Variational CLUB training algorithm associated with ($\text{P1}\!\!:\! \mathsf{S}$)}
    \label{Algorithm:P1_Supervised_DVCLUB}
\end{algorithm}
\end{spacing}
%\end{minipage} 
\end{center}
%---------------------------------------------------
%--------------------------------------------------
%

%---------------------------------------------------
%
%       TABLE:  Un-Supervised DVCLUB Training Algorithm
%
%---------------------------------------------------
%
\begin{center}
%\begin{minipage}{0.9\linewidth}
\centering
\begin{spacing}{1}
\begin{algorithm}
    \setstretch{1.2}
    \begin{algorithmic}[1]
        \State \textbf{Input:} Training Dataset: $\{ \left( \mathbf{s}_n, \mathbf{x}_n  \right) \}_{n=1}^{N}$; Hyper-Parameters: $\alpha,\beta$
        \State $\boldsymbol{\phi}, \boldsymbol{\theta}, \boldsymbol{\psi} , \boldsymbol{\varphi}, \boldsymbol{\eta}, \boldsymbol{\omega}\; \gets$ Initialize Network Parameters
        
        \Repeat
        
        \hspace{-15pt}(1) {\small\textbf{\textsf{Train the Encoder, Utility Decoder, Uncertainty Decoder}} $\left( \boldsymbol{\phi}, \boldsymbol{\theta}, \boldsymbol{\varphi} \right)$}
         \State  ~~Sample a mini-batch $\{ \mathbf{s}_m, \mathbf{x}_m \}_{m=1}^{M} \sim P_{\mathsf{D}} (\mathbf{X}) P_{\mathbf{S} 
         \mid \mathbf{X}}$
         \State  ~~Compute $\mathbf{z}_m \sim f_{\boldsymbol{\phi}} (\mathbf{x}_m), \forall m \in [M]$
        %  \Comment{\textcolor{blue!60!gray}{$f_{\boldsymbol{\phi}}$: Stochastic or Deterministic Encoder}}
         \State  ~~Sample $\{ \mathbf{\widehat{z}}_m \}_{m=1}^{M} \sim Q_{\boldsymbol{\psi}} (\mathbf{Z})$
         \State  ~~Compute $\mathbf{\widehat{x}}_m =  g_{\boldsymbol{\varphi}} (\mathbf{\widehat{z}}_m, \mathbf{s}_m), \forall m \in [M]$
         \State  ~~Back-propagate loss:\vspace{3pt}
         
                $ \mathcal{L} \! \left( \boldsymbol{\phi}, \! \boldsymbol{\theta}, \! \boldsymbol{\varphi} \right) \! = \!\! - \frac{1}{M} \! \sum_{m=1}^{M} \!\! \Big(\! \log P_{\boldsymbol{\theta}} ( \mathbf{x}_m \! \mid  \! \mathbf{z}_m ) \! - \! ( \beta + \alpha ) \D_{\mathrm{KL}} \! \left( P_{\boldsymbol{\phi}} (\mathbf{z}_m \! \mid \! \mathbf{x}_m ) \Vert  Q_{\boldsymbol{\psi}} (\mathbf{z}_m) \right) \! + \! \alpha \log P_{\boldsymbol{\varphi}} (\mathbf{\widehat{x}}_m \! \mid \! \mathbf{s}_m, \mathbf{z}_m) \Big)$
                
        \vspace{3pt}  
        
        \hspace{-15pt}(2) {\small\textbf{\textsf{Train the Latent Space Discriminator}} $ \boldsymbol{\eta} $}
        \State  ~~Sample $\{ \mathbf{x}_m \}_{m=1}^{M} \sim P_{\mathsf{D}} (\mathbf{X})$
        \State  ~~Sample $\{ \mathbf{n}_m \}_{m=1}^{M} \sim \mathcal{N}(\boldsymbol{0}, \mathbf{I})$
        \State  ~~Compute $\mathbf{z}_m \sim f_{\boldsymbol{\phi}} (\mathbf{x}_m), \forall m \in [M]$
        \State  ~~Compute $\mathbf{\widetilde{z}}_m \sim g_{\boldsymbol{\psi}} (\mathbf{n}_m), \forall m \in [M]$
        \State  ~~Back-propagate loss:\vspace{3pt}
        
                $\;\;\;\; \mathcal{L} \! \left( \boldsymbol{\eta} \right) \! = \!  \left( \beta + \alpha \right) \left[ - \frac{1}{M} \sum_{m=1}^{M} \! \big( \, \log D_{\boldsymbol{\eta}} (\mathbf{z}_m) +  \log \left( 1- D_{\boldsymbol{\eta}} (\, \mathbf{\widetilde{z}}_m \,) \right)\, \big) \right]$
      %  \State  ~~Clip discriminator $\boldsymbol{\eta}$ to $\left[ - \epsilon , \epsilon \right]^d$      
          
        \vspace{3pt}
        
        \hspace{-15pt}(3) {\small\textbf{\textsf{Train the Encoder and Prior Distribution Generator $\left( \boldsymbol{\phi}, \boldsymbol{\psi} \right)$ Adversarially}}} 
        \State  ~~Sample $\{ \mathbf{x}_m \}_{m=1}^{M} \sim P_{\mathsf{D}} (\mathbf{X})$
        \State  ~~Sample $\{ \mathbf{n}_m \}_{m=1}^{M} \sim \mathcal{N}(\boldsymbol{0}, \mathbf{I})$
        \State  ~~Compute $\mathbf{z}_m \sim f_{\boldsymbol{\phi}} (\mathbf{x}_m), \forall m \in [M]$
        \State  ~~Compute $\mathbf{\widetilde{z}}_m \sim g_{\boldsymbol{\psi}} (\mathbf{n}_m), \forall m \in [M]$
        \State  ~~Back-propagate loss:\vspace{3pt}
        
                $\;\;\;\; \mathcal{L} \! \left( \boldsymbol{\phi}, \boldsymbol{\psi} \right) \! = \! \left( \beta + \alpha \right) \left[ \frac{1}{M} \sum_{m=1}^{M} \! \big( \, \log D_{\boldsymbol{\eta}} (\mathbf{z}_m) +  \log \left( 1- D_{\boldsymbol{\eta}} (\, \mathbf{\widetilde{z}}_m \,) \right) \,\big) \right]$
        
        \vspace{3pt}  
        
        \hspace{-15pt}(4) {\small\textbf{\textsf{Train the Visible Space Discriminator}} $ \boldsymbol{\omega} $}
        \State  ~~Sample $\{ \mathbf{x}_m \}_{m=1}^{M} \sim P_{\mathsf{D}} (\mathbf{X})$
        \State  ~~Sample $\{ \mathbf{n}_m \}_{m=1}^{M} \sim \mathcal{N} \! \left( \boldsymbol{0}, \mathbf{I}\right)$
        %\State  ~~Compute $\mathbf{z}_m \sim g_{\boldsymbol{\psi}} (\mathbf{n}_m), \forall m \in [M]$
        \State  ~~Compute $\mathbf{\widetilde{x}}_m \sim g_{\boldsymbol{\theta}} (  g_{\boldsymbol{\psi}} (\mathbf{n}_m) ), \forall m \in [M]$
        \State  ~~Back-propagate loss:\vspace{3pt}
        
                $\;\;\;\; \mathcal{L} \! \left( \boldsymbol{\omega} \right) \! = \! -  \frac{1}{M} \sum_{m=1}^{M} \! \Big( \, \log D_{\boldsymbol{\omega}} (\mathbf{x}_m) + \log \left( 1- D_{\boldsymbol{\omega}} (\, \mathbf{\widetilde{x}}_m \,)\, \right) \Big)$
        
        \vspace{3pt} 
        
        \hspace{-15pt}(5) {\small\textbf{\textsf{Train the Prior Distribution Generator and Utility Decoder $\left( \boldsymbol{\psi}, \boldsymbol{\theta} \right)$ Adversarially}}}
        %\Comment{\textcolor{blue!60!gray}{Utility Decoder Task: Image (Re)generation}}
        \State  ~~Sample $\{ \mathbf{n}_m \}_{m=1}^{M} \sim \mathcal{N} \! \left( \boldsymbol{0}, \mathbf{I}\right)$
        %\State  ~~Compute $\mathbf{z}_m \sim g_{\boldsymbol{\psi}} (\mathbf{n}_m), \forall m \in [M]$
        \State  ~~Compute $\mathbf{\widetilde{x}}_m \sim g_{\boldsymbol{\theta}} \left( g_{\boldsymbol{\psi}} (\mathbf{n}_m) \right), \forall m \in [M]$
        \State  ~~Back-propagate loss:\vspace{3pt}
        
                $\;\;\;\; \mathcal{L} \! \left( \boldsymbol{\psi}, \boldsymbol{\theta} \right) \! = \!   \frac{1}{M} \sum_{m=1}^{M}   \log \left( 1- D_{\boldsymbol{\omega}} (\, \mathbf{\widetilde{x}}_m \,)  \right) $
        
        \vspace{3pt}

        \Until{Convergence}
        \State \textbf{return} $\boldsymbol{\phi}, \boldsymbol{\theta}, \boldsymbol{\psi}, \boldsymbol{\varphi}, \boldsymbol{\eta}, \boldsymbol{\omega}$
    \end{algorithmic}
    \caption{Unsupervised Deep Variational CLUB training algorithm associated with ($\text{P1}\!\!:\! \mathsf{U}$)}
    \label{Algorithm:P1_Unsupervised_DVCLUB}
\end{algorithm}
\end{spacing}
%\end{minipage} 
\end{center}
\section{Connections with other Problems}
\label{Sec:ConnectionsSOTA}

%---------------------------------------------------
%
%			Connection to Other IB Models
%
%---------------------------------------------------
\subsection{Connection with State-of-the-Art Bottleneck Models}
\label{Ssec:ConnectionToOtherProblems}
%\subsection{Connection to Other Models} 
%\subsection{Link to Source Coding Problems}

\begin{figure}[t!]
    \centering
  %  \hspace{-9pt} 
        \begin{subfigure}[h]{0.31\textwidth}
        \includegraphics[scale=0.7]{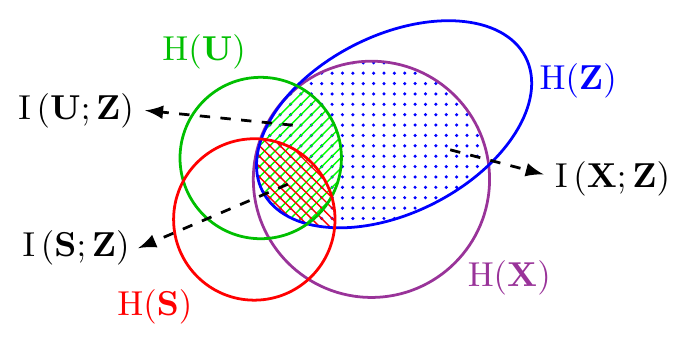}%
 %\includegraphics[width=cm, height=3.3cm]{TeXFig/GeneralDiag2.pdf}
%   \begin{subfigure}[h]{0.48\textwidth}
  %     \includegraphics[width=6cm, height=3.3cm]{TeXFig/GeneralDiag.pdf}%
  %        \vspace{-6pt}
        \caption{}
        %   \vspace{-10pt}
        \label{fig:Idiagram1}
    \end{subfigure}%
~
      \begin{subfigure}[h]{0.31\textwidth}
     %  \vspace{5pt}
        \includegraphics[scale=0.7]{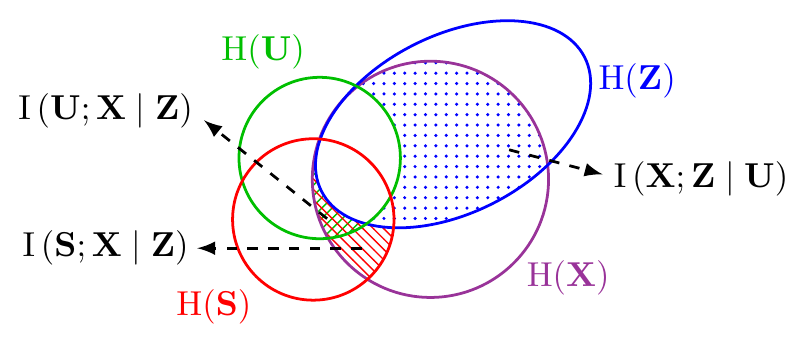}%
%     \begin{subfigure}[h]{0.48\textwidth}
  %   \includegraphics[width=6cm, height=3.3cm]{TeXFig/RegionOverlap.pdf}%
%\includegraphics[scale=0.2]{pdfFig/DisPreQuerySide.pdf}%
        %\vspace{-5pt}
        \caption{}
      %    \vspace{-10pt}
        \label{fig:Idiagram2}
    \end{subfigure}
~~~~~~
      \begin{subfigure}[h]{0.31\textwidth}
     %  \vspace{5pt}
        \includegraphics[scale=0.7]{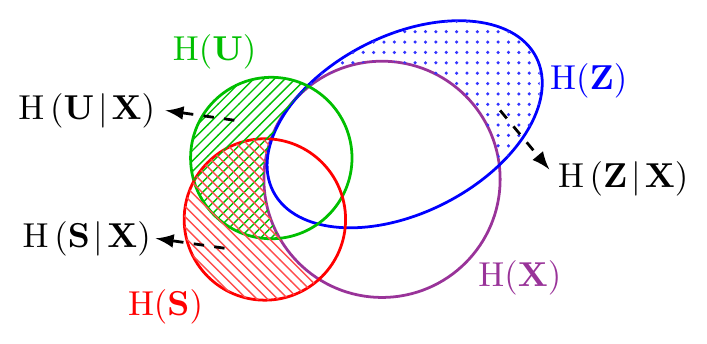}%
%     \begin{subfigure}[h]{0.48\textwidth}
  %   \includegraphics[width=6cm, height=3.3cm]{TeXFig/RegionOverlap.pdf}%
%\includegraphics[scale=0.2]{pdfFig/DisPreQuerySide.pdf}%
        %\vspace{-5pt}
        \caption{}
      %    \vspace{-10pt}
        \label{fig:Idiagram3}
    \end{subfigure}
  % \vspace{-6pt}
    \caption{Information diagrams for $\left( \mathbf{U}, \mathbf{S} \right) \markov \mathbf{X} \markov \mathbf{Z}$. (a) mutual information $\I \left(\mathbf{U}; \mathbf{Z}\right)$, $\I \left(\mathbf{X}; \mathbf{Z}\right)$ and $\I \left(\mathbf{S}; \mathbf{Z}\right)$; (b) redundant information $\I \left(\mathbf{X}; \mathbf{Z} \mid \mathbf{U}\right)$, residual information $\I \left(\mathbf{U}; \mathbf{X} \mid \mathbf{Z}\right)$ and residual information $\I \left(\mathbf{S}; \mathbf{X} \mid \mathbf{Z}\right)$; (c) utility attribute uncertainty $\H \left( \mathbf{U} \! \mid \! \mathbf{X} \right)$, private attribute uncertainty $\H \left( \mathbf{S} \! \mid \! \mathbf{X} \right)$ and encoding uncertainty $\H \left( \mathbf{Z} \! \mid \! \mathbf{X} \right)$.}
  %  \vspace{-15pt}
    \label{Fig:I-diagram}
%   \vspace{-13pt}
\end{figure}

In this subsection, we present the connection of CLUB model with the most relative state-of-the-art literature. We show that CLUB model generalizes most of the previously proposed models. 
To help interpretation, we use $\I$-Diagram \cite{yeung1991new}, an analogy between information theory and set theory, to visualize a clear connection between the different considered objectives. 
By constructing an unique measure, called $\I$-Measure, consistent with Shannon's information measure, we can geometrically represent the relationship among the Shannon's information measures. 
%Note that 
The Markov chain $ \left( \mathbf{U}, \mathbf{S} \right) \markov \mathbf{X} \markov \mathbf{Z}$ implies: 
%$P_{\mathbf{U}, \mathbf{S}, \mathbf{X}, \mathbf{Z}} = P_{\mathbf{U}, \mathbf{S}} P_{\mathbf{X} \mid \mathbf{U}, \mathbf{S}} P_{\mathbf{Z} \mid \mathbf{X}}$. 
\begin{eqnarray}
\I \left( \mathbf{U}; \mathbf{S}; \mathbf{Z} \mid \mathbf{X} \right) + \I \left( \mathbf{U}; \mathbf{Z} \mid \mathbf{S},  \mathbf{X} \right)  + \I \left( \mathbf{S}; \mathbf{Z} \mid  \mathbf{U},  \mathbf{X} \right) \!  \!  &=& \! \! 
\I \left( \mathbf{U}; \mathbf{Z}  \mid \mathbf{X} \right)  + 
\I \left( \mathbf{S}; \mathbf{Z}  \mid  \mathbf{U}, \mathbf{X} \right) \nonumber \\
&=& \! \!  \I \left( \mathbf{U}, \mathbf{S}; \mathbf{Z}  \mid \mathbf{X} \right) = 0. 
\end{eqnarray}
Considering the Markov chain $ \left( \mathbf{U}, \mathbf{S} \right) \markov \mathbf{X} \markov \mathbf{Z}$, the corresponding $\I$-Diagram is depicted in Fig.~\ref{Fig:I-diagram}, where various information quantities are also highlighted.

%---------------------------------------------------
%	    Information Bottleneck  
%---------------------------------------------------
%
\textbf{Information Bottleneck (IB).} 
The IB principle \cite{tishby2000information} 
%is a generalization of the sufficient statistic methods that 
formulates the problem of extracting the relevant information from the random variable $\mathbf{X}$ about the random variable $\mathbf{U}$ that is of interest. 
Given two correlated random variables $\mathbf{U}$ and $\mathbf{X}$ with joint distribution $P_{\mathbf{U,X}}$, the goal of \textit{original} IB is to find a representation $\mathbf{Z}$ of $\mathbf{X}$ using a stochastic mapping $P_{\mathbf{Z}\mid \mathbf{X}}$ such that: (i) $\mathbf{U} \markov \mathbf{X} \markov \mathbf{Z}$ and (ii) representation $\mathbf{Z}$ is maximally informative about $\mathbf{U}$ (maximizing $\I \left( \mathbf{U}; \mathbf{Z} \right)$) while being minimally informative about $\mathbf{X}$ (minimizing $\I \left( \mathbf{X}; \mathbf{Z}\right)$).
%  
%It allows a model to smoothly trade-off the maximality of the informativeness of the bottleneck variable ($\mathbf{Z}$) for the task at hand ($T$), against the compressiveness of the bottleneck variable ($\mathbf{Z}$) from data ($\mathbf{X}$). 
%
%Let $f: \left( 0, \infty \right) \rightarrow \mathbb{R}$ be a convex function which $f(1) =0$. The  $f$-information between two random variables $X$ and $Z$ is defined as $I_f \left( X ; Z \right) = D_f \left(  p(x,z) \| p(x) p(z) \right)$, where $D_f \left( \cdot \| \cdot \right)$ is $f$-divergence \cite{polyanskiy2014lecture}. 
%
This trade-off can be formulated by bottleneck functional: 
\begin{eqnarray}\label{Eq:IB_functional}
\mathsf{IB} \left( R, P_{\mathbf{U}, \mathbf{X}}\right)  \coloneqq \mathop{\sup}_{ \substack{P_{\mathbf{Z} \mid \mathbf{X}}:\\ \mathbf{U} \markov \mathbf{X} \markov \mathbf{Z}}} \I  \left( \mathbf{U}; \mathbf{Z} \right) \;\;  \mathrm{s.t.} \;\;  \I \left( \mathbf{X}; \mathbf{Z} \right) \leq R. 
\end{eqnarray}
In IB model, $\I  \left( \mathbf{U}; \mathbf{Z} \right)$ is referred to as the relevance of $\mathbf{Z}$ and $\I \left( \mathbf{X}; \mathbf{Z} \right)$ is referred to as the complexity of $\mathbf{Z}$. 
%Since mutual information is defined as Shannon information, the complexity here quantified by the minimum description length of compressed representation 
The values $\mathsf{IB} \left( R , P_{\mathbf{U}, \mathbf{X}}\right)$ for different $R$ specify the IB curve. 
Analogously, with the introduction of a Lagrange multiplier $\beta \in \left[ 0, 1\right]$ %\footnote{Note that by the DPI, $\I  \left( \mathbf{U}; \mathbf{Z} \right) \leq \I  \left( \mathbf{X}; \mathbf{Z} \right)$ and $\I  \left( \mathbf{S}; \mathbf{Z} \right) \leq \I  \left( \mathbf{X}; \mathbf{Z} \right)$. Hence, for $\beta > 1$, based on the considered Lagrangian functional the model may learn a trivial representation, independent of $\mathbf{X}$.}, 
we can quantify the IB problem by the associated Lagrangian functional 
$ \mathcal{L}_{\mathrm{IB}} \left( P_{\mathbf{Z}\mid \mathbf{X}}, \beta  \right)   \coloneqq   \I  \left( \mathbf{U}; \mathbf{Z} \right) - \beta \,   \I \left( \mathbf{X}; \mathbf{Z} \right)$. Clearly, IB is a specific case of CLUB model \eqref{CLUB_functional}, when the information leakage is disregarded, or equivalently, when $R^{\mathrm{s}} \geq \H \left( \mathbf{S} \right)$.

The underlying IB optimization problem dates back to the early 1970's, when Wyner and Ziv \cite{wyner1973theoremI} determined the value of IB functional \eqref{Eq:IB_functional} for the special case of binary random variables. See also \cite{wyner1973theoremII, witsenhausen1975conditional, ahlswede1977connection} for the related problems. 
Inspired by the formulation of IB method in \cite{tishby2000information}, abundant characterizations, generalizations, and applications have been proposed \cite{ makhdoumi2014information, tishby2015deep, alemi2016deep, strouse2017deterministic, vera2018collaborative, kolchinsky2018caveats, bang2019explaining, amjad2019learning, hu2019information, wu2019learnability, fischer2020conditional, federici2020learning, ding2019submodularity, hafez3information, hafez2020sample, kirsch2020unpacking}. We refer the reader to \cite{voloshynovskiyinformation, goldfeld2020information, zaidi2020information, asoodeh2020bottleneck} for a review of recent research on IB models.

%$\mathcal{L}_{\mathrm{PF}} \left( P_{\mathbf{Z}\mid \mathbf{X}}, \beta  \right)  \triangleq   \I  \left( \mathbf{S}; \mathbf{Z} \right) - \beta \,   \I \left( \mathbf{X}; \mathbf{Z} \right)$. 

%Considering Markov condition $X \rightarrow Z \rightarrow T$ and defining $f$-information between two random variables $X$ and $Z$ as $I_f \left( X; Z \right) = D_f \left( p(x,z) \| p(x) p(z)\right)$
%this trade-off can be formulated by bottleneck functional: 
%\begin{eqnarray}
%\mathop{\max}_{X \rightarrow Z \rightarrow T} I_{f_2} \left( X; T \right) \; \mathrm{s.t} \; I_{f_1} \left( X; Z \right) \leq R_1,
%\end{eqnarray}
%where $f_i: \! \left( 0, \infty \right) \! \rightarrow \mathbb{R}, f_i(1) =0,  \forall i$, are convex functions \cite{polyanskiy2014lecture}. 

%The  $f$-information between two random variables $X$ and $Z$ is defined as $I_f \left( X ; Z \right) = D_f \left(  p(x,z) \| p(x) p(z) \right)$, where $D_f \left( \cdot \| \cdot \right)$ is $f$-divergence \cite{polyanskiy2014lecture}. 

%Note that although all $D_f \left( \cdot \| \cdot \right)$ quantify the dissimilarity between a pair of distributions, their operational meanings are different. For example, $f(t) = \frac{1}{2} \vert t - 1 \vert$ results in Total Variation (TV) which is utilized in hypothesis testing, and $f(t) = (t-1)^2$ (or $t^2 - 1$) gives $\chi^2$-information which is useful in estimation problems \cite{du2017principal}. 

%---------------------------------------------------
%	       Figure: Information Diagrams
%---------------------------------------------------
%
%---------------------------------------------------
%---------------------------------------------------

%---------------------------------------------------
%	       Privacy Funnel
%---------------------------------------------------
%
\textbf{Privacy Funnel (PF).} 
In contrast to IB principle, which seeks to obtain a representation $\mathbf{Z}$ that is maximally expressive (information preserving) about $\mathbf{U}$ while maximally compressive about $\mathbf{X}$, 
considering Markov chain $\mathbf{S} \markov \mathbf{X} \markov \mathbf{Z}$, the goal of PF \cite{makhdoumi2014information} is to determine a representation $\mathbf{Z}$ that minimizes the information leakage between the private (sensitive) data $\mathbf{S}$ and the disclosed representation $\mathbf{Z}$, i.e., $\I \left( \mathbf{S}; \mathbf{Z} \right)$, while maximizes the amount of information between non-private (useful) data $\mathbf{X}$ and disclosed representation $\mathbf{Z}$, i.e., $\I \left( \mathbf{X}; \mathbf{Z} \right)$. 
Therefore, the PF method addresses the trade-off between the information leakage $\I \left( \mathbf{S}; \mathbf{Z} \right)$ and the revealed useful information $\I \left( \mathbf{X}; \mathbf{Z} \right)$. 
Analogously, this trade-off can be formulated by the PF functional: 
\begin{eqnarray}\label{Eq:PF_functional}
\mathsf{PF} \left( R, P_{\mathbf{S}, \mathbf{X}}\right)  \coloneqq \mathop{\inf}_{\substack{P_{\mathbf{Z} \mid \mathbf{X}}: \\  \mathbf{S} \markov \mathbf{X} \markov \mathbf{Z}}} \I \left( \mathbf{S}; \mathbf{Z} \right) \;\;  \mathrm{s.t.} \;\;  \I \left( \mathbf{X}; \mathbf{Z} \right) \geq R. 
\end{eqnarray}
The values $\mathsf{PF} \left( R, P_{\mathbf{S}, \mathbf{X}}\right)$ for different $R$ specify the PF curve. By introducing a Lagrange multiplier $\beta \in \left[ 0, 1\right]$ we can quantify the PF problem by the associated Lagrangian functional $\mathcal{L}_{\mathrm{PF}} \left( P_{\mathbf{Z}\mid \mathbf{X}}, \beta  \right)   \coloneqq   \I  \left( \mathbf{S}; \mathbf{Z} \right) - \beta \,   \I \left( \mathbf{X}; \mathbf{Z} \right)$. Setting $\mathbf{U} \equiv \mathbf{X}$ and $R^{\mathrm{z}} \geq \H \left( P_{\mathbf{X}}\right)$ in the CLUB objective \eqref{CLUB_functional}, the CLUB model reduces to the PF model. This corresponds to a scenario, for instance, in which the goal is to release facial images, without revealing a specific sensitive attribute.

In \cite{bertran2019adversarially}, the authors obtained a map $P_{\mathbf{Z} \mid \mathbf{X}}$ that minimizes $ \I \left( \mathbf{U}; \mathbf{X} \mid \mathbf{Z} \right)$ under an information leakage constraint $\I \left( \mathbf{S}; \mathbf{Z} \right)$. Considering a similar setting to ours, the following objective is considered:
\begin{equation}\label{Bertran_objective}
 \mathop{\inf}_{\substack{P_{\mathbf{Z} \mid \mathbf{X}}: \\\left( \mathbf{U}, \mathbf{S}\right) \markov \mathbf{X} \markov \mathbf{Z}}} \quad  \I \left( \mathbf{U}; \mathbf{X} \mid \mathbf{Z} \right) \quad \mathrm{s.t.} \quad \I \left( \mathbf{S}; \mathbf{Z} \right) \leq R^{\mathrm{s}}. 
\end{equation}
One can interpret $\I \left( \mathbf{U}; \mathbf{X} \mid \mathbf{Z} \right)$ as the amount of information about $\mathbf{U}$ we lose by observing the released representation $\mathbf{Z}$ instead of the original data $\mathbf{X}$. 
To see its connection to CLUB, note that, under the Markov chain $ \mathbf{U} \markov \mathbf{X} \markov \mathbf{Z}$, we have:
\begin{eqnarray}\label{MI_conditionalZ}
\I \left( \mathbf{U}; \mathbf{Z} \right) = \I \left( \mathbf{U}; \mathbf{X} \right) - \I \left( \mathbf{U} ; \mathbf{X} \mid \mathbf{Z} \right). 
\end{eqnarray}
Therefore, maximizing $\I \left( \mathbf{U}; \mathbf{Z} \right)$ is equivalent to minimizing $\I \left( \mathbf{U}; \mathbf{X} \mid \mathbf{Z} \right)$. 
%Compared with \cite{bertran2019adversarially}, the CLUB objective considers constraint on encoding rate  $\I \left( \mathbf{X}; \mathbf{Z} \right)$. 
Clearly, the objective \eqref{Bertran_objective} can be considered as a special case of the CLUB model in \eqref{CLUB_functional}. In particular, we highlight that our model addresses the key missed component in the considered formulation, i.e., the information complexity constraint.

Note that $P_{\mathbf{U}, \mathbf{X} \mid \mathbf{Z} } = P_{\mathbf{U} \mid \mathbf{X},  \mathbf{Z} } P_{\mathbf{X} \mid \mathbf{Z} }  = P_{\mathbf{U} \mid \mathbf{X} }  P_{\mathbf{X} \mid \mathbf{Z} } $, hence, 
\begin{eqnarray}
  \I \left( \mathbf{U}; \mathbf{X}  \mid \mathbf{Z} \right)  
& = &
 \mathbb{E}_{P_{\mathbf{U},\mathbf{X}, \mathbf{Z}}} \left[ \log \frac{P_{\mathbf{U},  \mathbf{X} \mid \mathbf{Z}}}{P_{\mathbf{U} \mid \mathbf{Z}} P_{\mathbf{X}  \mid \mathbf{Z}}}  \right] 
 =
  \mathbb{E}_{P_{\mathbf{U},\mathbf{X}, \mathbf{Z}}} \left[ \log \frac{P_{\mathbf{U} \mid  \mathbf{X} } P_{\mathbf{X} \mid  \mathbf{Z} } }{P_{\mathbf{U} \mid \mathbf{Z}} P_{\mathbf{X}  \mid \mathbf{Z}}}  \right] 
 =
  \mathbb{E}_{P_{\mathbf{X}, \mathbf{Z}}} \Big[ \mathbb{E}_{P_{\mathbf{U} \mid \mathbf{X}, \mathbf{Z}}} \big[  \log \frac{P_{\mathbf{U} \mid \mathbf{X}}}{P_{\mathbf{U} \mid \mathbf{Z}}}  \big] \Big] \nonumber \\
  & =&
 \mathbb{E}_{P_{\mathbf{X}, \mathbf{Z}}} \left[ 
\D_{\mathrm{KL}} \left( P_{\mathbf{U} \mid \mathbf{X}} \Vert P_{\mathbf{U} \mid \mathbf{Z}} \right)\right].
\end{eqnarray} 
This gives us another interpretation of the CLUB objective as a distribution matching problem, which aims at obtaining a stochastic map $P_{\mathbf{Z} \mid \mathbf{X}}$ such that $P_{\mathbf{U} \mid \mathbf{Z}} \approx P_{\mathbf{U} \mid \mathbf{X}}$. 
This means that the posterior distributions of the utility attribute $\mathbf{U}$ are similar conditioned on the released representation $\mathbf{Z}$ or the original data $\mathbf{X}$. 
Finally, we remark that Eq.~\eqref{MI_conditionalZ} is also related to the source coding problem with a common reconstruction constraint studied in \cite{steinberg2009coding, benammar2016rate}.

%---------------------------------------------------
%	       Deterministic Information Bottleneck
%---------------------------------------------------
%
\textbf{Deterministic IB (DIB).}  
In the information complexity decomposition $\I \left( \mathbf{X}; \mathbf{Z} \right) = \H \left( \mathbf{Z} \right) - \H \left( \mathbf{Z} \mid \mathbf{X} \right)$, the conditional entropy $\H \left( \mathbf{Z} \! \mid \! \mathbf{X} \right)$ measures the \textit{encoding uncertainty} (see Fig.~\ref{fig:Idiagram3}). The deterministic IB model \cite{strouse2017deterministic} is based on a deterministic encoder. In this case, we have $\H \left( \mathbf{Z} \! \mid \! \mathbf{X} \right) = 0$. Considering the Markov chain $\mathbf{U} \markov \mathbf{X} \markov \mathbf{Z}$, the following objective is considered:\vspace{-4pt}
\begin{equation}\label{Eq:DeterministicIB}
\mathsf{DIB} \left( R, P_{\mathbf{U}, \mathbf{X}}\right)  \coloneqq \mathop{\inf}_{\substack{P_{\mathbf{Z} \mid \mathbf{X}}: \\ \mathbf{U} \markov \mathbf{X} \markov \mathbf{Z}}} \quad  \H \left( \mathbf{Z} \right) \quad \mathrm{s.t.} \quad \I \left( \mathbf{U}; \mathbf{Z} \right) \geq R .
\end{equation}
In this case, the associated Lagrangian functional is given by $\mathcal{L}_{\mathrm{DIB}}(P_{\mathbf{Z} \mid \mathbf{X}}, \beta)  \coloneqq \H \left( \mathbf{Z} \right) - \beta \I \left( \mathbf{U}; \mathbf{Z} \right)$. Clearly, DIB is a specific case of the IB model; and hence, the CLUB model in \eqref{CLUB_functional_infimum}. 

% \footnote{Analogous to DIB, the deterministic CLUB (DCLUB) model encourages to have a deterministic encoding function. The DCLUB functional reads as:
% \begin{equation}
% \mathsf{DCLUB} \left( R^{\mathrm{u}}, R^{\mathrm{s}}, P_{\mathbf{U}, \mathbf{S}, \mathbf{X}}\right)  \coloneqq \mathop{\inf}_{\substack{ P_{\mathbf{Z} \mid \mathbf{X}}: \\\left( \mathbf{U}, \mathbf{S}\right) \markov \mathbf{X} \markov \mathbf{Z}}} 
% \H  \left( \mathbf{Z} \right)  \; \quad  \; \mathrm{s.t.}  \quad   \I \left( \mathbf{U}; \mathbf{Z} \right) \geq R^{\mathrm{u}}, \;\; 
% \I \left( \mathbf{S}; \mathbf{Z} \right) \leq R^{\mathrm{s}}. 
% \end{equation}
% Hence, the associated Lagrangian functional is given as:
% \begin{subequations}
% \begin{align}
% \mathcal{L}_{\mathrm{DCLUB}} \left( P_{\mathbf{Z}\mid \mathbf{X}}, \gamma , \lambda \right)   
% &  \coloneqq     \H  \left( \mathbf{Z} \right) - \gamma \,   \I \left( \mathbf{U}; \mathbf{Z} \right) +\lambda  \, \I \left( \mathbf{S}; \mathbf{Z} \right) \\
% & \equiv    \H  \left( \mathbf{Z} \right) + \gamma \,   \H \left( \mathbf{U} \mid \mathbf{Z} \right) - \lambda  \, \H \left( \mathbf{S} \mid \mathbf{Z} \right).% \\
% %& \equiv  \H \left( \mathbf{Z} \mid \mathbf{U} \right)  
% \end{align}
% \end{subequations}}. 
%%
%}

%---------------------------------------------------
%	       Conditional Entropy Bottleneck
%---------------------------------------------------
%
\textbf{Conditional Entropy Bottleneck (CEB).}  
The conditional entropy bottleneck (CEB) \cite{fischer2020conditional} is motivated from the principle of minimum necessary information, and considers the following objective:\vspace{-2pt}
\begin{equation}\label{Eq:ConditionalEntropyBottleneck}
\mathsf{CEB} \left( R, P_{\mathbf{U}, \mathbf{X}}\right)  \coloneqq \mathop{\sup}_{\substack{P_{\mathbf{Z} \mid \mathbf{X}}: \\ \mathbf{U} \markov \mathbf{X} \markov \mathbf{Z}}} \quad  \I \left( \mathbf{U}; \mathbf{Z} \right) \quad \mathrm{s.t.} \quad \I \left( \mathbf{X}; \mathbf{Z} \mid \mathbf{U} \right) \leq R. 
\end{equation}
One can interpret $\I \left( \mathbf{X}; \mathbf{Z} \mid \mathbf{U} \right)$ as the \textit{redundant} information in the released representation $\mathbf{Z}$ about the utility attribute $\mathbf{U}$. 
To see its connection to CLUB, let us consider the Markov chain $ \mathbf{U} \markov \mathbf{X} \markov \mathbf{Z}$, we have:\vspace{-2pt}
\begin{equation}\label{MI_XZgivenU}
\I \left( \mathbf{X}; \mathbf{Z} \right) = \I \left( \mathbf{X}; \mathbf{Z} \mid \mathbf{U} \right) + \I \left( \mathbf{U}; \mathbf{Z} \right). 
\end{equation}
Therefore, using \eqref{MI_XZgivenU}, the associated CEB Lagrangian $\mathcal{L}_{\mathrm{CEB}} \left( P_{\mathbf{Z} \mid \mathbf{X}} , \beta^{\prime} \right)  \coloneqq  \I \left( \mathbf{U}; \mathbf{Z} \right) - \beta^{\prime} \, \I \left( \mathbf{X}; \mathbf{Z} \mid \mathbf{U} \right)$ can recast as:
\begin{eqnarray}\label{CEB_Lagrangian_to_IB}
\mathcal{L}_{\mathrm{CEB}} \left( P_{\mathbf{Z} \mid \mathbf{X}}, \beta^{\prime} \right) &=& \left( \beta^{\prime} + 1\right) \I \left( \mathbf{U}; \mathbf{Z} \right) - \beta^{\prime} \I \left( \mathbf{X}; \mathbf{Z} \right) \nonumber \\
 &=& \left( \beta^{\prime} + 1\right)  \Big( \I \left( \mathbf{S}; \mathbf{Z} \right) -  \frac{\beta^{\prime}}{\beta^{\prime} + 1} \I \left( \mathbf{X}; \mathbf{Z} \right) \Big)
  \nonumber \\
 &=& 
 \left( \beta^{\prime} + 1\right)  \mathcal{L}_{\mathrm{IB}} \left( P_{\mathbf{Z} \mid \mathbf{X}}, \beta \right),
\end{eqnarray}
where $\beta = \beta^{\prime} / (\beta^{\prime} + 1)$. Hence, maximizing $\mathcal{L}_{\mathrm{CEB}} \left( P_{\mathbf{Z} \mid \mathbf{X}}, \beta^{\prime} \right)$ is equivalent to maximizing $\mathcal{L}_{\mathrm{IB}} \left( P_{\mathbf{Z} \mid \mathbf{X}}, \beta \right)$, which is a specific case of CLUB model \eqref{CLUB_functional}.

%---------------------------------------------------
%	       Conditional Privacy Funnel
%---------------------------------------------------
%
\textbf{Conditional PF (CPF).}  
Inspired by the conditional entropy bottleneck \cite{fischer2020conditional} and noting that the PF \eqref{Eq:PF_functional} is dual of the Information Bottleneck \eqref{Eq:IB_functional}, the recent work of \cite{rodriguez2020variational}, addressed the CPF as:
\begin{equation}\label{Eq:ConditionalPrivacyFunnel}
\mathsf{CPF} \left( R, P_{\mathbf{S}, \mathbf{X}}\right)  \coloneqq \mathop{\inf}_{\substack{P_{\mathbf{Z} \mid \mathbf{X}}: \\ \mathbf{S} \markov \mathbf{X} \markov \mathbf{Z}}} \quad  \I \left( \mathbf{S}; \mathbf{Z} \right) \quad \mathrm{s.t.} \quad \I \left( \mathbf{X}; \mathbf{Z} \mid \mathbf{S} \right) \geq R^{\mathrm{c} \mid \mathrm{s}}. 
\end{equation}
To see its connection to CLUB, we use the Markov chain $\mathbf{S} \markov \mathbf{X} \markov \mathbf{Z}$ to obtain:
\begin{equation}\label{MI_XZgivenS}
\I \left( \mathbf{X}; \mathbf{Z} \right) = \I \left( \mathbf{X}; \mathbf{Z} \mid \mathbf{S} \right) + \I \left( \mathbf{S}; \mathbf{Z} \right). 
\end{equation}
%$\I \left( \mathbf{X}; \mathbf{Z} \right) = \I \left( \mathbf{X}; \mathbf{Z} \mid \mathbf{S} \right) - \I \left( \mathbf{S}; \mathbf{Z} \right)$. 
Hence, CPF ignores the shared information in the released representation $\mathbf{Z}$ about the private attribute $\mathbf{S}$, and imposes the constraint on the \textit{residual} information in the released representation $\mathbf{Z}$ about the useful data $\mathbf{X}$\footnote{Note that in the PF model in \eqref{Eq:PF_functional}, $\I \left( \mathbf{X}; \mathbf{Z} \right)$ measures the `useful' information, which is of the designer's interest.  Hence, $\I \left( \mathbf{X}; \mathbf{Z} \mid \mathbf{S} \right)$ in PF quantifies the residual information, while $\I \left( \mathbf{X}; \mathbf{Z} \mid \mathbf{U} \right)$ in IB quantifies the redundant information.}. The associated CPF Lagrangian functional is given by $\mathcal{L}_{\mathrm{CPF}} \left( P_{\mathbf{Z} \mid \mathbf{X}}, \beta^{\prime} \right)  \coloneqq \I \left( \mathbf{S}; \mathbf{Z} \right) - \beta^{\prime}  \I \left( \mathbf{X}; \mathbf{Z} \mid \mathbf{S} \right)$. Analogous  to \eqref{CEB_Lagrangian_to_IB}, and using \eqref{MI_XZgivenS}, we have:
\begin{eqnarray}\label{CPF_Lagrangian_to_PF}
\mathcal{L}_{\mathrm{CPF}} \left( P_{\mathbf{Z} \mid \mathbf{X}}, \beta^{\prime} \right) =
 \left( \beta^{\prime} + 1\right)  \mathcal{L}_{\mathrm{PF}} \left( P_{\mathbf{Z} \mid \mathbf{X}}, \beta \right),
\end{eqnarray}
%\begin{eqnarray}\label{CPF_Lagrangian_to_PF}
%\mathcal{L}_{\mathrm{CPF}} \left( P_{\mathbf{Z} \mid \mathbf{X}}, \beta^{\prime} \right) &=& \left( \beta^{\prime} + 1\right) \I \left( \mathbf{S}; \mathbf{Z} \right) - \beta^{\prime} \I \left( \mathbf{X}; \mathbf{Z} \right) \nonumber \\
% &=& \left( \beta^{\prime} + 1\right)  \Big( \I \left( \mathbf{S}; \mathbf{Z} \right) -  \frac{\beta^{\prime}}{\beta^{\prime} + 1} \I \left( \mathbf{X}; \mathbf{Z} \right) \Big)
%  \nonumber \\
% &=& 
% \left( \beta^{\prime} + 1\right)  \mathcal{L}_{\mathrm{PF}} \left( P_{\mathbf{Z} \mid \mathbf{X}}, \beta \right),
%\end{eqnarray}
%$\mathcal{L}_{\mathrm{CPF}} \left( P_{\mathbf{Z} \mid \mathbf{X}}, \beta^{\prime} \right) = \left( \beta^{\prime} + 1\right) \I \left( \mathbf{S}; \mathbf{Z} \right) - \beta^{\prime} \I \left( \mathbf{X}; \mathbf{Z} \right) = $
where $\beta = \beta^{\prime} / (\beta^{\prime} + 1)$. Hence, minimizing $\mathcal{L}_{\mathrm{CPF}} \left( P_{\mathbf{Z} \mid \mathbf{X}}, \beta^{\prime} \right)$ is equivalent to minimizing $\mathcal{L}_{\mathrm{PF}} \left( P_{\mathbf{Z} \mid \mathbf{X}}, \beta \right)$, which is again a specific case of the CLUB model in \eqref{CLUB_functional}.

%---------------------------------------------------
%
%	  Connection to Other Generative Models
%
%---------------------------------------------------
%
\subsection{Connection with State-of-the-Art Generative Models}\label{Ssec:ConnectionOtherGenerativeModels}

%We now discuss the main implications of Deep Variational CLUB objective. 

We now compare the the DVCLUB functionals \eqref{Eq:DVCLUB_Supervised} and \eqref{Eq:DVCLUB_UnSupervised} to other generative modeling approaches in the literature. Note that the latent variable generative models, like the VAE and GAN families, aim at capturing data distributions by minimizing specific discrepancy measures $\mathsf{dist} ( P_{\mathbf{X}}, P_{\boldsymbol{\theta}} (\mathbf{X}) ) $ between the true (but unknown) data distribution $P_{\mathbf{X}}$ and the generated model distribution $P_{\boldsymbol{\theta}} (\mathbf{X})$. Furthermore, note that maximizing the objective \eqref{Eq:DVCLUB_UnSupervised} is equivalent to maximizing the information utility $\I_{\boldsymbol{\phi}, \boldsymbol{\theta}} \! \left( \mathbf{X}; \mathbf{Z} \right)$, which is equivalent to minimizing the $\mathrm{KL}$-divergence discrepancy measure $\D_{\mathrm{KL}} \! \left( P_{\mathsf{D}} (\mathbf{X}) \, \Vert \, P_{\boldsymbol{\theta}} (\mathbf{X}) \right)$. 
Finally note that although the CLUB model is formulated using Shannon's mutual information, which leads us to self-consistent equations, one can use other information measures with different operational meanings, as mentioned in Footnote~\ref{footnote_generalize_CLUB}. 

%
%
%Before addressing the connections between DVCLUB model and other related models, we present a remark. 

%---------------------------------------------------
%	          Relation to VAEs family
%---------------------------------------------------
%
%\textbf{Relation to Auto-Encoder Models:}

In $\beta$-VAE \cite{higgins2016beta} the Lagrangian functional is defined as:
\begin{eqnarray}\label{Eq:betaVAE_Lagrangian}
\mathcal{L}_{\beta\text{-VAE}} \left( \boldsymbol{\phi}, \boldsymbol{\theta} , \beta \right) 
 \coloneqq   \mathbb{E}_{P_{\mathsf{D}} (\mathbf{X})} \left[  \mathbb{E}_{P_{\boldsymbol{\phi}} \left( \mathbf{Z} \mid \mathbf{X} \right) } \left[ \log P_{\boldsymbol{\theta}} \! \left( \mathbf{X} \! \mid \! \mathbf{Z} \right) \right] \right] 
-  \beta \,  \D_{\mathrm{KL}} \! \left( P_{\boldsymbol{\phi}} (\mathbf{Z} \! \mid \! \mathbf{X}) \, \Vert \, Q_{\boldsymbol{\psi}}  (\mathbf{Z}) \mid P_{\mathsf{D}}(\mathbf{X}) \right).
\end{eqnarray}
In the original VAE \cite{kingma2014auto} framework the parameter $\beta$ associated with the information complexity is set to $1$. Hence, VAE and $\beta$-VAE force $P_{\boldsymbol{\phi}} (\mathbf{Z} \! \mid \! \mathbf{X} \!= \! \mathbf{x})$ to match the proposal prior $Q_{\boldsymbol{\psi}}  (\mathbf{Z})$ for all input samples $\mathbf{x} \in \mathcal{X}$ drawn from $P_{\mathsf{D}} (\mathbf{X})$. However, there is no constraint to penalize the discrepancy between the learned aggregated posterior $P_{\boldsymbol{\phi}} (\mathbf{Z})$ and the proposal prior distribution $Q_{\boldsymbol{\psi}}  (\mathbf{Z})$. Clearly, $\beta$-VAE functional \eqref{Eq:betaVAE_Lagrangian} is a specific case of DVCLUB functional \eqref{Eq:DVCLUB_UnSupervised}.

The InfoVAE \cite{zhao2017infovae}, considered the following Lagrangian functional:
\begin{multline}\label{Eq:infoVAE_Lagrangian}
\mathcal{L}_{\text{InfoVAE}} \left( \boldsymbol{\phi}, \boldsymbol{\theta} , \beta \right) 
 \coloneqq  \mathbb{E}_{P_{\mathsf{D}} (\mathbf{X})} \left[  \mathbb{E}_{P_{\boldsymbol{\phi}} \left( \mathbf{Z} \mid \mathbf{X} \right) } \left[ \log P_{\boldsymbol{\theta}} \! \left( \mathbf{X} \! \mid \! \mathbf{Z} \right) \right] \right]  \\
-  \beta \,  \D_{\mathrm{KL}} \! \left( P_{\boldsymbol{\phi}} (\mathbf{Z} \! \mid \! \mathbf{X}) \, \Vert \, Q_{\boldsymbol{\psi}}  (\mathbf{Z}) \mid P_{\mathsf{D}}(\mathbf{X}) \right)
+ (\beta - \tau ) \, \D_{\mathrm{KL}} \! \left( P_{\boldsymbol{\phi}}(\mathbf{Z}) \, \Vert \, Q_{\boldsymbol{\psi}} (\mathbf{Z})\right) ,
\end{multline}
where $\tau > 0$. When $\beta = \tau$, we get $\beta$-VAE. Let $\tau =0$ and consider DVCLUB Lagrangian functional \eqref{Eq:DVCLUB_UnSupervised}, therefore, the first term of InfoVAE functional corresponds to the first term of information utility lower bound $\I_{\boldsymbol{\phi}, \boldsymbol{\theta}}^{\mathrm{L}} \left( \mathbf{U}; \mathbf{Z} \right)$, while the second and third terms correspond to the information complexity $\I_{\boldsymbol{\phi}, \boldsymbol{\psi}}   \left( \mathbf{X}; \mathbf{Z} \right)$. 
Hence, there is no term to capture a discrepancy between the observed distribution $P_{\mathsf{D}} (\mathbf{X})$ and the generated model distribution $P_{\boldsymbol{\theta}} (\mathbf{X})$. 
%
% \vfill
% \pagebreak

%---------------------------------------------------
%	          Relation to AAE
%---------------------------------------------------
%

%\begin{eqnarray}\label{Eq:AAE_Lagrangian}
%\mathcal{L}_{\mathrm{AAE}} \left( \boldsymbol{\phi}, \boldsymbol{\theta}, \boldsymbol{\psi},  \beta \right) 
%= \mathbb{E}_{P_{\mathsf{D}} (\mathbf{X})} \left[  \mathbb{E}_{P_{\boldsymbol{\phi}} \left( \mathbf{Z} \mid \mathbf{X} \right) } \left[ \log P_{\boldsymbol{\theta}} \! \left( \mathbf{X} \! \mid \! \mathbf{Z} \right) \right] \right] -
%\beta \, \D \! \left( P_{\boldsymbol{\phi}}(\mathbf{Z}) \, \Vert \, Q_{\boldsymbol{\psi}} (\mathbf{Z})\right) 
%\end{eqnarray}

%---------------------------------------------------
%	          Relation to GANs family
%---------------------------------------------------
%
%\textbf{Relation to Generative Adversarial Models:}

%\textcolor{gray}{Behrooz: to say about original GAN, f-GANs, WGAN, MMD-GAN, Energy-based GAN, Quadratic GAN}

GANs compute the optimal generator $g_{\boldsymbol{\theta}}^{\ast}$ by minimizing a distance between the observed distribution $P_{\mathsf{D}} (\mathbf{X})$ and the generated distribution $P_{\boldsymbol{\theta}} (\mathbf{X})$, without considering an explicit probability model for the observed data. 
The original GAN \cite{goodfellow2014generative} problem, also called the vanilla GAN, considers the following objective:
\begin{equation}\label{Eq:GAN_objective}
 %   \mathop{\min}_{g}
\mathcal{L}_{\mathrm{GAN}} \left( \boldsymbol{\theta}, \boldsymbol{\omega} \right)  \coloneqq  \;\;
    \mathbb{E}_{P_{\mathsf{D}}(\mathbf{X})} \left[ \, \log D_{\boldsymbol{\omega}} (\mathbf{X})  \, \right] + \mathbb{E}_{Q_{\boldsymbol{\psi}} (\mathbf{Z})} \left[ \, \log \left( 1 - D_{\boldsymbol{\omega}} ( \, g_{\boldsymbol{\theta}}(\mathbf{Z} ) \, ) \right) \, \right], 
\end{equation}
where first a random code $\mathbf{Z} \in \mathcal{Z}$ is sampled from a fixed distribution $Q_{\psi} (\mathbf{Z})$, next $\mathbf{Z}$ is mapped to the $\mathbf{X} \in \mathcal{X}$ using a deterministic generator (decoder) $g_{\boldsymbol{\theta}}: \mathcal{Z} \rightarrow \mathcal{X}$. 
The generator $g_{\boldsymbol{\theta}}$ and the visual space discriminator $D_{\boldsymbol{\omega}}$ neural networks are trained adversarially:
\begin{equation}\label{Eq:GAN_objective_minmax}
   \mathop{\min}_{\boldsymbol{\theta}} \mathop{\max}_{\boldsymbol{\omega}} \;\;
\mathcal{L}_{\mathrm{GAN}} \left( \boldsymbol{\theta}, \boldsymbol{\omega} \right) .
\end{equation}
To see its relation to our DVCULB model, consider the visible space discriminator training step \eqref{Eq:TrainVisibleSpaceDiscriminator} and utility decoder adversarial training step \eqref{TrainPriorDistributionGeneratorAndUtilityDecoderAdversarially}. Inspired by the original formulation of the GAN problem \cite{goodfellow2014generative}, abundant characterizations, generalizations, and applications have since been proposed \cite{mirza2014conditional, makhzani2015adversarial, salimans2016improved, ho2016generative, dumoulin2016adversarially, zhao2016energy, nowozin2016f, isola2017image, zhu2017unpaired, arjovsky2017wasserstein, mao2017least, zhang2019self}.

\begin{remark}
In the Euclidean space we can represent any convex function as the point-wise supremum of a family of affine functions, and vice versa. Noting that $f$-divergence is a convex function of probability measures, we can represent it as a point-wise supremum of affine functions. Let $f: \mathcal{X} \rightarrow \mathbb{R}$ be a convex and lower semi-continuous function. The convex conjugate (also known as the Fenchel dual or Legendre transform) $f^{\ast}: \mathcal{X} \rightarrow \mathbb{R}$ of $f$ is defined by $f^{\ast} (\mathbf{v}) \coloneqq \mathop{\sup}_{\mathbf{c} \in \mathsf{dom} (f)} \, \left\{ \langle  \mathbf{c},  \mathbf{v} \rangle - f(\mathbf{c}) \right\}$, where $\mathsf{dom} (f)$ denotes the effective domain of $f$ which is given by $\mathsf{dom} (f) = \{ \mathbf{c} \mid f(\mathbf{c}) < \infty \} $ \cite{nguyen2010estimating, wu2017lecture, nowozin2016f}. Hence, for any $\mathbf{v} \in \mathsf{dom} ( f^{\ast} )$ and $\mathbf{c} \in \mathsf{dom} ( f  )$ we have $f(\mathbf{c}) \geq \mathbf{c} \mathbf{v} - f^{\ast} (\mathbf{v})$ (Fenchel-Young inequality). Using the notion of convex conjugate one can obtain a variational representation of a \textit{well-defined} $f$-divergence in terms of the convex conjugate of $f$:
\begin{subequations}\label{Eq:VariationalBound_f_divergences}
\begin{align}
    \D_f \! \left( P \Vert  Q \right) = \int_{\mathcal{X}} Q(\mathbf{x}) f\left( \frac{P(\mathbf{x})}{Q(\mathbf{x})} \right) \mathrm{d}\mathbf{x} &=  
    \int_{\mathcal{X}} Q(\mathbf{x})\, \mathop{\sup}_{\mathbf{c} \in \mathsf{dom} (f^{\ast})} \left\{ \mathbf{c} \; \frac{P(\mathbf{x})}{Q(\mathbf{x})} - f^{\ast} (\mathbf{c}) \right\} \mathrm{d}\mathbf{x}   \\
    & \geq    \mathop{\sup}_{C : \mathcal{X} \rightarrow \mathbb{R}} \left( \int_{\mathcal{X}} P(\mathbf{x})\, C(\mathbf{x})\, \mathrm{d}\mathbf{x}   - \int_{\mathcal{X}} Q(\mathbf{x})\, f^{\ast} (C(\mathbf{x})) \, \mathrm{d}\mathbf{x} \right)   \\
     & =     \mathop{\sup}_{C : \mathcal{X} \rightarrow \mathbb{R}} \left( \,  \mathbb{E}_{P} \left[ \,  C(\mathbf{X}) \,  \right]  + \mathbb{E}_{Q} \left[ \, - f^{\ast} (C(\mathbf{X})) \, \right] \, \right), \label{Eq:VariationalLowerBound_f_divergences}
\end{align}
\end{subequations} 
where the supremum is taken over all measurable functions $C: \mathcal{X} \rightarrow \mathbb{R}$. Under mild conditions on $f$ \cite{nguyen2010estimating}, the bound is tight for $C^{\ast} (x) = f^{\prime} \left( \frac{ p(x) }{ q(x) } \right)$, where $f^{\prime}$ denotes the first derivative of $f$. 
\end{remark}

Using the above remark, we now suppose $C ( \mathbf{x} )$ is a variational function parameterized by $\boldsymbol{\omega}$, and represent it as $C_{\boldsymbol{\omega}} ( \mathbf{x} ) = h \left( V_{\boldsymbol{\omega}} ( \mathbf{x} ) \right)$, where $V_{\boldsymbol{\omega}}: \mathcal{X} \rightarrow \mathbb{R}$, and $h: \mathbb{R} \rightarrow \mathsf{dom} (f^{\ast})$ is an output activation function. We call $C_{\boldsymbol{\omega}}: \mathcal{X} \rightarrow \mathbb{R}$ as the \textit{critic} function. Choosing $f(\mathbf{c}) = \mathbf{c} \log \mathbf{c} - (\mathbf{c} + 1) \log (\frac{\mathbf{c} + 1}{2})$ gives us the Jensen-Shannon's divergence (JSD) and we have $f^{\ast} (\mathbf{v}) = - \log \left( 1 - e^{\mathbf{v}} \right)$. 
Using the JSD and choosing the output activation function as $h( t ) = - \log (1 + e^{-t})$, the variational lower bound \eqref{Eq:VariationalLowerBound_f_divergences} reduces to the original GAN objective \eqref{Eq:GAN_objective}, where $C_{\boldsymbol{\omega}} ( \mathbf{x} ) = h(V_{\boldsymbol{\omega}} (\mathbf{x})) = \log D_{\boldsymbol{\omega}} (\mathbf{x})$, and $- f^{\ast} \left( \log D_{\boldsymbol{\omega}} (\mathbf{x}) \right) = \log ( 1 - D_{\boldsymbol{\omega}} (\mathbf{x}))$.

\pagebreak
 
%---------------------------------------------------
%
%	  Connection with Optimal Transport Cost
%
%---------------------------------------------------
%
%\textbf{Relation to Optimal Transport Cost, Wasserstein Distance:}

\begin{remark}[Optimal Transport Problem]
A recent body of work study generative models from an optimal transport (OT) point of view. 
Let $\mathcal{X}$ and $\mathcal{Y}$ be two measurable spaces, and let $\mathcal{P} (\mathcal{X})$ and $\mathcal{P} (\mathcal{Y})$ be the sets of all positive Radon probability measures on $\mathcal{X}$ and $\mathcal{Y}$, respectively. 
For any measurable non-negative cost function $\mathsf{c}: \mathcal{X} \times \mathcal{Y} \rightarrow \mathbb{R}^+$, the optimal transport problem (Kantorovich problem) between distributions $P \in \mathcal{P} (\mathcal{X})$ and $Q \in \mathcal{P} (\mathcal{Y})$ is defined as \cite{villani2008optimal, peyre2019computational}:\vspace{-4pt}
\begin{equation}\label{Eq:KantorovichProblem}
    \mathsf{OT}_{\mathsf{c}} \left( P , Q \right) \coloneqq  \mathop{\inf}_{\boldsymbol{\pi} \in \Pi ( P , Q )}  \int_{\mathcal{X} \times \mathcal{Y}} \mathsf{c}(\mathbf{x}, \mathbf{y}) \, \mathrm{d} \boldsymbol{\pi} (\mathbf{x}, \mathbf{y}) = \mathop{\inf}_{\boldsymbol{\pi} \in \Pi ( P , Q )} \mathbb{E}_{\boldsymbol{\pi}} \left[ \, \mathsf{c}(\mathbf{X}, \mathbf{Y}) \, \right],
\end{equation}
where $\Pi ( P  , Q )$ denotes the set of joint distributions (couplings) over the product space $\mathcal{X} \times \mathcal{Y}$ with marginals $P$ and $Q$, respectively. That is, for all measurable sets $\mathcal{A} \subset \mathcal{X} $ and $\mathcal{B} \subset \mathcal{Y}$, we have:\vspace{-4pt}
\begin{equation}
    \Pi ( P , Q)  \coloneqq  \left\{ \pi \in \mathcal{P} ( \mathcal{X} \times \mathcal{Y} ): \; \pi(\mathcal{A} \times \mathcal{Y}) = P (\mathcal{A}), \, \pi (\mathcal{X} \times \mathcal{B}) = Q(\mathcal{B}) \right\}. 
\end{equation}
The joint measures $\boldsymbol{\pi} \in \Pi (P, Q)$ are called the \textit{transport plans}. The cost function $\mathsf{c}$ represents the cost to move a unit of mass from $\mathbf{x}$ to $\mathbf{y}$. 
\end{remark}

\vspace{-12pt}

\begin{remark}[Dual Formulation]
The Kantorovich problem \eqref{Eq:KantorovichProblem} defines a constrained linear program, and hence admits an equivalent dual formulation:\vspace{-5pt}
\begin{subequations}\label{Eq:KantorovichProblem_dual}
\begin{align}
    \mathsf{OT}_{\mathsf{c}} \left( P , Q \right) & \coloneqq   \mathop{\sup}_{(h_1, h_2) \in \mathcal{R} (\mathsf{c})}  \int_{\mathcal{X}} h_1(\mathbf{x}) \; \mathrm{d}P(\mathbf{x}) + \int_{\mathcal{Y}} h_2(\mathbf{y})\; \mathrm{d}Q(\mathbf{y}) \\
    & =  \mathop{\sup}_{(h_1, h_2) \in \mathcal{R} (\mathsf{c})}  \mathbb{E}_{P} \left[ \,  h_1 (\mathbf{X})  \, \right] + \mathbb{E}_{Q} \left[ \,  h_2 (\mathbf{Y})  \, \right], 
\end{align}
\end{subequations} 
where for any $\mathsf{c} \in \mathcal{C} (\mathcal{X} \times \mathcal{Y})$, the set of admissible \textit{dual potentials} (also known as Kantorovich potentials) is:\vspace{-4pt}
\begin{equation}
    \mathcal{R} (\mathsf{c}) \coloneqq  \left\{ \,  (h_1 , h_2 ) \in \mathcal{C} (\mathcal{X}) \times \mathcal{C}(\mathcal{Y})  : \; h_1 (\mathbf{x}) + h_2 (\mathbf{y}) \leq \mathsf{c} (\mathbf{x}, \mathbf{y}), \, \forall (\mathbf{x}, \mathbf{y}) \in \mathcal{X} \times \mathcal{Y} \, \right\}, 
\end{equation}
where $\mathcal{C} (\mathcal{X})$ and $\mathcal{C} (\mathcal{Y})$ are the space of continuous (real-valued) functions on $\mathcal{X}$ and $\mathcal{Y}$, respectively. 
\end{remark}

For any $h_2 \in \mathcal{C}(\mathcal{Y})$, let us define its $\mathsf{c}$-transform\footnote{The $\mathsf{c}$-transform is a generalization of the Legendre transform from convex analysis. If $\mathsf{c} (\mathbf{x}, \mathbf{y}) = \langle  \mathbf{x},  \mathbf{y} \rangle $ on $\mathbb{R}^n \times \mathbb{R}^n$, the $\mathsf{c}$-transform coincides with the Legendre transform.} $h_1^{\mathsf{c}} \in \mathcal{C}(\mathcal{Y}) $ of $h_1 \in \mathcal{C}(\mathcal{X})$ as:\vspace{-4pt}
\begin{equation}
    h_1^{\mathsf{c}} (\mathbf{y}) \coloneqq  \mathop{\inf}_{\mathbf{x} \in \mathcal{X}} \mathsf{c}( \mathbf{x}, \mathbf{y} ) - h_1 (\mathbf{x}), \quad \forall \mathbf{y} \in \mathcal{Y}. 
\end{equation}
Given a candidate potential $h_1$ for the first variable, $ h_1^{\mathsf{c}}$ is the best possible potential that can be paired with $h_1$. 
A function $h_1^{\mathsf{c}}$ in the form above is called a $\mathsf{c}$-concave function. 
Note that Kantorovich potentials satisfy $h_2 = h_1^{\mathsf{c}}$. Denoting $h_1 = h$ we can rewrite the dual formulation in \eqref{Eq:KantorovichProblem_dual} as:\vspace{-4pt}
\begin{eqnarray}
    \mathsf{OT}_{\mathsf{c}} \left( P , Q \right) = \mathop{\sup}_{h \in  \mathcal{C}(\mathcal{X})  }   \mathbb{E}_{P} \left[ \,  h (\mathbf{X})  \, \right] + \mathbb{E}_{Q} \left[ \,  h^{\mathsf{c}} (\mathbf{Y})  \, \right]   .
\end{eqnarray}

% In general, the constraint set $ \mathcal{R} (\mathsf{c})$ is not compact, however, in the case $\mathsf{c} (\mathbf{x}, \mathbf{y}) = \mathsf{d}^p (\mathbf{x}, \mathbf{y})$ for some $p \geq 1$, where $\mathsf{d}$ is a distance function (metric), the optimal $(h_1 , h_2)$ are Lipschitz regular, which enables us to replace the constraint by a compact one \cite{peyre2019computational}.

% \pagebreak

\begin{remark}[Wasserstein Distance]
In the Kantorovich problem \eqref{Eq:KantorovichProblem}, assume $\mathcal{X} = \mathcal{Y}$, let $(\mathcal{X}, \mathsf{d})$ be a metric space, and consider $\mathsf{c} (\mathbf{x}, \mathbf{y}) = \mathsf{d}^p (\mathbf{x}, \mathbf{y})$ for some $p \geq 1$. In this case, the $p$-Wasserstein distance\footnote{The $p$-Wasserstein satisfies the three metric axioms, hence it defines an actual distance between $P$ and $Q$. Also, note that $ \mathcal{W}_p$ depends~on~$\mathsf{d}$. } on $\mathcal{X}$ is defined as:\vspace{-4pt}
\begin{equation}\label{Eq:p-Wasserstein}
    \mathcal{W}_p (P, Q) \coloneqq  {\mathsf{OT}_{\mathsf{d}^p} \left( P , Q \right)}^{1 / p} .
\end{equation}
\end{remark}

The cases $p=1$ and $p=2$ are particularly interesting. The $1$-Wasserstein distance is more flexible and easier to bound, moreover, the Kantorovich-Rubinstein duality holds for the $1$-Wasserstein distance. The $2$-Wasserstein distance is more appropriate to reflect the geometric features, moreover, it scales better with the dimension. 
Let $\mathrm{L}_{p}(h) = \mathop{\sup}_{\mathbf{x}, \mathbf{y} \in \mathcal{X} } \{ \frac{\vert h(\mathbf{x}) - h(\mathbf{y}) \vert}{\mathsf{d}^p (\mathbf{x}, \mathbf{y})} : \,   \mathbf{x} \neq \mathbf{y} \}$ denotes the Lipschitz constant of a function $h \in \mathcal{C} (\mathcal{X})$ with respect to cost $\mathsf{c} (\mathbf{x}, \mathbf{y}) = \mathsf{d}^p (\mathbf{x}, \mathbf{y})$. 
One can show that if $\mathrm{L}_{p}(h) \leq 1$, then $h^{\mathsf{c}} = - h$. 
Let $\mathcal{H}_{p,k} = \{ h \in \mathcal{C}(\mathcal{X}): \, \mathrm{L}_{p}(h) \leq k \}$ is the set of all bounded $\mathrm{L}_{p}(h)$-Lipschitz functions on $(\mathcal{X}, \mathsf{d})$ with $\mathsf{c} (\mathbf{x}, \mathbf{y}) = \mathsf{d}^p (\mathbf{x}, \mathbf{y})$ such that $\mathrm{L}_{p}(h) \leq k$. 
The $\mathcal{W}_1 (P, Q)$ can be rewritten as:\vspace{-6pt}
\begin{equation}\label{Eq:1-WassersteinDistance_dualFormulation}
    \mathcal{W}_1 (P, Q) = \mathop{\sup}_{h \in \mathcal{H}_{1,1} }   \mathbb{E}_{P} \left[ \,  h (\mathbf{X})  \, \right] - \mathbb{E}_{Q} \left[ \,  h (\mathbf{Y})  \, \right]. 
\end{equation}

Note that, as stated before, the latent variable generative models aim at capturing data distributions by minimizing specific discrepancy measures between the true (but unknown) data distribution $P_{\mathbf{X}}$ and the generated model distribution $P_{\boldsymbol{\theta}} (\mathbf{X})$. Moreover, in practice, only an empirical data distribution $P_{\mathsf{D}} (\mathbf{X})$ is available. 
In the Optimal Transport problem, one can factor the mapping from $\mathbf{X} \in \mathcal{X}$ to $\mathbf{Y} \in \mathcal{X}$, i.e., the couplings $\Pi (P, Q)$, through a latent code $\mathbf{Z} \in \mathcal{Z}$. This shed light on the connections between the latent variable generative models and Optimal Transport problem. 

Let $\mathbf{Y} = g_{\boldsymbol{\theta}} (\mathbf{Z}) \in \mathcal{X}$ denotes the generated samples by a generator $g_{\boldsymbol{\theta}}: \mathcal{Z} \rightarrow \mathcal{X}$ and consider the Kantorovich-Rubinstein  duality formulation \eqref{Eq:1-WassersteinDistance_dualFormulation}.  
By restricting the dual potential $h$ to have a parametric form, i.e., $h = D_{\boldsymbol{\omega}}: \mathcal{X} \rightarrow \mathbb{R} $, the Wasserstein GAN (WGAN) \cite{arjovsky2017wasserstein} considers the following objective:\vspace{-4pt}
\begin{equation}\label{Eq:WassersteinGAN_objective}
 %   \mathop{\min}_{g}
\mathop{\min}_{\boldsymbol{\theta}} \mathop{\max}_{\boldsymbol{\omega}} \;\; \mathcal{L}_{\mathrm{WGAN}} \left( \boldsymbol{\theta}, \boldsymbol{\omega} \right)  =  \;\;
    \mathbb{E}_{P_{\mathsf{D}}(\mathbf{X})} \left[ \,  D_{\boldsymbol{\omega}} (\mathbf{X})  \, \right] - \mathbb{E}_{Q_{\boldsymbol{\psi}} (\mathbf{Z})} \left[ \,    D_{\boldsymbol{\omega}} \left( \, g_{\boldsymbol{\theta}}(\mathbf{Z} ) \right)  \, \right], 
\end{equation}
where the $1$-Lipschitz constraint in \eqref{Eq:1-WassersteinDistance_dualFormulation} is satisfied by using a deep neural network with ReLu units.

Let $P_{\boldsymbol{\theta}} ( \mathbf{Y} \! \mid \! \mathbf{Z}= \mathbf{z} ) = \delta (\mathbf{y} - g_{\boldsymbol{\theta}} (\mathbf{z}) ), \forall \mathbf{z} \in \mathcal{Z}$, i.e., suppose $\mathbf{Y} \sim P_{\boldsymbol{\theta}} (\mathbf{X})$ is defined with a deterministic mapping, the parameterized Kantorovich problem associated with \eqref{Eq:KantorovichProblem} can be expressed as follows:\footnote{An almost similar formulation holds for the case in which $P_{\boldsymbol{\theta}} (\mathbf{Y} \! \mid \! \mathbf{Z})$ are not necessarily Dirac. We refer the reader to \cite{bousquet2017optimal, tolstikhin2018wasserstein} for more details.}:\vspace{-4pt}
\begin{equation}\label{Eq:KantorovichProblem_factorThroughZ}
    \mathsf{OT}_{\mathsf{c}} \left( P_{\mathsf{D}}(\mathbf{X}) , P_{\boldsymbol{\theta}} (\mathbf{X} ) \right) = \!\! \mathop{\inf}_{\boldsymbol{\pi} \in \Pi ( P_{\mathsf{D}}(\mathbf{X}) , P_{\boldsymbol{\theta}} (\mathbf{X} ) )} \mathbb{E}_{\boldsymbol{\pi}} \left[ \, \mathsf{c}(\mathbf{X}, \mathbf{Y}) \, \right] = 
    \!\! \mathop{\inf}_{\substack{P_{\boldsymbol{\phi}}  (\mathbf{Z} \mid \mathbf{X}):\\ P_{\boldsymbol{\phi}}(\mathbf{Z}) = Q_{\boldsymbol{\psi}}(\mathbf{Z}) }} \!\! \mathbb{E}_{P_{\mathsf{D}}(\mathbf{X})} \left[ \mathbb{E}_{P_{\boldsymbol{\phi}} (\mathbf{Z} \mid \mathbf{X})} \left[ \, \mathsf{c} \left( \mathbf{X}, g_{\boldsymbol{\theta}} (\mathbf{Z}) \right) \, \right] \right]. 
\end{equation}
The Wasserstein Auto-Encoder (WAE) \cite{tolstikhin2018wasserstein} is formulated as the relaxed unconstrained parameterized OT \eqref{Eq:KantorovichProblem_factorThroughZ}, and reads as:\vspace{-8pt}
\begin{equation}\label{Eq:WassersteinAE_objective}
 %   \mathop{\min}_{g}
\mathop{\min}_{\boldsymbol{\phi}, \boldsymbol{\theta}} \;\; \mathcal{L}_{\mathrm{WAE}} \left( \boldsymbol{\phi}, \boldsymbol{\theta} \right)   \coloneqq \;\;
    \mathbb{E}_{P_{\mathsf{D}} (\mathbf{X})} \left[  \mathbb{E}_{P_{\boldsymbol{\phi}} \left( \mathbf{Z} \mid \mathbf{X} \right) } \left[ \, \mathsf{c}(\mathbf{X}, g_{\boldsymbol{\theta}} (\mathbf{Z})) \, \right] \right] 
+  \lambda^\prime \,  \mathsf{dist} ( P_{\boldsymbol{\phi}} (\mathbf{Z}) , Q_{\boldsymbol{\psi}} (\mathbf{Z})  ) ,
\end{equation}
where $\lambda^{\prime} \geq 0$ is a regularization parameter, and $\mathsf{dist} ( P_{\boldsymbol{\phi}} (\mathbf{Z}) , Q_{\boldsymbol{\psi}} (\mathbf{Z}) )$ is a discrepancy between  $P_{\boldsymbol{\phi}} (\mathbf{Z})$ and $Q_{\boldsymbol{\psi}}  (\mathbf{Z})$. For instance, one can consider $\mathsf{dist} ( P_{\boldsymbol{\phi}} (\mathbf{Z}) , Q_{\boldsymbol{\psi}} (\mathbf{Z}) ) =  \D_f (P_{\boldsymbol{\phi}} (\mathbf{Z}) \Vert Q_{\boldsymbol{\psi}} (\mathbf{Z}) )$, or alternatively, one can use the Maximum Mean Discrepancy (MMD) for a characteristic positive-definite reproducing kernel \cite{tolstikhin2018wasserstein}. 
Note that, in contrast to $\beta$-VAE \eqref{Eq:betaVAE_Lagrangian}, the WAE \cite{tolstikhin2018wasserstein} directly captures discrepancy between aggregated posterior $P_{\boldsymbol{\phi}} (\mathbf{Z})$ and proposal prior $Q_{\boldsymbol{\psi}}  (\mathbf{Z})$, and ignoring the conditional relative entropy $\D_{\mathrm{KL}} \! \left( P_{\boldsymbol{\phi}} (\mathbf{Z} \! \mid \! \mathbf{X}) \, \Vert \, Q_{\boldsymbol{\psi}}  (\mathbf{Z}) \mid P_{\mathsf{D}}(\mathbf{X}) \right)$. To see its connection with CLUB model, consider the DVCLUB objective \eqref{Eq:DVCLUB_UnSupervised}. Note that maximizing the information utility $\I_{\boldsymbol{\phi}, \boldsymbol{\theta}}^{\mathrm{L}} (\mathbf{X}; \mathbf{Z})$ is equivalent to minimizing the divergence measure between $P_{\mathsf{D}} (\mathbf{X})$ and $P_{\boldsymbol{\theta}} (\mathbf{X})$, as well as, minimizing the \textit{negative} log-likelihood $ \mathbb{E}_{P_{\mathsf{D}} (\mathbf{X})} \left[  \mathbb{E}_{P_{\boldsymbol{\phi}} \left( \mathbf{Z} \mid \mathbf{X} \right) } \left[ - \log P_{\boldsymbol{\theta}} \! \left( \mathbf{X} \! \mid \! \mathbf{Z} \right) \right] \right]$.

In WAE objective \eqref{Eq:WassersteinAE_objective}, let $\mathsf{c} (\mathbf{x}, \mathbf{y}) = \mathsf{d}^2 (\mathbf{x}, \mathbf{y})$. This will lead us to the Adversarial Auto-Encoders (AAEs) model \cite{makhzani2015adversarial}. This means that the AAEs minimizes $2$-Wasserstein distance between $P_{\mathsf{D}} (\mathbf{X})$ and $P_{\boldsymbol{\theta}} (\mathbf{X})$. Hence, the AAE objective reads as:\vspace{-9pt}
\begin{equation}\label{Eq:AAE_objective}
\mathop{\min}_{\boldsymbol{\phi}, \boldsymbol{\theta}} \;\; \mathcal{L}_{\mathrm{AAE}} \left( \boldsymbol{\phi}, \boldsymbol{\theta} \right)   \coloneqq  \;\;
    \mathbb{E}_{P_{\mathsf{D}} (\mathbf{X})} \left[  \mathbb{E}_{P_{\boldsymbol{\phi}} \left( \mathbf{Z} \mid \mathbf{X} \right) } \left[ \,  - \log P_{\boldsymbol{\theta}} (\mathbf{X} \! \mid \! \mathbf{Z})  \, \right] \right] 
+  \lambda^\prime \,  \mathsf{dist} ( P_{\boldsymbol{\phi}} (\mathbf{Z}) , Q_{\boldsymbol{\psi}} (\mathbf{Z})  ).
\end{equation}

%Let ${\mathcal{L}}_1 (P)  \coloneqq \{  f: \mathcal{X} \rightarrow \mathbb{R}  \mid  \mathbb{E}_{P} \left[  f \right] < \infty \}$ denotes the set of integrable functions with respect to $P$. 
%Let the constraint set $\mathcal{R}(\mathsf{c})$ also satisfies $f_1 \in {\mathcal{L}}_1 (P)$ and $f_2 \in {\mathcal{L}}_1 (P)$. 
%

%\clearpage

%---------------------------------------------------
%
%	  Connection to Generative Compression
%
%---------------------------------------------------
%
\subsection{Connection with Modern Data Compression Models}
\label{Ssec:ConnectionModernCompression}

In the era of big data, with recent advances in modern computation environments coupled with growing concern about the `\textit{storage}', `\textit{communication}' and `\textit{process}' of data, it is desirable to emerge new models for data compression while simultaneously satisfying some \textit{privacy constraints}. Modern data compression research may try to study fundamental challenges in multi-terminal distributed compression models, find new perceptual metrics, address compression techniques of/with neural networks, study fundamentals of quantum compression of quantum or classical information, and pave the way toward data compression schemes in genomics and astronomy, to name~a~few. 

In general, the data compression schemes can be \textit{lossless} or \textit{lossy}. 
The principle engineering objective of lossless data compression schemes is to construct (assign) new representations (representatives) of given data with minimal possible description length, without information loss\footnote{Note that based on Definition~\ref{Def:MinimalSufficientStatistic} in Sec.~\ref{Ssec:Relevant Information}, a minimal sufficient statistics $\mathbf{Z}$ maximally compresses the information about $\mathbf{U}$ in the data~$\mathbf{X}$.}. 
Shannon introduced the theoretical discipline of treating data compression which is regardless of a specific coding method \cite{shannon1948mathematical}. The shortest description rate (achievable data compression limit) is the entropy $\H (\mathbf{X})$. The lossless compression schemes are only possible for discrete random variables and also require a priori knowledge of the data distribution.

% During compression, we remove all the redundancy in the data to form the most compressed version possible. 
% 

% as well as an entropy coding algorithm that transforms data to a bit-stream, using the distribution. 

Following Shannon's lead, lossy compression schemes are typically studied and analyzed through the lens of Shannon's rate-distortion theorem. 
Shannon's rate-distortion theorem gives us the minimal (infimal) rate required by an optimal encoder to achieve a particular distortion. 
The development of practical codes for lossy compression of data at rates approaching Shannon's rate-distortion bound is one of the significant problems in information theory.% \cite{}. 

The classical data compression techniques mostly rely on a \textit{transform coding} scheme \cite{goyal2001theoretical}, which has three components: (i) transform, (ii) quantization, and (iii) entropy coding, which operate successively and independently. The key idea of transform coding is that the data may be more effectively processed in the transform domain (latent space) than in the original data domain \cite{goyal2001theoretical}. The state-of-the-art literature on transform coding almost always studied and optimized linear transforms. For example, the well-known JPEG\footnote{JPEG stands for Joint Photographic Experts Group.} lossy image compression scheme is based on the discrete cosine transform (DCT) and JPEG~2000, an improved version of the JPEG, is based on the discrete wavelet transform (DWT). In a typical transform coding scheme, an encoder maps any data $\mathbf{X}$ into a transform (latent) space using an invertible analysis transform. Since the transform is invertible, it is information preserving. Next, the transform coefficients (latent space dimensions) are quantized independently using scalar quantization. Finally, the resulted discrete representations are encoded using a lossless entropy code, such as arithmetic coding or Huffman coding.

\pagebreak

% While this is a classic topic with roots in the 1980s,...

Although the idea of using neural networks for data compression dates back to the late 1980s \cite{sonehara1989image, luttrell1989image, anthony1990comparison, sicuranza1990artificial, setiono1994image, dony1995neural, namphol1996image, jiang1999image}, just very recently the research community turned its attention to learned data compression models \cite{toderici2015variable, gregor2016towards, balle2016end, balle2017variational,theis2017lossy, toderici2017full, agustsson2017soft, rippel2017real, santurkar2018generative, mentzer2018conditional, tschannen2018deep, minnen2018joint, balle2018variational, ulyanov2018deep, agustsson2019generative, lee2019context, choi2019variable, hu2021learning, yang2022introduction, gregorova2021learned}, thanks to recent advances in storage, communication, and computation facilities, as well as the introduction of deep generative models, such as autoregressive models \cite{larochelle2011neural}, GANs \cite{goodfellow2014generative} VAEs \cite{kingma2014auto}, and normalizing flows \cite{rezende2015variational}.

%%\cite{toderici2015variable, gregor2016towards, balle2016end, theis2017lossy, agustsson2017soft, santurkar2018generative, mentzer2018conditional, tschannen2018deep, minnen2018joint, balle2018variational, ulyanov2018deep, agustsson2019generative, lee2018context, hu2021learning, yang2022introduction}, 

% Distributed and multi terminal compression
%  Compression under new and emerging perceptual metrics
%  Rate - Distortion - Complexity - Delay tradeoffs
%  Joint source channel coding/Joint compression and error correction
%  Compression meets encryption/privacy/security
%  Random access and computation in the compressed domain
%  Compression with, for and of neural networks
%  Compression of graphs and non-traditional data structures
%  Specialized compression for: genomics, multimedia, VR, point clouds, sensors, astronomy, etc.
%  Quantum compression
%  Compression, learning, prediction and information processing
%  Theory of Compression
%  Model compression
%  Universality in data compression

In line with recent advances in neural data compression, we now relate our CLUB model to generative compression. The deterministic CLUB (DCLUB) model encourages to have a deterministic encoding function. Considering the Markov chain $( \mathbf{U}, \mathbf{S} )  \markov \mathbf{X} \markov \mathbf{Z}$, the DCLUB functional can be expressed as:\vspace{-6pt}
\begin{equation}\label{Eq:DeterministicCLUB}
\mathsf{DCLUB} \left( R^{\mathrm{u}}, R^{\mathrm{s}}, P_{\mathbf{U}, \mathbf{S}, \mathbf{X}}\right)  \coloneqq \mathop{\inf}_{\substack{ P_{\mathbf{Z} \mid \mathbf{X}}: \\\left( \mathbf{U}, \mathbf{S}\right) \markov \mathbf{X} \markov \mathbf{Z}}} 
\H  \left( \mathbf{Z} \right)  \; \quad  \; \mathrm{s.t.}  \quad   \I \left( \mathbf{U}; \mathbf{Z} \right) \geq R^{\mathrm{u}}, \;\; 
\I \left( \mathbf{S}; \mathbf{Z} \right) \leq R^{\mathrm{s}}.\vspace{-2pt}
\end{equation}
Hence, the associated Lagrangian functional is given as:\vspace{-8pt}
\begin{subequations}
\begin{align}
\mathcal{L}_{\mathrm{DCLUB}} \left( P_{\mathbf{Z}\mid \mathbf{X}}, \gamma , \lambda \right)   
&  \coloneqq     \H  \left( \mathbf{Z} \right) - \gamma \,   \I \left( \mathbf{U}; \mathbf{Z} \right) +\lambda  \, \I \left( \mathbf{S}; \mathbf{Z} \right) \\
& \equiv    \H  \left( \mathbf{Z} \right) + \gamma \,   \H \left( \mathbf{U} \mid \mathbf{Z} \right) - \lambda  \, \H \left( \mathbf{S} \mid \mathbf{Z} \right).\vspace{-5pt}% \\
%& \equiv  \H \left( \mathbf{Z} \mid \mathbf{U} \right)  
\end{align}
\end{subequations}
Clearly, DIB model \eqref{Eq:DeterministicIB} is a specific case of the DCLUB \eqref{Eq:DeterministicCLUB}; and hence, the CLUB model \eqref{CLUB_functional_infimum}\footnote{Motivated researchers can study and analyze deep variational DCLUB, taking similar steps as we addressed in Sec.~\ref{Sec:DVCLUB}. Furthermore, using the depicted $\I$-diagram in Fig.~\ref{Fig:I-diagram}, one can define \textit{residual CLUB (RCLUB)} and \textit{conditional CLUB (CCLUB)} models. The CCLUB model can be viewed as a generalization of the conditional entropy bottleneck (CEB) model \cite{fischer2020conditional}.}.

The associated deep variational DCLUB (DVDCLUB) Lagrangian functionals in the supervised and unsupervised scenarios can be obtained by simplifying corresponding parameterized variational approximations. 
Considering a deterministic encoding function, the utility attribute prediction fidelity $- \H_{\boldsymbol{\phi}, \boldsymbol{\theta}} \! \left( \mathbf{U} \! \mid \! \mathbf{Z} \right)$ in Eq.~\eqref{Eq:I_UZ_phi_theta_SecondDecomposition} is given as
$ \mathbb{E}_{P_{\mathbf{U}, \mathbf{X}}} \left[  \mathbb{E}_{P_{\boldsymbol{\phi}} \left( \mathbf{Z} \mid \mathbf{X} \right) } \left[ \log P_{\boldsymbol{\theta}} \! \left( \mathbf{U} \! \mid \! \mathbf{Z} \right) \right] \right] \equiv \mathbb{E}_{P_{\mathbf{U}, \mathbf{X}, \mathbf{Z}}}  \left[ \log P_{\boldsymbol{\theta}} \! \left( \mathbf{U} \! \mid \! \mathbf{Z} \right) \right] \eqqcolon - \H_{\boldsymbol{\theta}} \! \left( \mathbf{U} \! \mid \! \mathbf{Z} \right)  $. 
% $ \mathbb{E}_{P_{\mathsf{D}} (\mathbf{X})} \left[  \mathbb{E}_{P_{\boldsymbol{\phi}} \left( \mathbf{Z} \mid \mathbf{X} \right) } \left[ \log P_{\boldsymbol{\theta}} \! \left( \mathbf{U} \! \mid \! \mathbf{Z} \right) \right] \right] \equiv \mathbb{E}_{P_{\mathsf{D}} (\mathbf{X})}  \left[ \log P_{\boldsymbol{\theta}} \! \left( \mathbf{U} \! \mid \! \mathbf{Z} \right) \right] \eqqcolon - \H_{\boldsymbol{\theta}} \! \left( \mathbf{U} \! \mid \! \mathbf{Z} \right)  $. 
%  $\mathbb{E}_{P_{\mathsf{D}} (\mathbf{X})}  \left[ \log P_{\boldsymbol{\theta}} \! \left( \mathbf{X} \! \mid \! \mathbf{Z} \right) \right] $
Hence, $\I_{\boldsymbol{\phi}, \boldsymbol{\theta}} \! \left( \mathbf{U}; \mathbf{Z} \right) $ in Eq.~\eqref{Eq:I_UZ_phi_theta_SecondDecomposition} reduces to $ \I_{\boldsymbol{\theta}} \! \left( \mathbf{U}; \mathbf{Z} \right) \geq - \H_{ \boldsymbol{\theta}} \! \left( \mathbf{U} \! \mid  \! \mathbf{Z} \right) - \D_{\mathrm{KL}} \left( P_{\mathbf{U}} \, \Vert \, P_{\boldsymbol{\theta}} (\mathbf{U}) \right) \eqqcolon 
\I_{\boldsymbol{\theta}}^{\mathrm{L}} \! \left( \mathbf{U}; \mathbf{Z} \right) $. Analogously, $\I_{\boldsymbol{\phi}, \boldsymbol{\theta}} \! \left( \mathbf{X}; \mathbf{Z} \right) $ in Eq.~\eqref{Eq:I_XZ_phi_theta} reduces to $ \I_{\boldsymbol{\theta}} \! \left( \mathbf{X}; \mathbf{Z} \right) \geq - \H_{ \boldsymbol{\theta}} \! \left( \mathbf{X} \! \mid  \! \mathbf{Z} \right) - \D_{\mathrm{KL}} \left( P_{\mathsf{D}} (\mathbf{X}) \, \Vert \, P_{\boldsymbol{\theta}} (\mathbf{X}) \right) \eqqcolon 
\I_{\boldsymbol{\theta}}^{\mathrm{L}} \! \left( \mathbf{X}; \mathbf{Z} \right) $. 
%$\mathbb{E}_{P_{\mathsf{D}} (\mathbf{X})}  \left[ \log P_{\boldsymbol{\theta}} \! \left( \mathbf{X} \! \mid \! \mathbf{Z} \right) \right] $.
The information uncertainty upper bound $\H_{\boldsymbol{\phi}, \boldsymbol{\varphi}}^{\mathrm{U}} \left( \mathbf{X} \mid \mathbf{S}, \mathbf{Z} \right)$ in Eq.~\eqref{Eq:H_XgivenSZ_phi_varphi} is simplified as $- \mathbb{E}_{P_{\mathbf{S}, \mathbf{X}}} \left[  \log P_{\boldsymbol{\varphi}} \left( \mathbf{X} \mid \mathbf{S}, \mathbf{Z} \right)  \right] = - \mathbb{E}_{P_{\mathbf{S}, \mathbf{X}, \mathbf{Z}}} \left[  \log P_{\boldsymbol{\varphi}} \left( \mathbf{X} \mid \mathbf{S}, \mathbf{Z} \right)  \right] \eqqcolon \H_{\boldsymbol{\varphi}}^{\mathrm{U}} \left( \mathbf{X} \mid \mathbf{S}, \mathbf{Z} \right)$. Similarly, information uncertainty lower bound $\I_{\boldsymbol{\phi}, \boldsymbol{\xi}}^{\mathrm{L}} \! \left( \mathbf{S}; \mathbf{Z} \right)$ in Eq.~\eqref{Eq:I_SZ_phi_Xi_LowerUpperBounds} reduces to $\I_{\boldsymbol{\xi}}^{\mathrm{L}} \! \left( \mathbf{S}; \mathbf{Z} \right) \!  \coloneqq \! - \H_{\boldsymbol{\xi}} \left( \mathbf{S} \! \mid  \! \mathbf{Z} \right) 
 -  \D_{\mathrm{KL}} \left( P_{\mathbf{S}} \, \Vert \, P_{\boldsymbol{\xi}} (\mathbf{S}) \right) $. 
Using Eq.~\eqref{Eq:I_XZ_phi} and Eq.~\eqref{Eq:Lemma_InfoComplexity_KLD_CE}, in DCLUB model, the parameterized variational approximation of information complexity is simplified as:\vspace{-8pt}
\begin{equation}\label{Eq:I_XZ_phi_deterministic}
    \I_{\boldsymbol{\phi}, \boldsymbol{\psi}} \left( \mathbf{X}; \mathbf{Z} \right) = \H \left( P_{\boldsymbol{\phi}} (\mathbf{Z}) \, \Vert \, Q_{\boldsymbol{\psi}} (\mathbf{Z}) \right) -  \D_{\mathrm{KL}} \left(  P_{\boldsymbol{\phi}} (\mathbf{Z})\, \Vert\, Q_{\boldsymbol{\psi}} (\mathbf{Z}) \right) \eqqcolon \H_{\boldsymbol{\phi}, \boldsymbol{\psi}} \! \left( \mathbf{Z} \right).\vspace{-6pt}
\end{equation}

Using the above parameterized variational approximations of information quantities the Lagrangian functionals associated with ($\text{P1}\!\!:\! \mathsf{S}$), ($\text{P1}\!\!:\! \mathsf{U}$), and ($\text{P3}\!\!:\! \mathsf{S}$) are given as follows:\vspace{-6pt}
\begin{eqnarray}\label{Eq:DV_DCLUB_objectives}
\!\!(\text{P1}\!\!:\!  \mathsf{S})\!:\;  \mathcal{L}_{\mathrm{DVDCLUB}}^{\mathsf{S}} \left( \boldsymbol{\phi}, \boldsymbol{\theta}, \boldsymbol{\psi} , \boldsymbol{\varphi}, \beta, \alpha \right)  & \coloneqq & 
 \! \! \! \!\!\! \overbrace{\I_{\boldsymbol{\theta}}^{\mathrm{L}} \left( \mathbf{U}; \mathbf{Z} \right) }^{\textcolor{violet}{\mathrm{Information~Utility}}} 
 \! \! \! \!  \! \!  -  \left( \beta + \alpha \right) \! \! \! \! \! \! \! \! \!\!\!
\overbrace{\H_{\boldsymbol{\phi}, \boldsymbol{\psi}} \! \left( \mathbf{Z} \right)}^{\textcolor{blue}{\mathrm{Information~Complexity}}}
 \! \!  - \; \alpha  \! \! \! \! \! \overbrace{ \H_{\boldsymbol{\varphi}}^{\mathrm{U}}  \left( \mathbf{X} \mid  \mathbf{S}, \mathbf{Z} \right) }^{\textcolor{red}{\mathrm{Information~Uncertainty}}}\!\!\!\!\!\!,\\
\!\!(\text{P1}\!\!:\!  \mathsf{U})\!:\;  \mathcal{L}_{\mathrm{DVDCLUB}}^{\mathsf{U}} \left( \boldsymbol{\phi}, \boldsymbol{\theta}, \boldsymbol{\psi} , \boldsymbol{\varphi}, \beta, \alpha \right)  & \coloneqq & 
 \;\; \I_{\boldsymbol{\theta}}^{\mathrm{L}} \left( \mathbf{X}; \mathbf{Z} \right) 
  \; -  \left( \beta + \alpha \right)  \;\;
\H_{\boldsymbol{\phi}, \boldsymbol{\psi}} \! \left( \mathbf{Z} \right)
\;\;\;\;\;\;\; -  \alpha  \;\;\; \H_{\boldsymbol{\varphi}}^{\mathrm{U}}  \left( \mathbf{X} \mid  \mathbf{S}, \mathbf{Z} \right) ,\\
\!\!(\text{P3}\!\!:\!  \mathsf{S})\!:\;  \prescript{\mathrm{L}\!}{}{\mathcal{L}}_{\mathrm{DVDCLUB}}^{\mathsf{S}} \left( \boldsymbol{\phi}, \boldsymbol{\theta}, \boldsymbol{\psi} , \boldsymbol{\xi}, \gamma, \lambda \right)   & \coloneqq & 
\!\!\!\!\!\! - \gamma \; \I_{\boldsymbol{\theta}}^{\mathrm{L}} \left( \mathbf{U}; \mathbf{Z} \right) 
\; +  \quad \quad \quad \quad
\H_{\boldsymbol{\phi}, \boldsymbol{\psi}} \! \left( \mathbf{Z} \right)
\qquad + \lambda   \underbrace{\I_{\boldsymbol{\xi}}^{\mathrm{L}} \! \left( \mathbf{S}; \mathbf{Z} \right)}_{\textcolor{red}{\mathrm{Information~Leakage}}}.\vspace{-7pt}
\end{eqnarray}
Study and analysis of the DCLUB and corresponding DVDCLUB models are beyond the scope of this research. Here we briefly review the core idea of generative compression techniques.

% to ensure a good match between encoder and decoder distributions of both the quantized latents, and continuous-value latents.

%---------------------------------------------------
%
%	 Connection to Shannon's Rate-Distortion Theory
%
%---------------------------------------------------
%
%\subsection{Connection with Shannon's Rate-Distortion Theory}
%
Suppose the data $\mathbf{X}$ is approximated by a discrete codevector (center vector) $\mathbf{c}_i \in \mathcal{C}$ using an analysis transform $f: \mathbb{R}^{d_{\mathbf{x}}} \rightarrow \mathbb{R}^{d_{\mathbf{z}}}$ followed by quantization $\mathcal{Q}: \mathbb{R}^{d_{\mathbf{z}}} \rightarrow \mathbb{R}^{d_{\mathbf{c}}}$, and then assign to a unique binary sequence using an entropy coding, where $\mathcal{C} = \{ \mathbf{c}_i \in \mathbb{R}^{d_{\mathbf{c}}} \mid i \in [L]\}$ denotes a codebook in transform domain (latent space). 
Let $\mathbf{Z}_{\mathcal{Q}} = \mathcal{Q} (\mathbf{Z}) = \mathcal{Q} \left( f \left( \mathbf{X} \right) \right)$. 
Let $\mathsf{dist} \big( \mathbf{X}, \widehat{\mathbf{X}} \big) $ represents a well-defined distortion measure between the original data $\mathbf{X}$ and the reconstructed data $\widehat{\mathbf{X}}$, where $\widehat{\mathbf{X}} = g \left( \mathcal{Q} \left( f \left( \mathbf{X} \right)\right) \right)$. In the most general form, the classical data compression schemes goal is to minimize the expected distortion loss as well as the expected description length of the bit-stream (rate), optimizing the following trade-off:\vspace{-9pt}
\begin{equation}\label{Eq:Conventional_RD_tradeoff}
    \underbrace{\mathbb{E}_{P_{\mathbf{X}}} \left[ \, \mathsf{dist} ( \mathbf{X}, g \left( \mathcal{Q} \left( f \left( \mathbf{X} \right) \right) \right)  \; \right]}_{\mathrm{Distortion}} \, +\,  \beta^{'} \, \underbrace{\mathbb{E}_{P_{\mathbf{X}}} \left[ - \log \left( P_{\mathbf{Z}_{\mathcal{Q}}}\! \left( \mathbf{Z}_{\mathcal{Q}} \right) \right) \right]}_{\mathrm{Rate}},\vspace{-4pt}
\end{equation}
where $\beta^{'} \!  \geq \! 0$ is a Lagrangian multiplier and $P_{\mathbf{Z}_{\mathcal{Q}}}$ is a \textit{discrete entropy model}. Note that the rate term is the cross-entropy between the marginal probability distribution of codevector index $i$ (shared entropy model) and the true marginal distribution of index $i$. 
%
%
% According to Shannon's (noiseless) source coding theorem \cite{shannon1948mathematical}, the entropy of discrete codevector $\mathbf{Z}_{\mathcal{Q}}$ is given as 
% %\begin{equation}
% $\H \left( \mathbf{Z}_{\mathcal{Q}} \right) = - \sum_{\mathbf{c}_i \in \mathcal{C}} P_{\mathbf{Z}_{\mathcal{Q}}} \! \left( \mathbf{Z}_{\mathcal{Q}} = \mathbf{c}_i \right)\; \log \left( P_{\mathbf{Z}_{\mathcal{Q}}}\! \left( \mathbf{Z}_{\mathcal{Q}} = \mathbf{c}_i \right) \right)$. 
% %\end{equation} 
%
%
One can use some parametric nonlinear \textit{analysis} and \textit{synthesis} transforms $f_{\boldsymbol{\phi}}\!: \mathbb{R}^{d_{\mathbf{x}}} \!\rightarrow \mathbb{R}^{d_{\mathbf{z}}}$ and $g_{\boldsymbol{\theta}}\!: \mathbb{R}^{d_{\mathbf{z}}} \!\rightarrow \mathbb{R}^{d_{\mathbf{x}}}$, respectively, instead of conventional linear transforms. This is the core idea of generative compression techniques. 
For instance, one can consider $\mathbf{c}_i, i \in [L]$ as center vectors (centroids) generated from model prior probability distribution $Q_{\boldsymbol{\psi}} (\mathbf{Z})$ (the parameterized entropy model). Considering a parameterized deterministic encoder $f_{\boldsymbol{\phi}}\!: \mathbb{R}^{d_{\mathbf{x}}} \! \rightarrow \mathbb{R}^{d_{\mathbf{z}}}$, the input data $\mathbf{X}$ is mapped to one of the code vectors in $\mathcal{C}$ using a neural network\footnote{This can be achieved using a vector quantizer, where $\mathbf{Z}_{\boldsymbol{\mathcal{Q}}} = \mathcal{Q} \left( f_{\boldsymbol{\phi}} \! \left( \mathbf{X} \right) \right) \coloneqq \mathop{\arg \min}_{i \in [L]} {\Vert f_{\boldsymbol{\phi}} \! \left( \mathbf{X} \right) - \mathbf{c}_i\Vert}_2 $.}.
However, there are two challenges when optimizing the rate-distortion trade-off \eqref{Eq:Conventional_RD_tradeoff} using deep neural networks \cite{agustsson2017soft}: (i) tackling the non-differentiable cost function derived by quantization, and (ii) attaining an accurate and differentiable estimate of cross-entropy. 
To tackle the non-differentiability of the optimization functional, the proposed approaches are basically based on stochastic approximations \cite{balle2016end, theis2017lossy, balle2017variational, choi2019variable, lee2019context}. For example, \cite{balle2016end} modeled the quantization error as additive uniform stochastic noise, \cite{choi2019variable} employed universal quantization, and \cite{theis2017lossy} proposed a stochastic rounding operation with a smooth derivative approximation. To tackle the entropy estimation challenge, a common approach is to (a) model the marginal distribution with a parametric model, such as a piece-wise linear model \cite{balle2016end}, or a Gaussian mixture model \cite{theis2017lossy, ullrich2017soft}; and (b) consider i.i.d assumption for the symbol stream. In \cite{agustsson2017soft}, the authors proposed a soft-to-hard quantization approach to tackle both above~challenges.

\vspace{-7pt}

%---------------------------------------------------
%
%			Connection to Fairness Models
%
%---------------------------------------------------
\subsection{Connection with Fair Machine Learning Models}
\label{Ssec:ConnectionToFairnessModels}
\vspace{-3pt}

A common classical task in computer vision is to find the features that are \textit{invariant} with respect to some covariate factors, e.g., invariant to image translation, scaling, and rotation \cite{lowe1999object}. 
%A similar model is the `Fader Network' \cite{lample2017fader} in which its encoder is adversarially trained to learn the deterministic feature representations invariant to the facial attribute. 
More recently, the algorithmic fairness community \cite{dwork2012fairness, kamiran2012data, zemel2013learning, hardt2016equality, louizos2015variational, calmon2017optimized, zafar2017fairness, louppe2017learning, wadsworth2018achieving, zhang2018mitigating, madras2019fairness, hebert2017calibration, madras2018learning, madras2018predict, kearns2018preventing, kim2019multiaccuracy, celis2019classification, bellamy2019ai, chen2019fairness, grover2019fair, mehrabi2019survey, mary2019fairness, creager2019flexibly, song2019learning, nachum2019group, zhao2019inherent, adel2019one, wei2019optimized, liu2019implicit, serrurier2019fairness, yurochkin2019learning, grari2020adversarial, zhang2020fairness, dutta2020fairness, ragonesi2020learning} endeavors to quantify and mitigate undesired biases against some variable $\mathbf{S}$ (e.g., gender, ethnicity, sexual orientation, disability) from the results of the model. 
The mitigation approaches can be categorized into three groups: (i) pre-processing \cite{kamiran2012data, calmon2017optimized, bellamy2019ai}, post-processing\cite{kamiran2010discrimination, hardt2016equality,pleiss2017fairness, menon2018cost, chen2019fairness, alghamdi2022beyond}, and in-processing \cite{louizos2015variational, louppe2017learning, wadsworth2018achieving, zafar2017fairness, zhang2018mitigating, celis2019classification} approaches. 
The objective of in-processing approaches is to mitigate the undesired bias during the training phase of a machine learning model. 
%Hence, the objective of fair representation learning is to obtain the invariant representations with respect to changes in $\mathbf{S}$ \cite{louizos2015variational, moyer2018invariant}. 
%From this perspective, this term encourages the model to learn representations $\mathbf{Z}$ that are invariant to variable $\mathbf{S}$. 
From this perspective, fair representation learning encourages the model to learn a representation $\mathbf{Z}$ that is \textit{invariant} to variable $\mathbf{S}$. Hence, the CLUB model can contribute to fair representation learning under information complexity and utility constraints.

\pagebreak

%---------------------------------------------------
%		 Experiments
%---------------------------------------------------
\section{Experiments}
\label{Sec:Experiments}

We conduct experiments on the large-scale Colored-MNIST and CelebA datasets.
The Colored-MNIST is \textit{our modified} version of the MNIST \cite{lecun-mnisthandwrittendigit-2010} data-set, which is a collection of $70,000$ `colored' digits of size $3 \times 28 \times 28$. The digits are randomly colored into $\mathsf{Red}$, $\mathsf{Green}$, and $\mathsf{Blue}$ using \textit{uniform} and \textit{non-uniform (biased)} distributions. 
The CelebA \cite{liu2015faceattributes} data-set contains around $200,000$ images of size $3 \times 128 \times 128$. 
%We randomly split the dataset and picked $80 \%$ of the images only for training the network, and the rest only for testing. 
We used TensorFlow $2.3$ \cite{abadi2016tensorflow} with Integrated Keras API to implement and train the proposed DVCLUB models. 
%We trained the network for around $40$ epochs using the standard Adam optimizer \cite{kingma2014adam} and settings. 

\subsection{Alternative DVCLUB Objectives}

We can obtain the DVCLUB Lagrangian functionals associated to \eqref{CLUB_vs_VCLUB_2} in both the supervised and unsupervised scenarios. 
In the supervised setup, we can recast $\mathcal{L}_{\mathrm{DVCLUB}}^{\mathsf{S}} \left( \boldsymbol{\phi}, \boldsymbol{\theta}, \boldsymbol{\psi} , \boldsymbol{\varphi}, \gamma, \lambda \right)$ as follows:%\vspace{-4pt}
\begin{multline}\label{Eq:DVCLUB_Supervised_2}
\!\!(\text{P2}\!\!:\!  \mathsf{S})\!:\;  \mathcal{L}_{\mathrm{DVCLUB}}^{\mathsf{S}} \left( \boldsymbol{\phi}, \boldsymbol{\theta}, \boldsymbol{\psi} , \boldsymbol{\varphi}, \gamma, \lambda \right)   \coloneqq (1 + \lambda )\! 
\overbrace{ \Big( \D_{\mathrm{KL}} \! \left( P_{\boldsymbol{\phi}} (\mathbf{Z} \! \mid \! \mathbf{X}) \, \Vert \, Q_{\boldsymbol{\psi}}   (\mathbf{Z}) \mid P_{\mathsf{D}}(\mathbf{X}) \right) \! - \!
 \D_{\mathrm{KL}} \! \left( P_{\boldsymbol{\phi}}(\mathbf{Z}) \, \Vert \, Q_{\boldsymbol{\psi}} (\mathbf{Z})\right) \! \Big) }^{\textcolor{blue}{\mathrm{Information~Complexity:}~\I_{\boldsymbol{\phi}, \boldsymbol{\psi}}   \left( \mathbf{X}; \mathbf{Z} \right)}} \\
\qquad  \qquad \qquad \qquad \qquad \qquad \qquad \quad  \; - \gamma \;
\underbrace{  \Big( \mathbb{E}_{P_{\mathbf{U}, \mathbf{X}}} \left[ \mathbb{E}_{P_{\boldsymbol{\phi}} \left( \mathbf{Z} \mid \mathbf{X} \right) } \left[ \log P_{\boldsymbol{\theta}} \! \left( \mathbf{U} \! \mid \! \mathbf{Z} \right) \right] \right] - \D_{\mathrm{KL}} \left( P_{\mathbf{U}} \, \Vert \, P_{\boldsymbol{\theta}} (\mathbf{U}) \right) \!  \Big) }_{\textcolor{violet}{\mathrm{Information~Utility:}~  \I_{\boldsymbol{\phi}, \boldsymbol{\theta}}^{\mathrm{L}} \left( \mathbf{U}; \mathbf{Z} \right) }}
 \\
\qquad \;\; + \,   \lambda \, \underbrace{ \left( -\, \mathbb{E}_{P_{\mathbf{S}, \mathbf{X}}} \left[  
\mathbb{E}_{P_{\boldsymbol{\phi}} \left( \mathbf{Z}  \mid \mathbf{X} \right)} \left[ \log P_{\boldsymbol{\varphi}} \! \left( \mathbf{X} \! \mid \! \mathbf{S}, \mathbf{Z} \right) \right] \right] \right)}_{\textcolor{red}{\mathrm{Information~Uncertainty:}~\H_{\boldsymbol{\phi}, \boldsymbol{\varphi}}^{\mathrm{U}}  \left( \mathbf{X} \mid  \mathbf{S}, \mathbf{Z} \right) }} . 
\end{multline}
Likewise  consider $(\text{P2}\!\!:\! \mathsf{U})$ associated with $\mathcal{L}_{\mathrm{DVCLUB}}^{\mathsf{U}} \left( \boldsymbol{\phi}, \boldsymbol{\theta}, \boldsymbol{\psi} , \boldsymbol{\varphi}, \gamma, \lambda \right)$. 
Note that in maximizing the DVCLUB Lagrangian functionals $\mathcal{L}_{\mathrm{DVCLUB}}^{\mathsf{S}} \left( \boldsymbol{\phi}, \boldsymbol{\theta}, \boldsymbol{\psi} , \boldsymbol{\varphi}, \beta, \alpha \right)$ and $\mathcal{L}_{\mathrm{DVCLUB}}^{\mathsf{U}} \left( \boldsymbol{\phi}, \boldsymbol{\theta}, \boldsymbol{\psi} , \boldsymbol{\varphi}, \beta, \alpha \right)$ we explicitly trade-off the opposing desiderata of information complexity and information leakage, by tuning the regularization parameters $\beta$ and $\alpha$. While in minimizing the DVCLUB Lagrangian functionals $\mathcal{L}_{\mathrm{DVCLUB}}^{\mathsf{S}} \left( \boldsymbol{\phi}, \boldsymbol{\theta}, \boldsymbol{\psi} , \boldsymbol{\varphi}, \gamma, \lambda \right)$ and $\mathcal{L}_{\mathrm{DVCLUB}}^{\mathsf{U}} \left( \boldsymbol{\phi}, \boldsymbol{\theta}, \boldsymbol{\psi} , \boldsymbol{\varphi}, \gamma, \lambda \right)$ we explicitly trade-off the opposing desiderata of information utility and information leakage, by tuning the regularization parameters $\gamma$ and $\lambda$. Finally, note that in these objectives we minimize the upper bound of information leakage in \eqref{Eq:I_SZ_phi_Xi_LowerUpperBounds}, i.e., we minimize the average maximal (worse-case) leakage. Alternatively, one can consider the lower bound on the information leakage in \eqref{Eq:I_SZ_phi_Xi_LowerUpperBounds}. In this case, one can consider the following \textit{minimization} objective:
\begin{multline}\label{Eq:DVCLUB_Supervised_SLowerBound}
\!\! (\text{P3}\!\!:\! \mathsf{S})\!:\;  \prescript{\mathrm{L}\!}{}{\mathcal{L}}_{\mathrm{DVCLUB}}^{\mathsf{S}} \left( \boldsymbol{\phi}, \boldsymbol{\theta}, \boldsymbol{\psi} , \boldsymbol{\xi}, \gamma, \lambda \right)  \coloneqq
\overbrace{   \D_{\mathrm{KL}} \! \left( P_{\boldsymbol{\phi}} (\mathbf{Z} \! \mid \! \mathbf{X}) \, \Vert \, Q_{\boldsymbol{\psi}}   (\mathbf{Z}) \mid P_{\mathsf{D}}(\mathbf{X}) \right)- 
 \D_{\mathrm{KL}} \! \left( P_{\boldsymbol{\phi}}(\mathbf{Z}) \, \Vert \, Q_{\boldsymbol{\psi}} (\mathbf{Z})\right)}^{\textcolor{blue}{\mathrm{Information~Complexity:}~\I_{\boldsymbol{\phi}, \boldsymbol{\psi}}   \left( \mathbf{X}; \mathbf{Z} \right)}} \\
\qquad \qquad \qquad \qquad \qquad \qquad  \;  - \gamma \;
\underbrace{  \Big( \mathbb{E}_{P_{\mathbf{U}, \mathbf{X}}} \left[ \mathbb{E}_{P_{\boldsymbol{\phi}} \left( \mathbf{Z} \mid \mathbf{X} \right) } \left[ \log P_{\boldsymbol{\theta}} \! \left( \mathbf{U} \! \mid \! \mathbf{Z} \right) \right] \right] - \D_{\mathrm{KL}} \left( P_{\mathbf{U}} \, \Vert \, P_{\boldsymbol{\theta}} (\mathbf{U}) \right)  \Big) }_{\textcolor{violet}{\mathrm{Information~Utility:}~  \I_{\boldsymbol{\phi}, \boldsymbol{\theta}}^{\mathrm{L}} \left( \mathbf{U}; \mathbf{Z} \right) }}
 \\
\;\;\;\; \; + \, \lambda \, \underbrace{ \Big(  \mathbb{E}_{P_{\mathbf{S}, \mathbf{X}}} \left[ \mathbb{E}_{P_{\boldsymbol{\phi}} \left( \mathbf{Z} \mid \mathbf{X} \right) } \left[ \log P_{\boldsymbol{\xi}} \! \left( \mathbf{S} \! \mid \! \mathbf{Z} \right) \right] \right] - \D_{\mathrm{KL}} \left( P_{\mathbf{S}} \, \Vert \, P_{\boldsymbol{\xi}} (\mathbf{S}) \right) \Big) }_{\textcolor{red}{\mathrm{Information~Leakage:}~\I_{\boldsymbol{\phi}, \boldsymbol{\xi}}^{\mathrm{L}}  \left( \mathbf{S} ; \mathbf{Z} \right) }} . 
\end{multline}
Likewise consider $(\text{P3}\!\!:\! \mathsf{U})$ associated with the objective $\prescript{\mathrm{L}\!}{}{\mathcal{L}}_{\mathrm{DVCLUB}}^{\mathsf{U}} \left( \boldsymbol{\phi}, \boldsymbol{\theta}, \boldsymbol{\psi} , \boldsymbol{\xi}, \gamma, \lambda \right)$ in unsupervised setup. 
Note that the obtained Lagrangian functionals based on the lower bound of information leakage \eqref{Eq:I_SZ_phi_Xi_LowerUpperBounds} cannot be included in the variational bounds \eqref{Eq:VCLUB_vs_DVCLUB_max} or \eqref{Eq:VCLUB_vs_DVCLUB_min}. 
%We address this setup in Sec.~\ref{Ssec:Algorithm}. 
%

\begin{figure}[t!]
    \centering
   %  \hspace{-9pt} 
        \begin{subfigure}[h]{0.485\textwidth}
        \includegraphics[scale=0.32]{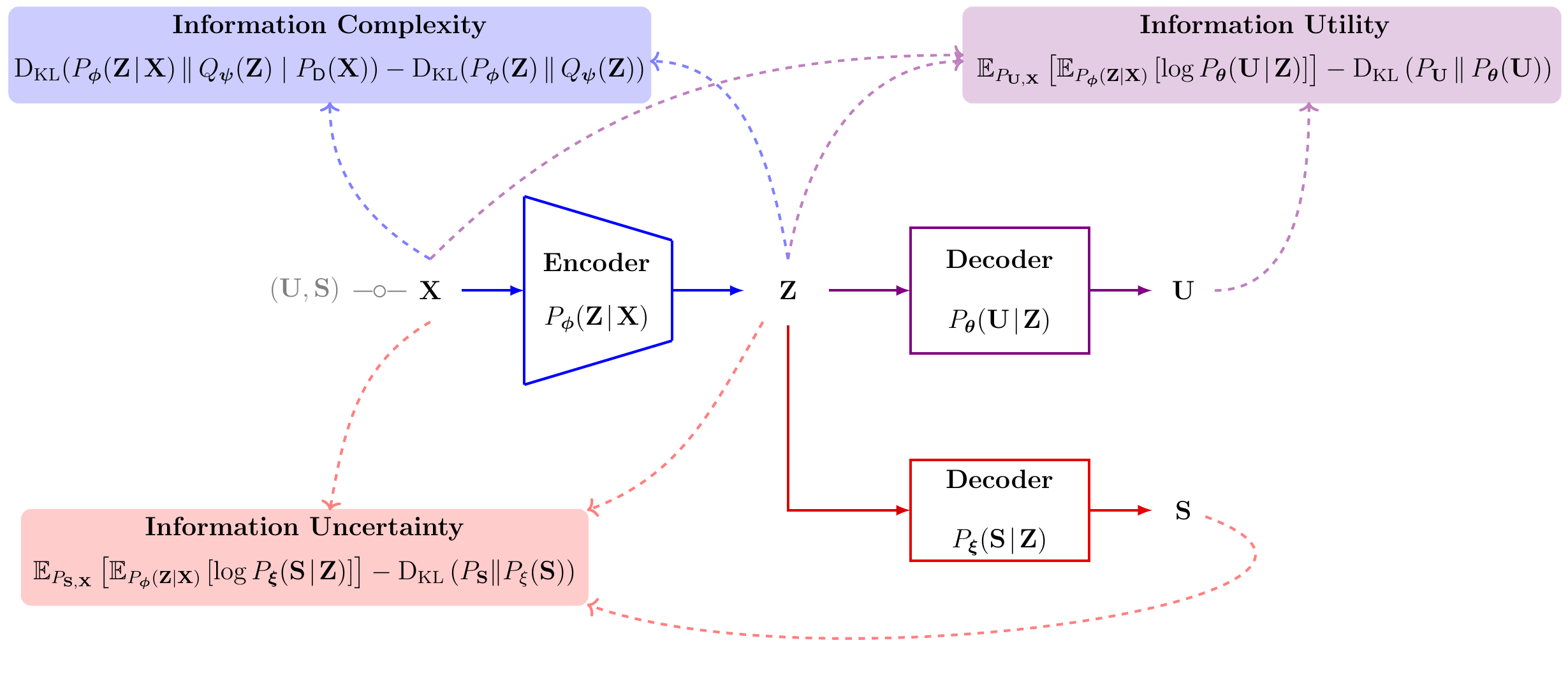}%
 %\includegraphics[width=cm, height=3.3cm]{TeXFig/GeneralDiag2.pdf}
%   \begin{subfigure}[h]{0.48\textwidth}
   %     \includegraphics[width=6cm, height=3.3cm]{TeXFig/GeneralDiag.pdf}%
   %        \vspace{-6pt}
        \caption{}
        %   \vspace{-10pt}
        \label{fig:DeepVariationalCLUB_SupervisedLowerBound}
    \end{subfigure}%
~
       \begin{subfigure}[h]{0.485\textwidth}
     %  \vspace{5pt}
        \includegraphics[scale=0.32]{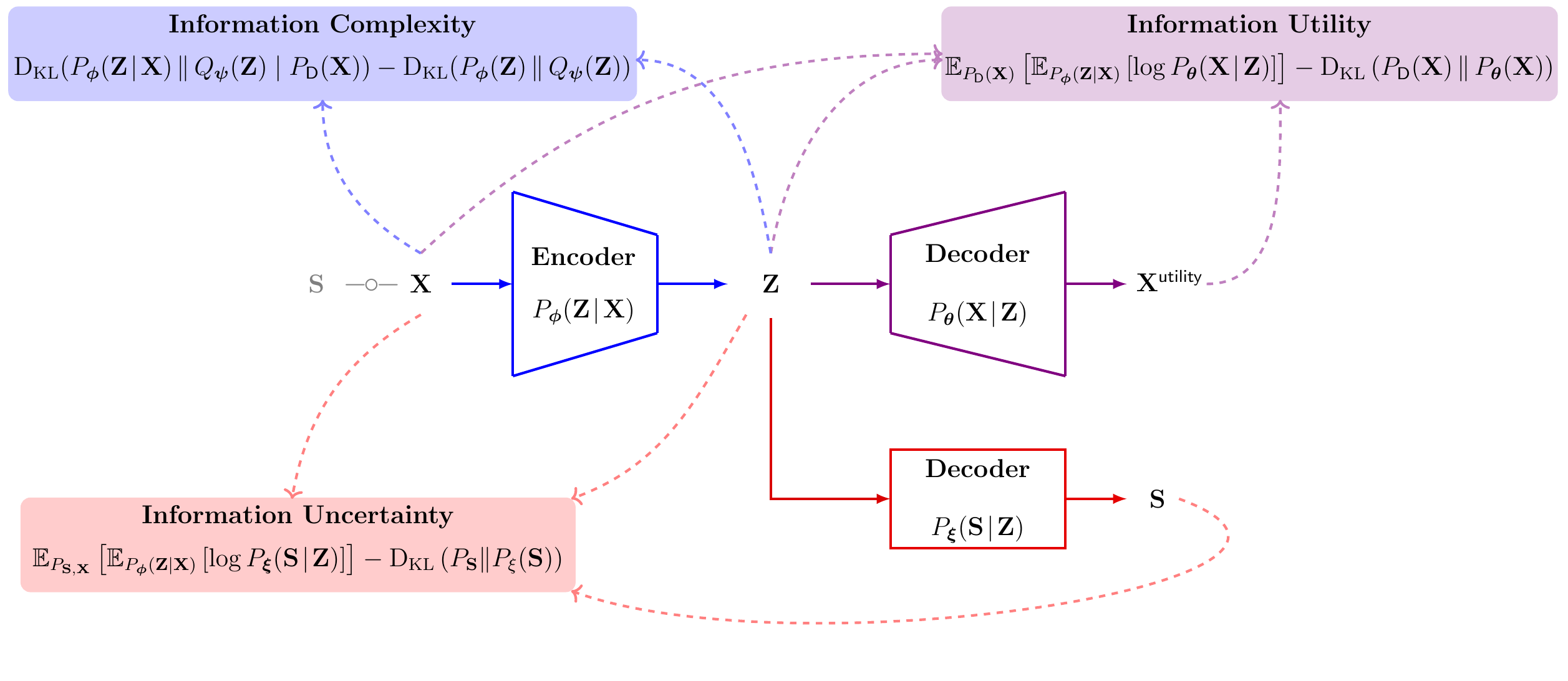}%
%     \begin{subfigure}[h]{0.48\textwidth}
  %   \includegraphics[width=6cm, height=3.3cm]{TeXFig/RegionOverlap.pdf}%
%\includegraphics[scale=0.2]{pdfFig/DisPreQuerySide.pdf}%
        %\vspace{-5pt}
        \caption{}
       %    \vspace{-10pt}
        \label{fig:DeepVariationalCLUB_UnSupervisedLowerBound}
    \end{subfigure}
   % \vspace{-6pt}
    \caption{Block diagram of the DVCLUB associated with $(\text{P3})$ in the (a) supervised setup and (b) unsupervised setup.}
  %  \vspace{-15pt}
    \label{Fig:DeepVariationalCLUB_P5_P6}
%   \vspace{-13pt}
\end{figure}

\begin{figure}[!t]
\centering
\includegraphics[height=5.8cm]{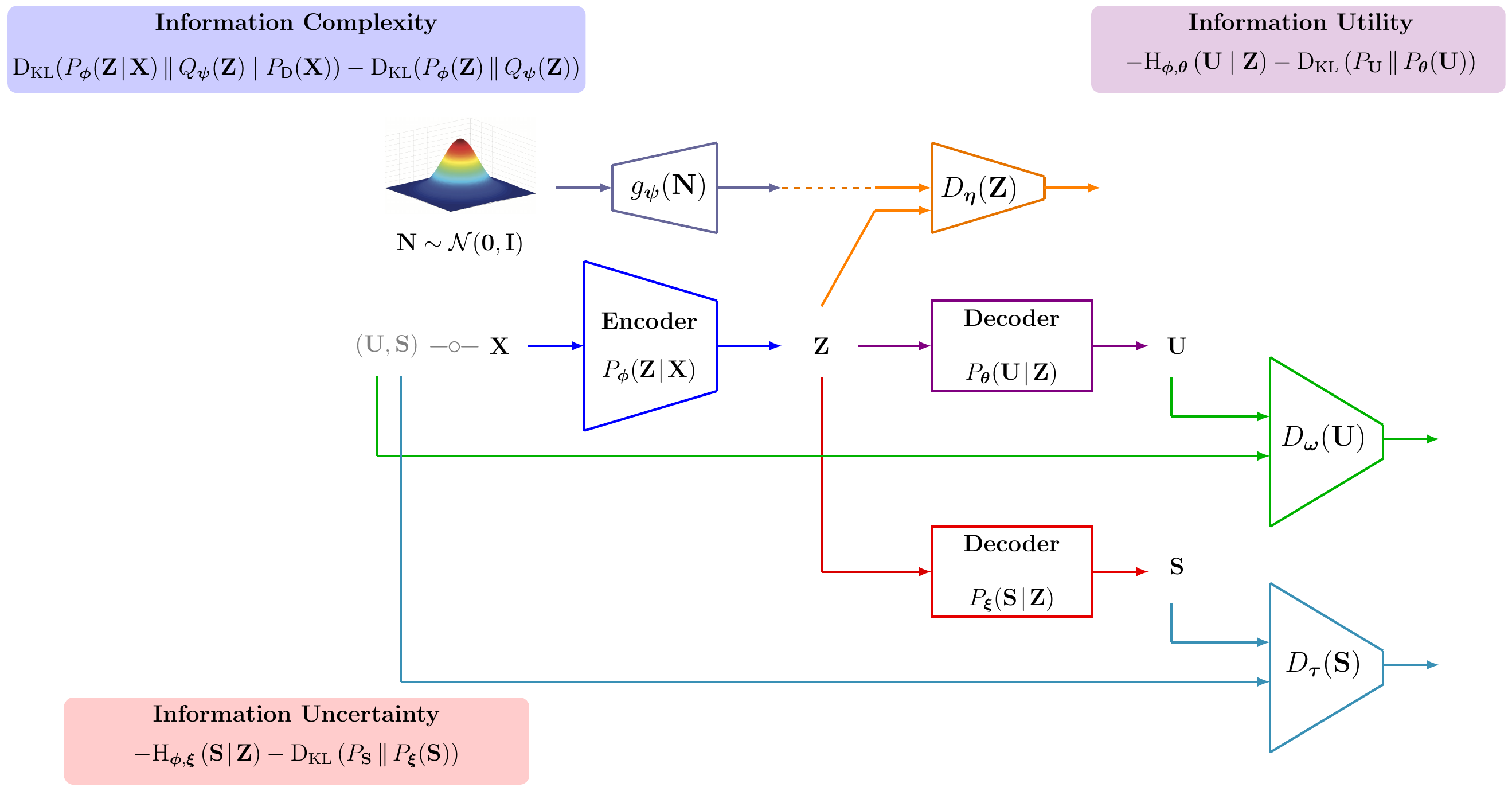}
%\vspace{-15pt}
\caption{Supervised DVCLUB training architecture: considering the lower bound of information leakage.}
\label{Fig:DVCLUB_Supervised_complete_SLowerBound}
%\vspace{-10pt} 
\end{figure}

\begin{figure}[!t]
\centering
\includegraphics[height=5.8cm]{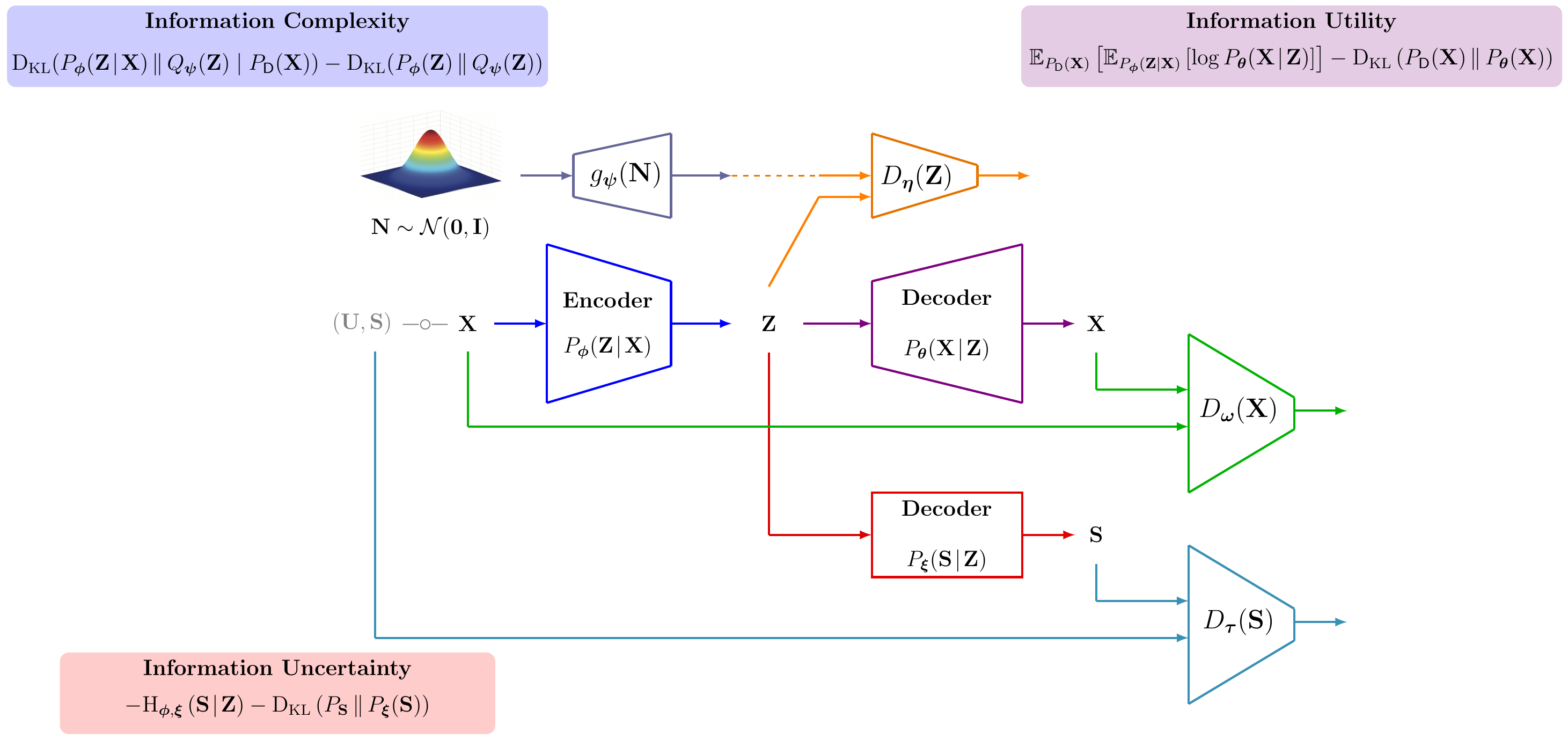}
%\vspace{-15pt}
\caption{Unsupervised DVCLUB training architecture: considering the lower bound of information leakage.}
\label{Fig:DVCLUB_UnSupervised_complete_SLowerBound}
%\vspace{-10pt} 
\end{figure}

Fig.~\ref{Fig:DeepVariationalCLUB_P5_P6} demonstrates the general block diagram for the supervised and unsupervised DVCLUB models associated with $(\text{P3})$. 
The training architecture associated with problems $(\text{P3}\!\!:\! \mathsf{S})$ and $(\text{P3}\!\!:\!  \mathsf{U})$ are depicted in Fig.~\ref{Fig:DVCLUB_Supervised_complete_SLowerBound} and Fig.~\ref{Fig:DVCLUB_UnSupervised_complete_SLowerBound}, respectively. The associated learning algorithm are provided in Appendix~\ref{Appendix:SupplementaryResults}.

%In practice, we need to train the model using alternating block coordinate descent algorithms. 
% In the following, we gradually build the fundamentals of the learning model and establish its connections to other generative models.

%----------------------------------------------------------
%      Figure:  Deep Variational CLUB  >> P5 and P6
%----------------------------------------------------------
%
%---------------------------------------------------
%---------------------------------------------------

%---------------------------------------------------
%	  Figure: Unsupervised DVCLUB >> Complete Diagram
%---------------------------------------------------
%
%---------------------------------------------------
%---------------------------------------------------
%
%---------------------------------------------------
%	  Figure: Unsupervised DVCLUB >> Complete Diagram
%---------------------------------------------------
%
%---------------------------------------------------
%---------------------------------------------------

%% ---------------------------------------------
%
%% ------------    Color MNIST Results   ------------  
%
%% ---------------------------------------------

\clearpage
\subsection{Colored-MNIST Experiments}

\begin{figure}[b!]
     \centering
     \includegraphics[width=0.6\textwidth]{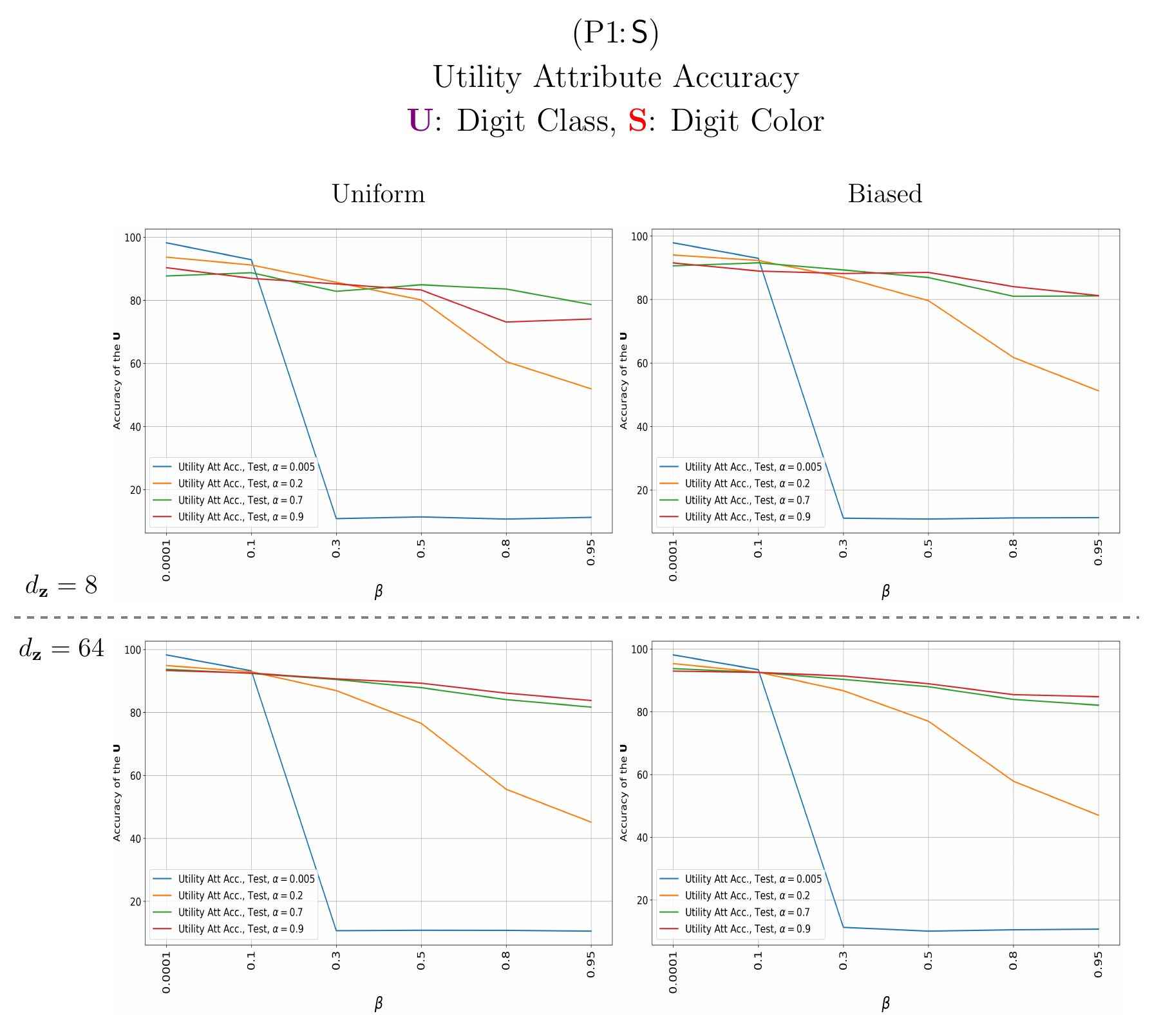}
     %\vspace{-6pt}
     \caption[]
     {Recognition accuracy of the utility attribute $\mathbf{U}$ for supervised CLUB model ($\text{P1}\!\!:\! \mathsf{S}$) on `Test' dataset of Colored-MNIST, considering $d_{\mathbf{z}}=8$ (First Row) and $d_{\mathbf{z}}=64$ (Second Row), setting $Q_{\boldsymbol{\psi}}\! \left( \mathbf{Z} \right) = \mathcal{N} \! \left( \boldsymbol{0}, \mathbf{I}_{d_{\mathbf{z}}} \right)$, for different information complexity weights $\beta$ and information leakage weights $\alpha$. (Left Column): setting $P (\mathsf{Red}) \!= \!P (\mathsf{Green}) \!= \! P (\mathsf{Blue}) \! = \! \frac{1}{3}$;
(Right Column): setting $P (\mathsf{Red}) \! = \! \frac{1}{2}$, $P (\mathsf{Green})\! = \! \frac{1}{6}$, $P (\mathsf{Blue}) \!=\! \frac{1}{3}$.}
     \label{Fig:AccU_ColMNIST_P1Supervised_Scolor_d8}
\end{figure}

\begin{figure}[htp]
\centering
\includegraphics[width=0.91\textwidth]{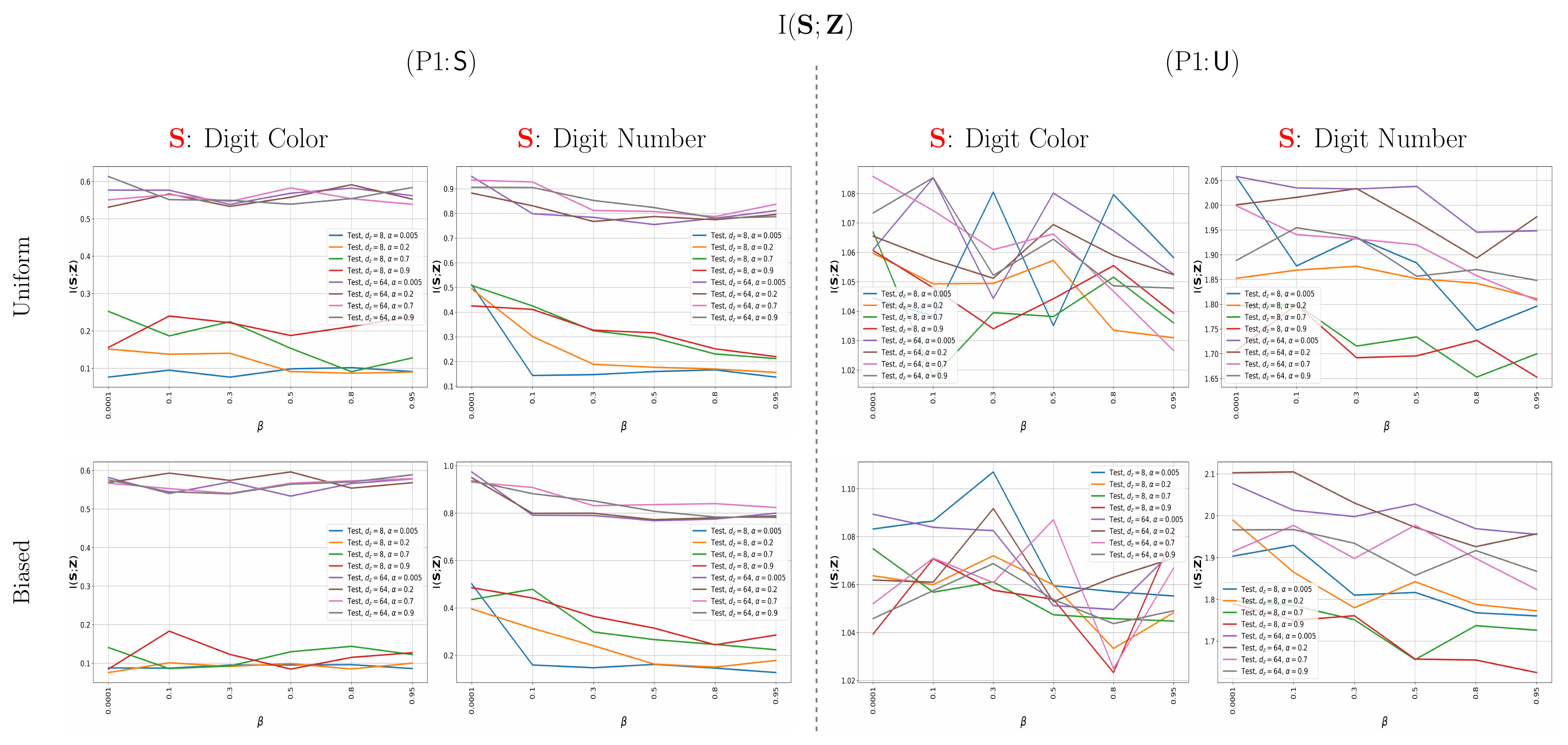}
%% Mutual Information I(S;Z)
%% P1 Supervised/Unsupervised
%
\vspace{-10pt}
\caption{Estimated information leakage $\I(\mathbf{S}; \mathbf{Z})$ on Colored-MNIST dataset using MINE, considering supervised and unsupervised scenarios ($\text{P1}\!\!:\! \mathsf{S}$) (Left Panel) and ($\text{P1}\!\!:\! \mathsf{U}$) (Right Panel), respectively, for $d_{\mathrm{z}} \in \{ 8, 64 \}$. 
(First Row): setting $P (\mathsf{Red}) \!= \!P (\mathsf{Green}) \!= \! P (\mathsf{Blue}) \! = \! \frac{1}{3}$;
(Second Row): setting $P (\mathsf{Red}) \! = \! \frac{1}{2}$, $P (\mathsf{Green})\! = \! \frac{1}{6}$, $P (\mathsf{Blue}) \!=\! \frac{1}{3}$.}
\vspace{-10pt}
\label{Fig:MIsz_ColMNIST_P1}
\end{figure}

\begin{figure}[htp]
\centering
\includegraphics[width=0.93\textwidth]{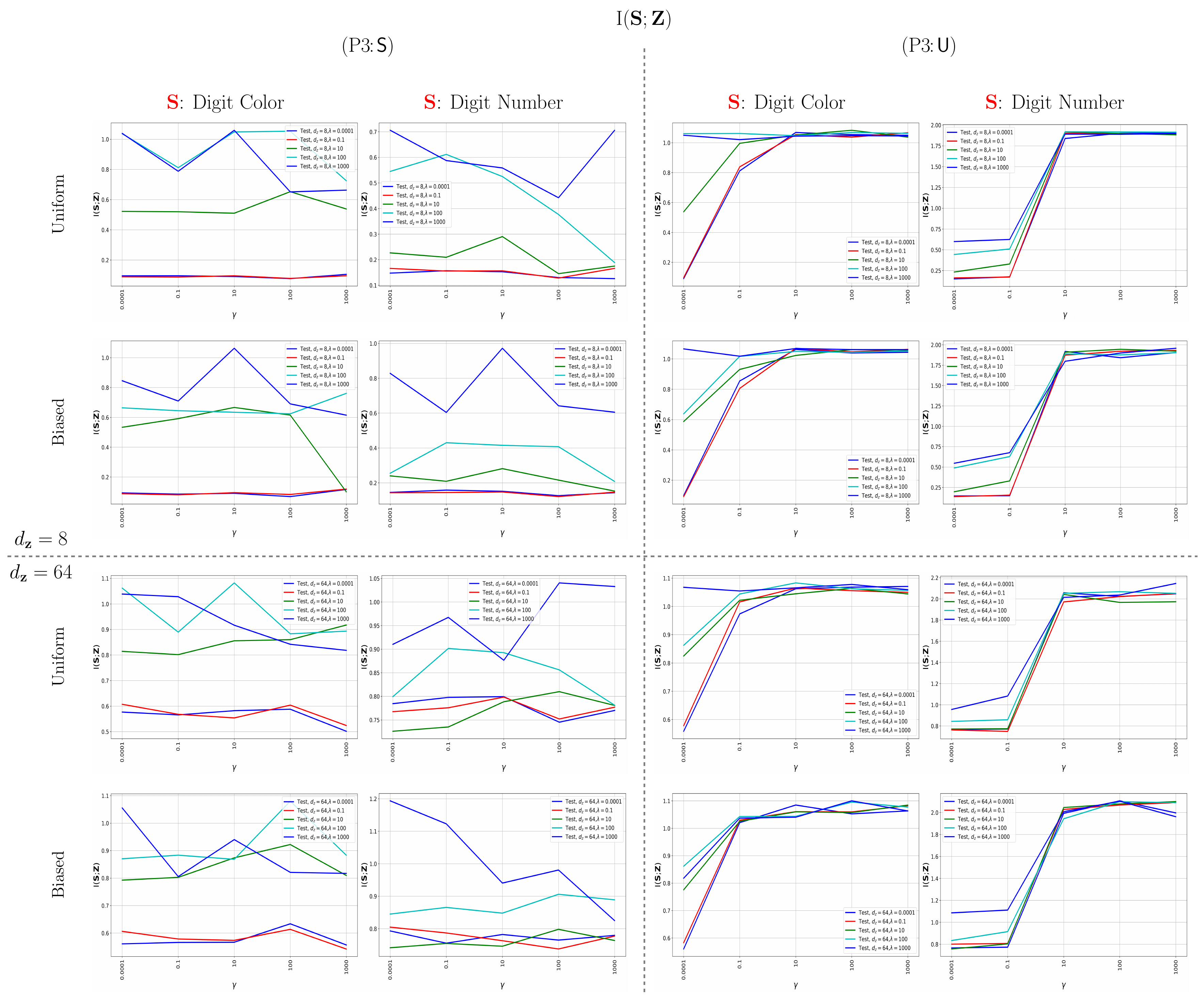}
%% Mutual Information I(S;Z)
%% P3 Supervised/Unsupervised
%
\vspace{-10pt}
\caption{Estimated information leakage $\I(\mathbf{S}; \mathbf{Z})$ on Colored-MNIST dataset using MINE, considering supervised and unsupervised scenarios ($\text{P3}\!\!:\! \mathsf{S}$) (Left Panel) and ($\text{P3}\!\!:\! \mathsf{U}$) (Right Panel), respectively, for $d_{\mathrm{z}} = 8$ (Up Panel) and $d_{\mathrm{z}} =64$ (Down Panel). In each horizontal panel, 
(First Row): setting $P (\mathsf{Red}) \!= \!P (\mathsf{Green}) \!= \! P (\mathsf{Blue}) \! = \! \frac{1}{3}$;
(Second Row): setting $P (\mathsf{Red}) \! = \! \frac{1}{2}$, $P (\mathsf{Green})\! = \! \frac{1}{6}$, $P (\mathsf{Blue}) \!=\! \frac{1}{3}$.}
\label{Fig:MIsz_ColMNIST_P3}
\end{figure}

\begin{figure}[htp]
\centering
\includegraphics[width=0.91\textwidth]{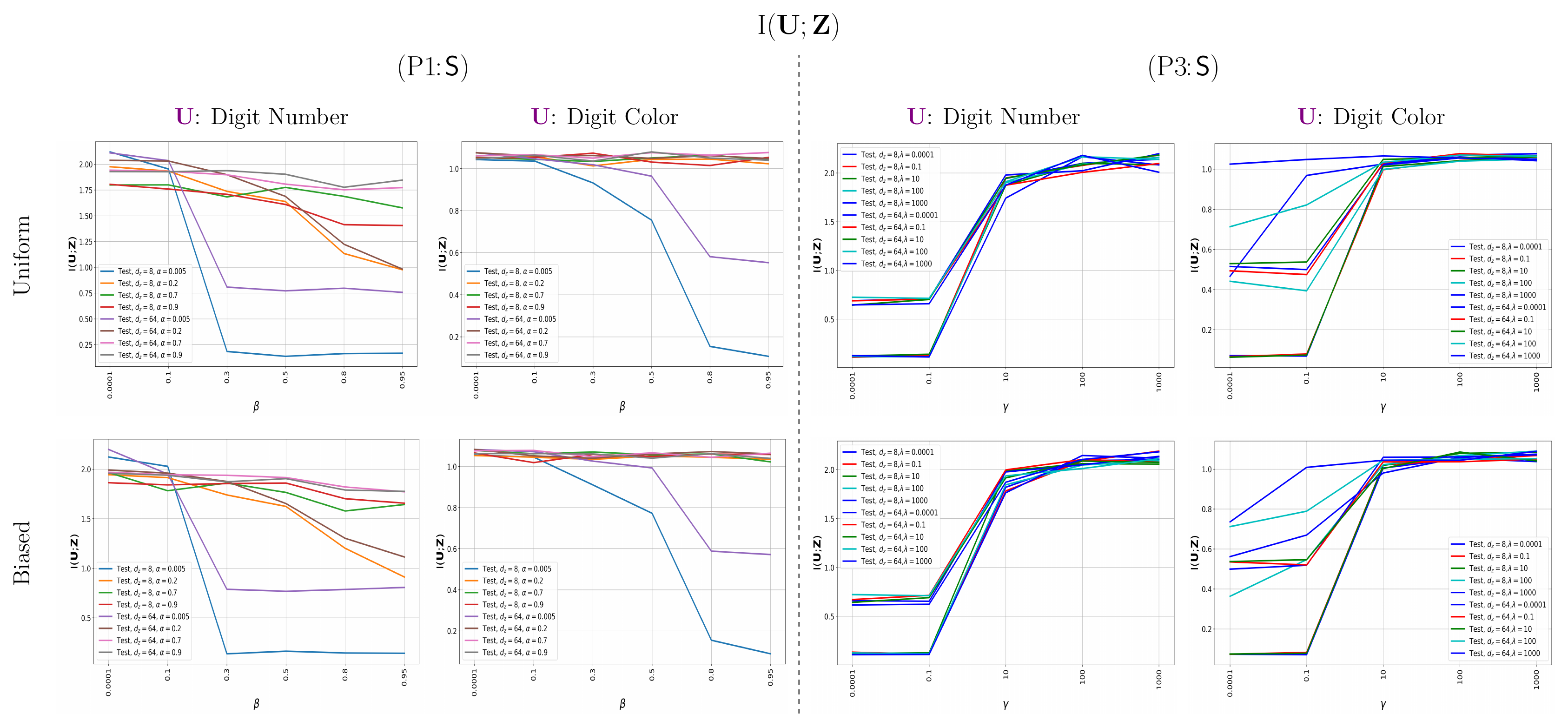}
%% Mutual Information I(U;Z)
%% P1 and P3 Supervised
%
\vspace{-10pt}
\caption{Estimated information utility $\I(\mathbf{U}; \mathbf{Z})$ on Colored-MNIST dataset using MINE, considering supervised scenarios ($\text{P1}\!\!:\! \mathsf{S}$) (Left Panel) and ($\text{P3}\!\!:\! \mathsf{S}$) (Right Panel), for $d_{\mathrm{z}} \in \{ 8, 64 \}$. (First Row): setting $P (\mathsf{Red}) \!= \!P (\mathsf{Green}) \!= \! P (\mathsf{Blue}) \! = \! \frac{1}{3}$;
(Second Row): setting $P (\mathsf{Red}) \! = \! \frac{1}{2}$, $P (\mathsf{Green})\! = \! \frac{1}{6}$, $P (\mathsf{Blue}) \!=\! \frac{1}{3}$.}
\vspace{-10pt}
\label{Fig:MIuz_ColMNIST_P1P3_supervised}
\end{figure}

\begin{figure}[htp]
\centering
\includegraphics[width=0.91\textwidth]{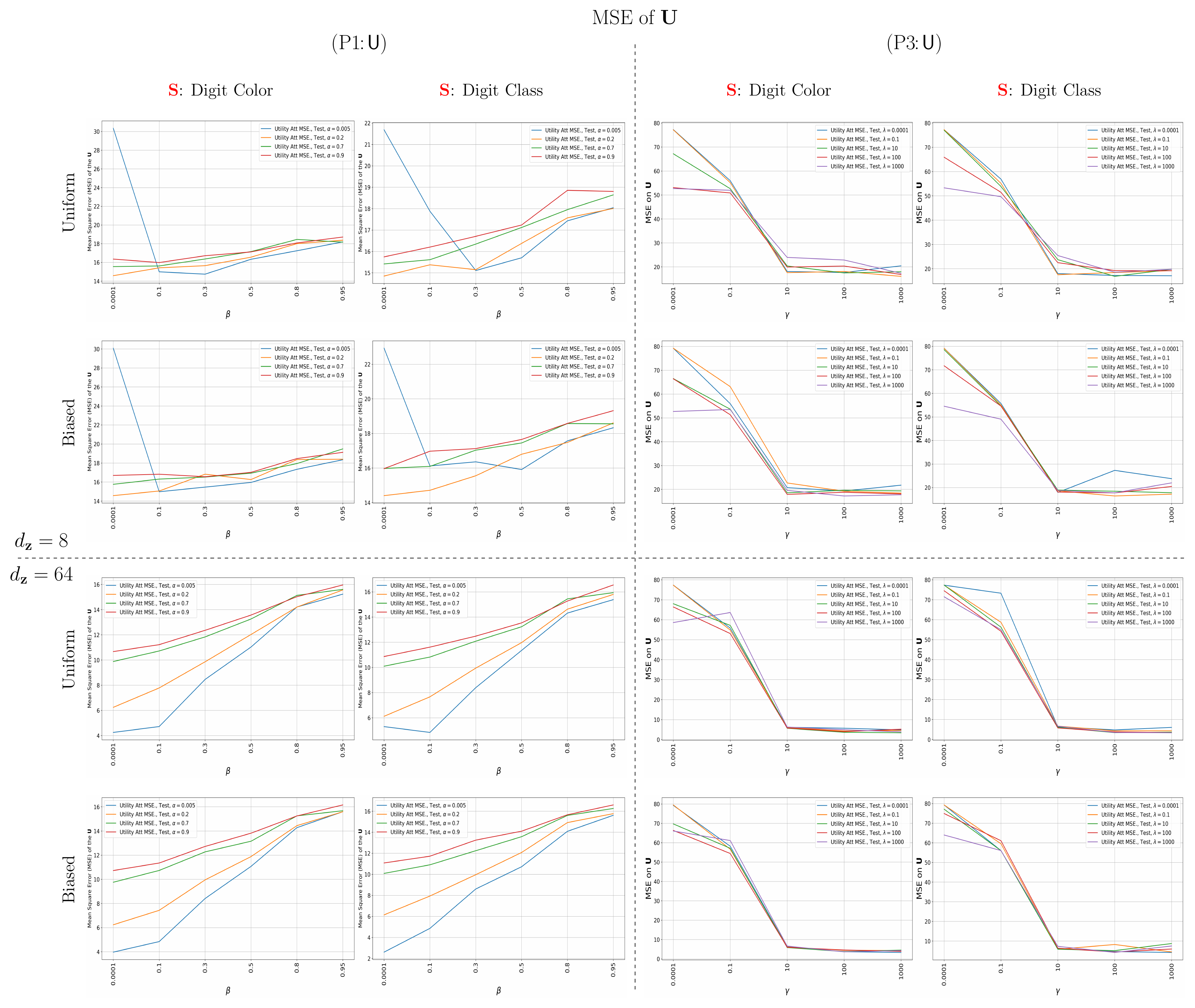}
%% Mutual Information I(S;Z)
%% P1 and P3 Supervised/Unsupervised
%% S: Color
\vspace{-10pt}
\caption{Mean Square Error of utility data $\mathbf{U} \equiv \mathbf{X}$ on Colored-MNIST dataset, considering unsupervised scenarios ($\text{P1}\!\!:\!\mathsf{U}$) (Left Panel) and ($\text{P3}\!\!:\!\mathsf{U}$) (Right Panel), for $d_{\mathrm{z}} = 8$ (Up Panel) and $d_{\mathrm{z}} =64$ (Down Panel). In each horizontal panel, 
(First Row): setting $P (\mathsf{Red}) \!= \!P (\mathsf{Green}) \!= \! P (\mathsf{Blue}) \! = \! \frac{1}{3}$;
(Second Row): setting $P (\mathsf{Red}) \! = \! \frac{1}{2}$, $P (\mathsf{Green})\! = \! \frac{1}{6}$, $P (\mathsf{Blue}) \!=\! \frac{1}{3}$.}
\label{Fig:MSEu_ColMNIST_P1P3_unsupervised}
\end{figure}

\begin{figure}[!p]
     \centering
     \includegraphics[width=0.75\textwidth]{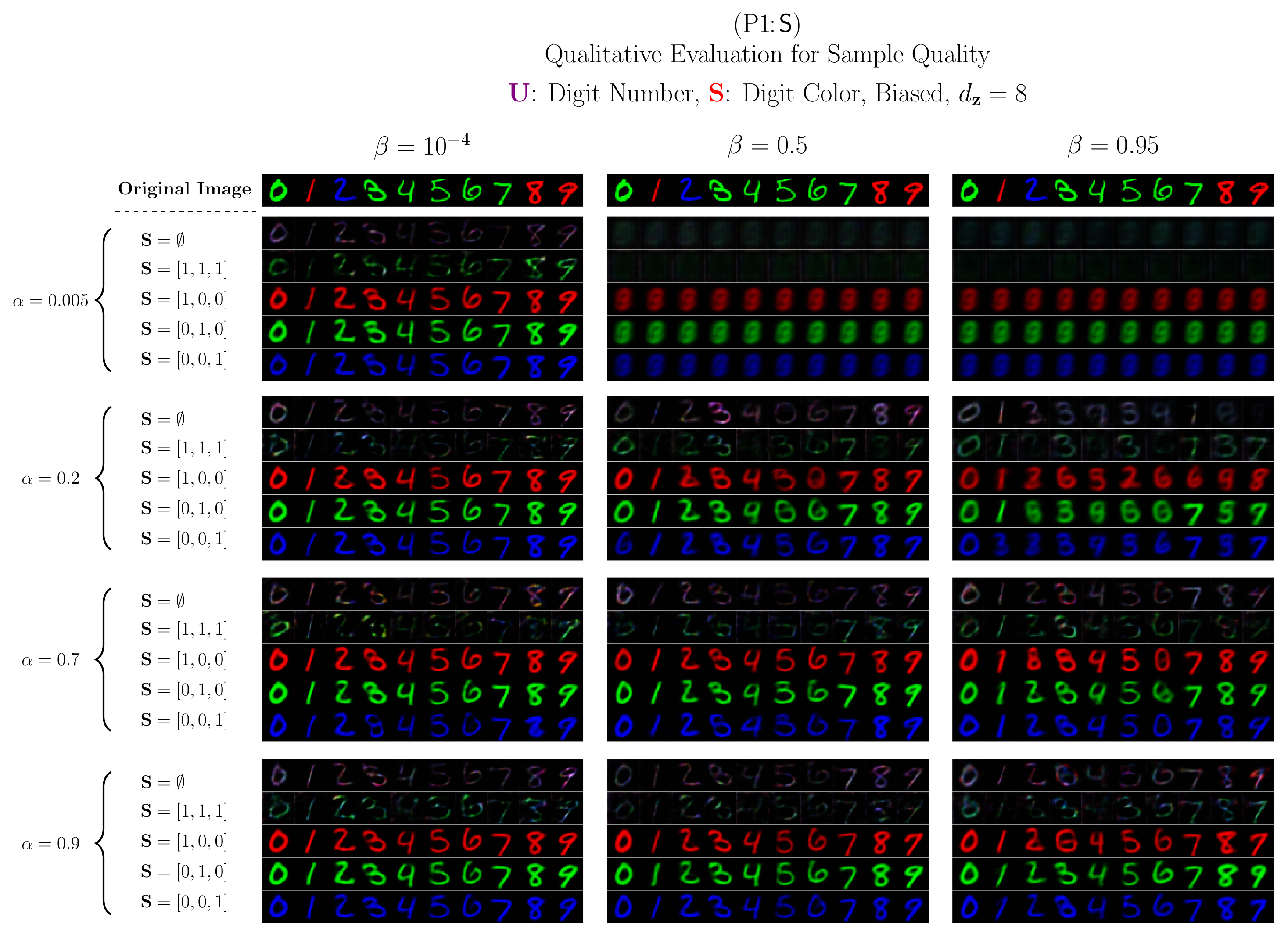}
     \vspace{-10pt}
     \caption[]
     {Qualitative evaluation for sample quality at the inferential adversary for supervised CLUB model ($\text{P1}\!\!:\! \mathsf{S}$) on Colored-MNIST, setting $d_{\mathbf{z}}=8$, $\mathbf{U}$: digit number, $\mathbf{S}$: digit color, $P_S (\mathsf{Red}) = \frac{1}{2}$, $P_S (\mathsf{Green}) = \frac{1}{6}$, $P_S (\mathsf{Blue}) = \frac{1}{3}$, and $Q_{\boldsymbol{\psi}}\! \left( \mathbf{Z} \right) = \mathcal{N} \! \left( \boldsymbol{0}, \mathbf{I}_d \right)$, for different trade-offs between $\beta$ and $\alpha$. From left to right: increasing information complexity weight $\beta$. From top to down: increasing information leakage weight $\alpha$. Five scenarios are considered to investigate the adversary's beliefs about sensitive attribute $\mathbf{S}$. $\mathbf{S} =  \emptyset$ corresponds to the case in  which the adversary recovers $\mathbf{X}$ without any assumption on the digit color; $\mathbf{S} =  \left[ 1, 1, 1 \right]$ corresponds to a possible, probably meaningless, adversary's belief about sensitive attribute to infer $\mathbf{S}$ from reconstructed $\mathbf{X}$; $\mathbf{S} =  \left[ 1, 0, 0 \right]$, $\mathbf{S} =  \left[ 0, 1, 0 \right]$, and $\mathbf{S} =  \left[ 0, 0, 1 \right]$ correspond to different adversary's beliefs about sensitive attribute, i.e., the digit color, associated with $\mathsf{Red}$, $\mathsf{Green}$ and $\mathsf{Blue}$ colors, respectively. The results show that the adversary cannot revise his belief about $\mathbf{S}$.}
\vspace{-20pt}
\label{Fig:ColMNIST_VisualRec_P1S_UclassScolor_biased_d8}
\end{figure}

\begin{figure}[p]
     \centering
     \includegraphics[width=0.75\textwidth]{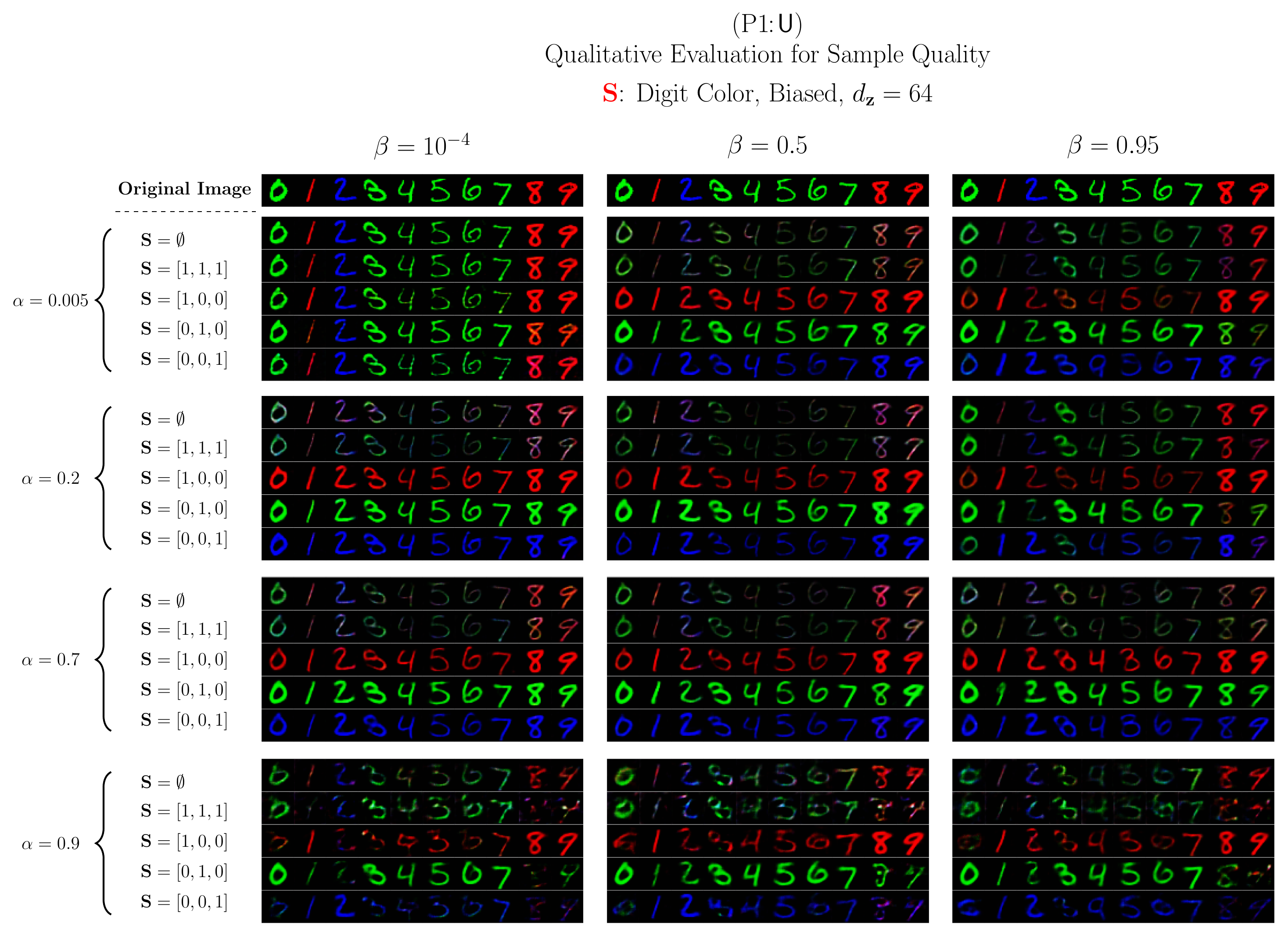}
     \vspace{-10pt}
     \caption[]
     {Qualitative evaluation for sample quality at the inferential adversary for supervised CLUB model ($\text{P1}\!\!:\! \mathsf{U}$) on Colored-MNIST, setting $d_{\mathbf{z}}=64$, keeping the same setup as considered in Fig.~\ref{Fig:ColMNIST_VisualRec_P1S_UclassScolor_biased_d8}.}
     \label{Fig:ColMNIST_VisualRec_P1U_Scolor_biased_d64}
\end{figure}

\begin{figure}[p]
     \centering
     \includegraphics[width=0.75\textwidth]{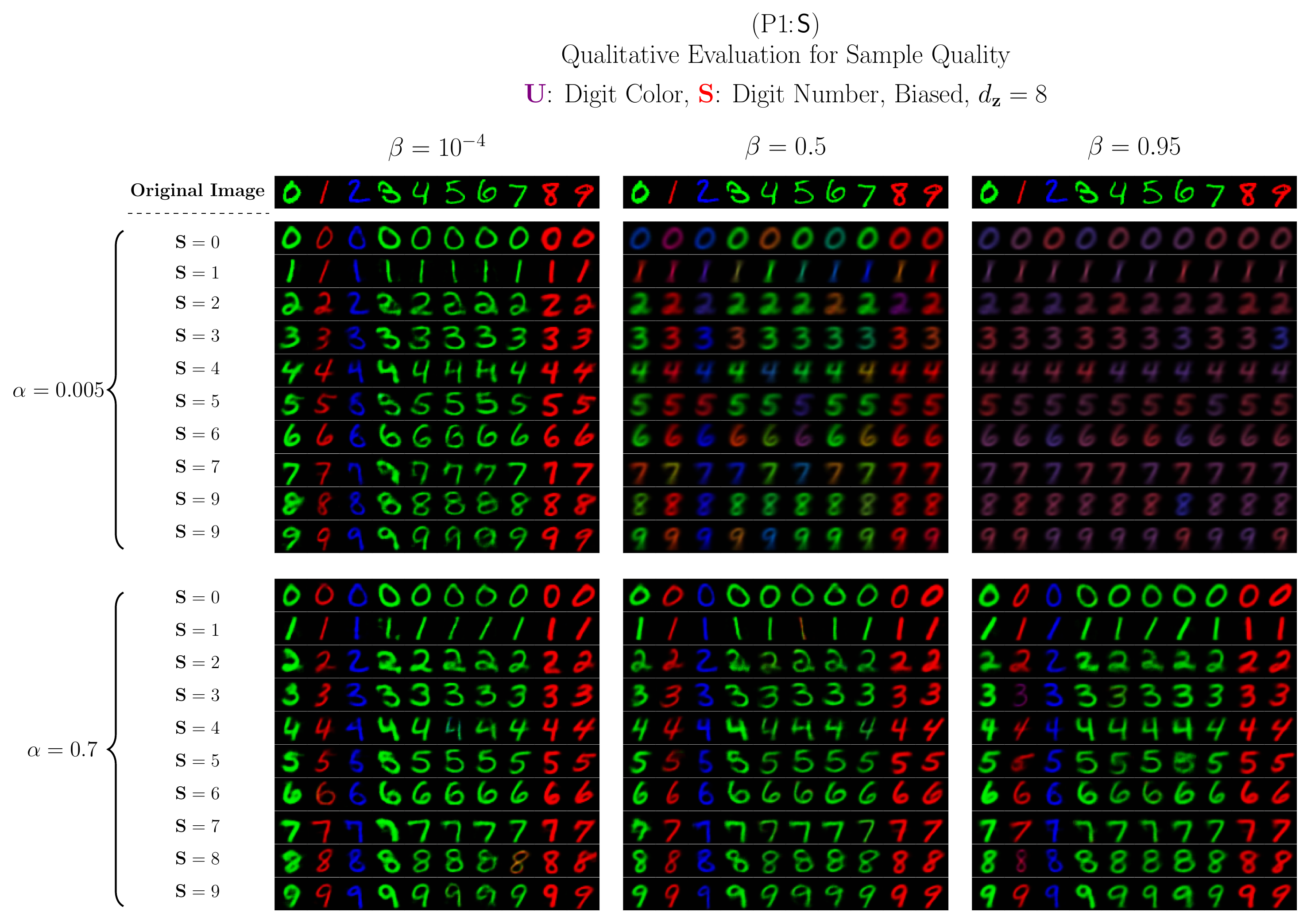}
     \vspace{-10pt}
     \caption[]
     {Qualitative evaluation for sample quality at the inferential adversary for supervised CLUB model ($\text{P1}\!\!:\! \mathsf{S}$) on Colored-MNIST, setting $d_{\mathbf{z}}=8$, $\mathbf{U}$: digit color, $\mathbf{S}$: digit number, $P_U (\mathsf{Red}) = \frac{1}{2}$, $P_U (\mathsf{Green}) = \frac{1}{6}$, $P_U (\mathsf{Blue}) = \frac{1}{3}$, and $Q_{\boldsymbol{\psi}}\! \left( \mathbf{Z} \right) = \mathcal{N} \! \left( \boldsymbol{0}, \mathbf{I}_d \right)$, for different trade-offs between $\beta$ and $\alpha$. From left to right: increasing information complexity weight $\beta$. From top to down: increasing information leakage weight $\alpha$. Ten scenarios are considered to investigate the adversary's beliefs about sensitive attribute $\mathbf{S}$. Each scenario corresponds to the case in which the adversary recovers $\mathbf{X}$ with a assumption on the digit number. The results show that the adversary cannot revise his belief about $\mathbf{S}$.}
  \vspace{-20pt}
  \label{Fig:ColMNIST_VisualRec_P1S_UcolorSclass_biased_d8}
\end{figure}

\begin{figure}[p]
     \centering
     \includegraphics[width=0.75\textwidth]{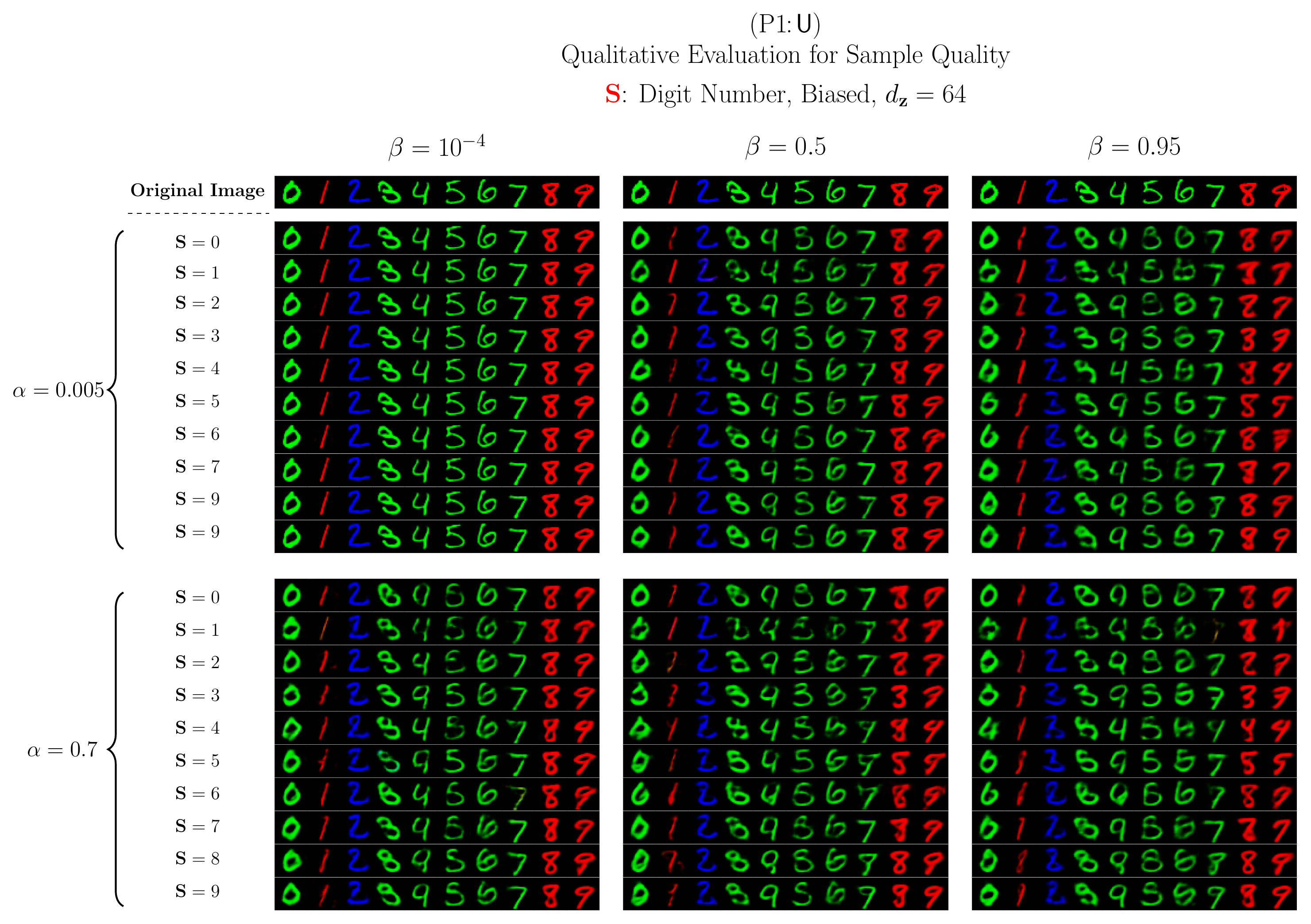}
    % \vspace{-10pt}
     \caption[]
     {Qualitative evaluation for sample quality at the inferential adversary for unsupervised CLUB model ($\text{P1}\!\!:\! \mathsf{U}$) on Colored MNIST, setting $d_{\mathbf{z}}=64$, keeping the same setup as considered in Fig.~\ref{Fig:ColMNIST_VisualRec_P1S_UcolorSclass_biased_d8}.}
     \label{Fig:ColMNIST_VisualRec_P1U_Sclass_biased_d64}
\end{figure}

The experiments on colored-MNIST dataset are depicted in this subsection. 
In order to study the impact of possible biases in the distribution of $\mathbf{S}$, we consider two scenarios for a digit color. In the \textit{uniform} scenario the digits are randomly colored with the probabilities $P_S (\mathsf{Red}) =  P_S (\mathsf{Green}) =  P_S (\mathsf{Blue}) = \frac{1}{3}$, while in the \textit{biased} scenario the digits are randomly colored with the probabilities $P_S (\mathsf{Red}) = \frac{1}{2}$, $P_S (\mathsf{Green}) = \frac{1}{6}$, $P_S (\mathsf{Blue}) = \frac{1}{3}$, and set $Q_{\boldsymbol{\psi}}\! \left( \mathbf{Z} \right) = \mathcal{N} \! \left( \boldsymbol{0}, \mathbf{I}_d \right)$. 
The recognition accuracy of the utility attribute for the supervised CLUB model ($\text{P1}\!\!:\! \mathsf{S}$) is depicted in Fig.~\ref{Fig:AccU_ColMNIST_P1Supervised_Scolor_d8}. 
The estimated information leakage for CLUB models ($\text{P1}$) and ($\text{P3}$) are depicted in Fig.~\ref{Fig:MIsz_ColMNIST_P1} and Fig.~\ref{Fig:MIsz_ColMNIST_P3}, respectively, for the supervised and unsupervised setup. 
Fig.~\ref{Fig:MIuz_ColMNIST_P1P3_supervised} depicts estimated information utility for the CLUB models ($\text{P1}\!\!:\! \mathsf{S}$) and ($\text{P3}\!\!:\! \mathsf{S}$). 
The MSE results of utility data $\mathbf{U} \equiv \mathbf{X}$ for the unsupervised scenarios ($\text{P1}\!\!:\! \mathsf{U}$) and ($\text{P3}\!\!:\! \mathsf{U}$), are depicted in Fig.~\ref{Fig:MSEu_ColMNIST_P1P3_unsupervised}. 
We provide a qualitative evaluation for sample quality at the inferential adversary for different CLUB models in Fig.~\ref{Fig:ColMNIST_VisualRec_P1S_UclassScolor_biased_d8}, Fig.~\ref{Fig:ColMNIST_VisualRec_P1U_Scolor_biased_d64}, Fig.~\ref{Fig:ColMNIST_VisualRec_P1S_UcolorSclass_biased_d8}, and Fig.~\ref{Fig:ColMNIST_VisualRec_P1U_Sclass_biased_d64}. 
The training details and network architectures are provided in Appendix.~\ref{Appendix:NetworkArchitecture_TrainingDetails} and Appendix~\ref{Appendix:NetworksArchitecture}.

%---------------------------------------------------
%   Figure: Colored MNIST : Utility Accuracy
%---------------------------------------------------
%
%---------------------------------------------------%---------------------------------------------------

%---------------------------------------------------
%  Figure: Mutual Information I(S;Z) 
%  P1  
%---------------------------------------------------
%
%
%---------------------------------------------------
%---------------------------------------------------

%---------------------------------------------------
%  Figure: Mutual Information I(S;Z) 
%  P3 
%---------------------------------------------------
%
%
%---------------------------------------------------
%---------------------------------------------------

%---------------------------------------------------
%  Figure: 
%---------------------------------------------------
%
%
%---------------------------------------------------
%---------------------------------------------------

%---------------------------------------------------
%  Figure: MSE on U
%  P1 and P3 Unsupervised
%---------------------------------------------------
%
%---------------------------------------------------
%---------------------------------------------------

%---------------------------------------------------
%   Figure: Qualitative evaluation Colored MNIST d=4
%   P1 Unsupervised
%   Biased
%   d =64
%   S: Color
%---------------------------------------------------
%
%---------------------------------------------------
%---------------------------------------------------

%---------------------------------------------------
%   Figure: Qualitative evaluation Colored MNIST d=8
%   P1 Supervised
%   Biased
%   d = 8
%   U: Color
%   S: Class
%---------------------------------------------------
%
%---------------------------------------------------%---------------------------------------------------

%---------------------------------------------------
%   Figure: Qualitative evaluation Colored MNIST d=4
%   P1 Unsupervised
%   Biased
%   d = 64
%   S: Class
%---------------------------------------------------
%
%---------------------------------------------------
%---------------------------------------------------

% %---------------------------------------------------
% %   Figure: Colored MNIST : Reconstruction Loss
% %---------------------------------------------------
% %
% \begin{figure}[t!]
%     \centering
%     \includegraphics[width=0.9\textwidth]{ResultFig/chart_colored_mnist_rec_loss.png}
%     \caption[]
%     {\small .... }
%     \label{Fig:ColoredMNIST_chart_rec_loss}
% \end{figure}
% %---------------------------------------------------
% %---------------------------------------------------

%% ---------------------------------------------
%
%% ------------  	 	CelebA Results  	   ------------  
%
%% ---------------------------------------------

\clearpage
\subsection{CelebA Experiments}

\begin{figure}[!b]
\centering
\includegraphics[width=\textwidth]{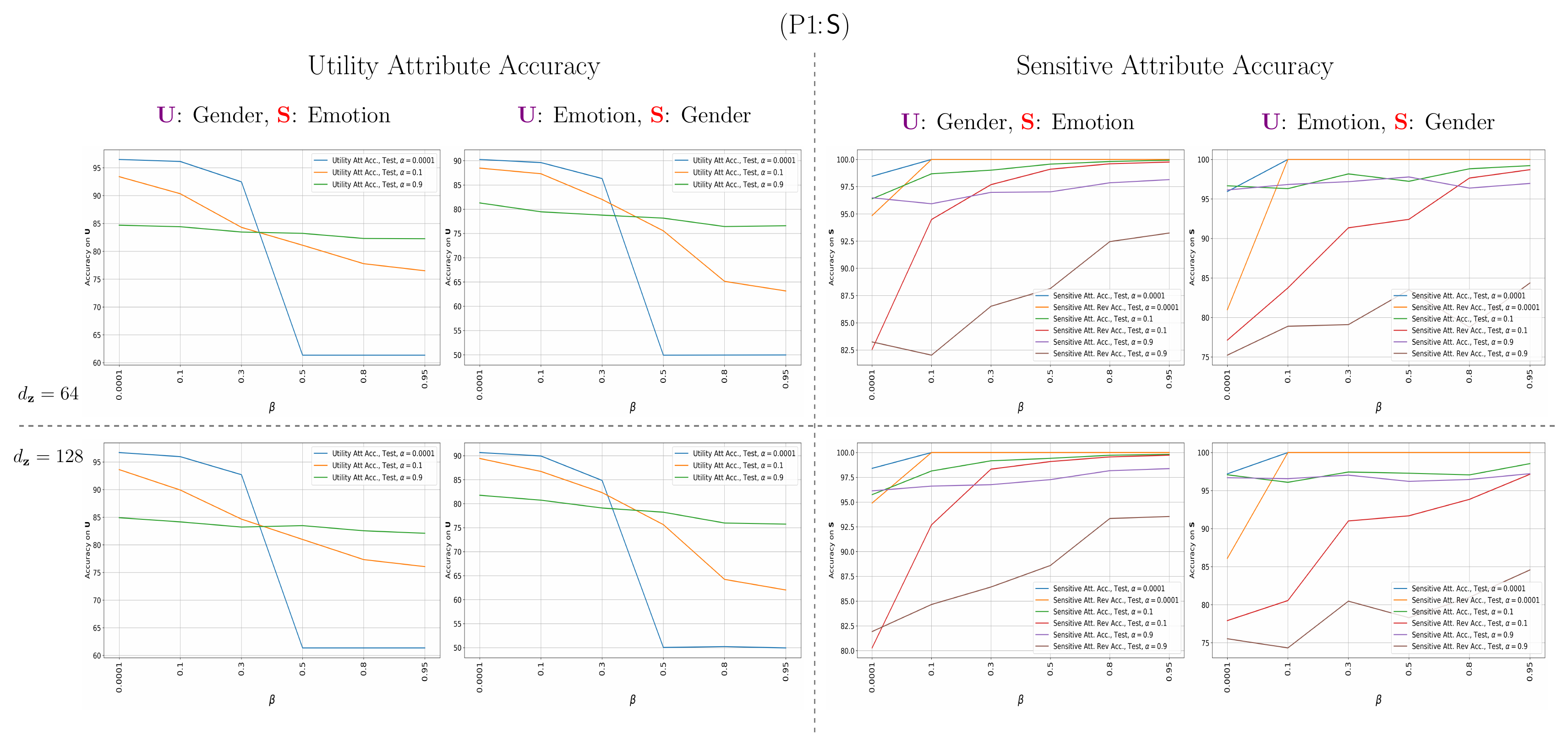}
%% Accuracy on U and S
%% P1 Supervised
%% d: 64 & 128
\caption{Recognition accuracy of the utility attribute $\mathbf{U}$ (Left Panel) and sensitive attribute $\mathbf{S}$ (Right Panel) for supervised CLUB model ($\text{P1}\!\!:\! \mathsf{S}$) on `Test' dataset of CelebA, considering $d_{\mathbf{z}}=64$ (First Row) and $d_{\mathbf{z}}=128$ (Second Row), setting $Q_{\boldsymbol{\psi}}\! \left( \mathbf{Z} \right) = \mathcal{N} \! \left( \boldsymbol{0}, \mathbf{I}_{d_{\mathbf{z}}} \right)$, for different information complexity weights $\beta$ and information leakage weights $\alpha$. For each panel, (Left Column): utility task is gender recognition, sensitive attribute is emotion (smiling); (Right Column): utility task is emotion (smiling) recognition, sensitive attribute is gender.}
\label{Fig:AccUandS_CelebA_P1supervised}
\end{figure}

\begin{figure}[!t]
\centering
\includegraphics[width=0.6\textwidth]{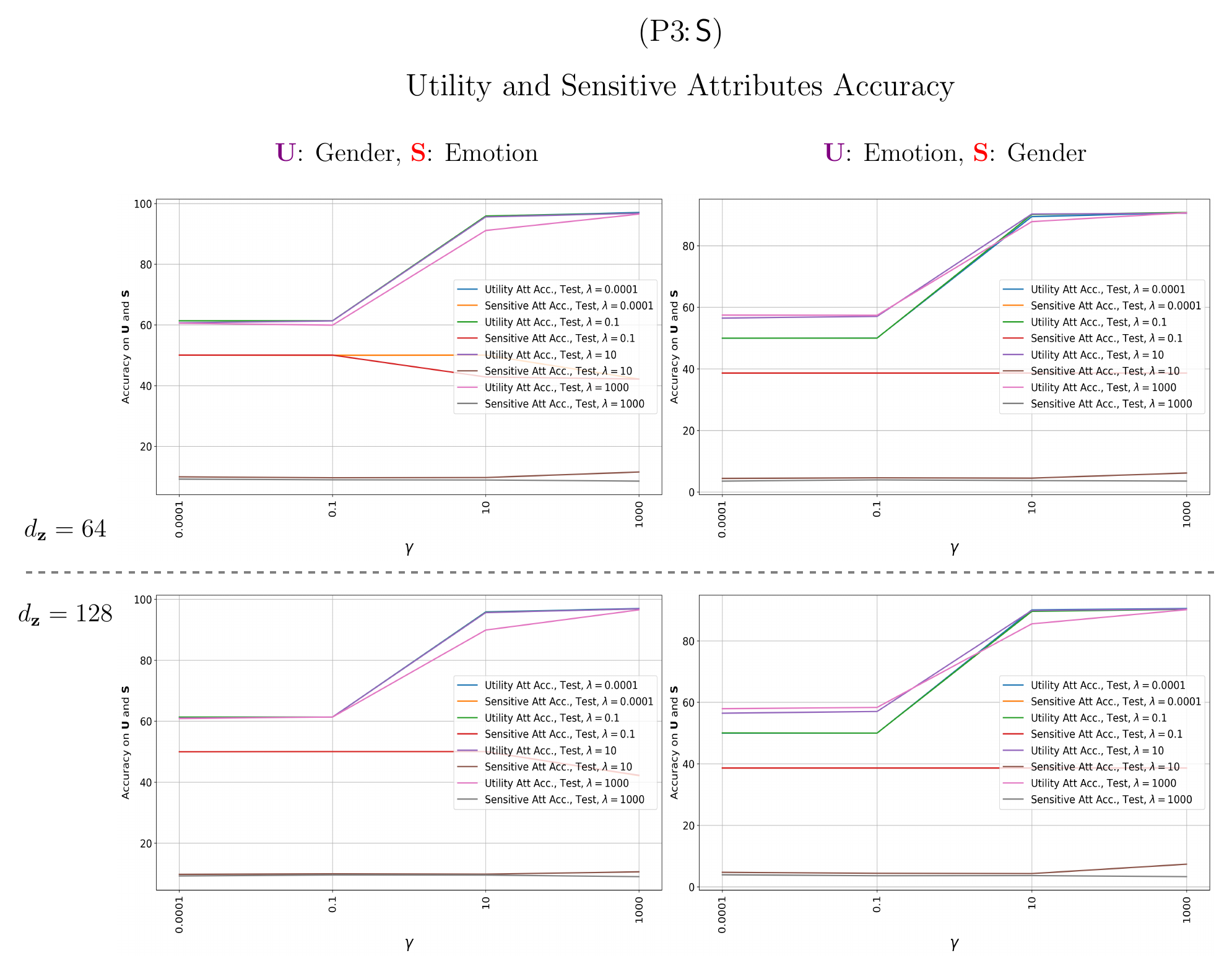}
%% Accuracy on U and S
%% P3 Supervised
%% d: 64 & 128
\caption{Recognition accuracy of the utility attribute $\mathbf{U}$ and sensitive attribute $\mathbf{S}$ for supervised CLUB model ($\text{P3}\!\!:\! \mathsf{S}$) on `Test' dataset of CelebA, considering $d_{\mathbf{z}}=64$ (First Row) and $d_{\mathbf{z}}=128$ (Second Row), setting $Q_{\boldsymbol{\psi}}\! \left( \mathbf{Z} \right) = \mathcal{N} \! \left( \boldsymbol{0}, \mathbf{I}_{d_{\mathbf{z}}} \right)$, for different information utility weights $\gamma$ and information leakage weights $\lambda$. (Left Column): utility task is gender recognition, sensitive attribute is emotion (smiling); (Right Column): utility task is emotion (smiling) recognition, sensitive attribute is gender.}
\label{Fig:AccUandS_CelebA_P3supervised}
\end{figure}

\begin{figure}[!t]
\centering
\includegraphics[width=0.49\textwidth]{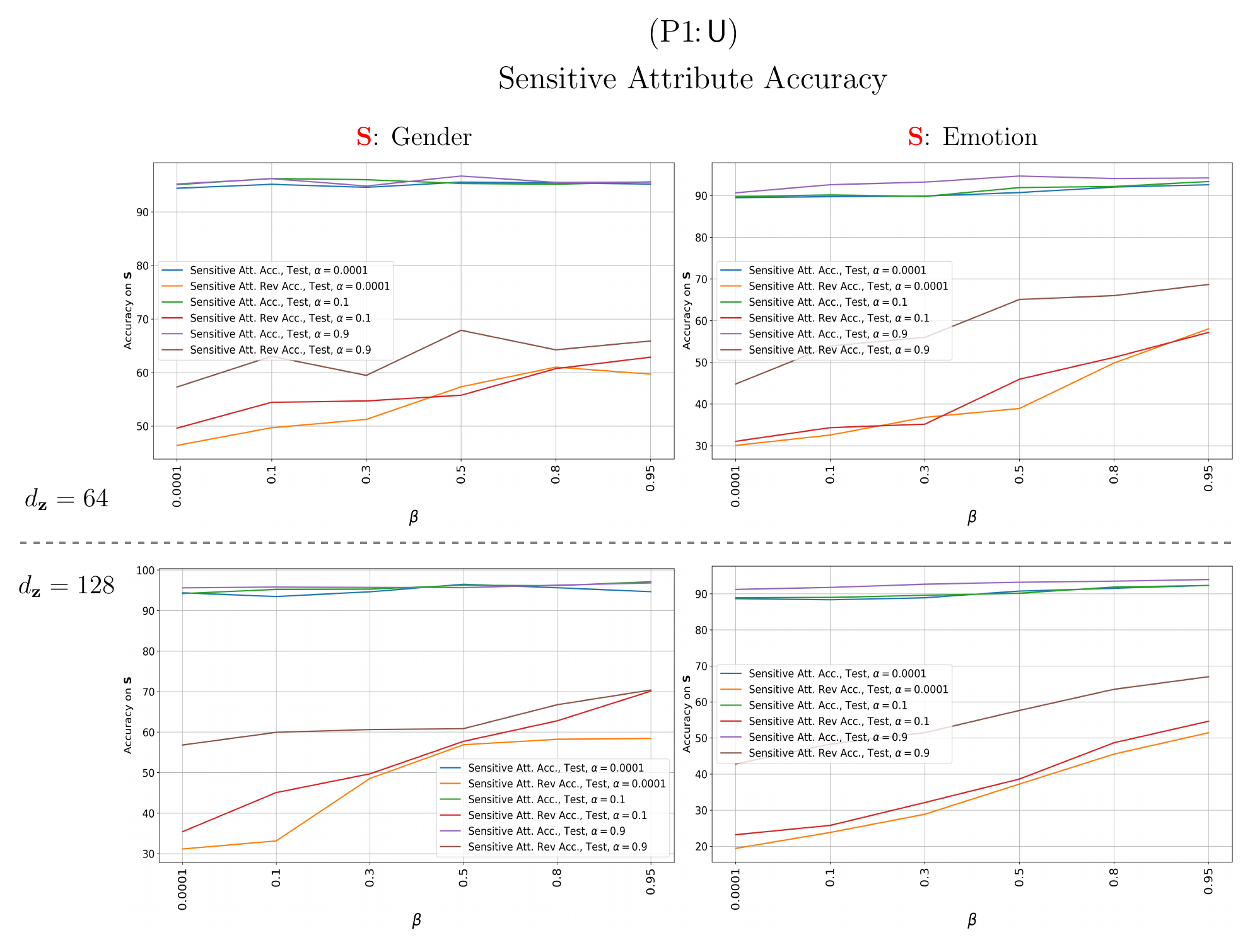}
~
\includegraphics[width=0.49\textwidth]{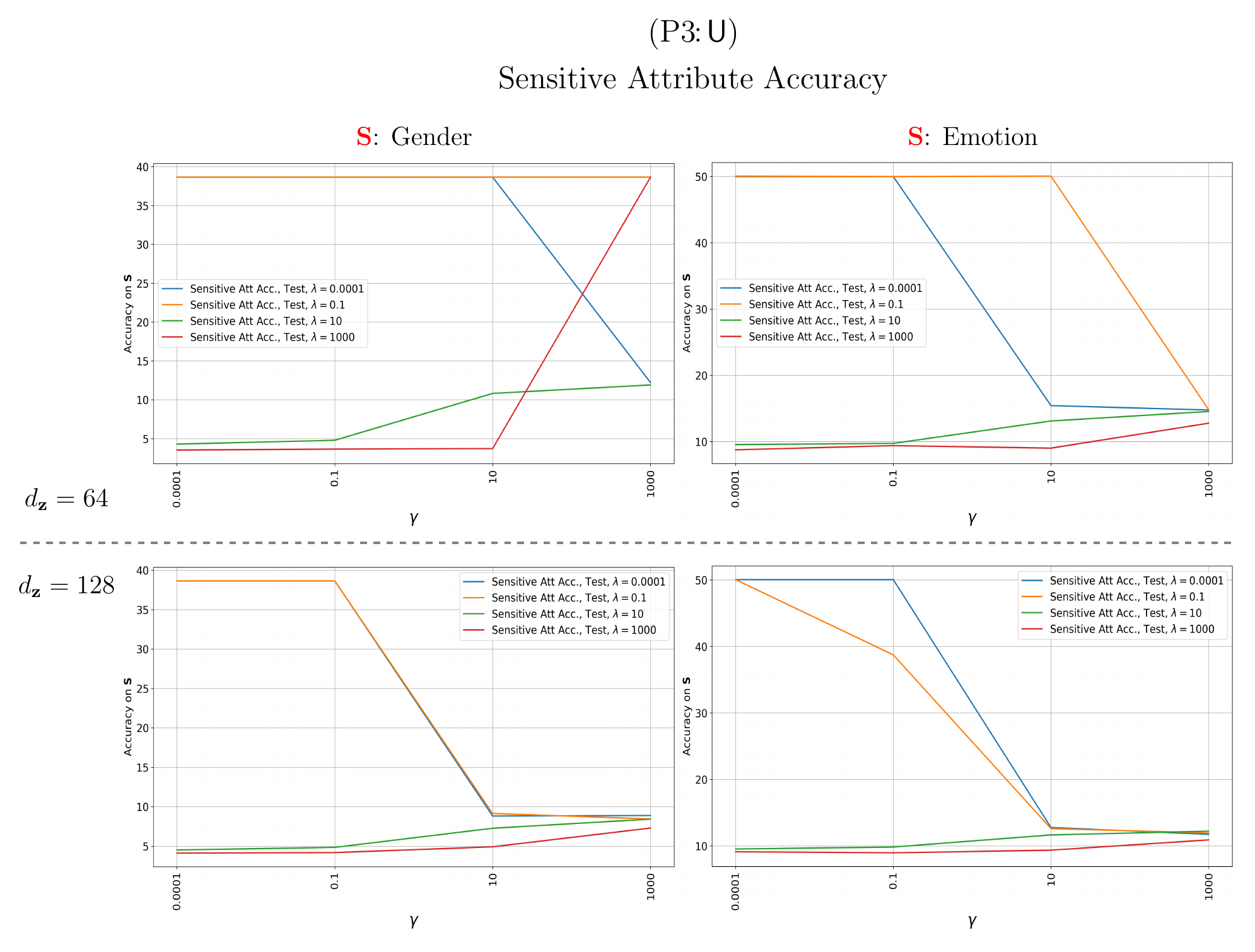}
%% Accuracy on S
%% P1 UnSupervised
%% d: 64 & 128
\caption{Recognition accuracy of the sensitive attribute $\mathbf{S}$ for unsupervised CLUB models ($\text{P1}\!\!:\! \mathsf{U}$) (Left Panel) and ($\text{P3}\!\!:\! \mathsf{U}$) (Right Panel) on `Test' dataset of CelebA, considering $d_{\mathbf{z}}=64$ (First Row) and $d_{\mathbf{z}}=128$ (Second Row), setting $Q_{\boldsymbol{\psi}}\! \left( \mathbf{Z} \right) = \mathcal{N} \! \left( \boldsymbol{0}, \mathbf{I}_{d_{\mathbf{z}}} \right)$; for different information complexity weights $\beta$ and information leakage weights $\alpha$ in ($\text{P1}\!\!:\! \mathsf{U}$), and for different information utility weights $\gamma$ and information leakage weights $\lambda$ in ($\text{P3}\!\!:\! \mathsf{U}$). For each panel, (Left Column): sensitive attribute is gender; (Right Column): sensitive attribute is emotion (smiling).}
\label{Fig:AccS_CelebA_P1unsupervised}
\end{figure}

\begin{figure}[!t]
\centering
\includegraphics[width=\textwidth]{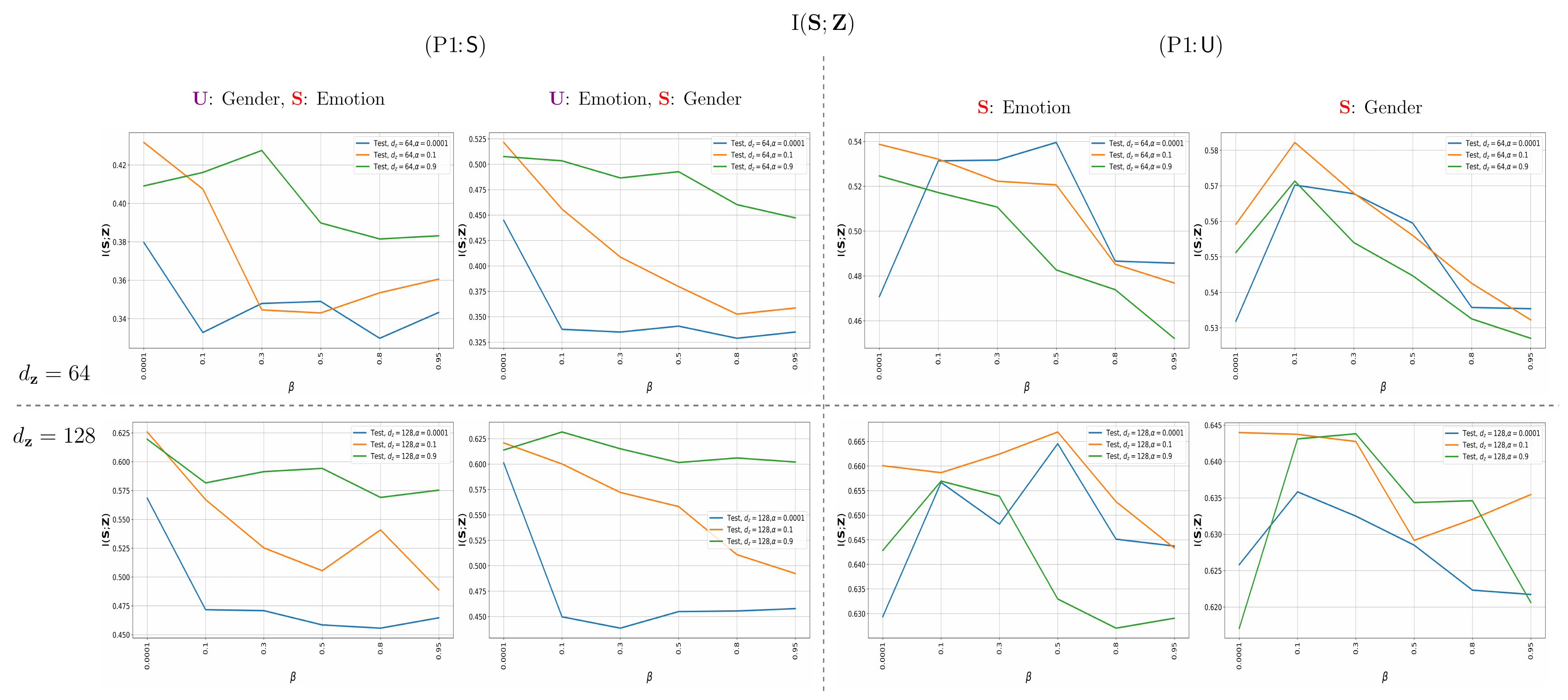}
%% I(S ; Z)
%% P1 Supervised and UnSupervised
%% d: 64 & 128
\caption{Estimated information leakage $\I (\mathbf{S}; \mathbf{Z})$ for CLUB models ($\text{P1}\!\!:\! \mathsf{S}$) (Left Panel) and ($\text{P1}\!\!:\! \mathsf{U}$) (Right Panel) on `Test' dataset of CelebA using MINE, considering $d_{\mathbf{z}}=64$ (First Row) and $d_{\mathbf{z}}=128$ (Second Row), setting $Q_{\boldsymbol{\psi}}\! \left( \mathbf{Z} \right) = \mathcal{N} \! \left( \boldsymbol{0}, \mathbf{I}_{d_{\mathbf{z}}} \right)$, for different information complexity weights $\beta$ and information leakage weights $\alpha$.}
\label{Fig:MIsz_CelebA_P1SandU}
\end{figure}

\begin{figure}[!t]
\centering
\includegraphics[width=\textwidth]{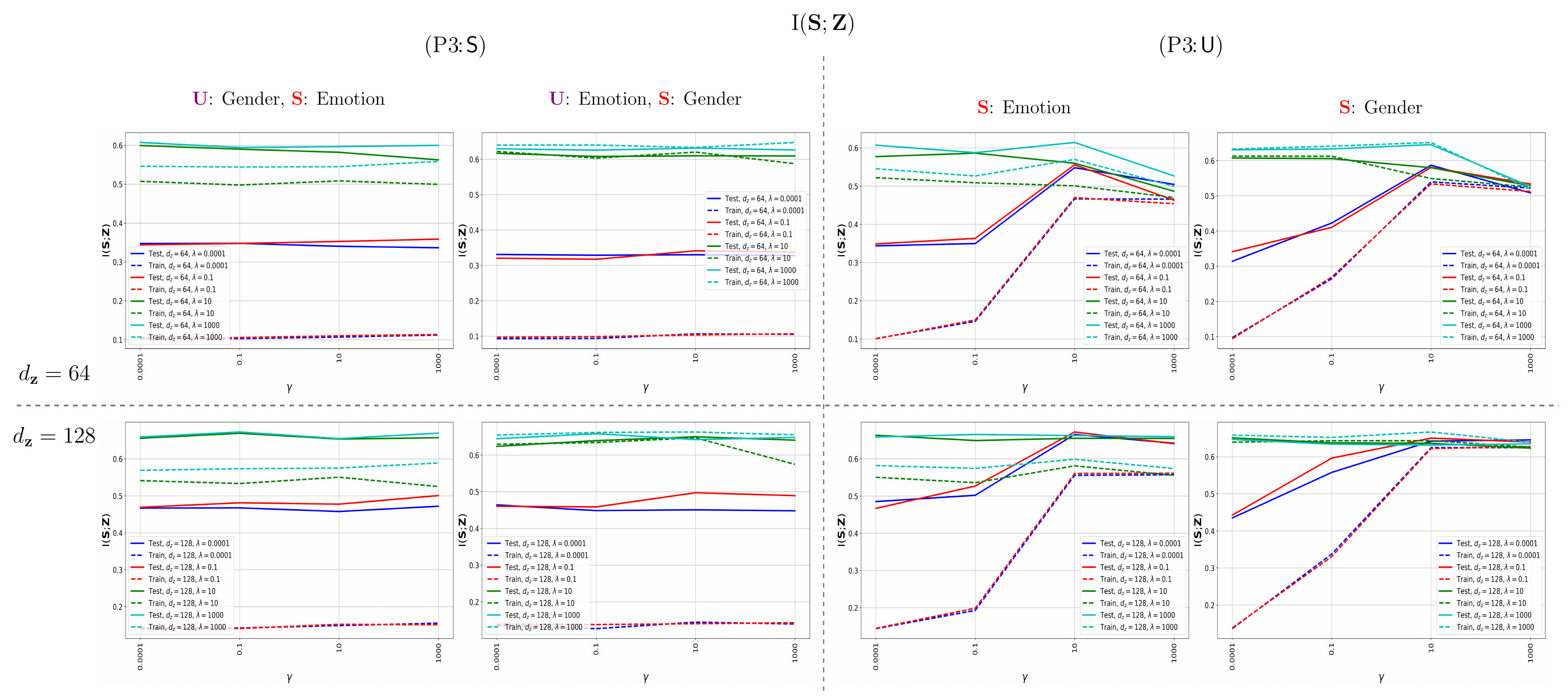}
%% I(S ; Z)
%% P3 Supervised and UnSupervised
%% d: 64 & 128
\caption{Estimated information leakage $\I (\mathbf{S}; \mathbf{Z})$ for CLUB models ($\text{P3}\!\!:\! \mathsf{S}$) (Left Panel) and ($\text{P3}\!\!:\! \mathsf{U}$) (Right Panel) on `Test' and `Train' datasets of CelebA using MINE, considering $d_{\mathbf{z}}=64$ (First Row) and $d_{\mathbf{z}}=128$ (Second Row), setting $Q_{\boldsymbol{\psi}}\! \left( \mathbf{Z} \right) = \mathcal{N} \! \left( \boldsymbol{0}, \mathbf{I}_{d_{\mathbf{z}}} \right)$, for different information utility weights $\gamma$ and information leakage weights $\lambda$.}
\label{Fig:MIsz_CelebA_P3SandU}
\end{figure}

\begin{figure}[!t]
\centering
\includegraphics[width=\textwidth]{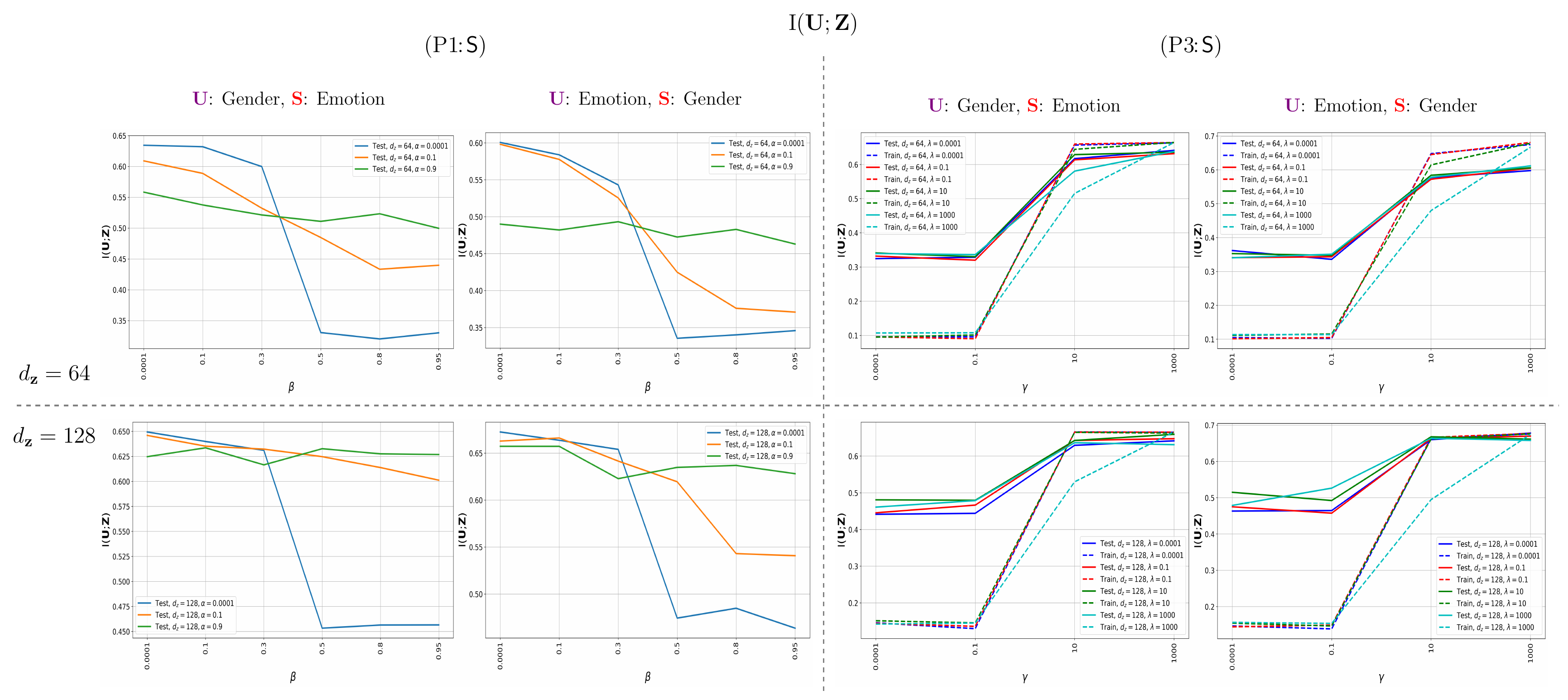}
%% I(U ; Z)
%% P1 Supervised and P3 Supervised
%% d: 64 & 128
\caption{Estimated information utility $\I ( \mathbf{U}; \mathbf{Z} )$ for supervised CLUB models ($\text{P1}\!\!:\! \mathsf{S}$) (Left Panel) and ($\text{P3}\!\!:\! \mathsf{S}$) (Right Panel) on `Test' and `Train' datasets of CelebA using MINE, considering $d_{\mathbf{z}}=64$ (First Row) and $d_{\mathbf{z}}=128$ (Second Row), setting $Q_{\boldsymbol{\psi}}\! \left( \mathbf{Z} \right) = \mathcal{N} \! \left( \boldsymbol{0}, \mathbf{I}_{d_{\mathbf{z}}} \right)$, for different information complexity weights $\beta$ and information leakage weights $\alpha$ in ($\text{P1}\!\!:\! \mathsf{S}$), and for different information utility weights $\gamma$ and information leakage weights $\lambda$ in ($\text{P3}\!\!:\! \mathsf{S}$).}
\label{Fig:MIuz_CelebA_P1P3S}
\end{figure}

\begin{figure}[!t]
\centering
\includegraphics[width=\textwidth]{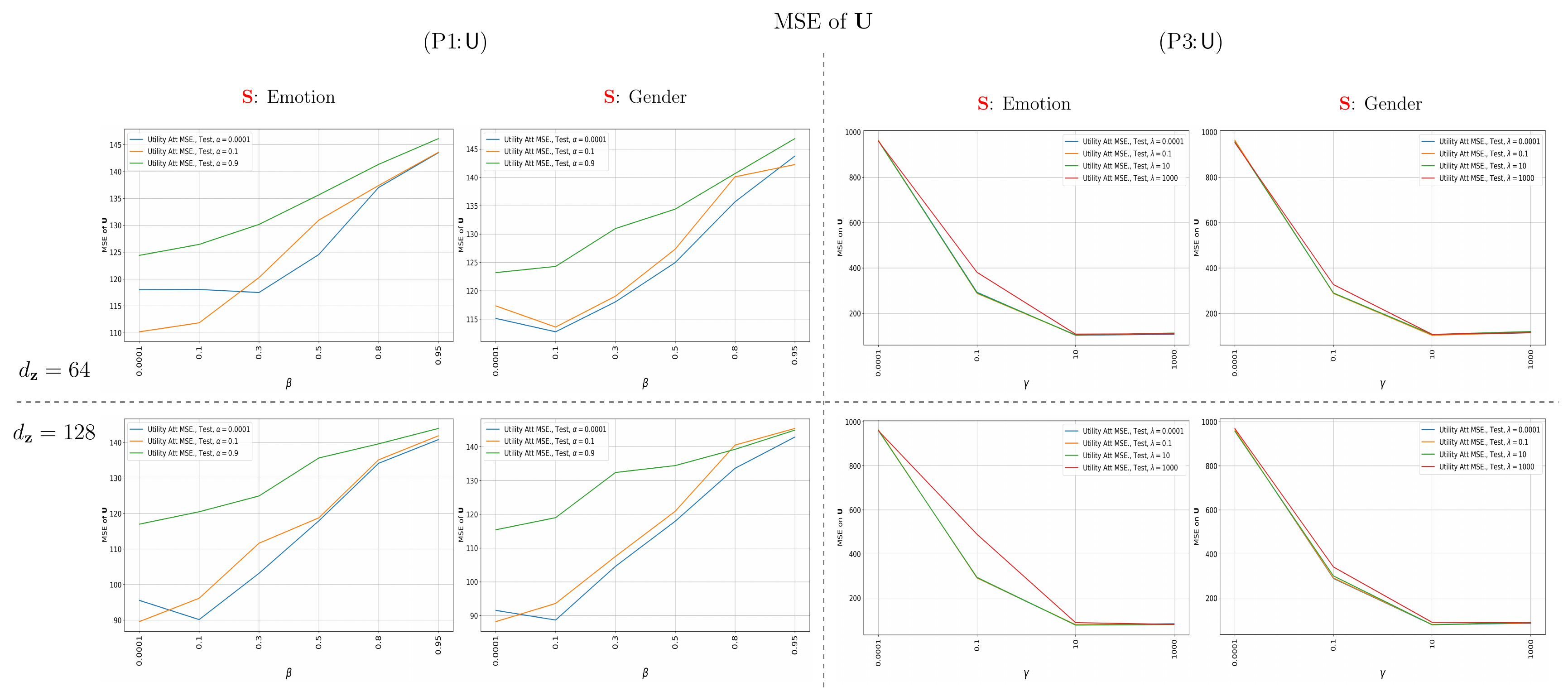}
%% MSE of U
%% P1 and P3 UnSupervised
%% d: 64 & 128
\caption{Mean Square Error (MSE) of reconstructed $\mathbf{U} \equiv \mathbf{X}$ for unsupervised CLUB models ($\text{P1}\!\!:\! \mathsf{U}$) (Left Panel) and ($\text{P3}\!\!:\! \mathsf{U}$) (Right Panel) on `Test' dataset of CelebA, considering $d_{\mathbf{z}}=64$ (First Row) and $d_{\mathbf{z}}=128$ (Second Row), setting $Q_{\boldsymbol{\psi}}\! \left( \mathbf{Z} \right) = \mathcal{N} \! \left( \boldsymbol{0}, \mathbf{I}_{d_{\mathbf{z}}} \right)$, for different information complexity weights $\beta$ and information leakage weights $\alpha$ in ($\text{P1}\!\!:\! \mathsf{U}$), and for different information utility weights $\gamma$ and information  leakage weights $\lambda$ in ($\text{P3}\!\!:\! \mathsf{U}$).}
\label{Fig:MSEu_CelebA_P1P3U}
\end{figure}

\begin{figure}[t]
    \centering
    \includegraphics[width=\textwidth]{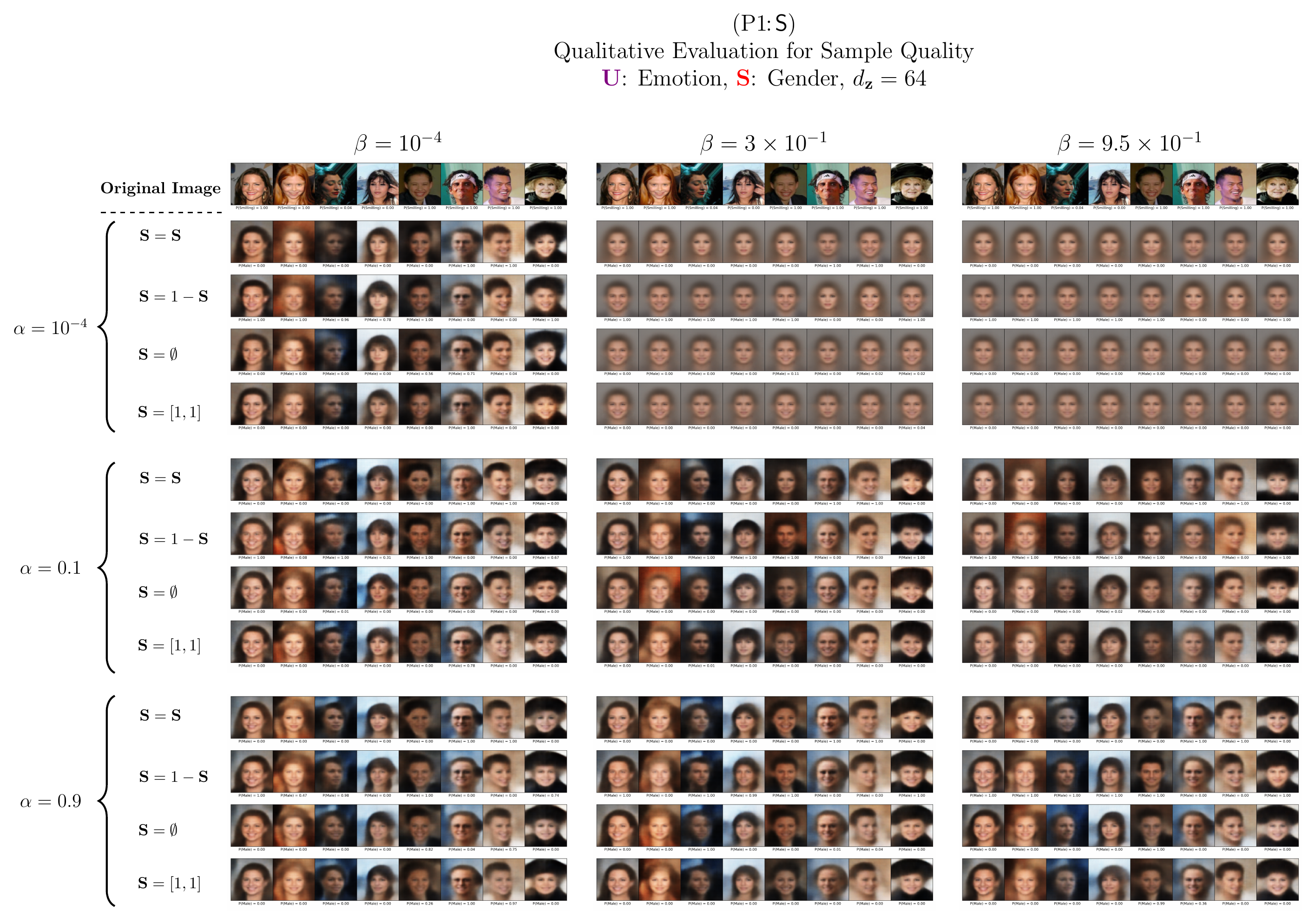}
%    \vspace{-10pt}
    \caption[]
    {Qualitative evaluation for sample quality at the inferential adversary for supervised CLUB model ($\text{P1}\!\!:\! \mathsf{S}$) on CelebA, setting $d_{\mathbf{z}}=64$, $\mathbf{U}$: Emotion, $\mathbf{S}$: Gender, $Q_{\boldsymbol{\psi}}\! \left( \mathbf{Z} \right) = \mathcal{N} \! \left( \boldsymbol{0}, \mathbf{I}_d \right)$, for different trade-offs between $\beta$ and $\alpha$. From left to right: increasing information complexity weight $\beta$. From top to down: increasing information uncertainly weight $\alpha$. Four scenarios are considered to investigate the adversary's beliefs about sensitive attribute $\mathbf{S}$. $\mathbf{S} =  \mathbf{S}$ corresponds to the case in  which the adversary recovers $\mathbf{X}$ when his belief coincides with the original $\mathbf{S}$ (i.e., $\mathbf{S} = \left[ 1, 0 \right]$ for man, and $\mathbf{S} = \left[ 0, 1 \right]$ for woman); $\mathbf{S} = 1 - \mathbf{S}$ corresponds to the case in which the adversary recovers $\mathbf{X}$ when his belief coincides with the inverse of original $\mathbf{S}$; $\mathbf{S} =  \emptyset $ corresponds to the case in which the adversary recovers $\mathbf{X}$ without any assumption on the sensitive attribute $\mathbf{S}$; $\mathbf{S} =  \left[ 1, 1 \right]$ correspond to a possible, probably meaningless, adversary's belief about sensitive attribute, to infer the sensitive attribute $\mathbf{S}$ from reconstructed $\mathbf{X}$. Gender probabilities are computed and depicted under each image. These probabilities show how accurately an adversary can revise his belief about $\mathbf{S}$, at different information leakage weights $\alpha$ and information complexity weights $\beta$.}
%    \vspace{-20pt}
    \label{Fig:CelebA_VisualRec_P1S_UemotionSgender_d64}
\end{figure}

The experiments on CelebA dataset are depicted in this subsection. 
The recognition accuracy of the utility attribute for the supervised CLUB models ($\text{P1}\!\!:\! \mathsf{S}$) and ($\text{P3}\!\!:\! \mathsf{S}$) are depicted in Fig.~\ref{Fig:AccUandS_CelebA_P1supervised} and Fig.~\ref{Fig:AccUandS_CelebA_P3supervised}, respectively. The recognition accuracy of the sensitive attribute for the unsupervised CLUB models ($\text{P1}\!\!:\! \mathsf{U}$) and ($\text{P3}\!\!:\! \mathsf{U}$) is depicted in Fig.~\ref{Fig:AccS_CelebA_P1unsupervised}. 
The estimated information leakage for CLUB models ($\text{P1}$) and ($\text{P3}$) are depicted in Fig.~\ref{Fig:MIsz_CelebA_P1SandU} and Fig.~\ref{Fig:MIsz_CelebA_P3SandU}, respectively, for the supervised and unsupervised setup. 
Fig.~\ref{Fig:MIuz_CelebA_P1P3S} depicts estimated information utility for the CLUB models ($\text{P1}\!\!:\! \mathsf{S}$) and ($\text{P3}\!\!:\! \mathsf{S}$). 
The MSE results of utility data $\mathbf{U} \equiv \mathbf{X}$ for the unsupervised scenarios ($\text{P1}\!\!:\! \mathsf{U}$) and ($\text{P3}\!\!:\! \mathsf{U}$), are depicted in Fig.~\ref{Fig:MSEu_CelebA_P1P3U}. 
We provide a qualitative evaluation for sample quality at the inferential adversary for different CLUB models in Fig.~\ref{Fig:CelebA_VisualRec_P1S_UemotionSgender_d64} and Fig.~\ref{Fig:CelebA_VisualRec_P1U_Semotion_d64}. 
The training details and network architectures are provided in Appendix.~\ref{Appendix:NetworkArchitecture_TrainingDetails} and Appendix~\ref{Appendix:NetworksArchitecture}.

% The experiments on CelebA consider the scenarios in which the attributes $\mathbf{U}$ and $\mathbf{S}$ are correlated, while $\vert \mathcal{U} \vert = \vert \mathcal{S} \vert = 2$.  We provide utility accuracy curves for (i) training set, (ii) validation set, and (iii) test set. 
% As we have argued, there is a direct relationship between information complexity and intrinsic information leakage. Note that, as $\beta$ increases, the information complexity is reduced, and we observe that this also results in a reduction in the information leakage. We also see that the leakage is further reduced when the dimension of the released representation $\mathbf{Z}$, i.e., $d_{\mathrm{z}}$, is reduced. This forces the data owner to obtain a more succinct representation of the utility variable, removing any extra information.

%---------------------------------------------------
% Main Body
%---------------------------------------------------
%  Figure: CelebA >>>>> Acc on U and S
%  P1 Supervised
%---------------------------------------------------
%
%
%---------------------------------------------------
%---------------------------------------------------

% %---------------------------------------------------
% %---------------------------------------------------
% %  Figure: CelebA >>>>> Acc on S  
% %  P1 Supervised
% %---------------------------------------------------
% %
% \begin{figure}[!t]
% \centering
% \includegraphics[width=0.7\textwidth]{ResultTeX_CelebA/AccS_CelebA_P1supervised}
% %% Accuracy on S
% %% P1 Supervised
% %% d: 64 & 128
% \caption{Recognition accuracy of the sensitive attribute $\mathbf{S}$ for supervised CLUB model ($\text{P1}\!\!:\! \mathsf{S}$) on `Test' dataset of CelebA, considering $d_{\mathbf{z}}=64$ (First Row) and $d_{\mathbf{z}}=128$ (Second Row), setting $Q_{\boldsymbol{\psi}}\! \left( \mathbf{Z} \right) = \mathcal{N} \! \left( \boldsymbol{0}, \mathbf{I}_{d_{\mathbf{z}}} \right)$, for different information complexity weights $\beta$ and information uncertainty weights $\alpha$. (Left Column): utility task is gender recognition, sensitive attribute is emotion (smiling); (Right Column): utility task is emotion (smiling) recognition, sensitive attribute is gender.}
% \label{Fig:AccS_CelebA_P1supervised}
% \end{figure}
% %
% %---------------------------------------------------
% %---------------------------------------------------

%---------------------------------------------------
% Main Body
%---------------------------------------------------
%  Figure: CelebA >>>>> Acc on U and S
%  P3 Supervised
%---------------------------------------------------
%
%
%---------------------------------------------------
%---------------------------------------------------

%---------------------------------------------------
%---------------------------------------------------
%  Figure: CelebA >>>>> Acc on S  
%  P1 UnSupervised
%---------------------------------------------------
%
%
%---------------------------------------------------
%---------------------------------------------------

%---------------------------------------------------
% Main Body
%---------------------------------------------------
%  Figure: CelebA >>>>> I(S ; Z)
%  P3 Supervised and UnSupervised
%---------------------------------------------------
%
%
%---------------------------------------------------
%---------------------------------------------------

%---------------------------------------------------
% Mian Body
%---------------------------------------------------
%  Figure: CelebA >>>>> I(U ; Z)
%  P1 Supervised and P3 Supervised
%---------------------------------------------------
%
%
%---------------------------------------------------
%---------------------------------------------------

%---------------------------------------------------
% Main Body
%---------------------------------------------------
%  Figure: CelebA >>>>> MSE on U
%  P1 and P3 UnSupervised
%---------------------------------------------------
%
%
%---------------------------------------------------
%---------------------------------------------------

%---------------------------------------------------
%   Figure: Qualitative evaluation CelebA d=64
%   P1 Supervised
%   U: Emoion
%   S: Gender
%   d_z : 64
%---------------------------------------------------
%
%---------------------------------------------------
%---------------------------------------------------

%---------------------------------------------------
%   Figure: Qualitative evaluation CelebA 
%   P1 Unsupervised
%   S: Emotion
%   d_z : 64
%---------------------------------------------------
%
\begin{figure}[p]
    \centering
    \includegraphics[width=\textwidth]{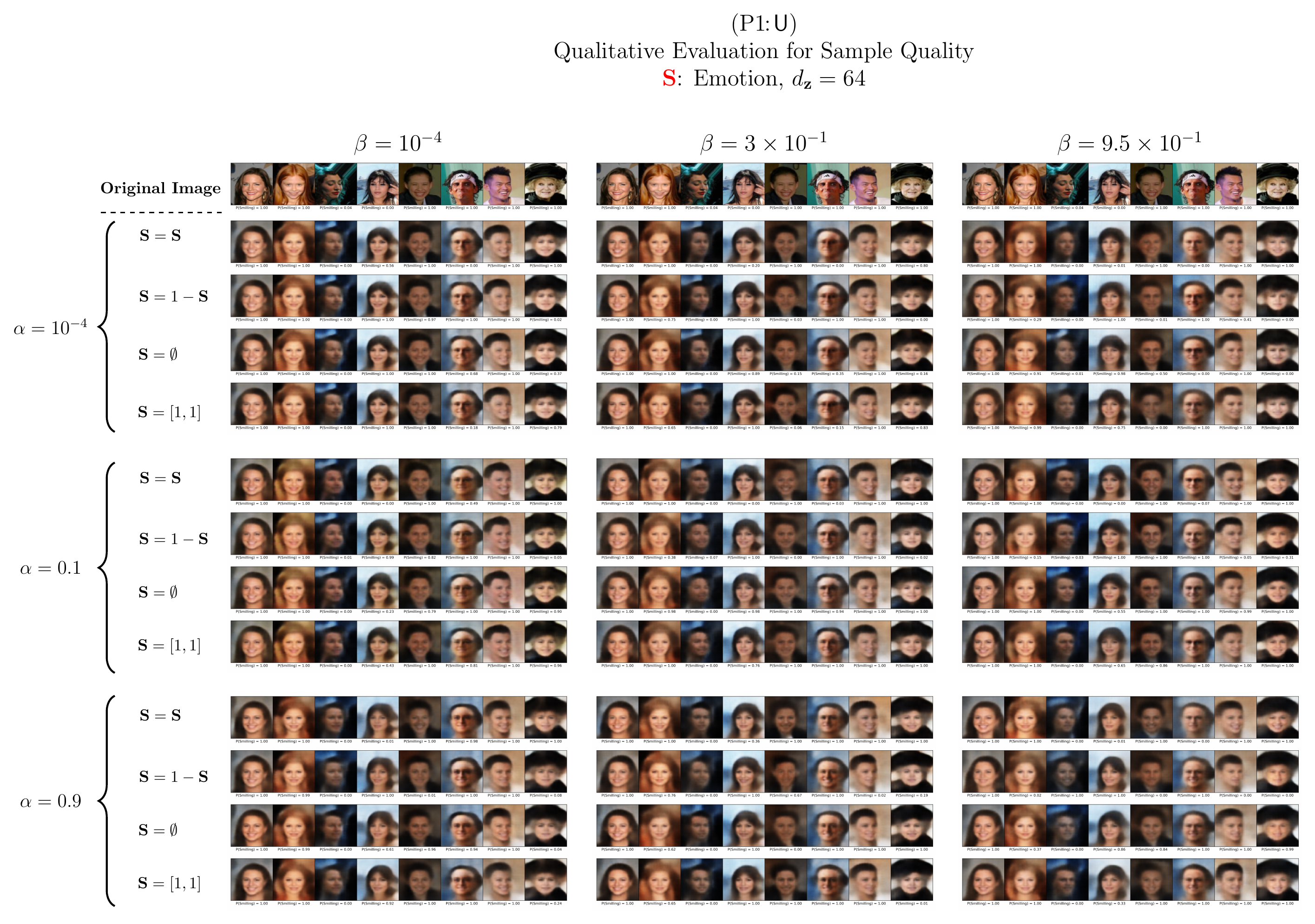}
%    \vspace{-10pt}
    \caption[]
    {Qualitative evaluation for sample quality at the inferential adversary for unsupervised CLUB model ($\text{P1}\!\!:\! \mathsf{U}$) on CelebA, setting $\mathbf{S}$: Emotion, and keeping the same setup as considered in Fig.~\ref{Fig:CelebA_VisualRec_P1S_UemotionSgender_d64}. Emotion probabilities are computed and depicted under each image.}
%   {Qualitative evaluation for sample quality at the inferential adversary for unsupervised CLUB model $\text{P1}\!\!:\! \mathsf{U}$ on CelebA, setting $d=32$, and keeping the same setup as considered in the previous figure.}
%    \vspace{-20pt}
    \label{Fig:CelebA_VisualRec_P1U_Semotion_d64}
\end{figure}

\clearpage
\pagebreak
\newpage

%---------------------------------------------------
%	             	 Conclusion
%---------------------------------------------------
\section{Conclusion}
\label{Sec:Conclusion}

%Following the recent surge in the use of information bottleneck
We proposed a general family of optimization problems that unified and generalized most of the state-of-the-art information-theoretic privacy models. 
We first addressed obfuscation and utility measures under logarithmic loss, and then expressed the concept of relevant information from information-theoretic, statistical, and measure-theoretic perspectives.  
We then introduced and characterized the CLUB optimization problem and established a variational lower bound to optimize the associated CLUB Lagrangian functional. 
We constructed the DVCLUB models by employing neural networks to parameterize variational approximations of the information complexity, information leakage, and information utility quantities. 
We also proposed alternating block coordinate descent training algorithms associated with the proposed DVCLUB models. 
The DVCLUB model sheds light on the connections between information theory and generative models, deep compression models, as well as, fair machine learning models.
In addition, this unifying perspective allows us to relate the CLUB model to several information-theoretic coding problems.
Constructing an $\I$-measure, consistent with Shannon's information measures, we geometrically represent the relationship among Shannon's information measures in state-of-the-art bottleneck problems, interpreting the differences in their objectives. 
Although the CLUB model is formulated using Shannon's mutual information, which leads us to self-consistent equations, one can use other information measures with different operational meanings.

%---------------------------------------------------
%	             	 Acknowledgement
%---------------------------------------------------
\section*{Acknowledgement}
Behrooz Razeghi would like to thank Amir A. Atashin~\orcid{0000-0002-6788-4002} for valuable discussions and contributions in implementing the proposed model. The authors would like to thank Dr.~Shahab~Asoodeh~\orcid{0000-0003-4960-6081} for insightful suggestions leading to improvements of this manuscript.

%---------------------------------------------------
%	                   Bibliography
%--------------------------------------------------- 
\clearpage
\pagebreak
\newpage

\bibliographystyle{IEEEtran}
\bibliography{references}

%-----------------------------------------------------------------
%                        Appendices
%-----------------------------------------------------------------
%\appendix
\appendices
\addtocontents{toc}{\protect\setcounter{tocdepth}{1}}
\clearpage
%\goodbreak
\clearpage
\newpage

\counterwithin{table}{section}
\counterwithin{figure}{section}

\section{Training Details}
\label{Appendix:NetworkArchitecture_TrainingDetails}

\vspace{12pt}

All of the experiments in the study were carried out in the following order:

\vspace{7pt}

\subsubsection{Warm-up Phase}\hfill \break
We utilize the warm-up phase before running training Algorithms \ref{Algorithm:P1_Supervised_DVCLUB},\ref{Algorithm:P1_Unsupervised_DVCLUB},\ref{Algorithm:P3_Supervised_DVCLUB_S_LowerBound},\ref{Algorithm:P3_UnSupervised_DVCLUB_S_LowerBound} for all experiments. In the warm-up phase, we pre-trained encoder $\left(f_{\boldsymbol{\phi}}\right)$ and utility-decoder $\left(g_{\boldsymbol{\theta}}\right)$ together for a few epochs via backpropagation with the Adam optimizer \cite{kingma2014adam}. We found out the warm-up stage was helpful in method convergence speed. Therefore, we initialize the encoder and the utility-decoder weights with the obtained values rather than random or zero initialization. 
For each experiment, the hyper-parameters of the learning algorithm in this phase are depicted in TABLE~\ref{table:phase1_parameters}.

\vspace{7pt}

\begin{table}[h]
\centering
\resizebox{0.65\linewidth}{!}{%
 \begin{tabular}{c c c c} 
\toprule 
 Experiment Dataset & Learning Rate & Max Iteration & Batch Size \\ %[0.5ex] 
 \hline
 Colored-MNIST & \multirow{2}{*}{0.001} & \multirow{2}{*}{50} & \multirow{2}{*}{1024} \\
(both version) &  & &\\ 
 \hline
 CelebA & 0.005 & 50 & 1024 \\
\toprule 
\end{tabular}
}
\caption{Models' hyper-parameters in the Warm-Up phase for the experiments.}
\label{table:phase1_parameters}
\end{table}

\vspace{7pt}

\subsubsection{Main Block-wised Training Phase}\hfill \break
In contrast to most neural network models that only have one forward step and a backward step in each epoch to update all network weights, each of our learning algorithms consists of several blocks in which forward and backward steps have been done on different paths. The model's parameters update based on the corresponding path.

%-------------------------------------------------------------
%
%        To Add.  >>>>   Figure of paths in the algorithm 1 
%
%-----------------------------------------------------------

Since it was not possible for us to use the default model training function of the Keras API, we implement All six training Algorithms \ref{Algorithm:P1_Supervised_DVCLUB},\ref{Algorithm:P1_Unsupervised_DVCLUB},\ref{Algorithm:P3_Supervised_DVCLUB_S_LowerBound},\ref{Algorithm:P3_UnSupervised_DVCLUB_S_LowerBound} from scratch in Tensorflow. It is important to remember that we initialize all parameters to zero except for the $\left( \boldsymbol{\phi}, \boldsymbol{\theta} \right)$ values acquired in the previous stage. Furthermore, we set the learning rate of the block (1) in all Algorithms to five times larger than other blocks. The hyper-parameters of models for each experiment are shown in TABLE~\ref{table:phase2_parameters}.

\vspace{5pt}

\begin{table}[h]
\centering
\resizebox{0.65\linewidth}{!}{%
 \begin{tabular}{c c c c} 
\toprule 
 Experiment Dataset & Learning Rate & Max Iteration & Batch Size \\ %[0.5ex] 
                    &  [blocks (2)-(5)]&           & \\
 \hline 
 Colored-MNIST & \multirow{2}{*}{0.0001} & \multirow{2}{*}{500} & \multirow{2}{*}{2048} \\
 (both version) & & & \\ 
 \hline
 CelebA & 0.00001 & 500 & 1024 \\
\toprule 
\end{tabular}
}
\caption{Models' hyper-parameters in the main phase for the experiments.}
\label{table:phase2_parameters}
\end{table}

%\vspace{5pt}

\pagebreak

%\appendix{Networks's Architecture}
\section{Network Architecture}
\label{Appendix:NetworksArchitecture}

%\subsection{Networks' Architecture}

\subsubsection{Colored-MNIST}\hfill \break
In the Colored-MNIST experiment, we consider two setups for data utility and information leakage evaluation. 
In the first scenario, we set the utility data to the class label of the input image and consider the color of the input image as sensitive data, and for the second one, we did vice versa. It is worth mentioning both balanced and unbalanced Colored-MNIST datasets are applied with the same architecture. The architecture of networks is given in Table~\ref{table:Architecture_Colored-MNIST} and \ref{table:Architecture_Colored-MNIST_PartII}.

\vspace{-5pt}

%-------------------------------------------------------------
%
%      TABLE: Architecture  >>  Colored-MNIST ::: Encoders and decoders
%
%-----------------------------------------------------------
%
\begin{table}[!h]
\scshape
\small 
\centering
{\renewcommand{\arraystretch}{1.14}%
\begin{tabular}{ p{0.45\linewidth}}
\toprule 
%------------
%%%% Encoder
%------------
\multicolumn{1}{c}{Encoder ($f_{\boldsymbol{\phi}}$) - Both Setups} \\
\hline
%%%% row 1
Input~~$\mathbf{x} \in \mathbb{R}^{28 \times 28 \times 3}$ Color Image \\
%%%% row 2
Conv(64,5,2), BN, LeakyReLU   \\
%%%% row 3
Conv(128,5,2), BN, LeakyReLU   \\
%%%% row 4
   Flatten  \\
%%%% row 5
FC($d_z\times4$), BN, Tanh \\
%%%% row 6
   $\mu$= FC($d_z$), $\sigma$= FC($d_z$)   \\
%%%% row 7
   $z$=SamplingWithReparameterizationTrick[$\mu$,$\sigma$] \\
\toprule 
%------------
%%%% Decoder : Supervised
%------------
\multicolumn{1}{c}{Utility Decoder ($g_{\boldsymbol{\theta}}$) - Supervised Setup} \\
\hline
%%%% row 1
Input~~$\mathbf{z} \in \mathbb{R}^{d_z}$ Code\\
%%%% row 2
FC($d_z \times 4$), BN, LeakyReLU \\
%%%% row 3
FC($\vert \mathcal{U} \vert$), SOFTMAX \\
\toprule
%------------
%%%% Decoder : Un-Supervised
%------------
\multicolumn{1}{c}{Utility Decoder ($g_{\boldsymbol{\theta}}$) - Unsupervised Setup} \\
\hline
%%%% row 1
Input~~$\mathbf{z} \in \mathbb{R}^{d_{\mathrm{z}}}$ Code\\
%%%% row 2
FC($6272$), BN, LeakyReLU\\
%%%% row 3
Reshape($(7,7,128)$)\\
 %%%% row 3
DeConv($64,5,2$), BN, LeakyReLU \\
 %%%% row 4
DeConv($3,5,2$), Sigmoid\\
\toprule
%------------
%%%% Uncertainty Decoder :: S
%------------
\multicolumn{1}{c}{Uncertainty Decoder ($g_{\boldsymbol{\varphi}}$)} \\
\hline
%%%% row 1
Input~~$\mathbf{z} \in \mathbb{R}^{d_{\mathrm{z}}}$ Code\\
%%%% row 2
FC($6272$), BN, LeakyReLU\\
%%%% row 3
Reshape($(7,7,128)$)\\
 %%%% row 3
DeConv($64,5,2$), BN, LeakyReLU \\
 %%%% row 4
DeConv($3,5,2$), Sigmoid \\
\toprule
%------------
%%%% Uncertainty Decoder :: X^hat obfuscated
%------------
\multicolumn{1}{c}{Uncertainty Decoder ($g_{\boldsymbol{\xi}}$)} \\
\hline
%%%% row 1
Input~~$\mathbf{z} \in \mathbb{R}^{d_{\mathrm{z}}}$ Code, $\mathbf{s}\in \mathbb{R}^{\vert \mathcal{U}\vert}$\\
%%%% row 2
$\mathbf{zs}$ = concatenate($\mathbf{z}, \mathbf{s}$)\\
FC($2\times2\times256$), BN, LeakyReLU\\
%%%% row 3
Reshape($(256,2,2)$)\\
%%%% row 3
DeConv($(128,3,2)$), BN, LeakyReLU\\
DeConv($(64,3,2)$), BN, LeakyReLU\\
DeConv($(32,3,2)$), BN, LeakyReLU\\
DeConv($(16,3,2)$), BN, LeakyReLU\\
DeConv($(8,3,2)$), BN, LeakyReLU\\
%%%% row 4
Conv($3,3,2$), Sigmoid\\
\toprule
%---------------------------------------
%%%% Prior Distribution Generator
%---------------------------------------
\multicolumn{1}{c}{Prior Distribution Generator ($g_{\boldsymbol{\psi}}$) - Both Setups} \\
\hline
%%%% row 1
Input~~Sample Noise$ \sim \mathcal{N} \! \left( \boldsymbol{0},\mathbf{I}\right)$ \\
%%%% row 2
FC($100$), BN, LeakyReLU \\
%%%% row 3
FC($100$), BN, LeakyReLU \\
%%%% row 4
FC($\vert d_z\vert$), Sigmoid \\
\toprule 
\end{tabular}}
\caption{The encoders and decoders architecture for the Colored-MNIST experiments.}
\label{table:Architecture_Colored-MNIST}
\end{table}
%-------------------------------------------------------------

%-------------------------------------------------------------
%
%  TABLE: Architecture  >>  Colored-MNIST  ::: Discriminators
%
%-----------------------------------------------------------
%
\begin{table}[!h]
\scshape
\small 
\centering
{\renewcommand{\arraystretch}{1.2}%
\begin{tabular}{ p{0.45\linewidth}}
\toprule 
%-------------------------------
%%%% Latent Space Discriminator
%-------------------------------
\multicolumn{1}{c}{Latent Space Discriminator ($D_{\boldsymbol{\eta}}$)} \\
\hline
%%%% row 1
Input~~$\mathbf{z} \in \mathbb{R}^{d_z}$ Code \\
%%%% row 2
FC($128$), BN, LeakyReLU \\
%%%% row 3
FC($64$), BN, LeakyReLU \\
%%%% row 4
FC($1$), Sigmoid \\
\toprule 
%---------------------------------------
%%%% Utility Attribute Class Discriminator
%---------------------------------------
\multicolumn{1}{c}{Utility Attribute Class Discriminator ($D_{\boldsymbol{\omega}}$)} \\
\hline
%%%% row 1
Input~~$\mathbf{u} \in \mathbb{R}^{\vert \mathcal{U} \vert}$ \\
%%%% row 2
FC($\vert \mathcal{U} \vert \times 8$), BN, LeakyReLU \\
%%%% row 3
FC($\vert \mathcal{U} \vert \times 8$), BN, LeakyReLU \\
%%%% row 4
FC($1$), Sigmoid \\
\toprule 
%---------------------------------------
%%%% Visible Space Discriminator
%---------------------------------------
\multicolumn{1}{c}{Visible Space Discriminator ($D_{\boldsymbol{\omega}}$)} \\
\hline
%%%%% row 1
Input~~$\mathbf{\widehat{x}} \in \mathbb{R}^{28\times28\times3}$~~Color Image \\
%%%%% row 2
Conv($32,3,2$), BN, LeakyReLU  \\
%%%%% row 3
Conv($64,3,2$), BN, LeakyReLU  \\
%%%%% row 4
Conv($128,3,2$), BN, LeakyReLU  \\
%%%%% row 5
Flatten  \\
%%%%% row 6
FC($128$), BN, LeakyReLU \\
%%%%% row 7
FC($1$), Sigmoid   \\
%%%%%
\toprule 
%%%%
\end{tabular}}
\caption{The discriminators architecture details for the Colored-MNIST experiments.}
\label{table:Architecture_Colored-MNIST_PartII}
\end{table}
%---------------------------------------------------
%---------------------------------------------------

%-------------------------------------------------------------
%
%               MINE
%
%-----------------------------------------------------------
%
\subsubsection{Mutual Information Estimation}\hfill \break
For all experiments in this paper, we report estimation of MI between the released representation and sensitive attribute, i.e., $\I \left(\mathbf{S};\mathbf{Z}\right)$, as well as the MI between the released representation and utility attribute, i.e., $\I \left(\mathbf{U};\mathbf{Z}\right)$. To approximate mutual information, we employed the MINE model \cite{belghazi2018mutual}. The architecture of the model is depicted in \ref{table:Architecture_MINE}. Note that MINE's network for estimating $\I \left(\mathbf{S};\mathbf{Z} \right)$ has the same architecture as shown in Table \ref{table:Architecture_MINE}.

\begin{table}[h]
\scshape
\small 
\centering
{\renewcommand{\arraystretch}{1.3}%
\begin{tabular}{ p{0.4\linewidth}}
\toprule 
\multicolumn{1}{c}{MINE~~~~$\I \left(\mathbf{U};\mathbf{Z} \right)$} \\
\hline
%%%% row 1
Input~~$\mathbf{z} \in \mathbb{R}^{d_z}$ Code; $\mathbf{u} \in \mathbb{R}^{\vert \mathcal{U} \vert}$ \\
x = Concatenate([z, u]) \\
%%%% row 2
FC(100), ELU   \\
%%%% row 3
FC(100), ELU   \\
%%%% row 4
FC(100), ELU   \\
%%%% row 4
FC(1)  \\
\toprule 
\end{tabular}}
\vspace{-3pt}
\caption{The architecture of MINE network for mutual information estimation.}
\label{table:Architecture_MINE}
\end{table}
%---------------------------------------------------
%---------------------------------------------------

% \pagebreak

%---------------------------------------------------
%       CelebA
%--------------------------------------------------
%
\subsubsection{CelebA}\hfill \break
In this experiment, we considered two scenarios for data utility and sensitive attributes. These setups are shown in Table~\ref{table:CelebA_Scenarios}. 
Note that all of the utility and sensitive attributes are binary data. 
The architecture of networks is given in Table~\ref{table:Architecture_CelebA_PartI}.

\begin{table}[h]
\centering
\resizebox{0.6\linewidth}{!}{%
 \begin{tabular}{c c c} 
\toprule 
Scenario Number & Utility Attribute & Sensitive Attribute\\
 \hline
 1 & Gender & Smiling\\
 \hline
 2 & Smiling & Gender\\
\toprule 
\end{tabular}
}
\caption{The considered scenarios for the CelebA experiments.}
\vspace{-3pt}
\label{table:CelebA_Scenarios}
\end{table}
%--------------------------------------------------

%-------------------------------------------------------------
%
%    TABLE: Architecture  >>> CelebA  ::: encoder and decoders
%
%-------------------------------------------------------------
%
\begin{table}[h]
\scshape
\small 
\centering
{\renewcommand{\arraystretch}{1.3}%
\begin{tabular}{ p{0.45\linewidth}}
\toprule 
%---------------------------------------
%%%%   Encoder
%---------------------------------------
\multicolumn{1}{c}{Encoder $f_{\boldsymbol{\phi}}$ - Both Setups} \\
\hline
%%%% row 1
Input~~$\mathbf{x} \in \mathbb{R}^{64 \times 64 \times 3}$ Color Image \\
%%%% row 2
Conv(64,5,2), BN, LeakyReLU   \\
Conv(128,5,2), BN, LeakyReLU   \\
%%%% row 3
Flatten  \\
%%%% row 4
FC($d_z\times4$), BN, Tanh \\
%%%% row 5
$\mu$= FC($d_z$), $\sigma$= FC($d_z$)   \\
%%%% row 6
$z$=SamplingWithReparameterizationTrick[$\mu$,$\sigma$] \\
\toprule 
%---------------------------------------
%%%%   Utility Decoder : Supervised
%---------------------------------------
\multicolumn{1}{c}{Utility Decoder ($g_{\boldsymbol{\theta}}$) - Supervised Setups} \\
\hline
%%%% row 1
Input~~$\mathbf{z} \in \mathbb{R}^{d_z}$ Code\\
%%%% row 2
FC($d_z\times4$), BN, LeakyReLU \\
%%%% row 3
FC($\vert \mathcal{U} \vert$), SOFTMAX \\
%%%% row 7
%
\toprule 
%---------------------------------------
%%%%   Utility Decoder : Un-Supervised
%---------------------------------------
\multicolumn{1}{c}{Utility Decoder ($g_{\boldsymbol{\theta}}$) - Unsupervised Setups} \\
\hline
%%%% row 1
Input~~$\mathbf{z} \in \mathbb{R}^{d_{\mathrm{z}}}$ Code\\
%%%% row 2
FC($7\times7\times128$), BN, LeakyReLU\\
%%%% row 3
Reshape($(7,7,128)$)\\
 %%%% row 3
DeConv($64,5,2$), BN, LeakyReLU \\
 %%%% row 4
DeConv($3,5,2$), Sigmoid\\
\toprule 
%---------------------------------------
%%%%   Uncertainty Decoder : S^hat
%---------------------------------------
\multicolumn{1}{c}{Uncertainty Decoder ($g_{\boldsymbol{\varphi}}$) - Supervised Setups} \\
\hline
%%%% row 1
Input~~$\mathbf{z} \in \mathbb{R}^{d_{\mathrm{z}}}$ Code, $\mathbf{s}\in \mathbb{R}^{\vert \mathcal{U}\vert}$\\
%%%% row 2
$\mathbf{zs}$ = concatenate($\mathbf{z}, \mathbf{s}$)\\
FC($7\times7\times128$), BN, LeakyReLU\\
%%%% row 3
Reshape($(128,7,7)$), BN, LeakyReLU\\
%%%% row 3
DeConv($(64,5,2)$), BN, LeakyReLU\\
DeConv($(3,5,2)$), Sigmoid\\
\toprule 
%---------------------------------------
%%%%   Uncertainty Decoder : X^hat  obfuscated
%---------------------------------------
\multicolumn{1}{c}{Uncertainty Decoder ($g_{\boldsymbol{\xi}}$)} \\
\hline
Input~~$\mathbf{z} \in \mathbb{R}^{d_{\mathrm{z}}}$ Code\\
FC($d_{\mathrm{z}}\times4$), BN, LeakyReLU\\
FC($\vert \mathcal{S} \vert$), Softmax\\
\toprule 
%---------------------------------------
%%%% Prior Distribution Generator
%---------------------------------------
\multicolumn{1}{c}{Prior Distribution Generator ($g_{\boldsymbol{\psi}}$)} \\
\hline
%%%% row 1
Input~~Sample Noise$ \sim \mathcal{N} \! \left( \boldsymbol{0},\mathbf{I}\right)$ \\
%%%% row 2
FC($100$), BN, LeakyReLU \\
%%%% row 3
FC($100$), BN, LeakyReLU \\
%%%% row 4
FC($\vert d_z\vert$), Sigmoid \\
\toprule 
%%%
\end{tabular}}
\caption{The encoders and decoders architecture for the CelebA experiments.}
\label{table:Architecture_CelebA_PartI}
\end{table}
%---------------------------------------------------

%-------------------------------------------------------------
%
%    TABLE: Architecture  >>> CelebA  ::: encoder and decoders
%
%-------------------------------------------------------------
%
\begin{table}[h!]
\scshape
\small 
\centering
{\renewcommand{\arraystretch}{1.3}%
\begin{tabular}{ p{0.45\linewidth}}
\toprule 
%---------------------------------------
%%%%   Latent Space Discriminator
%---------------------------------------
\multicolumn{1}{c}{Latent Space Discriminator ($D_{\boldsymbol{\eta}}$)} \\
\hline
%%%% row 1
Input~~$\mathbf{z} \in \mathbb{R}^{d_z}$ Code \\
%%%% row 2
FC($512$), BN, LeakyReLU \\
%%%% row 3
FC($128$), BN, LeakyReLU \\
%%%% row 4
FC($1$), Sigmoid \\
\toprule 
%------------------------------------------
%%%%  Utility Attribute Class Discriminator
%------------------------------------------
\multicolumn{1}{c}{Utility Attribute Class Discriminator ($D_{\boldsymbol{\omega}}$)} \\
\hline
%%%% row 1
Input~~$\mathbf{u} \in \mathbb{R}^{\vert \mathcal{U} \vert}$ \\
%%%% row 2
FC($\vert \mathcal{U} \vert \times 4$), BN, LeakyReLU \\
%%%% row 3
FC($\vert \mathcal{U} \vert \times 4$), BN, LeakyReLU \\
%%%% row 4
FC($1$), Sigmoid \\
\toprule 
%------------------------------------------
%%%%  Visible Space Discriminator
%------------------------------------------
\multicolumn{1}{c}{Visible Space Discriminator ($D_{\boldsymbol{\omega}}$)} \\
\hline
%%%%% row 1
%%%% row 1
Input~~$\mathbf{x} \in \mathbb{R}^{64 \times 64 \times 3}$ Color Image \\
%%%% row 2
Conv(32,3,2), BN, LeakyReLU   \\
Conv(64,3,2), BN, LeakyReLU   \\
Conv(128,3,2), BN, LeakyReLU   \\
%%%% row 3
Flatten  \\
%%%% row 4
FC($128$), BN, LeakyReLU \\
%%%% row 5
FC($1$), Sigmoid\\
\toprule 
%%%
\end{tabular}}
\caption{The discriminators architecture for the CelebA experiments.}
\label{table:Architecture_CelebA_PartII}
\end{table}

\clearpage

\section{Implementation Overview}\label{Appendix:ImplementationOverview}

\begin{figure}[!h]
    \centering
    \begin{subfigure}{0.31\textwidth}
        \centering
        \includegraphics[width=\textwidth]{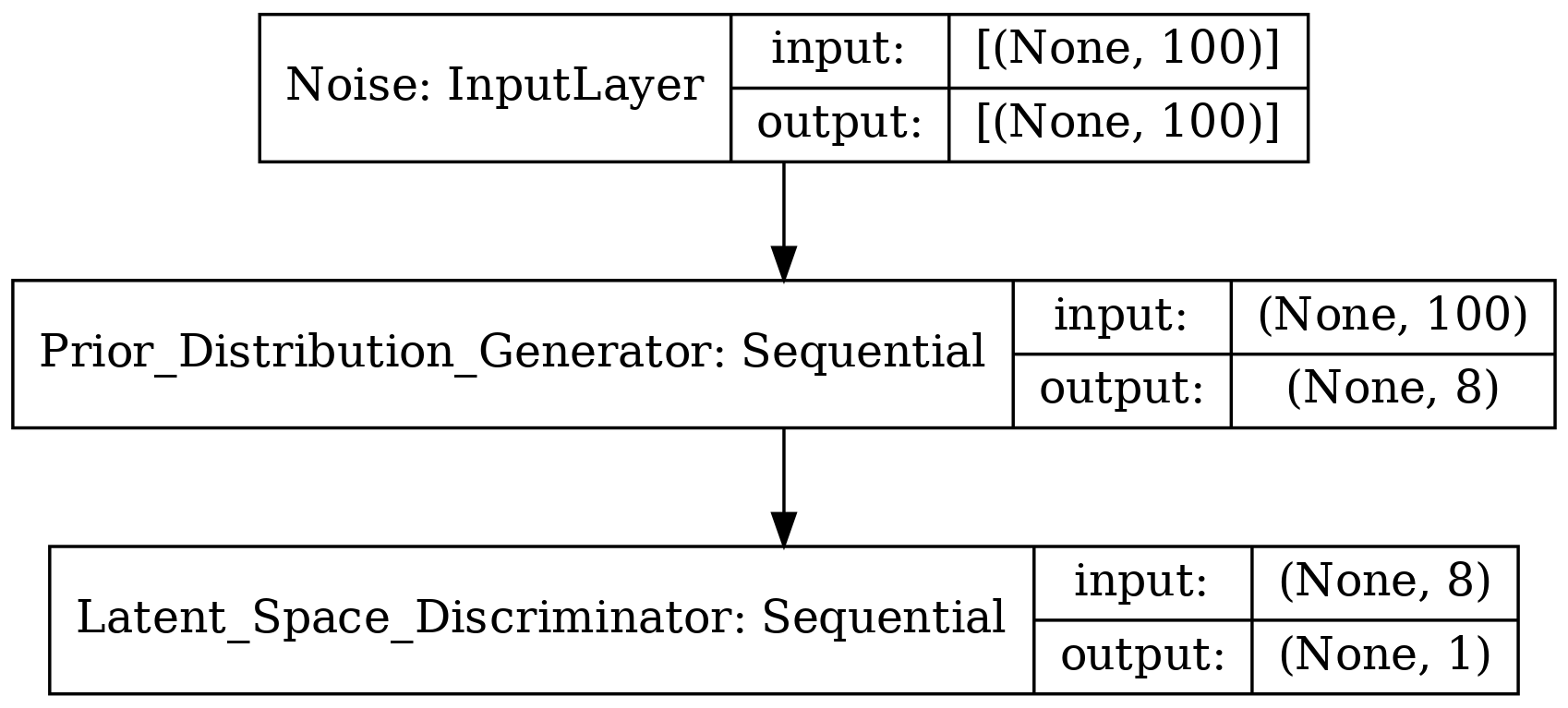}
        \caption{Prior-Generator adjoint Latent-Space-Discriminator module}
        \label{fig:P1_a_mnist}
    \end{subfigure}
    \hfill
    \begin{subfigure}{0.31\textwidth}
        \centering
        \includegraphics[width=\textwidth]{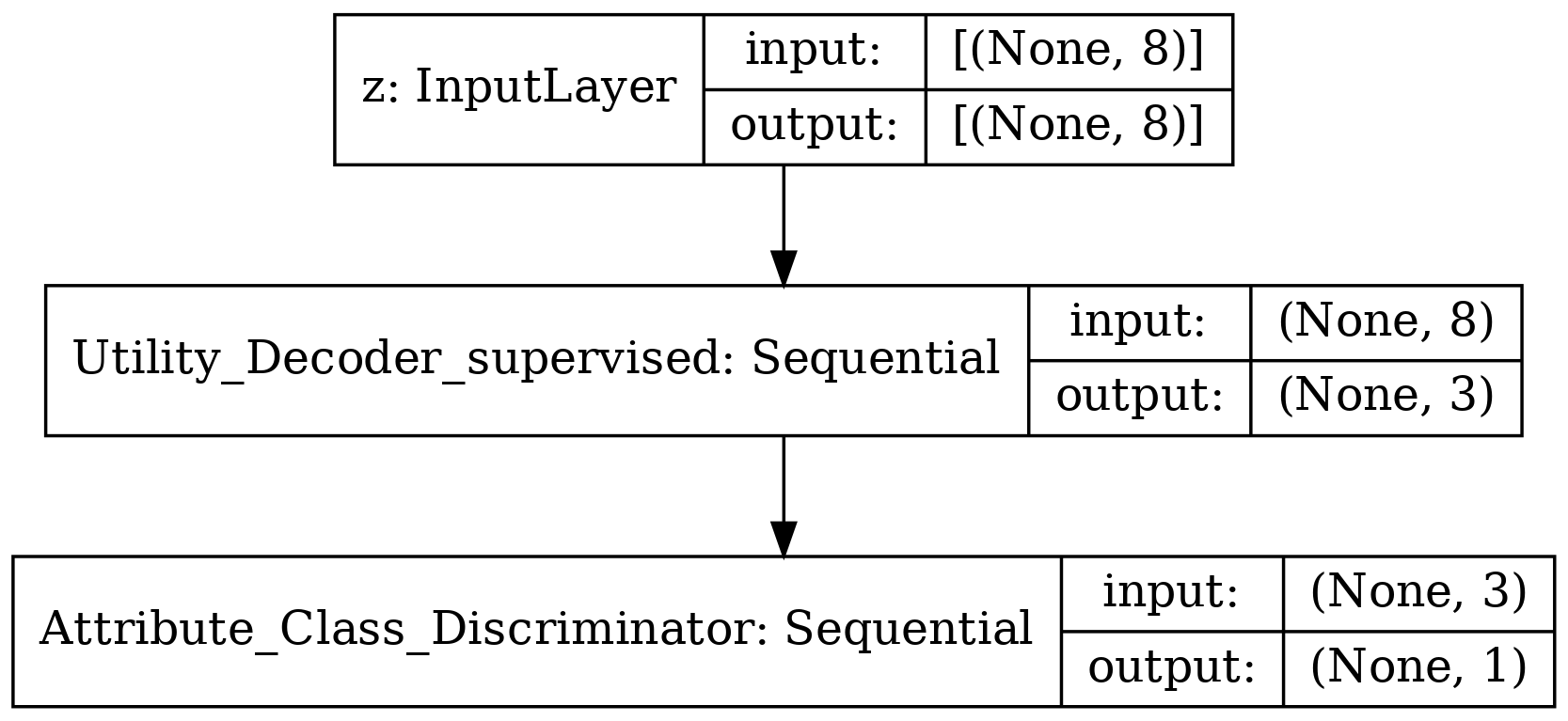}
        \caption{Utility-Decoder adjoint Attribute-Class-Discriminator module}
        \label{fig:P1_b_mnist}
    \end{subfigure}
    \hfill
    \begin{subfigure}{0.31\textwidth}
        \centering
        \includegraphics[width=\textwidth]{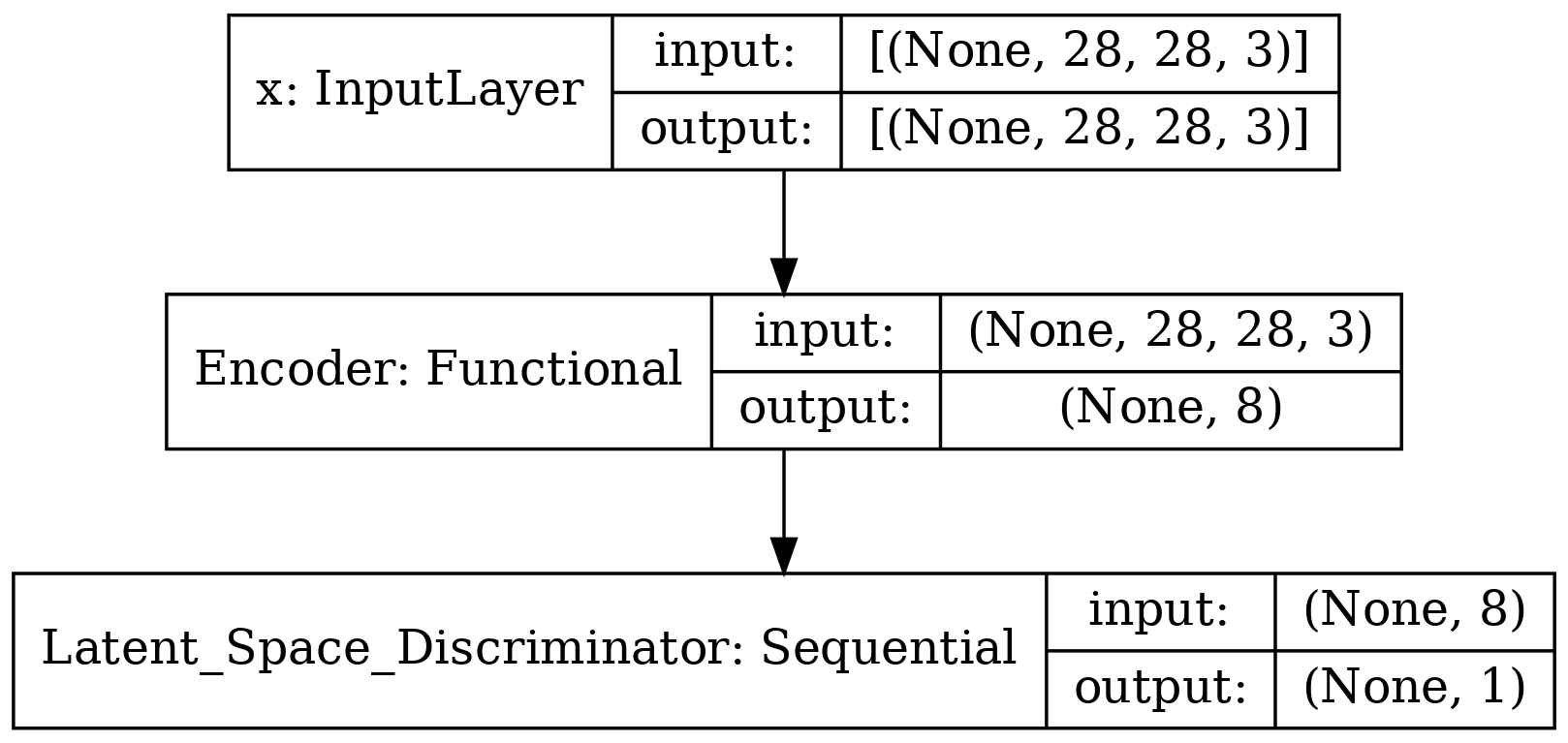}
        \caption{Encoder adjoint Latent-Space-Discriminator module}
        \label{fig:P1_c_mnist}
    \end{subfigure}
    \vfill
    \vspace{0.2cm}
    \begin{subfigure}{0.85\textwidth}
        \centering
        \includegraphics[width=\textwidth]{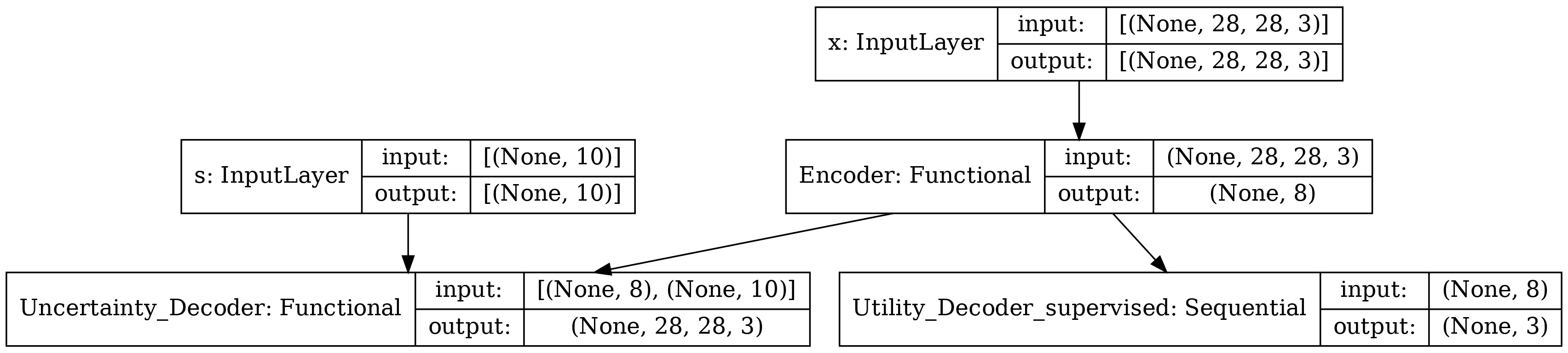}
        \caption{The auto-encoder module}
        \label{fig:P1_d_mnist}
    \end{subfigure}
    \caption{All model modules for supervised CLUB algorithm associated with ($\text{P1}\!\!:\! \mathsf{S}$) on Colored-MNIST dataset, setting $\mathbf{U}$: Digit Color, $\mathbf{S}$: Digit Class. The encoder output is set to 8 neurons.}
    \label{Fig:Full_Model_mnist_s_class_u_color_Isotropic_biased_supervised_P1}
\end{figure}

\begin{figure}[!h]
    \centering
    \begin{subfigure}{0.31\textwidth}
        \centering
        \includegraphics[width=\textwidth]{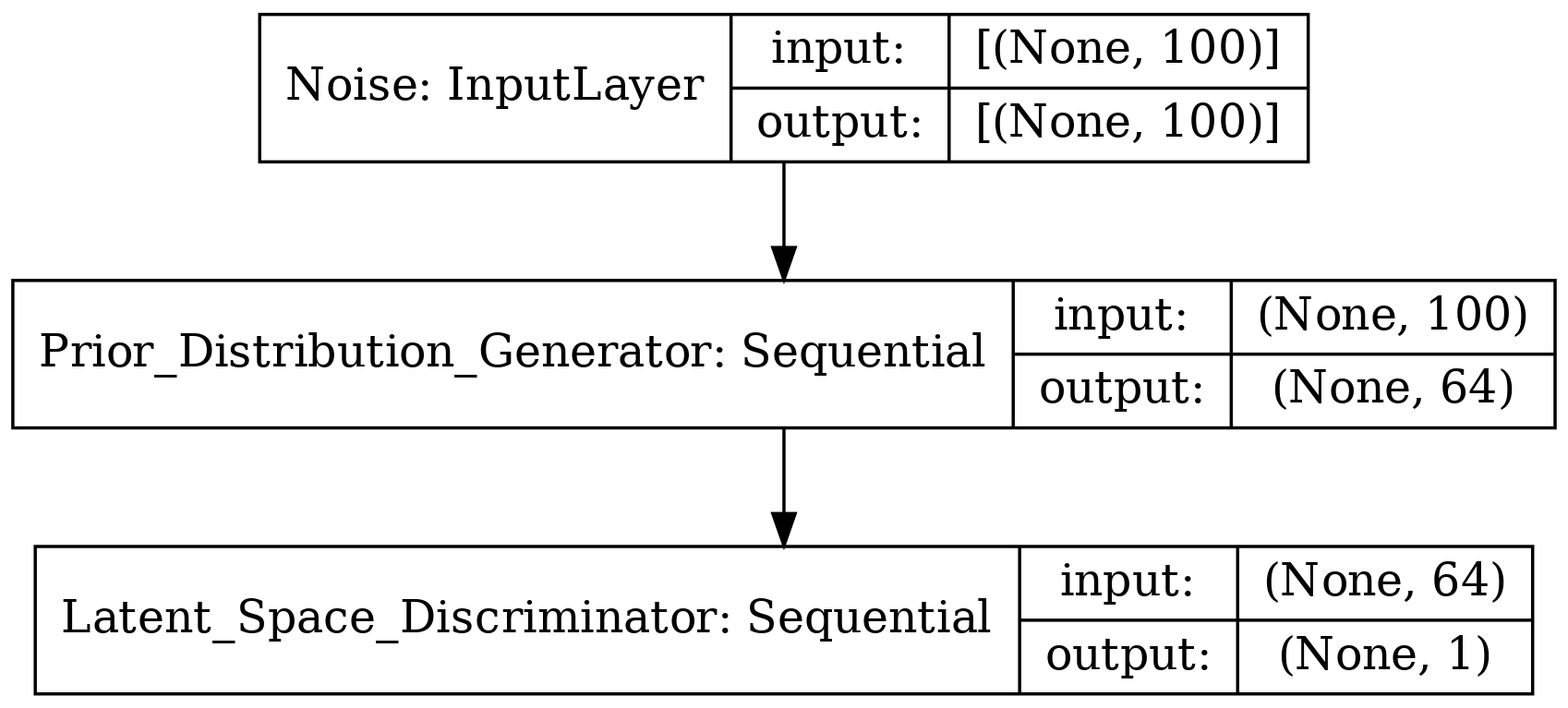}
        \caption{Prior-Generator adjoint Latent-Space-Discriminator module}
        \label{fig:P3_a_mnist}
    \end{subfigure}
    \hfill
    \begin{subfigure}{0.31\textwidth}
        \centering
        \includegraphics[width=\textwidth]{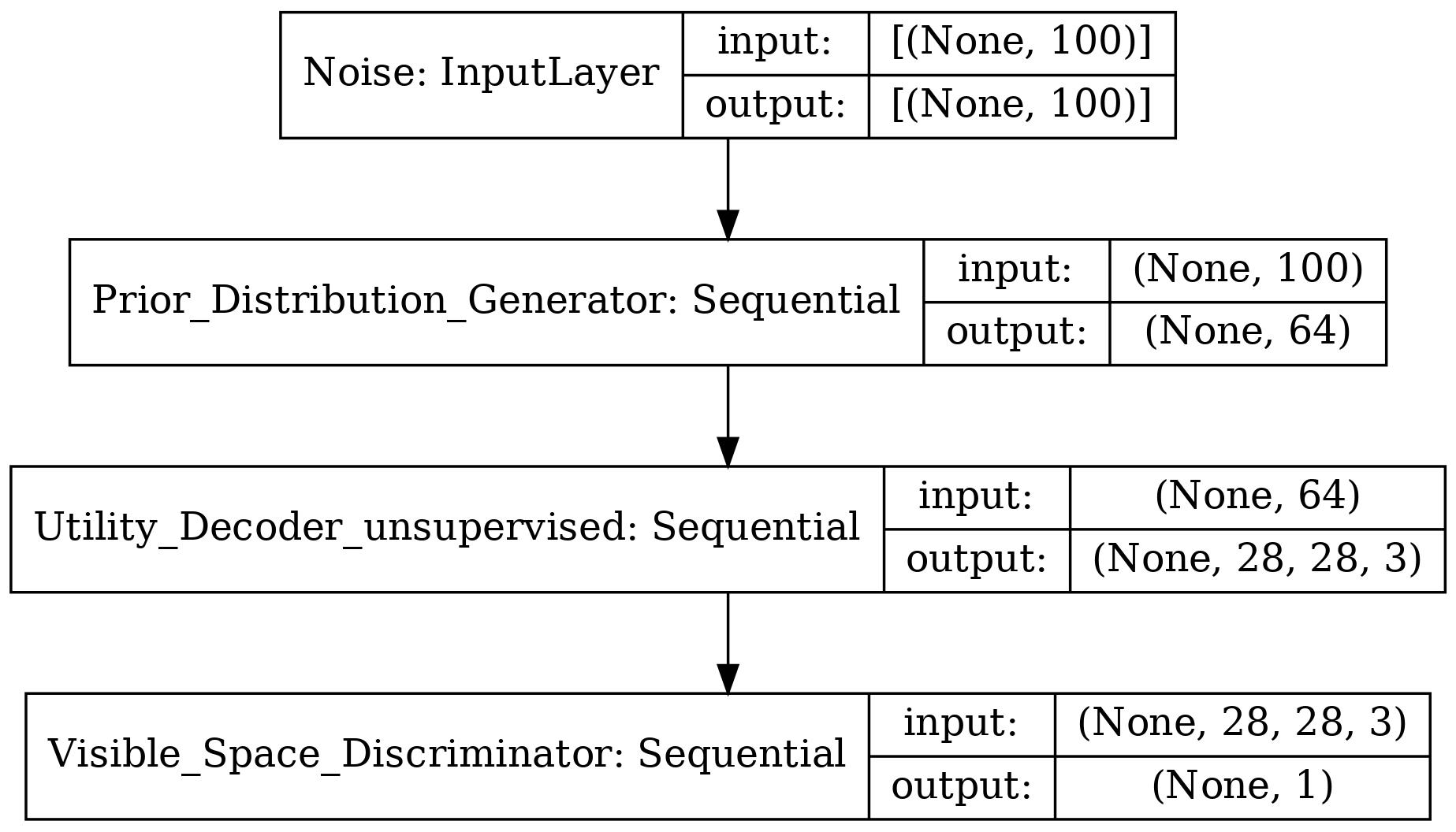}
        \caption{Utility-Decoder adjoint Attribute-Class-Discriminator module}
        \label{fig:P3_b_mnist}
    \end{subfigure}
    \hfill
    \begin{subfigure}{0.31\textwidth}
        \centering
        \includegraphics[width=\textwidth]{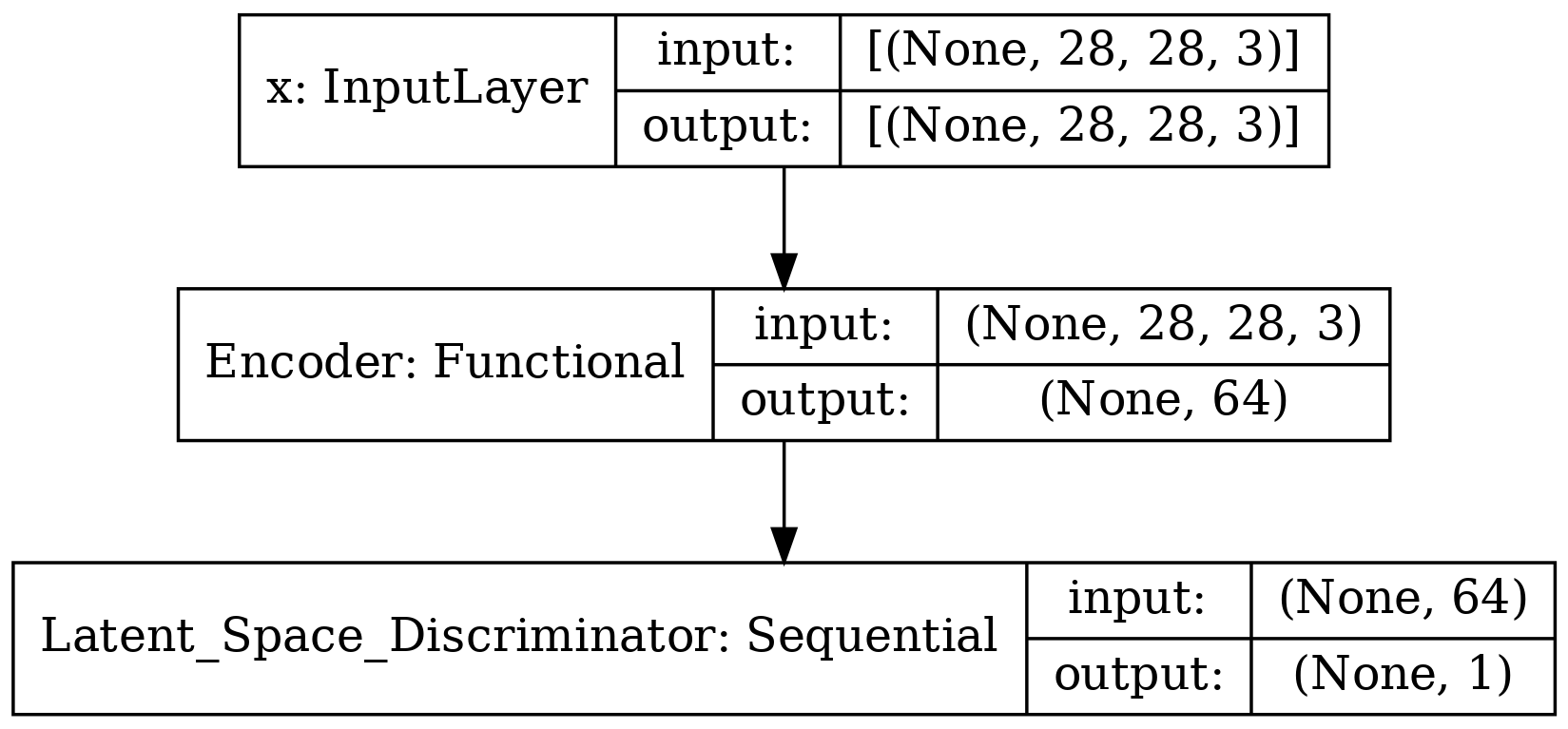}
        \caption{Encoder adjoint Latent-Space-Discriminator module}
        \label{fig:P3_c_mnist}
    \end{subfigure}
    \vfill
    \vspace{0.2cm}
    \begin{subfigure}{0.85\textwidth}
        \centering
        \includegraphics[width=\textwidth]{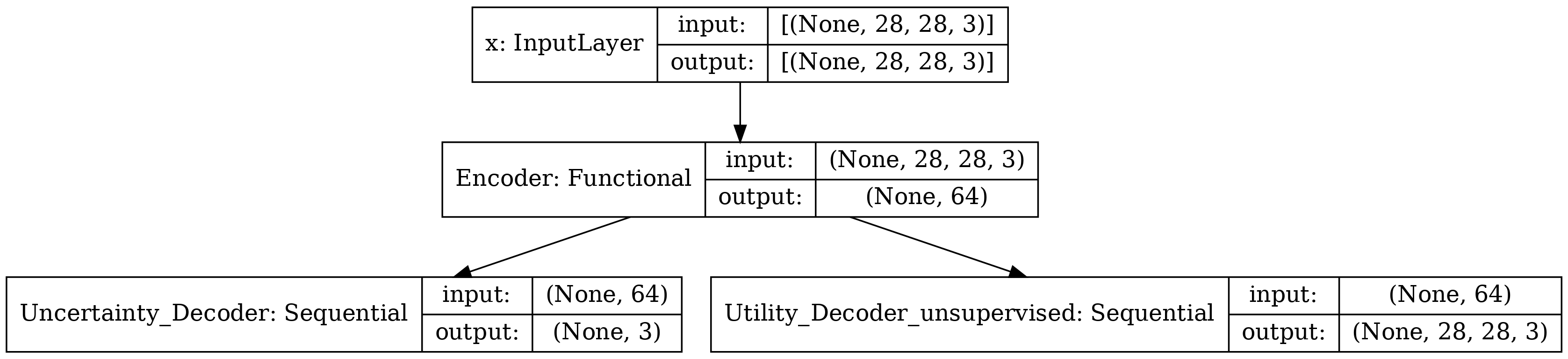}
        \caption{The auto-encoder module}
        \label{fig:P3_d_mnist}
    \end{subfigure}
    \caption{All model modules for unsupervised CLUB algorithm associated with ($\text{P3}\!\!:\! \mathsf{U}$) on Colored-MNIST dataset, setting $\mathbf{U}$: Digit Class, $\mathbf{S}$: Digit Color. The encoder output is set to 64 neurons.}
    \label{Fig:Full_Model_mnist_s_class_u_color_Isotropic_biased_unsupervised_P3}
\end{figure}

\begin{figure}[!h]
    \centering
    \begin{subfigure}{0.31\textwidth}
        \centering
        \includegraphics[width=\textwidth]{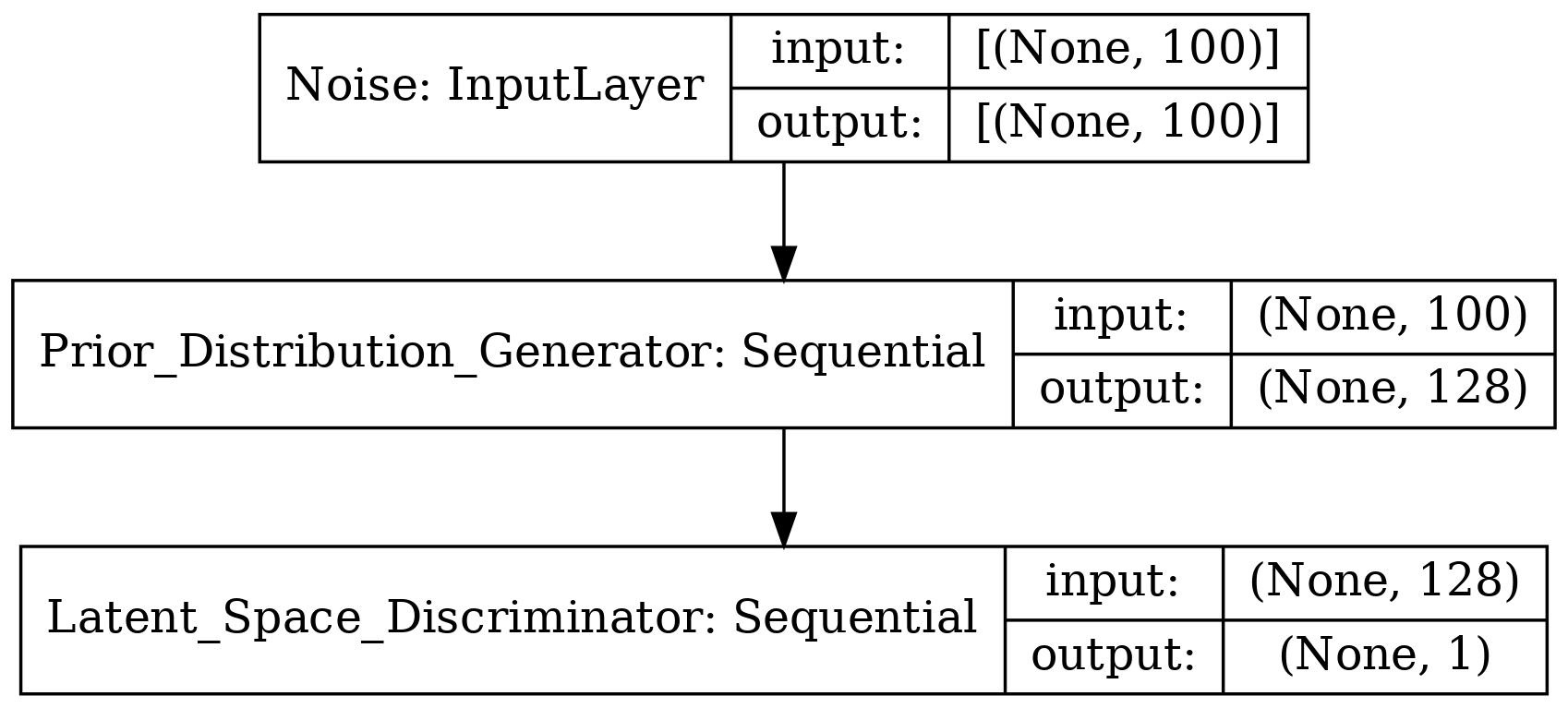}
        \caption{Prior-Generator adjoint Latent-Space-Discriminator module}
        \label{fig:P1_a_celeba}
    \end{subfigure}
    \hfill
    \begin{subfigure}{0.31\textwidth}
        \centering
        \includegraphics[width=\textwidth]{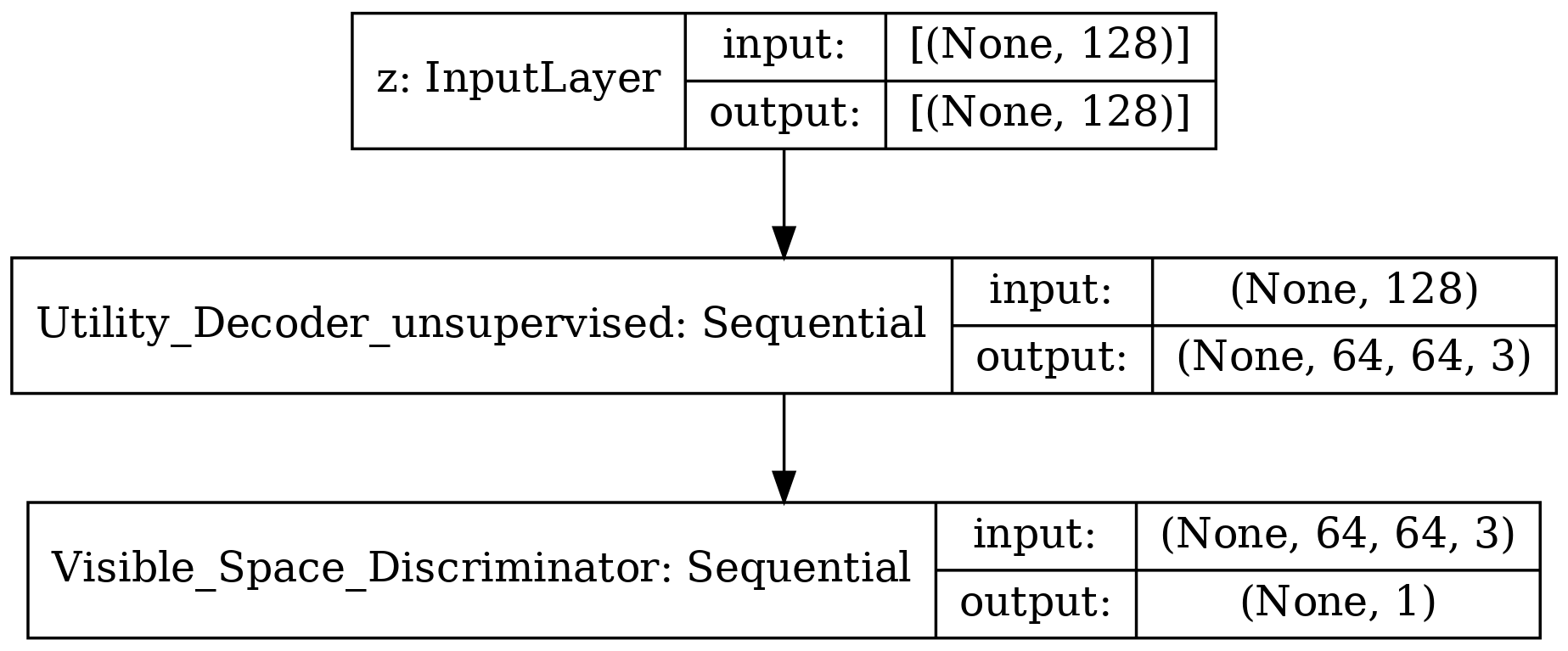}
        \caption{Utility-Decoder adjoint Attribute-Class-Discriminator module}
        \label{fig:P1_b_celeba}
    \end{subfigure}
    \hfill
    \begin{subfigure}{0.31\textwidth}
        \centering
        \includegraphics[width=\textwidth]{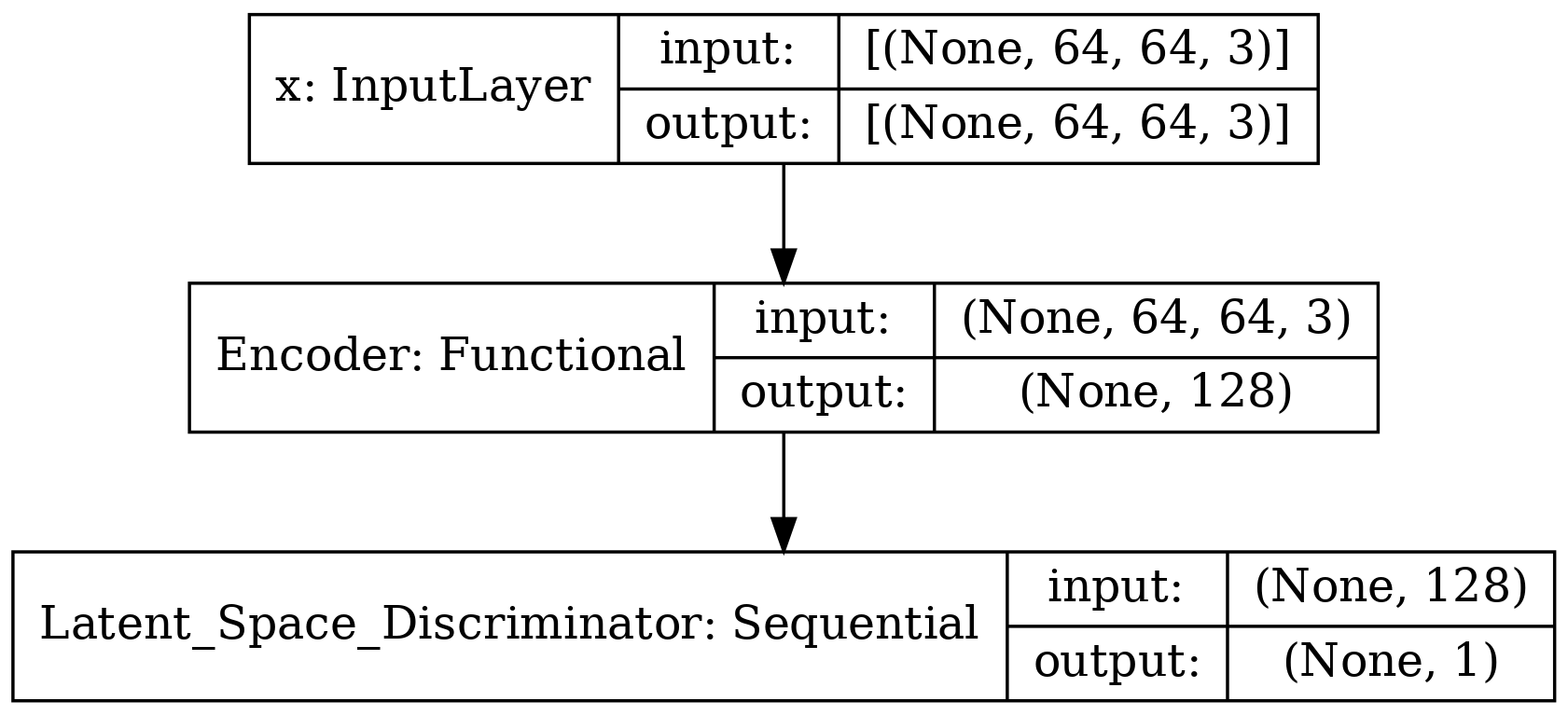}
        \caption{Encoder adjoint Latent-Space-Discriminator module}
        \label{fig:P1_c_celeba}
    \end{subfigure}
    \vfill
    \vspace{0.2cm}
    \begin{subfigure}{0.85\textwidth}
        \centering
        \includegraphics[width=\textwidth]{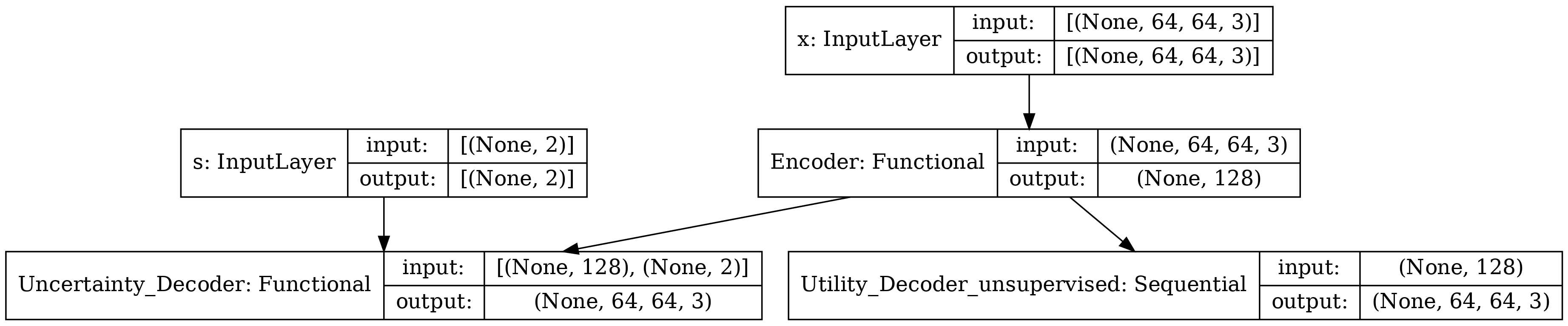}
        \caption{The auto-encoder module}
        \label{fig:P1_d_celeba}
    \end{subfigure}
    \caption{All model modules for unsupervised CLUB algorithm associated with ($\text{P1}\!\!:\! \mathsf{U}$) on CelebA dataset, setting $\mathbf{U}$: Gender, $\mathbf{S}$: Emotion. The encoder output is set to 128 neurons.}
    \label{Fig:Full_Model_celeba_s_class_u_color_Isotropic_unsupervised_P1}
\end{figure}

\begin{figure}[!h]
    \centering
    \begin{subfigure}{0.31\textwidth}
        \centering
        \includegraphics[width=\textwidth]{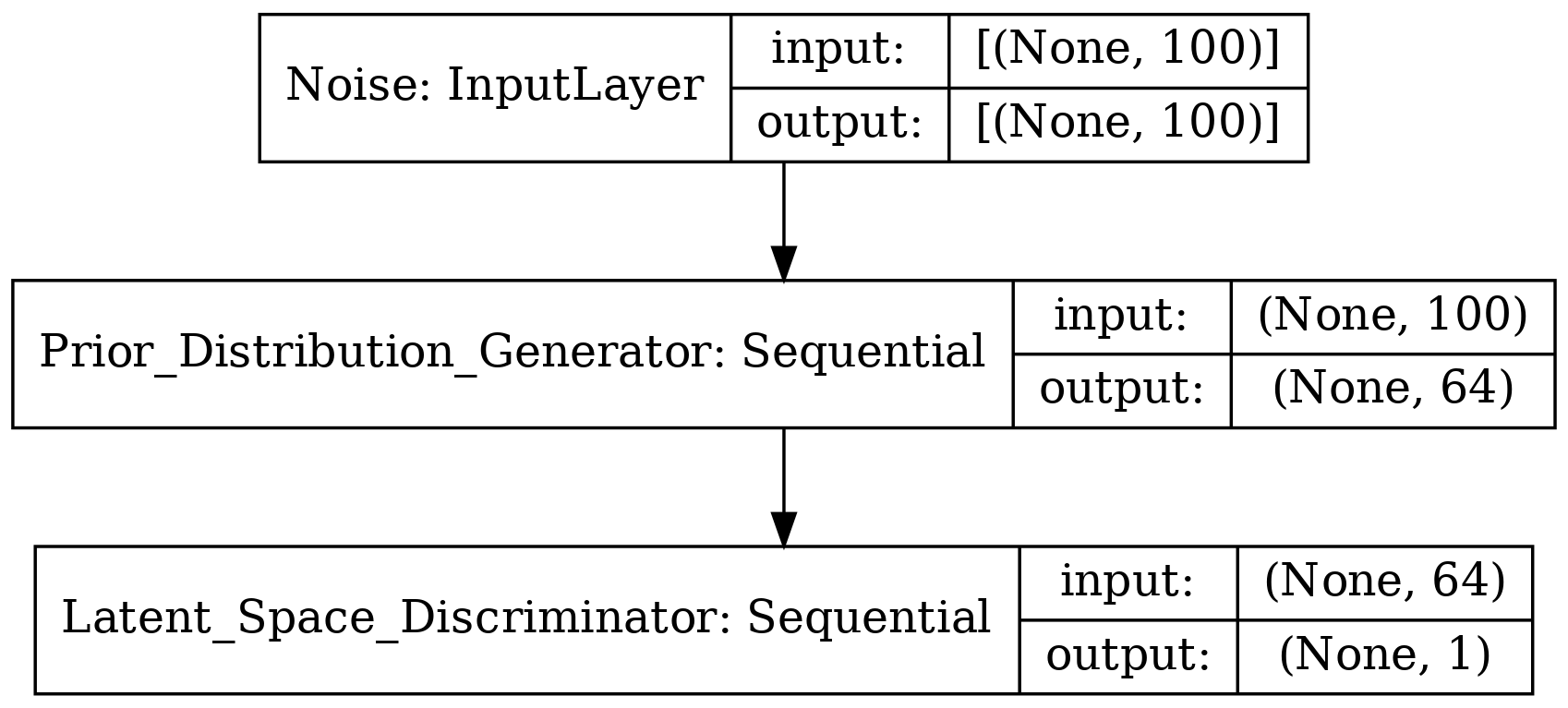}
        \caption{Prior-Generator adjoint Latent-Space-Discriminator module}
        \label{fig:P3_a_celeba}
    \end{subfigure}
    \hfill
    \begin{subfigure}{0.31\textwidth}
        \centering
        \includegraphics[width=\textwidth]{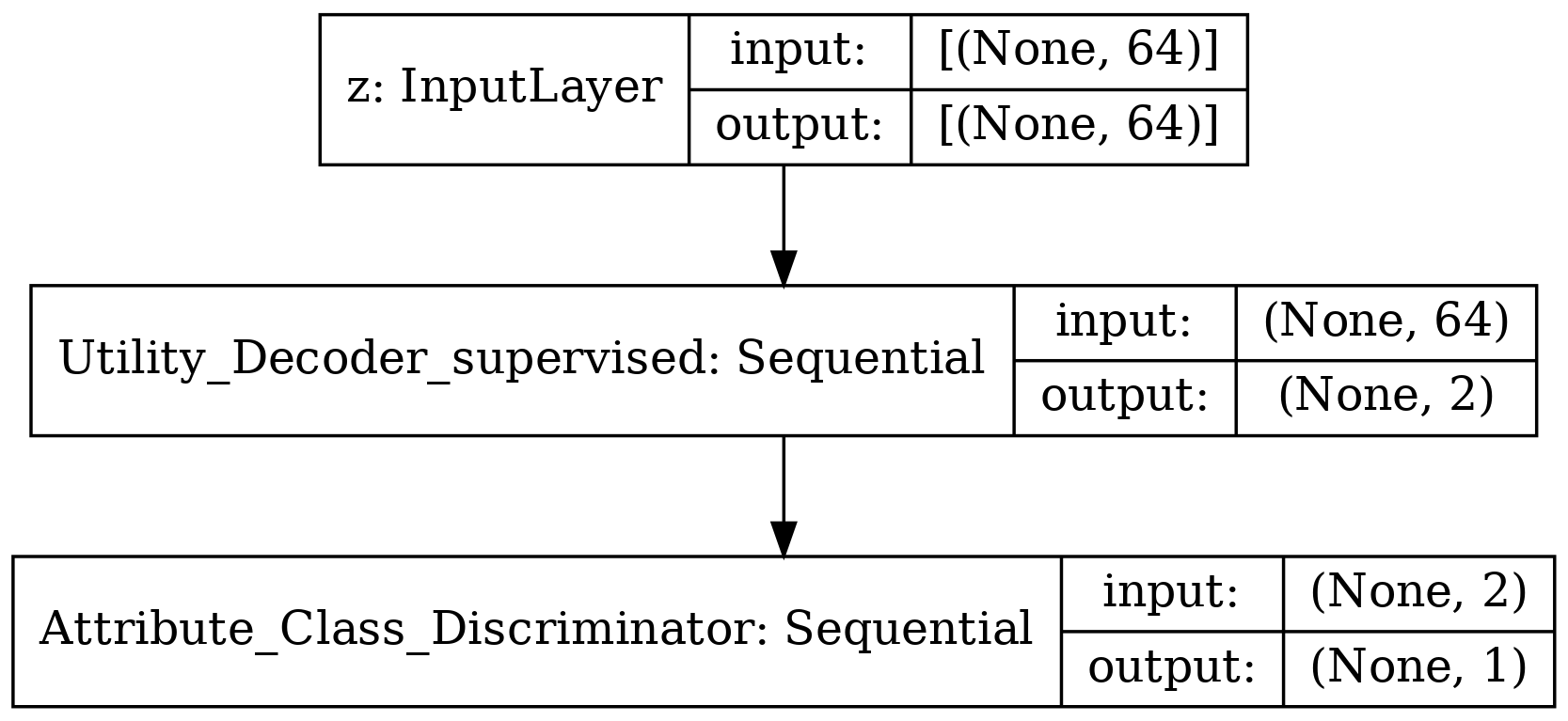}
        \caption{Utility-Decoder adjoint Attribute-Class-Discriminator module}
        \label{fig:P3_b_celeba}
    \end{subfigure}
    \hfill
    \begin{subfigure}{0.31\textwidth}
        \centering
        \includegraphics[width=\textwidth]{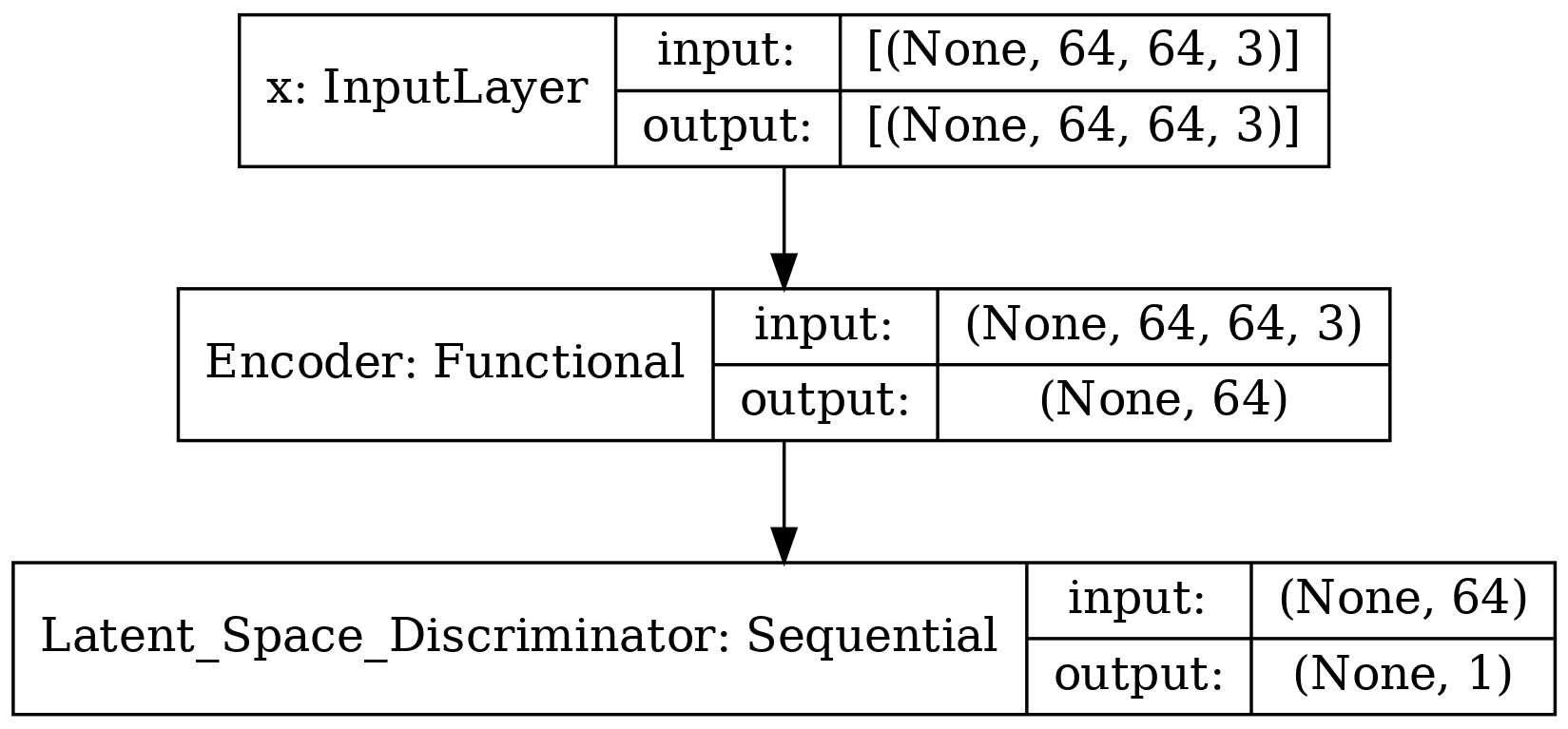}
        \caption{Encoder adjoint Latent-Space-Discriminator module}
        \label{fig:P3_c_celeba}
    \end{subfigure}
    \vfill
    \vspace{0.2cm}
    \begin{subfigure}{0.85\textwidth}
        \centering
        \includegraphics[width=\textwidth]{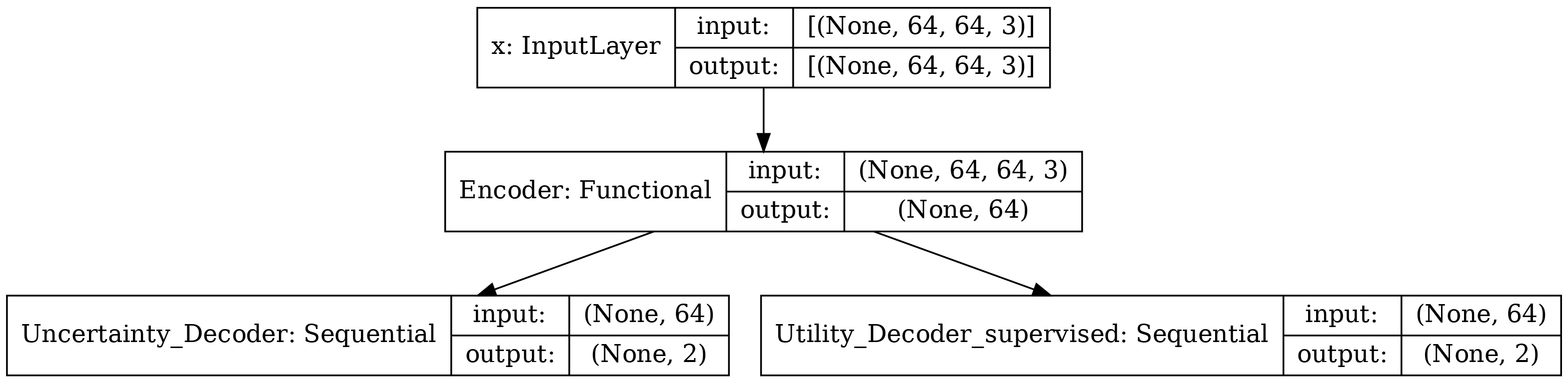}
        \caption{The auto-encoder module}
        \label{fig:P3_d_celeba}
    \end{subfigure}
    \caption{All model modules for supervised CLUB algorithm associated with ($\text{P3}\!\!:\! \mathsf{S}$) on CelebA dataset, setting $\mathbf{U}$: Emotion, $\mathbf{S}$: Gender. The encoder output is set to 64 neurons.}
    \label{Fig:Full_Model_celeba_s_class_u_color_Isotropic_supervised_P3}
\end{figure}

Considering Algorithm~\ref{Algorithm:P1_Supervised_DVCLUB} and Algorithm~\ref{Algorithm:P1_Unsupervised_DVCLUB}, the main loop consists of `five' blocks where only some networks are used in the \textit{forward} phase, and mostly one of them would update their parameters via \textit{back propagation} in each block. Therefore, we shattered the model into four sub-modules in the training stage for simplicity and performance. 
For the sake of visualization, we demonstrate a collection of sub-networks of the CLUB models\footnote{Note that due to technical reasons to efficiently implement the training algorithms, our CLUB models are not defined and saved in this form.} in Fig.~\ref{Fig:Full_Model_mnist_s_class_u_color_Isotropic_biased_supervised_P1},  Fig.~\ref{Fig:Full_Model_mnist_s_class_u_color_Isotropic_biased_unsupervised_P3}, Fig.~\ref{Fig:Full_Model_celeba_s_class_u_color_Isotropic_unsupervised_P1}, and Fig.~\ref{Fig:Full_Model_celeba_s_class_u_color_Isotropic_supervised_P3}.

%Fig.~\ref{} shows the corresponding sub-modules of Fig.~\ref{}, which are used in our implementation. 
During training, all of the sub-module (d) parameters, call the "auto-encoder part" would update with BP after each forward step. 
For the sub-modules (a), (b), and (c), only the parameters of one network are updated when the corresponding error function values back-propagate, and we freeze the other network parameters in the sub-module. 
For example, block (3) of Algorithm~\ref{Algorithm:P1_Supervised_DVCLUB} is related to the (c) sub-module, but at the BP step, the latent space discriminator is frozen to prevent its parameters from updating. 
%This procedure is vice versa for module (b) at block (2).

It should be mentioned that during our experiments, we found out that before running our main algorithm, it is beneficial to pre-train the auto-encoder sub-module since we need to sample from the latent space, which uses in other parts of the main model during training. We justify this by mentioning that sampling meaningful data rather than random ones from latent variables from the beginning of learning helps the model to converge better and faster in comparison with starting one of our DVCLUB algorithms with a randomly initiated auto-encoder.

%---------------------------------------------------
%  Figure: 
%---------------------------------------------------
%
%---------------------------------------------------
%---------------------------------------------------
%---------------------------------------------------
%---------------------------------------------------
%---------------------------------------------------
%---------------------------------------------------

\clearpage

\section{Supplementary Results}
\label{Appendix:SupplementaryResults}

\subsection{Supplementary Results of Colored-MNIST Experiments}

%---------------------------------------------------
% Tail (Appendix)
%---------------------------------------------------
%  Figure: CelebA >>>>> Acc on S  
%  P3 UnSupervised
% S: Emotion
%---------------------------------------------------
%
\begin{figure}[!h]
\centering
\includegraphics[width=\textwidth]{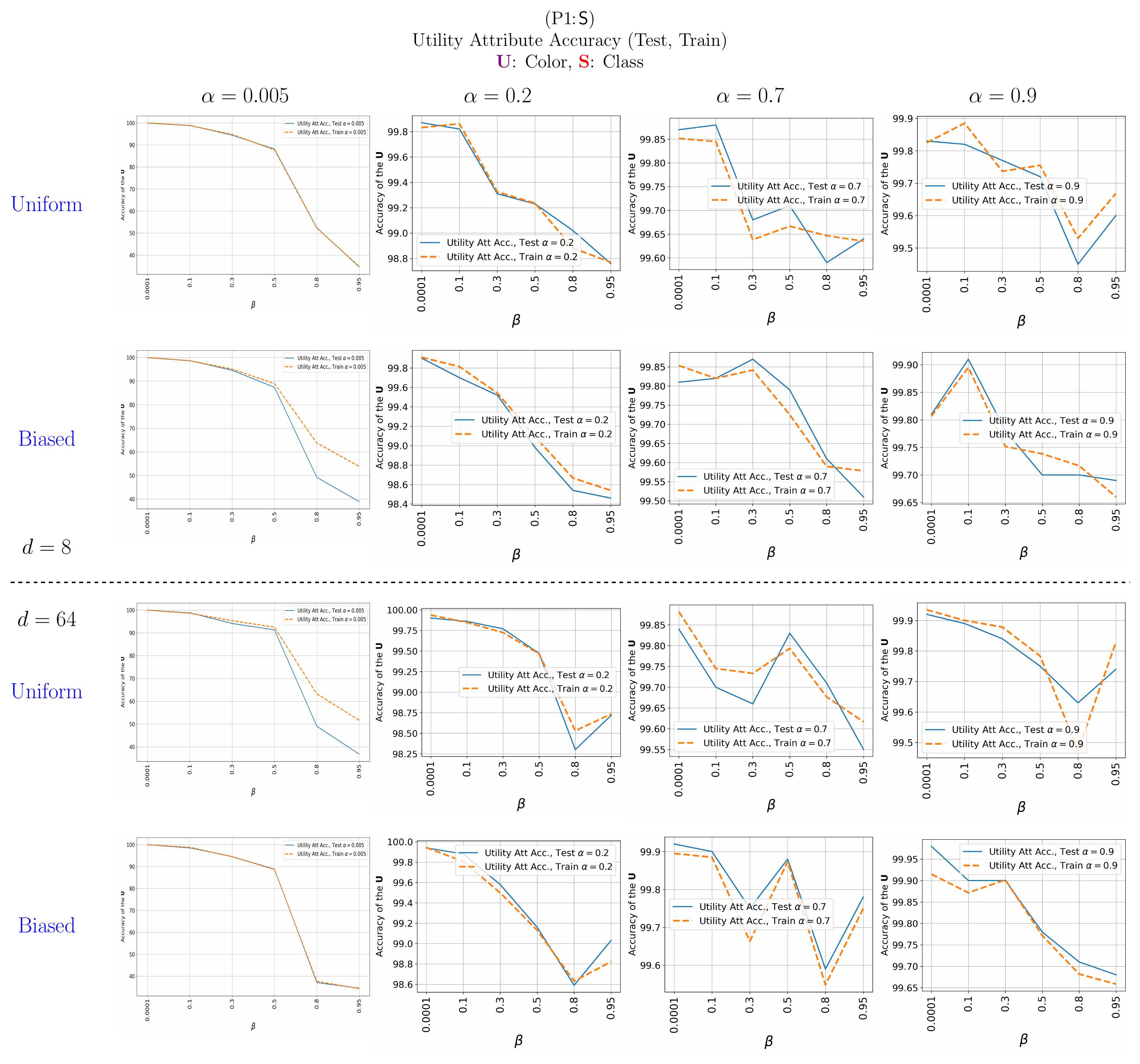}
%% Accuracy on S
%% P3 UnSupervised
%% S: Emotion
%% d: 64 & 128
\caption{Recognition accuracy of the utility attribute $\mathbf{U}$ for supervised CLUB model ($\text{P1}\!\!:\! \mathsf{S}$) on `Test' and `Train' datasets of Colored-MNIST, considering $d_{\mathbf{z}}=8$ (Up Panel) and $d_{\mathbf{z}}=64$ (Down Panel), setting $Q_{\boldsymbol{\psi}}\! \left( \mathbf{Z} \right) = \mathcal{N} \! \left( \boldsymbol{0}, \mathbf{I}_{d_{\mathbf{z}}} \right)$, for different information complexity weights $\beta$ and information leakage weights $\alpha$.}
\label{Fig:AccU_ColMNIST_P1supervised_SclassUcolor}
\end{figure}

\clearpage

\subsection{Supplementary Results of CelebA Experiments}

%---------------------------------------------------
% Tail (Appendix)
%---------------------------------------------------
%  Figure: CelebA >>>>> Acc on S  
%  P3 UnSupervised
% S: Emotion
%---------------------------------------------------
%
\begin{figure}[!h]
\centering
\includegraphics[width=\textwidth]{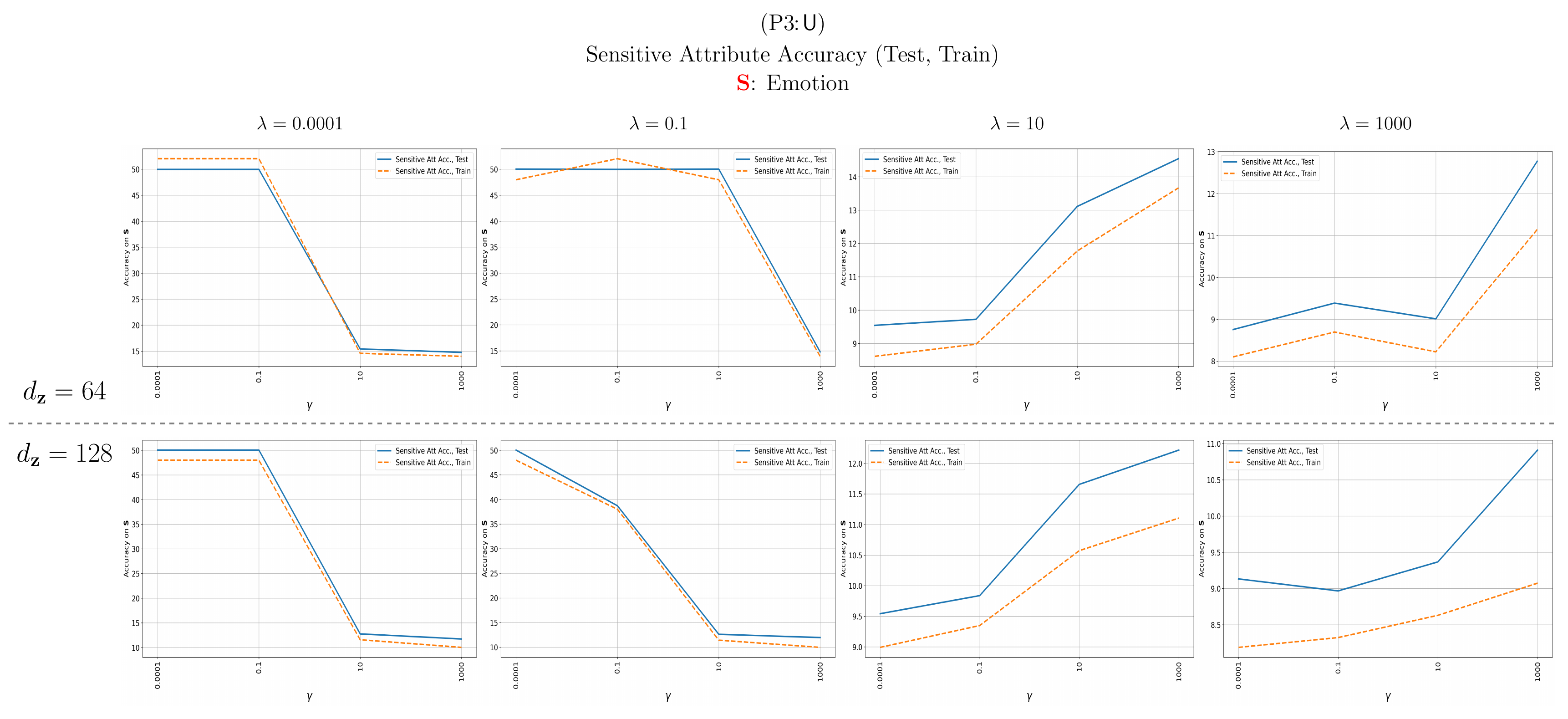}
%% Accuracy on S
%% P3 UnSupervised
%% S: Emotion
%% d: 64 & 128
\caption{Recognition accuracy of the sensitive attribute $\mathbf{S}$ for unsupervised CLUB model ($\text{P3}\!\!:\! \mathsf{U}$) on `Test' and `Train' datasets of CelebA, considering $d_{\mathbf{z}}=64$ (First Row) and $d_{\mathbf{z}}=128$ (Second Row), setting $\mathbf{S}$: Emotion (smiling), $Q_{\boldsymbol{\psi}}\! \left( \mathbf{Z} \right) = \mathcal{N} \! \left( \boldsymbol{0}, \mathbf{I}_{d_{\mathbf{z}}} \right)$, for different information utility weights $\gamma$ and information leakage weights $\lambda$.}
\label{Fig:AccS_CelebA_P3unsupervised_Semotion}
\end{figure}
%
%---------------------------------------------------
%---------------------------------------------------

%---------------------------------------------------
%  Tail (Appendix)
%---------------------------------------------------
%  Figure: CelebA >>>>> Acc on S  
%  P3 UnSupervised
%  S: Gender
%---------------------------------------------------
%
\begin{figure}[!hh]
\centering
\includegraphics[width=\textwidth]{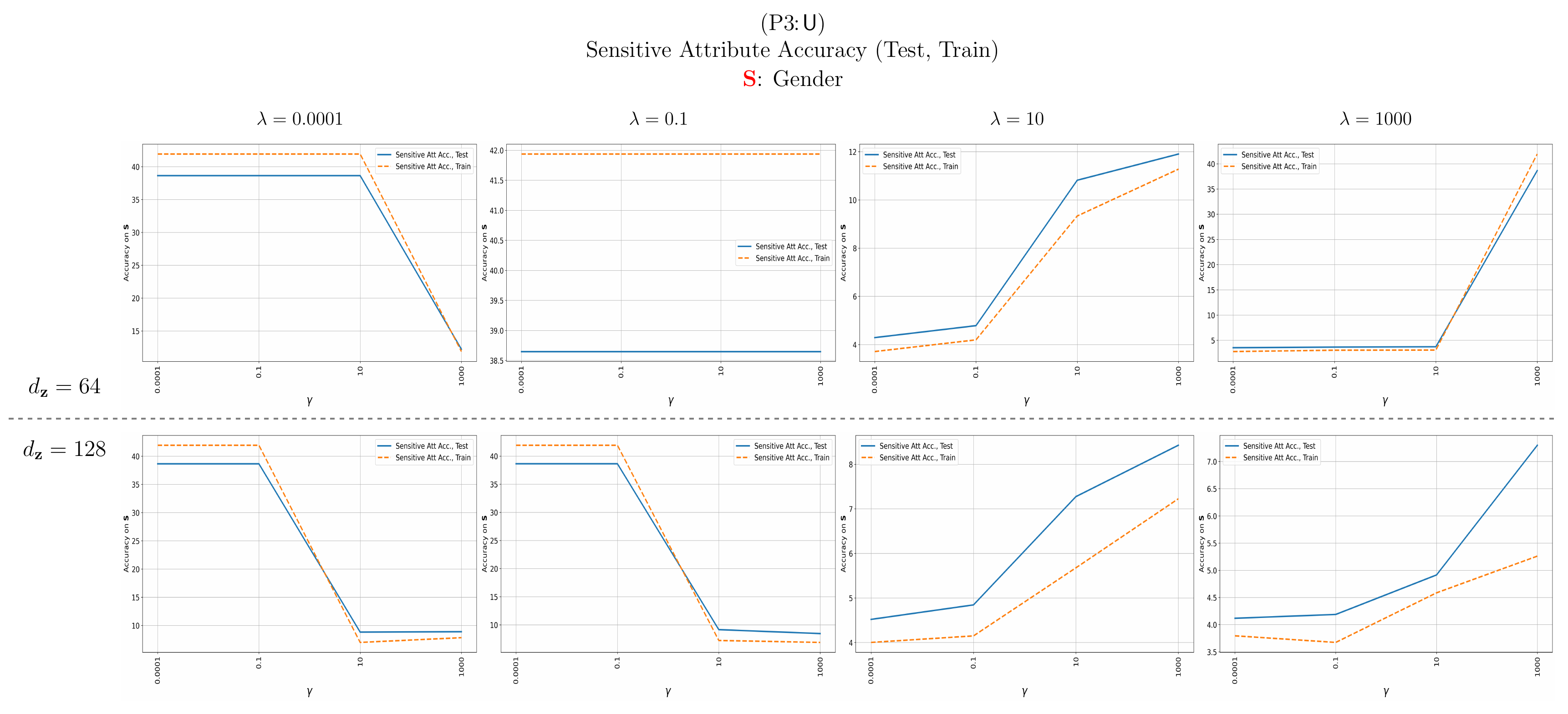}
%% Accuracy on S
%% P3 UnSupervised
%% S: Gender
%% d: 64 & 128
\caption{Recognition accuracy of the sensitive attribute $\mathbf{S}$ for unsupervised CLUB model ($\text{P3}\!\!:\! \mathsf{U}$) on `Test' and `Train' datasets of CelebA, considering $d_{\mathbf{z}}=64$ (First Row) and $d_{\mathbf{z}}=128$ (Second Row), setting $\mathbf{S}$: Gender, $Q_{\boldsymbol{\psi}}\! \left( \mathbf{Z} \right) = \mathcal{N} \! \left( \boldsymbol{0}, \mathbf{I}_{d_{\mathbf{z}}} \right)$, for different information utility weights $\gamma$ and information leakage weights $\lambda$.}
\label{Fig:AccS_CelebA_P3unsupervised_Sgender}
\end{figure}
%
%---------------------------------------------------
%---------------------------------------------------

%---------------------------------------------------
% Tail (Appendix)
%---------------------------------------------------
%  Figure: CelebA >>>>> Acc on U and S
%  P3 Supervised
%  S: Emotion
%  U: Gender
%---------------------------------------------------
%
\begin{figure}[!t]
\centering
\includegraphics[width=\textwidth]{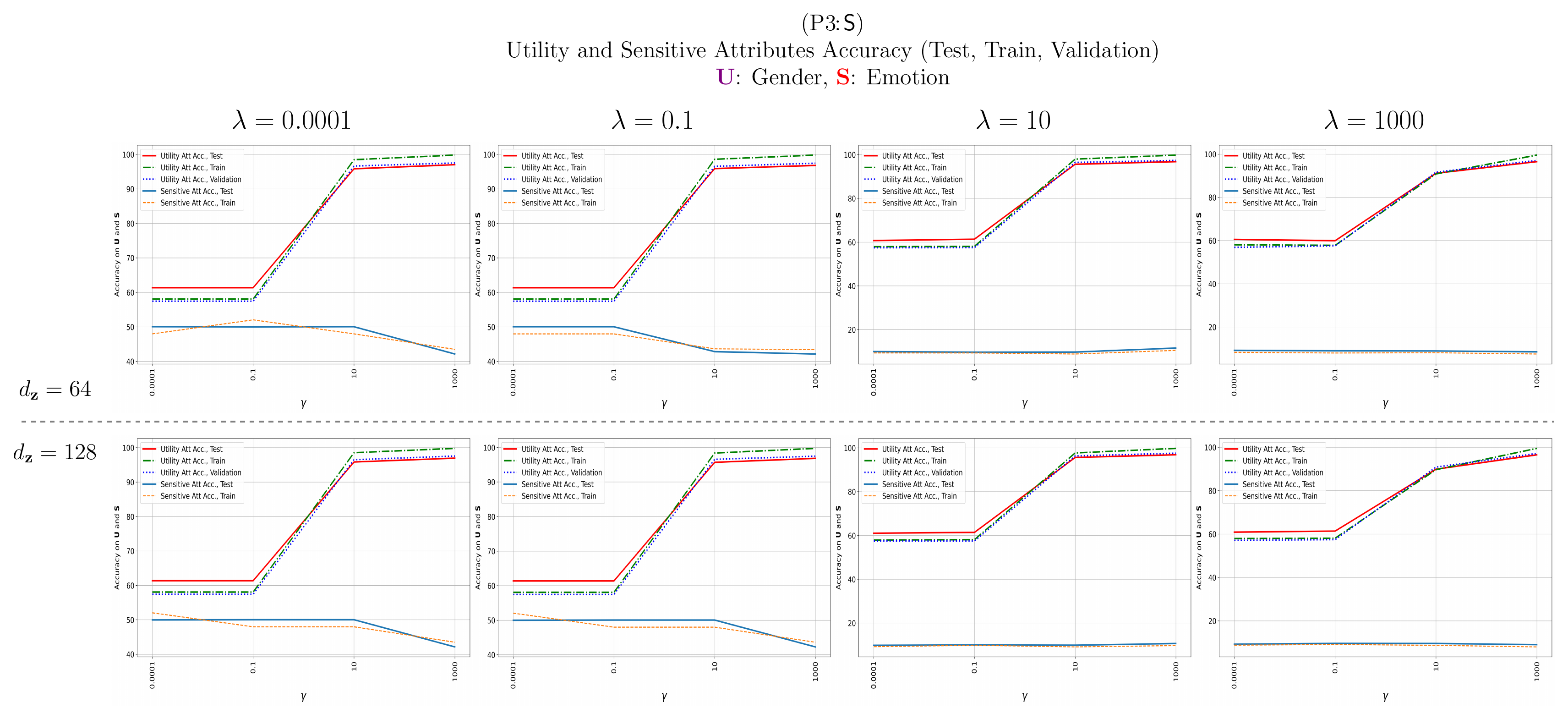}
%% Accuracy on U and S
%% P3 Supervised
%% S: Emotion
%% U: Gender
%% d: 64 & 128
\caption{Recognition accuracy of the utility attribute $\mathbf{U}$ and sensitive attribute $\mathbf{S}$ for supervised CLUB model ($\text{P3}\!\!:\! \mathsf{S}$) on `Test', `Train' and `Validation' datasets of CelebA, considering $d_{\mathbf{z}}=64$ (First Row) and $d_{\mathbf{z}}=128$ (Second Row), setting $\mathbf{U}$: Gender, $\mathbf{S}$: Emotion (smiling), $Q_{\boldsymbol{\psi}}\! \left( \mathbf{Z} \right) = \mathcal{N} \! \left( \boldsymbol{0}, \mathbf{I}_{d_{\mathbf{z}}} \right)$, for different information utility weights $\gamma$ and information leakage weights $\lambda$.}
\label{Fig:AccUandS_CelebA_P3supervised_SemotionUgender}
\end{figure}
%
%---------------------------------------------------
%---------------------------------------------------

%---------------------------------------------------
% Tail (Appendix)
%---------------------------------------------------
%  Figure: CelebA >>>>> Acc on U and S
%  P3 Supervised
%  U: Emotion
%  S: Gender
%---------------------------------------------------
%
\begin{figure}[!t]
\centering
\includegraphics[width=\textwidth]{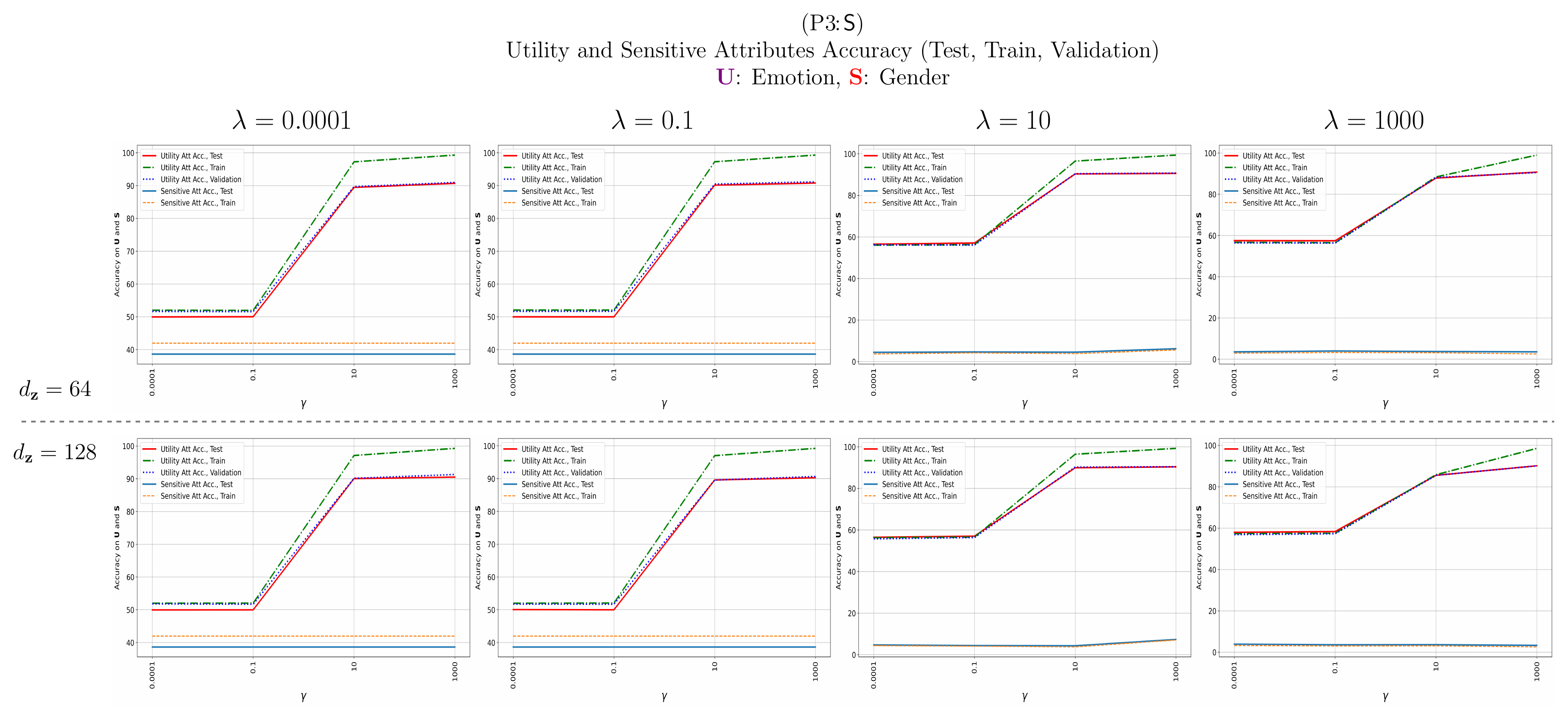}
%% Accuracy on U and S
%% P3 Supervised
%% U: Emotion
%% S: Gender
%% d: 64 & 128
\caption{Recognition accuracy of the utility attribute $\mathbf{U}$ and sensitive attribute $\mathbf{S}$ for supervised CLUB model ($\text{P3}\!\!:\! \mathsf{S}$) on `Test', `Train' and `Validation' datasets of CelebA, considering $d_{\mathbf{z}}=64$ (First Row) and $d_{\mathbf{z}}=128$ (Second Row), setting $\mathbf{U}$: Emotion (smiling), $\mathbf{S}$: Gender, $Q_{\boldsymbol{\psi}}\! \left( \mathbf{Z} \right) = \mathcal{N} \! \left( \boldsymbol{0}, \mathbf{I}_{d_{\mathbf{z}}} \right)$, for different information utility weights $\gamma$ and information leakage weights $\lambda$.}
\label{Fig:AccUandS_CelebA_P3supervised_SgenderUemotion}
\end{figure}
%
%---------------------------------------------------
%---------------------------------------------------

%---------------------------------------------------
% Tail (Appendix)
%---------------------------------------------------
%  Figure: CelebA >>>>> Acc on U  
%  P1 Supervised
%  S: Emotion
%  U: Gender
%---------------------------------------------------
%
\begin{figure}[!t]
\centering
\includegraphics[width=\textwidth]{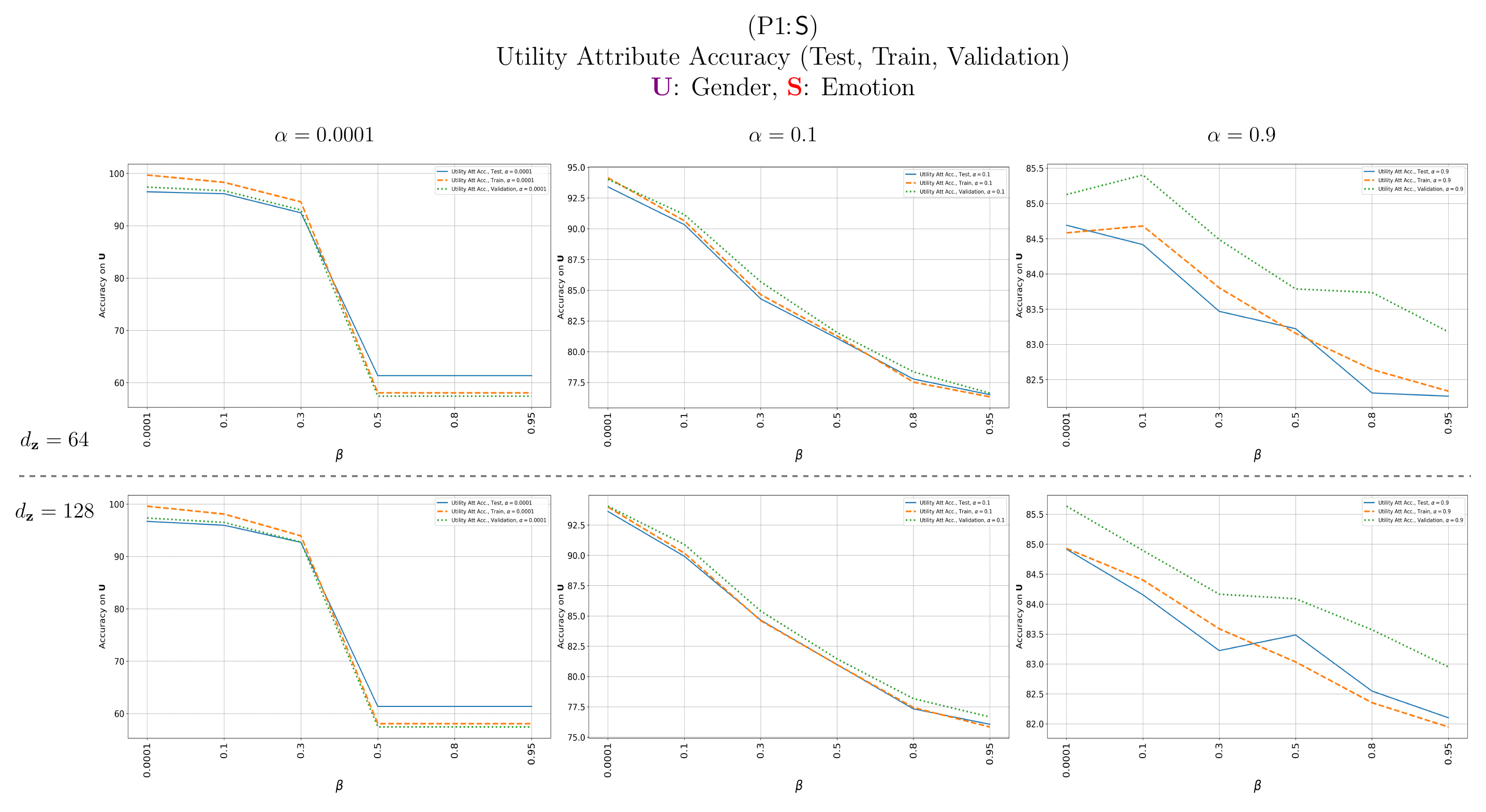}
%% Accuracy on U
%% P1 Supervised
%% U: Gender
%% S: Emotion
%% d: 64 & 128
\caption{Recognition accuracy of the utility attribute $\mathbf{U}$ for supervised CLUB model ($\text{P1}\!\!:\! \mathsf{S}$) on `Test', `Train' and `Validation' datasets of CelebA, considering $d_{\mathbf{z}}=64$ (First Row) and $d_{\mathbf{z}}=128$ (Second Row), setting $\mathbf{U}$: Gender, $\mathbf{S}$: Emotion (smiling), $Q_{\boldsymbol{\psi}}\! \left( \mathbf{Z} \right) = \mathcal{N} \! \left( \boldsymbol{0}, \mathbf{I}_{d_{\mathbf{z}}} \right)$, for different information complexity weights $\beta$ and information leakage weights $\alpha$.}
\label{Fig:AccU_CelebA_P1supervised_SemotionUgender}
\end{figure}
%
%---------------------------------------------------
%---------------------------------------------------

%---------------------------------------------------
% Tail (Appendix)
%---------------------------------------------------
%  Figure: CelebA >>>>> Acc on U  
%  P1 Supervised
%  S: Gender
%  U: Emotion
%---------------------------------------------------
%
\begin{figure}[!t]
\centering
\includegraphics[width=\textwidth]{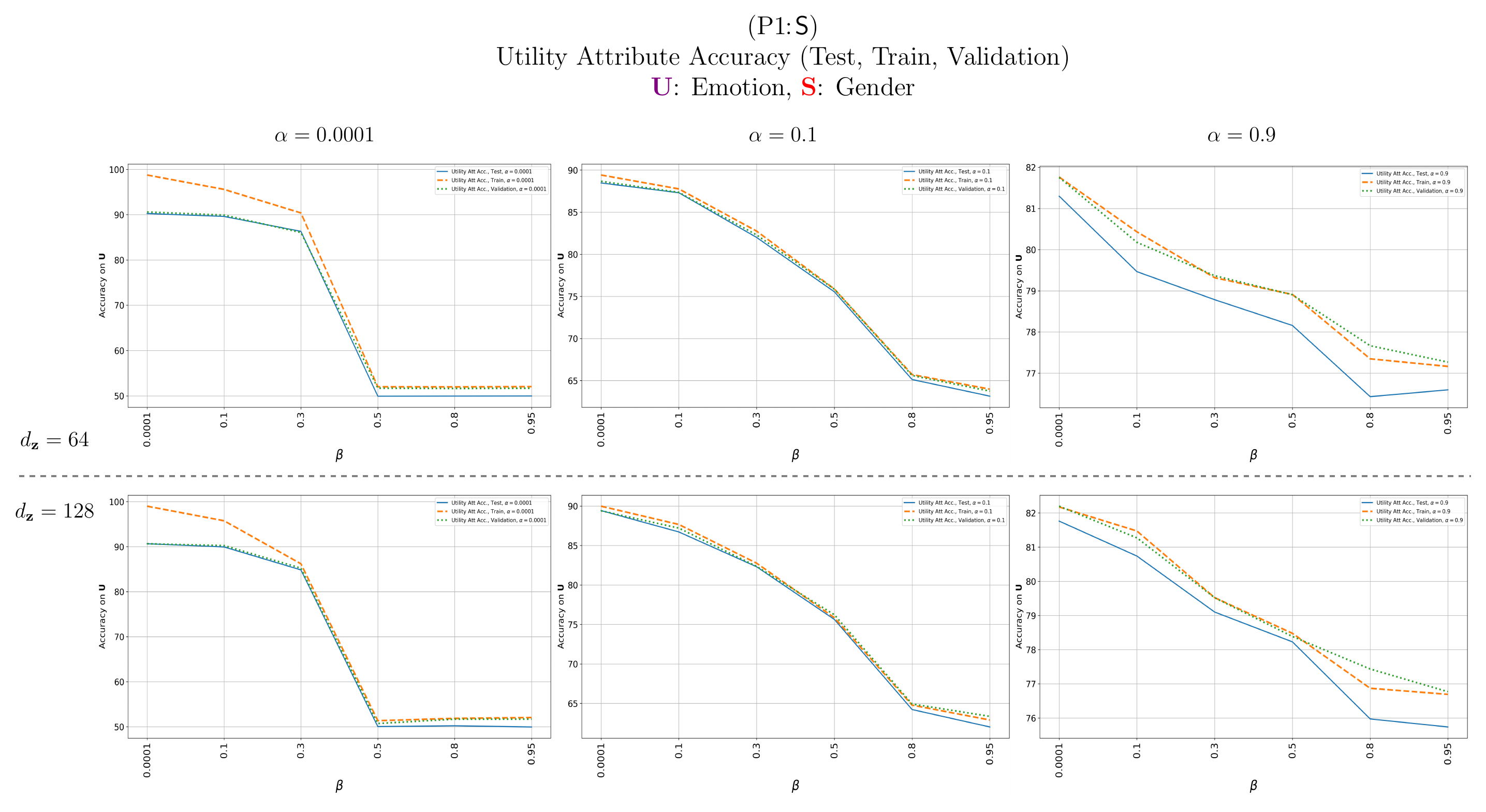}
%% Accuracy on U
%% P1 Supervised
%% U: Emotion
%% S: Gender
%% d: 64 & 128
\caption{Recognition accuracy of the utility attribute $\mathbf{U}$ for supervised CLUB model ($\text{P1}\!\!:\! \mathsf{S}$) on `Test', `Train' and `Validation' datasets of CelebA, considering $d_{\mathbf{z}}=64$ (First Row) and $d_{\mathbf{z}}=128$ (Second Row), setting $\mathbf{U}$: Emotion (smiling), $\mathbf{S}$: Gender, $Q_{\boldsymbol{\psi}}\! \left( \mathbf{Z} \right) = \mathcal{N} \! \left( \boldsymbol{0}, \mathbf{I}_{d_{\mathbf{z}}} \right)$, for different information complexity weights $\beta$ and information leakage weights $\alpha$.}
\label{Fig:AccU_CelebA_P1supervised_SgenderUemotion}
\end{figure}
%
%---------------------------------------------------
%---------------------------------------------------

\clearpage

\subsection{Training Algorithms}

%---------------------------------------------------
%
%       TABLE:  Supervised DVCLUB Training Algorithm
%
%---------------------------------------------------
%
\begin{center}
%\begin{minipage}{0.9\linewidth}
\centering
\begin{spacing}{1}
\begin{algorithm}
    \small
    \setstretch{1.2}
    \begin{algorithmic}[1]
        \State \textbf{Input:} Training Dataset: $\{ \left( \mathbf{u}_n, \mathbf{s}_n, \mathbf{x}_n  \right) \}_{n=1}^{N}$; Hyper-Parameters: $\gamma,\lambda$
        \State $\boldsymbol{\phi}, \boldsymbol{\theta}, \boldsymbol{\psi}, \boldsymbol{\xi}, \boldsymbol{\eta}, \boldsymbol{\omega}, \boldsymbol{\tau} \; \gets$ Initialize Network Parameters
        
        \Repeat
        
        \hspace{-15pt}(1) {\small\textbf{\textsf{Train the Encoder, Utility Decoder, Uncertainty Decoder}} $\left( \boldsymbol{\phi}, \boldsymbol{\theta}, \boldsymbol{\xi} \right)$}
         \State  ~~Sample a mini-batch $\{ \mathbf{u}_m, \mathbf{s}_m, \mathbf{x}_m \}_{m=1}^{M} \sim P_{\mathsf{D}} (\mathbf{X}) P_{\mathbf{U}, \mathbf{S} 
         \mid \mathbf{X}}$
         \State  ~~Compute $\mathbf{z}_m \sim f_{\boldsymbol{\phi}} (\mathbf{x}_m), \forall m \in [M]$
        %  \Comment{\textcolor{blue!60!gray}{$f_{\boldsymbol{\phi}}$: Stochastic or Deterministic Encoder}}
         %\State  ~~Sample $\{ \mathbf{\widehat{z}}_m \}_{m=1}^{M} \sim Q_{\boldsymbol{\psi}} (\mathbf{Z})$
         %\State  ~~Compute $\mathbf{\widehat{x}}_m =  g_{\boldsymbol{\xi}} (\mathbf{\widehat{z}}_m, \mathbf{s}_m), \forall m \in [M]$
         \State  ~~Back-propagate loss:\vspace{3pt}
         
                $ \mathcal{L} \! \left( \boldsymbol{\phi}, \! \boldsymbol{\theta}, \! \boldsymbol{\xi} \right) \! = \!\! - \frac{1}{M} \! \sum_{m=1}^{M} \!\! \big( \D_{\mathrm{KL}} \! \left( P_{\boldsymbol{\phi}} (\mathbf{z}_m \! \mid \! \mathbf{x}_m ) \Vert  Q_{\boldsymbol{\psi}} (\mathbf{z}_m) \right) - \gamma \log P_{\boldsymbol{\theta}} ( \mathbf{u}_m \! \mid  \! \mathbf{z}_m )   \! + \! \lambda \log P_{\boldsymbol{\xi}} (\mathbf{s}_m \! \mid \! \mathbf{z}_m) \big)$
                
        \vspace{3pt}  
        
        \hspace{-15pt}(2) {\small\textbf{\textsf{Train the Latent Space Discriminator}} $ \boldsymbol{\eta} $}
        \State  ~~Sample $\{ \mathbf{x}_m \}_{m=1}^{M} \sim P_{\mathsf{D}} (\mathbf{X})$
        \State  ~~Sample $\{ \mathbf{n}_m \}_{m=1}^{M} \sim \mathcal{N}(\boldsymbol{0}, \mathbf{I})$
        \State  ~~Compute $\mathbf{z}_m \sim f_{\boldsymbol{\phi}} (\mathbf{x}_m), \forall m \in [M]$
        \State  ~~Compute $\mathbf{\widetilde{z}}_m \sim g_{\boldsymbol{\psi}} (\mathbf{n}_m), \forall m \in [M]$
        \State  ~~Back-propagate loss:\vspace{3pt}
        
                $\;\;\;\; \mathcal{L} \! \left( \boldsymbol{\eta} \right) \! = \!   -\frac{1}{M} \sum_{m=1}^{M} \! \left( \log D_{\boldsymbol{\eta}} (\mathbf{z}_m) \right) - \frac{1}{M} \sum_{m=1}^{M} \! \left( \log \left( 1- D_{\boldsymbol{\eta}} (\, \mathbf{\widetilde{z}}_m \,) \right) \right) $
       % \State  ~~Clip discriminator $\boldsymbol{\eta}$ to $\left[ - \epsilon , \epsilon \right]^d$      
          
        \vspace{3pt}
        
        \hspace{-15pt}(3) {\small\textbf{\textsf{Train the Encoder and Prior Distribution Generator $\left( \boldsymbol{\phi}, \boldsymbol{\psi} \right)$ Adversarially}}} 
        \State  ~~Sample $\{ \mathbf{x}_m \}_{m=1}^{M} \sim P_{\mathsf{D}} (\mathbf{X})$
        \State  ~~Sample $\{ \mathbf{n}_m \}_{m=1}^{M} \sim \mathcal{N}(\boldsymbol{0}, \mathbf{I})$
        \State  ~~Compute $\mathbf{z}_m \sim f_{\boldsymbol{\phi}} (\mathbf{x}_m), \forall m \in [M]$
        \State  ~~Compute $\mathbf{\widetilde{z}}_m \sim g_{\boldsymbol{\psi}} (\mathbf{n}_m), \forall m \in [M]$
        \State  ~~Back-propagate loss:\vspace{3pt}
        
                $\;\;\;\; \mathcal{L} \! \left( \boldsymbol{\phi}, \boldsymbol{\psi} \right) \! = \!  \frac{1}{M} \sum_{m=1}^{M} \! \left( \log D_{\boldsymbol{\eta}} (\mathbf{z}_m) \right) + \frac{1}{M} \sum_{m=1}^{M} \! \left( \log \left( 1- D_{\boldsymbol{\eta}} (\, \mathbf{\widetilde{z}}_m \,) \right) \right)$
        \vspace{3pt}

        \hspace{-15pt}(4) {\small\textbf{\textsf{Train the Utility Attribute Class Discriminator}} $ \boldsymbol{\omega} $}
        \State  ~~Sample $\{ \mathbf{u}_m \}_{m=1}^{M} \sim P_{\mathbf{U}} $
        \State  ~~Sample $\{ \mathbf{n}_m \}_{m=1}^{M} \sim \mathcal{N} \! \left( \boldsymbol{0}, \mathbf{I}\right)$
        %\State  ~~Compute $\mathbf{z}_m \sim g_{\boldsymbol{\psi}} (\mathbf{n}_m), \forall m \in [M]$
        \State  ~~Compute $\mathbf{\widetilde{u}}_m \sim g_{\boldsymbol{\theta}} \left( g_{\boldsymbol{\psi}} (\mathbf{n}_m) \right), \forall m \in [M]$
        \State  ~~Back-propagate loss:\vspace{3pt}
        
                $\;\;\;\; \mathcal{L} \! \left( \boldsymbol{\omega} \right) \! = \!   \gamma \left[ - \frac{1}{M} \sum_{m=1}^{M} \! \left( \log D_{\boldsymbol{\omega}} (\mathbf{u}_m) \right) - \frac{1}{M} \sum_{m=1}^{M} \! \left( \log \left( 1- D_{\boldsymbol{\omega}} (\, \mathbf{\widetilde{u}}_m \,) \right) \right) \right] $
        
        \vspace{3pt}

        \hspace{-15pt}(5) {\small\textbf{\textsf{Train the Sensitive Attribute Class Discriminator}} $ \boldsymbol{\tau} $}
        \State  ~~Sample $\{ \mathbf{s}_m \}_{m=1}^{M} \sim P_{\mathbf{S}} $
        \State  ~~Sample $\{ \mathbf{n}_m \}_{m=1}^{M} \sim \mathcal{N} \! \left( \boldsymbol{0}, \mathbf{I}\right)$
        %\State  ~~Compute $\mathbf{z}_m \sim g_{\boldsymbol{\psi}} (\mathbf{n}_m), \forall m \in [M]$
        \State  ~~Compute $\mathbf{\widetilde{s}}_m \sim g_{\boldsymbol{\xi}} \left( g_{\boldsymbol{\psi}} (\mathbf{n}_m) \right), \forall m \in [M]$
        \State  ~~Back-propagate loss:\vspace{3pt}
        
                $\;\;\;\; \mathcal{L} \! \left( \boldsymbol{\tau} \right) \! = \!   \lambda \left[ \frac{1}{M} \sum_{m=1}^{M} \! \left( \log D_{\boldsymbol{\tau}} (\mathbf{s}_m) \right) + \frac{1}{M} \sum_{m=1}^{M} \! \left( \log \left( 1- D_{\boldsymbol{\tau}} (\, \mathbf{\widetilde{s}}_m \,) \right) \right) \right]$
        
        \vspace{3pt} 
        
        \hspace{-15pt}(6) {\small\textbf{\textsf{Train the Prior Distribution Generator, Utility Decoder, and Uncertainty Decoder $\left( \boldsymbol{\psi}, \boldsymbol{\theta}, \boldsymbol{\xi} \right)$ Adversarially}}}
        %\Comment{\textcolor{blue!60!gray}{Utility Decoder Task: Attribute Classification}}
        \State  ~~Sample $\{ \mathbf{n}_m \}_{m=1}^{M} \sim \mathcal{N} \! \left( \boldsymbol{0}, \mathbf{I}\right)$
        %\State  ~~Compute $\mathbf{z}_m \sim g_{\boldsymbol{\psi}} (\mathbf{n}_m), \forall m \in [M]$
        \State  ~~Compute $\mathbf{\widetilde{u}}_m \sim g_{\boldsymbol{\theta}} \left( g_{\boldsymbol{\psi}} (\mathbf{n}_m) \right), \forall m \in [M]$
        \State  ~~Compute $\mathbf{\widetilde{s}}_m \sim g_{\boldsymbol{\xi}} \left( g_{\boldsymbol{\psi}} (\mathbf{n}_m) \right), \forall m \in [M]$
        \State  ~~Back-propagate loss:\vspace{3pt}
        
                $\;\;\;\; \mathcal{L} \! \left( \boldsymbol{\psi}, \boldsymbol{\theta}, \boldsymbol{\xi} \right) \! = \!  \gamma \,  \frac{1}{M} \sum_{m=1}^{M} \! \left( \log \left( 1- D_{\boldsymbol{\omega}} (\, \mathbf{\widetilde{u}}_m \,) \right) \right) - \lambda \, \frac{1}{M} \sum_{m=1}^{M} \! \left( \log \left( 1- D_{\boldsymbol{\tau}} (\, \mathbf{\widetilde{s}}_m \,) \right) \right)$
        
        \vspace{3pt}  
        
        \Until{Convergence}
        \State \textbf{return} $\boldsymbol{\phi}, \boldsymbol{\theta}, \boldsymbol{\psi}, \boldsymbol{\xi}, \boldsymbol{\eta}, \boldsymbol{\omega}, \boldsymbol{\tau}$
    \end{algorithmic}
    \caption{Supervised Deep Variational CLUB training algorithm associated with ($\text{P3}\!\!:\! \mathsf{S}$)}
    \label{Algorithm:P3_Supervised_DVCLUB_S_LowerBound}
\end{algorithm}
\end{spacing}
%\end{minipage} 
\end{center}
%
%---------------------------------------------------
%---------------------------------------------------

%---------------------------------------------------
%
%       TABLE:  Un-Supervised DVCLUB Training Algorithm
%
%---------------------------------------------------
%
\begin{center}
%\begin{minipage}{0.9\linewidth}
\centering
\begin{spacing}{1}
\begin{algorithm}
    \small
    \setstretch{1.2}
    \begin{algorithmic}[1]
        \State \textbf{Input:} Training Dataset: $\{ \left( \mathbf{s}_n, \mathbf{x}_n  \right) \}_{n=1}^{N}$; Hyper-Parameters: $\gamma,\lambda$
        \State $\boldsymbol{\phi}, \boldsymbol{\theta}, \boldsymbol{\psi}, \boldsymbol{\xi}, \boldsymbol{\eta}, \boldsymbol{\omega}, \boldsymbol{\tau} \; \gets$ Initialize Network Parameters
        
        \Repeat
        
        \hspace{-15pt}(1) {\small\textbf{\textsf{Train the Encoder, Utility Decoder, Uncertainty Decoder}} $\left( \boldsymbol{\phi}, \boldsymbol{\theta}, \boldsymbol{\xi} \right)$}
         \State  ~~Sample a mini-batch $\{ \mathbf{s}_m, \mathbf{x}_m \}_{m=1}^{M} \sim P_{\mathsf{D}} (\mathbf{X}) P_{\mathbf{S} 
         \mid \mathbf{X}}$
         \State  ~~Compute $\mathbf{z}_m \sim f_{\boldsymbol{\phi}} (\mathbf{x}_m), \forall m \in [M]$
        %  \Comment{\textcolor{blue!60!gray}{$f_{\boldsymbol{\phi}}$: Stochastic or Deterministic Encoder}}
         %\State  ~~Sample $\{ \mathbf{\widehat{z}}_m \}_{m=1}^{M} \sim Q_{\boldsymbol{\psi}} (\mathbf{Z})$
         %\State  ~~Compute $\mathbf{\widehat{x}}_m =  g_{\boldsymbol{\xi}} (\mathbf{\widehat{z}}_m, \mathbf{s}_m), \forall m \in [M]$
         \State  ~~Back-propagate loss:\vspace{3pt}
         
                $ \mathcal{L} \! \left( \boldsymbol{\phi}, \! \boldsymbol{\theta}, \! \boldsymbol{\xi} \right) \! = \!\! - \frac{1}{M} \! \sum_{m=1}^{M} \!\! \big( \D_{\mathrm{KL}} \! \left( P_{\boldsymbol{\phi}} (\mathbf{z}_m \! \mid \! \mathbf{x}_m ) \Vert  Q_{\boldsymbol{\psi}} (\mathbf{z}_m) \right) - \gamma \log P_{\boldsymbol{\theta}} ( \mathbf{x}_m \! \mid  \! \mathbf{z}_m )   \! + \! \lambda \log P_{\boldsymbol{\xi}} (\mathbf{s}_m \! \mid \! \mathbf{z}_m) \big)$
                
        \vspace{3pt}  
        
        \hspace{-15pt}(2) {\small\textbf{\textsf{Train the Latent Space Discriminator}} $ \boldsymbol{\eta} $}
        \State  ~~Sample $\{ \mathbf{x}_m \}_{m=1}^{M} \sim P_{\mathsf{D}} (\mathbf{X})$
        \State  ~~Sample $\{ \mathbf{n}_m \}_{m=1}^{M} \sim \mathcal{N}(\boldsymbol{0}, \mathbf{I})$
        \State  ~~Compute $\mathbf{z}_m \sim f_{\boldsymbol{\phi}} (\mathbf{x}_m), \forall m \in [M]$
        \State  ~~Compute $\mathbf{\widetilde{z}}_m \sim g_{\boldsymbol{\psi}} (\mathbf{n}_m), \forall m \in [M]$
        \State  ~~Back-propagate loss:\vspace{3pt}
        
                $\;\;\;\; \mathcal{L} \! \left( \boldsymbol{\eta} \right) \! = \!   -\frac{1}{M} \sum_{m=1}^{M} \! \left( \log D_{\boldsymbol{\eta}} (\mathbf{z}_m) \right) - \frac{1}{M} \sum_{m=1}^{M} \! \left( \log \left( 1- D_{\boldsymbol{\eta}} (\, \mathbf{\widetilde{z}}_m \,) \right) \right) $
       % \State  ~~Clip discriminator $\boldsymbol{\eta}$ to $\left[ - \epsilon , \epsilon \right]^d$      
          
        \vspace{3pt}
        
        \hspace{-15pt}(3) {\small\textbf{\textsf{Train the Encoder and Prior Distribution Generator $\left( \boldsymbol{\phi}, \boldsymbol{\psi} \right)$ Adversarially}}} 
        \State  ~~Sample $\{ \mathbf{x}_m \}_{m=1}^{M} \sim P_{\mathsf{D}} (\mathbf{X})$
        \State  ~~Sample $\{ \mathbf{n}_m \}_{m=1}^{M} \sim \mathcal{N}(\boldsymbol{0}, \mathbf{I})$
        \State  ~~Compute $\mathbf{z}_m \sim f_{\boldsymbol{\phi}} (\mathbf{x}_m), \forall m \in [M]$
        \State  ~~Compute $\mathbf{\widetilde{z}}_m \sim g_{\boldsymbol{\psi}} (\mathbf{n}_m), \forall m \in [M]$
        \State  ~~Back-propagate loss:\vspace{3pt}
        
                $\;\;\;\; \mathcal{L} \! \left( \boldsymbol{\phi}, \boldsymbol{\psi} \right) \! = \!  \frac{1}{M} \sum_{m=1}^{M} \! \left( \log D_{\boldsymbol{\eta}} (\mathbf{z}_m) \right) + \frac{1}{M} \sum_{m=1}^{M} \! \left( \log \left( 1- D_{\boldsymbol{\eta}} (\, \mathbf{\widetilde{z}}_m \,) \right) \right)$
        \vspace{3pt}

        \hspace{-15pt}(4) {\small\textbf{\textsf{Train the Visible Space Discriminator}} $ \boldsymbol{\omega} $}
        \State  ~~Sample $\{ \mathbf{x}_m \}_{m=1}^{M} \sim P_{\mathsf{D}} (\mathbf{X}) $
        \State  ~~Sample $\{ \mathbf{n}_m \}_{m=1}^{M} \sim \mathcal{N} \! \left( \boldsymbol{0}, \mathbf{I}\right)$
        %\State  ~~Compute $\mathbf{z}_m \sim g_{\boldsymbol{\psi}} (\mathbf{n}_m), \forall m \in [M]$
        \State  ~~Compute $\mathbf{\widetilde{x}}_m \sim g_{\boldsymbol{\theta}} \left( g_{\boldsymbol{\psi}} (\mathbf{n}_m) \right), \forall m \in [M]$
        \State  ~~Back-propagate loss:\vspace{3pt}
        
                $\;\;\;\; \mathcal{L} \! \left( \boldsymbol{\omega} \right) \! = \!   \gamma \left[ - \frac{1}{M} \sum_{m=1}^{M} \! \left( \log D_{\boldsymbol{\omega}} (\mathbf{x}_m) \right) - \frac{1}{M} \sum_{m=1}^{M} \! \left( \log \left( 1- D_{\boldsymbol{\omega}} (\, \mathbf{\widetilde{x}}_m \,) \right) \right) \right] $
        
        \vspace{3pt}

        \hspace{-15pt}(5) {\small\textbf{\textsf{Train the Sensitive Attribute Class Discriminator}} $ \boldsymbol{\tau} $}
        \State  ~~Sample $\{ \mathbf{s}_m \}_{m=1}^{M} \sim P_{\mathbf{S}} $
        \State  ~~Sample $\{ \mathbf{n}_m \}_{m=1}^{M} \sim \mathcal{N} \! \left( \boldsymbol{0}, \mathbf{I}\right)$
        %\State  ~~Compute $\mathbf{z}_m \sim g_{\boldsymbol{\psi}} (\mathbf{n}_m), \forall m \in [M]$
        \State  ~~Compute $\mathbf{\widetilde{s}}_m \sim g_{\boldsymbol{\xi}} \left( g_{\boldsymbol{\psi}} (\mathbf{n}_m) \right), \forall m \in [M]$
        \State  ~~Back-propagate loss:\vspace{3pt}
        
                $\;\;\;\; \mathcal{L} \! \left( \boldsymbol{\tau} \right) \! = \!   \lambda \left[ \frac{1}{M} \sum_{m=1}^{M} \! \left( \log D_{\boldsymbol{\tau}} (\mathbf{s}_m) \right) + \frac{1}{M} \sum_{m=1}^{M} \! \left( \log \left( 1- D_{\boldsymbol{\tau}} (\, \mathbf{\widetilde{s}}_m \,) \right) \right) \right]$
        
        \vspace{3pt} 
        
        \hspace{-15pt}(6) {\small\textbf{\textsf{Train the Prior Distribution Generator, Utility Decoder, and Uncertainty Decoder $\left( \boldsymbol{\psi}, \boldsymbol{\theta}, \boldsymbol{\xi} \right)$ Adversarially}}}
        %\Comment{\textcolor{blue!60!gray}{Utility Decoder Task: Attribute Classification}}
        \State  ~~Sample $\{ \mathbf{n}_m \}_{m=1}^{M} \sim \mathcal{N} \! \left( \boldsymbol{0}, \mathbf{I}\right)$
        %\State  ~~Compute $\mathbf{z}_m \sim g_{\boldsymbol{\psi}} (\mathbf{n}_m), \forall m \in [M]$
        \State  ~~Compute $\mathbf{\widetilde{x}}_m \sim g_{\boldsymbol{\theta}} \left( g_{\boldsymbol{\psi}} (\mathbf{n}_m) \right), \forall m \in [M]$
        \State  ~~Compute $\mathbf{\widetilde{s}}_m \sim g_{\boldsymbol{\xi}} \left( g_{\boldsymbol{\psi}} (\mathbf{n}_m) \right), \forall m \in [M]$
        \State  ~~Back-propagate loss:\vspace{3pt}
        
                $\;\;\;\; \mathcal{L} \! \left( \boldsymbol{\psi}, \boldsymbol{\theta} , \boldsymbol{\xi} \right) \! = \!  \gamma \,  \frac{1}{M} \sum_{m=1}^{M} \! \left( \log \left( 1- D_{\boldsymbol{\omega}} (\, \mathbf{\widetilde{x}}_m \,) \right) \right) - \lambda \, \frac{1}{M} \sum_{m=1}^{M} \! \left( \log \left( 1- D_{\boldsymbol{\tau}} (\, \mathbf{\widetilde{s}}_m \,) \right) \right)$
        
        \vspace{3pt}  
        
        \Until{Convergence}
        \State \textbf{return} $\boldsymbol{\phi}, \boldsymbol{\theta}, \boldsymbol{\psi}, \boldsymbol{\xi}, \boldsymbol{\eta}, \boldsymbol{\omega}, \boldsymbol{\tau}$
    \end{algorithmic}
    \caption{Unsupervised Deep Variational CLUB training algorithm associated with ($\text{P3}\!\!:\!\mathsf{U} $)}
    \label{Algorithm:P3_UnSupervised_DVCLUB_S_LowerBound}
\end{algorithm}
\end{spacing}
%\end{minipage} 
\end{center}
%
%---------------------------------------------------
%---------------------------------------------------

\end{document}